%% file: main.tex
\documentclass{article}

\PassOptionsToPackage{round}{natbib}

\usepackage{xurl}

\usepackage[main, final]{neurips_2026}

\usepackage[utf8]{inputenc} 
\usepackage[T1]{fontenc}   
\usepackage{url}            
\usepackage{booktabs}     
\usepackage{amsfonts}      
\usepackage{nicefrac}      
\usepackage{microtype}      
\usepackage{graphicx}
\usepackage{wrapfig}

\usepackage{amsmath}
\usepackage{subcaption}
\usepackage{hyperref}
\usepackage{comment}
\usepackage{etoc}

\hypersetup{
    colorlinks,
    linkcolor={red!50!black},
    citecolor={blue!50!black},
    urlcolor={blue!80!black}
}

\usepackage[table]{xcolor}
\usepackage{colortbl}

\usepackage{xskak}
\usepackage{tikz,array}
\usetikzlibrary{fit,backgrounds}

\definecolor{heatmapcolor}{HTML}{2E86AB}

\usepackage{booktabs}
\usepackage{xcolor}
\usepackage{tikz}
\usepackage{array}
\usepackage{collcell}

\definecolor{heatmapcolor}{HTML}{0984E3}

\newcolumntype{C}[1]{>{\centering\arraybackslash}p{#1}}

\definecolor{first_target}{HTML}{00B894}
\definecolor{third_target}{HTML}{0984E3}
\definecolor{corrupted}{HTML}{FF7675}
\definecolor{other}{HTML}{495057}

\newcommand{\firsttarget}[1]{\textbf{\textcolor{first_target}{#1}}}
\newcommand{\thirdtarget}[1]{\textbf{\textcolor{third_target}{#1}}}
\newcommand{\corrupted}[1]{\textbf{\textcolor{corrupted}{#1}}}
\newcommand{\othersquare}[1]{\textbf{\textcolor{other}{#1}}}

\newcommand{\first}{1st}

\newcommand{\third}{3rd}

\definecolor{king_safety_opp}{HTML}{E69F00}
\definecolor{threats_mine}{HTML}{CC79A7}
\definecolor{king_safety_mine}{HTML}{0984E3}   %
\definecolor{threats_opp}{HTML}{00B894}         %
\definecolor{baseline_unsteered}{HTML}{FF7675}
\definecolor{random_direction}{HTML}{495057}

\newcommand{\kingsafetyopp}[1]{\textbf{\textcolor{king_safety_opp}{#1}}}
\newcommand{\threatsmine}[1]{\textbf{\textcolor{threats_mine}{#1}}}
\newcommand{\kingsafetymine}[1]{\textbf{\textcolor{king_safety_mine}{#1}}}
\newcommand{\threatsopp}[1]{\textbf{\textcolor{threats_opp}{#1}}}
\newcommand{\baselineunsteered}[1]{\textbf{\textcolor{baseline_unsteered}{#1}}}
\newcommand{\randomdirection}[1]{\textbf{\textcolor{random_direction}{#1}}}

\definecolor{at_layer}{HTML}{0984E3}
\definecolor{by_any_layer}{HTML}{00B894}
\definecolor{first_at_layer}{HTML}{E69F00}
\definecolor{forgotten_hatch}{HTML}{FF7675}

\newcommand{\atlayer}[1]{\textbf{\textcolor{at_layer}{#1}}}
\newcommand{\byanylayer}[1]{\textbf{\textcolor{by_any_layer}{#1}}}
\newcommand{\firstatlayer}[1]{\textbf{\textcolor{first_at_layer}{#1}}}
\newcommand{\forgotten}[1]{\textbf{\textcolor{forgotten_hatch}{#1}}}

\definecolor{training_target}{HTML}{CC79A7}
\definecolor{penultimate_layer}{HTML}{E69F00}
\definecolor{final_layer}{HTML}{0984E3}
\definecolor{solved_gain_hatch}{HTML}{00B894}

\newcommand{\trainingtarget}[1]{\textbf{\textcolor{training_target}{#1}}}
\newcommand{\penultimatelayer}[1]{\textbf{\textcolor{penultimate_layer}{#1}}}
\newcommand{\finallayer}[1]{\textbf{\textcolor{final_layer}{#1}}}
\newcommand{\solvedgain}[1]{\textbf{\textcolor{solved_gain_hatch}{#1}}}

\definecolor{steer_pos}{HTML}{3783BB}
\definecolor{steer_neg}{HTML}{C94741}
\newcommand{\steerpos}[1]{\textbf{\textcolor{steer_pos}{#1}}}
\newcommand{\steerneg}[1]{\textbf{\textcolor{steer_neg}{#1}}}

\newcommand{\gradcell}[3]{%
  \hspace*{-\tabcolsep}%
  \hspace*{-2.75pt}%
  \tikz[baseline=(t.base)]{%
    \node[
      rectangle,
      minimum width=2.33em+2\tabcolsep-3pt,
      text depth=.25ex,  %
      text height=1.5ex,
      inner sep=2pt,  %
      outer sep=0pt,
      path picture={
        \shade[left color=heatmapcolor!#1!white,
               right color=heatmapcolor!#2!white]
          (path picture bounding box.south west) 
          rectangle 
          (path picture bounding box.north east);
      }
    ] (t) {#3};  %
  }%
  \hspace*{-\tabcolsep}%
  \hspace*{-2.75pt}%
}

\title{The Algorithm Is Not the Behavior:\\Learned Priors Override Look-Ahead\\in a Chess-Playing Neural Network}

\author{%
  Elias Sandmann\\
  Fraunhofer HHI 
  \And
  Sebastian Lapuschkin\\
  Fraunhofer HHI \\
  TU Dublin
  \And
  Wojciech Samek\\
  Fraunhofer HHI\\
  TU Berlin
  \and 
  \texttt{firstname.lastname@hhi.fraunhofer.de}
}

\begin{document}

\maketitle

\begin{abstract}Recent mechanistic work has uncovered learned algorithms within neural networks, from modular arithmetic to search and planning in game-playing agents. But does algorithmic structure guarantee algorithmic behavior? We investigate this in Leela Chess Zero, the strongest neural chess engine, where prior work identified learned look-ahead. By extending the logit lens to its move-selecting policy network, we discover that correct puzzle solutions—including immediate checkmates—often appear in intermediate layers but are systematically overridden in the final output, a phenomenon we term \emph{forgotten puzzles}. Replicating prior analyses on these positions, we find that look-ahead operates normally—future moves of the correct continuation are represented, causally important, and linearly decodable—ruling out a failure of the algorithm. Instead, late layers increasingly shift toward safe play over aggression. To test whether this shift drives the override, we steer the model against these preferences and recover $71.7\%$ of forgotten puzzles, providing causal evidence that safety priors override algorithmically computed solutions. These findings show that algorithmic structure does not guarantee algorithmic behavior: a model can internally solve a problem and still output the wrong answer.
\end{abstract}

\etocdepthtag.toc{mtmain}
\section{Introduction}

When mechanistic analysis identifies a learned algorithm inside a neural network, does that algorithm determine the network's output? Several works have provided evidence for algorithmic computation within neural networks, from modular arithmetic \citep{nanda2023progress} and search in synthetic tasks \citep{brinkmann2024mechanistic} to planning in Sokoban \citep{bush2025interpreting, taufeeque2026path} and look-ahead in a chess-playing neural network \citep{jenner2024evidencelearnedlookaheadchessplaying}. These works establish that learned algorithms exist and are causally active, focusing on how they function when they succeed. Less is known about how they can fail: can a model internally compute the correct solution through a learned algorithm only to output the wrong answer—and if so, what overrides it?

We investigate this by extending the logit lens \citep{nostalgebraist2020logitlens}—which projects intermediate representations through a model's output head \citep{jastrzębski2018residualconnectionsencourageiterative} to reveal what the network would predict at each layer—to the move-selecting policy network of Leela Chess Zero \citep{leela}, the model in which \citet{jenner2024evidencelearnedlookaheadchessplaying} identified learned look-ahead and the strongest neural chess engine available today \citep{tcec}. Leela's encoder-only transformer architecture provides distinct interpretability advantages: it treats each chessboard square as a token and, unlike decoder-only language models where the logit lens is limited to decoding from a single token position, uses all tokens simultaneously for prediction, giving the logit lens direct access to every intermediate activation contributing to the model's output. 

Applying the logit lens reveals interpretable policies at every layer, with move preferences following a structured progression through recognizable chess heuristics of increasing complexity rather than converging smoothly on the final answer (Figure~\ref{fig:puzzle_example}). But the layer-wise view also exposes a surprising phenomenon: in chess puzzles with a single clearly winning move, the correct solution often appears in intermediate layers only to be overridden in the final output, including immediate checkmates where even a single step of look-ahead should suffice (Figure~\ref{fig:forgotten_puzzle}).

Do these \emph{forgotten puzzles} arise despite look-ahead functioning normally, or does the algorithm simply fail in these positions? By replicating the analyses of \citet{jenner2024evidencelearnedlookaheadchessplaying} on forgotten puzzles, we find that look-ahead operates as expected: future moves of the correct continuation are represented, causally important, and linearly decodable. If the algorithm is not at fault, what overrides it? Characterization of these positions reveals that the model overwhelmingly replaces tactical solutions, such as sacrifices, with quiet positional alternatives, suggesting that learned priors for safe play may be responsible. Such priors would reflect sensible generalizations from the training distribution that conflict with the correct move in rare tactical positions. To test for such priors, we measure which human-interpretable chess concepts, such as king safety and threats, each layer prioritizes, and find that late layers increasingly favor safer concepts over more aggressive ones. Gradient-based steering against these preferences recovers $71.7\%$ of forgotten puzzles, providing causal evidence that \emph{safety priors} override algorithmically computed solutions.

Our contributions are: (1) We extend the logit lens to the Post-LN architecture of Leela's policy network, yielding highly interpretable intermediate policies whose playing strength improves consistently with depth. (2) Using this, we identify forgotten puzzles—positions where intermediate layers find the correct solution but the final output does not—and verify that look-ahead functions normally in these positions. (3) We provide causal evidence that learned safety priors override these algorithmically computed solutions, recovering a majority of forgotten puzzles through gradient-based steering against safety concepts. (4) We show that forgotten puzzles are genuine failures of the network rather than of its objective, as the policy's own training target would place almost all of its weight on the solution. Since both the algorithm and the priors emerge from the same likelihood-based training objective, these results raise the question of whether such training can  produce models that reliably prefer algorithmically computed solutions over distributional  heuristics, even in domains with verifiable ground truth. More broadly, our findings imply that models may possess capabilities their outputs do not reflect, and that mechanistic understanding can identify such failure modes and suggest principled interventions to address them. \href{https://github.com/hartigel/leela-logit-lens}{Our code is available on GitHub}.

\vspace{-0pt}
\begin{figure}[t]
\centering
\includegraphics[width=1.0\textwidth]{Figures/forgotten_puzzle_figure.pdf}
\vspace{-22pt}
\caption{Our extended logit lens maps intermediate activations to policy distributions across transformer layers in Leela Chess Zero, revealing a forgotten mate-in-two puzzle with optimal line of play—the \emph{principal variation}—1.~\protect\wmove{Qxf1+} \protect\wmove{Kxf1} 2.~\protect\wmove{Rd1\mate}. \textbf{Left:} The \firsttarget{correct} queen sacrifice \protect\wmove{Qxf1+} maintains high probability through most layers, peaking at layer $12$, before dropping sharply in the final layers. The \corrupted{losing} move \protect\wmove{Qc8}—a quiet queen retreat to the back rank—remains near zero throughout the network and rises only in the final layer to become the top choice. \textbf{Right:} The value head, trained to predict game outcomes, correctly assesses both resulting positions, indicating that the policy head's final move selection contradicts the model's own evaluation of the consequences. At the puzzle position itself, the value head does not see the win. Full probabilities are provided in Appendix~\ref{app:forgotten_puzzles_example}, with additional examples in Appendix~\ref{app:forgotten_puzzles}.}
\label{fig:forgotten_puzzle}
\end{figure}

\section{Background}

\begin{figure}[t]
\vspace{0pt}

\centering
\includegraphics[width=1.0\textwidth]{Figures/main_figure.pdf}
\vspace{-18pt}
\caption{Our extended logit lens reveals progressive policy refinement across layers for a mate-in-two puzzle with principal variation 1.~\protect\wmove{Ng3+} \protect\wmove{hxg3} 2.~\protect\wmove{Rh6\mate}. Move preferences follow a structured progression: from queen moves reflecting piece priors before any inter-square interaction in the input encoding, through greedy captures in early layers, to forcing checks by layer $9$ and a positional rook lift at layer $12$, before converging on the correct knight sacrifice \protect\wmove{Ng3+} in the final output. While the intermediate preferences shown are all losing, the trajectory is non-arbitrary, with each stage reflecting recognizable chess heuristics of increasing complexity. Full probabilities are provided in Appendix~\ref{app:probs_example_puzzle}, with additional examples in Appendix~\ref{app:puzzles}.}
\label{fig:puzzle_example}
\vspace{-0pt}

\end{figure}

\subsection{Leela Chess Zero}

We analyze the \texttt{T82-768x15x24h} transformer model from Leela Chess Zero \citep{monroe2026chessformer}, the network studied by \citet{jenner2024evidencelearnedlookaheadchessplaying}. This model uses a Post-LN architecture similar to the original transformer \citep{vaswani2017attention} with DeepNorm scaling \citep{deepnet}, featuring a $15$-layer transformer encoder with $768$-dimensional embeddings and $24$ attention heads per layer. The shared encoder feeds two output heads: a policy head that outputs a distribution over legal moves and a value head that predicts game outcomes. Chess positions are encoded as $8 \times 8$ grids where each square corresponds to a token position. Architectural details are provided in Appendix~\ref{app:model_architecture}.

Leela is trained following the AlphaZero paradigm \citep{silver2018general} and normally functions in tandem with Monte Carlo tree search (MCTS) as a chess engine. Its training games are generated by self-play, in which each move is chosen by MCTS guided by a Leela network: the policy head supplies the prior over which moves to explore, and the value head predicts the game outcome for each position the search reaches. Moves with a higher prior or a better predicted outcome receive more visits, and the engine ultimately plays the move with the most visits at the root. For each position played in these games, the policy head is trained by cross-entropy against the normalized root visit counts of the search, and the value head against the final game outcome. Our main claims concern the policy network, which already demonstrates strong play even without external search \citep{leela_vs_gdm}, and we use search only to examine the network's training target and how its prior affects the engine in deployment. Search details are provided in Appendix~\ref{app:search_mechanics}. 

\subsection{Encoder-only Post-LN logit lens}

The logit lens \citep{nostalgebraist2020logitlens} projects intermediate activations after layer $\ell$  through the final layer normalization and unembedding matrix to obtain layer-wise predictions. This approach works seamlessly for Pre-LN transformers, where layer normalization precedes each sublayer and leaves the residual stream free of normalization operations. Post-LN architectures create a challenge by applying normalization after residual connections instead, transforming the residual stream at each layer and creating non-linear dependencies that prevent straightforward application of the logit lens.

Functionally, the Pre-LN logit lens is equivalent to applying zero ablation to all sublayer outputs beyond a given layer $\ell$ \citep{belrose2025elicitinglatentpredictionstransformers}. We extend this principle to Post-LN models by applying the same zero ablation while preserving the subsequent layer normalizations. For consistency we also ablate layer normalization biases for layers beyond $\ell$, as justified in Appendix~\ref{app:post_ln_logit_lens}.

Since Leela is an encoder that uses representations from all tokens simultaneously, we project the intermediate representations of all $64$ squares through the policy head to obtain policies at each layer.

\subsection{Evaluation setup} To verify that our Post-LN logit lens produces meaningful intermediate predictions, we evaluate the general playing strength and tactical puzzle-solving ability of layer-wise policies following \citet{ruoss2024amortizedplanninglargescaletransformers}. Additionally, we track distributional properties of intermediate policies across layers.

\paragraph{Playing strength assessment} We conduct a round-robin tournament between policies derived from representations at the input and all $15$ transformer layers. Each pairing plays $200$ Encyclopedia of Chess Openings \citep{matanovic1978encyclopaedia} positions with one game per side using argmax move selection and five games per side using a temperature of $\tau = 1.0$. Elo ratings are computed using BayesElo \citep{coulom208whole}. We include the Leela policy network from \citet{ruoss2024amortizedplanninglargescaletransformers} as a fixed anchor using argmax selection to match their evaluation and obtain absolute Elo scores. Additionally, we deploy the layer-wise policies as bots on Lichess \citep{lichess2025} across multiple time controls, using a temperature of $1.0$ for the first five full-moves to introduce opening diversity.

\paragraph{Puzzle-solving performance} We evaluate tactical understanding using the \emph{general dataset} of $10{,}000$ Lichess puzzles from \citet{ruoss2024amortizedplanninglargescaletransformers}, each constructed with a single clear winning line—the \emph{principal variation}—while all other moves are significantly inferior, though not necessarily losing. For each layer's policy, we use argmax selection to predict moves and consider a puzzle solved if this reproduces all the player's moves of the principal variation correctly or leads to checkmate.

\paragraph{Intermediate policy dynamics} To characterize how layer-wise policies evolve across depth, we track policy entropy and Kendall's $\tau$ ranking correlation between intermediate and final move rankings at each layer. Additional metrics are provided in Appendix~\ref{app:policy_dynamics}.

\subsection{Learned look-ahead}
\label{sec:look_ahead}
\citet{jenner2024evidencelearnedlookaheadchessplaying} demonstrated that Leela internally represents future moves of correct puzzle continuations and that these representations are causally important for its output. They analyzed positions where the model correctly solves tactical puzzles with a clear principal variation and presented three lines of evidence, which we briefly summarize below. We replicate their analyses on forgotten puzzles to determine whether this forgetting reflects a failure of the look-ahead mechanism itself or whether look-ahead operates normally but is overridden.

\paragraph{Dataset}
To study look-ahead, they filter for positions where the model is especially likely to employ look-ahead rather than simple heuristics—tactical puzzles with a unique best line that are difficult to evaluate without considering future moves. They focus on puzzles the model solves correctly; applying their criteria yields $23{,}601$ positions (the \emph{solved dataset}). We apply identical filtering criteria but select for forgotten puzzles, yielding $8{,}648$ positions (the \emph{forgotten dataset}), indicating that forgotten puzzles are far from an edge case. To ensure these represent genuine forgetting rather than absence of look-ahead, we also require the full model to assign $>50\%$ probability to all the player's subsequent moves in the principal variation, given the preceding correct moves, confirming that the model demonstrably understands the winning continuation. Neither dataset filters for the presence of look-ahead mechanisms; any effects observed are emergent rather than selected for.

\paragraph{Activation patching} To measure the causal importance of individual square representations, \citet{jenner2024evidencelearnedlookaheadchessplaying} employ activation patching. For each position, a corrupted board state is generated by applying a small modification—such as adding or removing a single piece—selected to substantially affect the model's output while minimally affecting a weaker model, to target mechanisms beyond simple heuristics. Representations from the corrupted forward pass are then patched into the clean forward pass at each layer one square at a time. The resulting change in log-odds of the correct move measures each square's causal importance at that depth. They find that target squares of future moves in the principal variation exhibit outsized causal effects, indicating that look-ahead information is actively used. Although the model does not select the correct move in forgotten puzzles, it still assigns some probability to it. If this residual probability originates from look-ahead, causal interventions targeting look-ahead should reduce it, and log-odds reduction remains a meaningful metric.

\paragraph{Attention head ablation} \citet{jenner2024evidencelearnedlookaheadchessplaying} identified specific attention heads that transfer information between squares involved in future moves—in particular, heads that attend from the first move's target square to the third move's target square, effectively relaying look-ahead information backward along the principal variation. To test the causal role of this information flow, they zero-ablate the single attention entry corresponding to this query-key pair and find that it has an outsized effect on the log-odds of the correct move.

\paragraph{Probing} To test whether information about future moves is explicitly encoded in the model's representations, \citet{jenner2024evidencelearnedlookaheadchessplaying} train bilinear probes \citep{hewitt2019designing}, which score pairs of squares through a bilinear form of their activations, to predict the third move in the principal variation in two steps: first predicting the target square of the third move from activations at the first move's target square, then predicting the source square conditioned on this predicted target. Full methodological details for all look-ahead analyses are provided in Appendix~\ref{app:look_ahead}.

\section{Results}

\subsection{Capability progression and policy dynamics}

\paragraph{Tournament strength}

Table~\ref{tab:tournament_results_full} reports Elo ratings across layers. Playing strength increases with depth but suggests a three-phase progression rather than uniform improvement. Early layers show rapid gains through layer $5$, middle layers form a performance plateau through approximately layer $10$, and late layers demonstrate sharp strengthening beginning around layer $11$. This pattern holds consistently under both deterministic $(\tau = 0)$ and stochastic $(\tau = 1 )$ move selection. Real-world Lichess deployment shows similar trends with clear late-layer strengthening (Appendix~\ref{app:tournament}).

\begin{table*}[h]
\centering
\caption{Playing strength (Elo rating) across transformer layers and evaluation methods}
\label{tab:tournament_results_full}
\resizebox{\textwidth}{!}{%
\begin{tabular}{l*{16}{c}c}
\toprule
\textbf{Evaluation} & \textbf{Input} & \textbf{L0} & \textbf{L1} & \textbf{L2} & \textbf{L3} & \textbf{L4} & \textbf{L5} & \textbf{L6} & \textbf{L7} & \textbf{L8} & \textbf{L9} & \textbf{L10} & \textbf{L11} & \textbf{L12} & \textbf{L13} & \textbf{Full} & \textbf{Anchor} \\
\midrule
Internal Tournament ($\tau=0$) &
\gradcell{10}{17}{351} &
\gradcell{17}{19}{609} &
\gradcell{19}{22}{655} &
\gradcell{22}{25}{730} &
\gradcell{25}{30}{832} &
\gradcell{30}{34}{908} &
\gradcell{34}{36}{993} &
\gradcell{36}{36}{997} &
\gradcell{36}{36}{1028} &
\gradcell{36}{36}{1030} &
\gradcell{36}{37}{1055} &
\gradcell{37}{38}{1082} &
\gradcell{38}{41}{1117} &
\gradcell{41}{54}{1356} &
\gradcell{54}{70}{1670} &
\gradcell{70}{100}{2268} &
2292 \\[2pt]
Internal Tournament ($\tau=1$) &
\gradcell{10}{33}{46} &
\gradcell{33}{34}{387} &
\gradcell{34}{39}{391} &
\gradcell{39}{47}{469} &
\gradcell{47}{56}{585} &
\gradcell{56}{62}{718} &
\gradcell{62}{61}{810} &
\gradcell{61}{63}{799} &
\gradcell{63}{62}{816} &
\gradcell{62}{65}{813} &
\gradcell{65}{66}{851} &
\gradcell{66}{67}{872} &
\gradcell{67}{79}{876} &
\gradcell{79}{83}{1065} &
\gradcell{83}{92}{1119} &
\gradcell{92}{100}{1367} &
-- \\
\midrule
Lichess Bullet & 817 & 843 & 907 & 930 & 961 & 998 & 1071 & 1085 & 1087 & 1096 & 1094 & 1124 & 1159 & 1399 & 1622 & 2306 & -- \\
Lichess Blitz  & 545 & 631 & 611 & 655 & 740 & 723 & 815 & 815 & 827 & 906 & 952 & 860 & 946 & 1205 & 1482 & 2099 & -- \\
Lichess Rapid  & 492 & 617 & 569 & 669 & 789 & 791 & 871 & 907 & 954 & 954 & 985 & 1008 & 1072 & 1324 & 1637 & 2237 & -- \\
\bottomrule
\end{tabular}%
}
\end{table*}

\begin{figure}[h]
\centering
\includegraphics[width=1.0\textwidth]{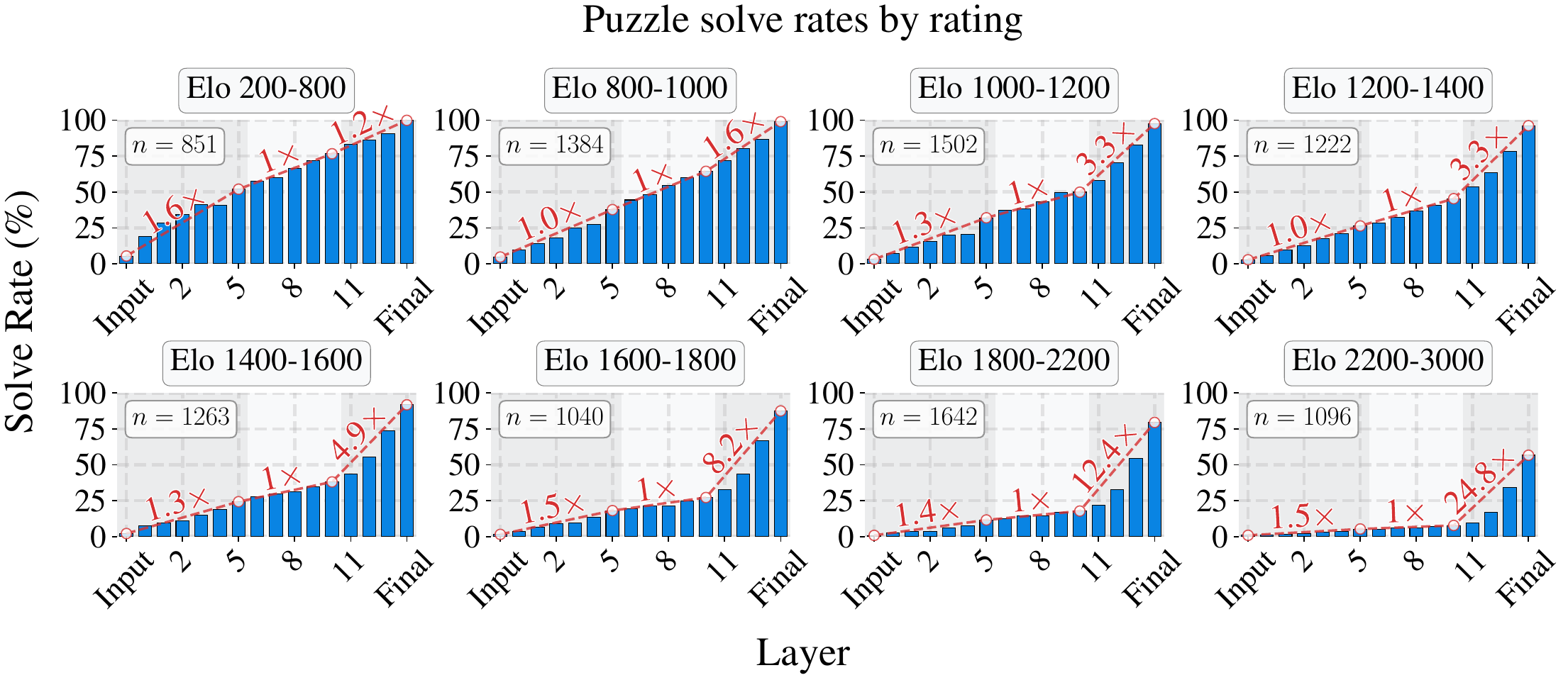}
\vspace{-12pt}
\caption{Puzzle-solving performance across layers, stratified by Elo rating. Red dashed lines mark phase boundaries derived from tournament analysis, indicating phase-specific improvement rates.}
\label{fig:puzzle_performance}
\end{figure}

\paragraph{Puzzle-solving capabilities} 
Figure~\ref{fig:puzzle_performance} shows clear improvement in puzzle-solving ability across network layers within each Elo range, and consistent decrease in performance across increasing difficulty levels for all layers. Dashed lines and shading mark phase boundaries with annotated slope ratios indicating relative improvement rates between phases. While the early-to-middle phase distinction is less pronounced than in tournaments, the final-phase acceleration is clearly visible, particularly for harder puzzles where improvement rates reach $24.8$ times the middle phase.

\newpage

\paragraph{Distributional analysis} Kendall's $\tau$ correlation between intermediate and final move rankings stays near zero through middle layers and rises sharply only in the final layers (Appendix~\ref{app:policy_dynamics}). Policy entropy remains approximately constant across layers, indicating that intermediate predictions reflect meaningful preference distributions. Constant entropy with changing rankings implies that the model reorganizes move preferences across layers rather than gradually sharpening them.

\paragraph{Solution discovery and forgetting}

\begin{wrapfigure}{r}{0.5\textwidth} 
    \centering
    \vspace{-18pt}
    \includegraphics{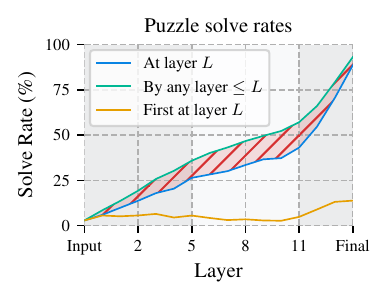}

    \vspace{-5pt}
    \caption{Layer-wise puzzle-solving performance across network depth. The hatched region indicates puzzles solved by at least one previous layer but not by the current layer—forgotten solutions. }
    \label{fig:solve_dynamics}
    \vspace{-20pt}
\end{wrapfigure}

Figure~\ref{fig:solve_dynamics} tracks three metrics across layers: the \atlayer{current} solve rate at each layer, the \byanylayer{cumulative} solve rate across all layers up to the current one, and the rate of puzzles solved for the \firstatlayer{first time} at each layer. The gap between \atlayer{current} and \byanylayer{cumulative} rates—highlighted by the \forgotten{hatched} region—shows that solutions are frequently discovered and subsequently discarded. The \firstatlayer{discovery} rate follows the three-phase pattern from playing strength, with reduced discoveries during the middle-layer plateau and renewed acceleration in late layers where many puzzles are solved for the first time. The gap between the final \byanylayer{cumulative} and \atlayer{current} solve rates represents \forgotten{forgotten} puzzles:  solutions discovered at intermediate layers but lost in the output.

\subsection{Forgotten puzzles and the limits of learned look-ahead}
\label{sec:limits-look-ahead}

Having established that our logit lens produces meaningful intermediate policies, we turn to forgotten puzzles—and find a systematic pattern rather than random failures. In the general dataset, $57\%$ of forgotten puzzles involve a sacrifice, compared to $26\%$ of solved ones. Within the forgotten dataset, $87.3\%$ of puzzles involve a sacrifice\footnote{A puzzle involves a sacrifice if the player's material balance drops below its starting value along the principal variation.} and the correct moves are overwhelmingly forcing ($76.9\%$ checks, $77.9\%$ captures), yet the model replaces them with quiet moves in $68.3\%$ of cases. The model's own value head assigns higher win probability to the position after the correct move than to the position after the model's chosen move in $97.0\%$ of forgotten puzzles (Appendix~\ref{app:value_head}). Notably, $45.6\%$ are puzzles whose principal variation ends in checkmate, about half of them mate-in-two positions where even a single step of reliable look-ahead should guarantee playing the correct move.

\begin{figure}[t]
\centering
\begin{subfigure}[b]{\textwidth}
    \centering
    \includegraphics[width=\textwidth]{Figures/forgotten_act_patching.pdf}
    \vspace{-20pt}
    \caption{Activation patching}
    \label{fig:patching_forgotten}
\end{subfigure}

\begin{subfigure}[b]{0.58\textwidth}
    \centering
    \includegraphics[height=4.5cm]{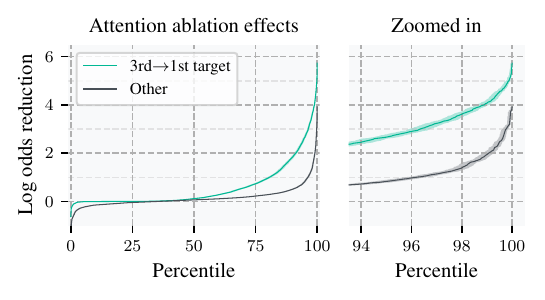}
    \vspace{-15pt}
    \caption{Attention head ablation}
    \label{fig:ablation_forgotten}
\end{subfigure}%
\hfill
\begin{subfigure}[b]{0.42\textwidth}
    \centering
    \includegraphics[height=4.5cm]{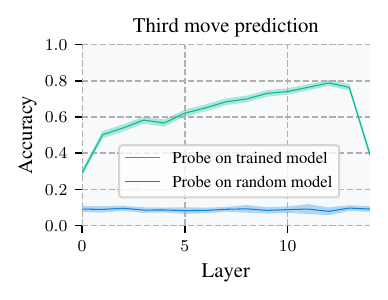}
    \vspace{-15pt}
    \caption{Probing}
    \label{fig:probing_forgotten}
\end{subfigure}

\caption{Look-ahead analysis from \citet{jenner2024evidencelearnedlookaheadchessplaying} on forgotten puzzles. \textbf{(a)} Mean activation patching effects in the residual stream across layers. The \firsttarget{\first{} move target} and \thirdtarget{\third{} move target} squares show outsized effects compared to the \othersquare{other} squares line. The \corrupted{corrupted} square(s) effect diminishes in late layers. Error bars are two times the standard error of the mean. \textbf{(b)} Zero-ablation of the attention entry from the third to the first move target square in L12H12. The \firsttarget{3rd$\rightarrow$1st target} line shows ablation of this single entry, while the \othersquare{other} line shows ablation of all remaining $4{,}095$ entries simultaneously. Lines show the effect at a given percentile of puzzles sorted by effect size. Error bars are $95\%$ CIs. \textbf{(c)} Bilinear probe accuracy for predicting the third move across layers on a \firsttarget{trained} and \thirdtarget{randomly} initialized model. Error bars combine variance across five probe training runs with finite-sample uncertainty in accuracy estimates.} 
\label{fig:look_ahead_forgotten}
\vspace{-5pt}
\end{figure}

\paragraph{Look-ahead analysis}
Could these puzzles be solved purely by shallow heuristics or chance in intermediate layers, without any contribution from look-ahead? The localization of the override suggests otherwise: $63.4\%$ of forgotten puzzles remain solved after the final attention sublayer, indicating that the override is concentrated in the final feed-forward layer. To directly test whether look-ahead is present and causally contributes to these solutions, we apply the methodology described in Section~\ref{sec:look_ahead} to the forgotten dataset. Figure~\ref{fig:look_ahead_forgotten} shows that all three lines of evidence yield qualitatively similar patterns to those reported by \citet{jenner2024evidencelearnedlookaheadchessplaying} on solved puzzles, albeit with somewhat smaller effect sizes for the causal interventions, as forgotten puzzles have lower baseline probability on the correct move and thus less log-odds to lose. Activation patching reveals that the \thirdtarget{\third{} move target} square exhibits outsized causal effects in middle to late layers, diminishing after layer $11$ as look-ahead information is transferred to the \firsttarget{\first{} move target} square, the same pattern as in solved puzzles. Zero-ablating the single attention entry in L12H12 that relays information from the \thirdtarget{\third{}} to the \firsttarget{\first{} move target} square reduces the log-odds of the correct move by more than $1.5$ in more than $10\%$ of forgotten puzzles, and in $61.6\%$ of cases this single entry has a larger effect than ablating all remaining $4{,}095$ entries simultaneously. For the example in Figure~\ref{fig:forgotten_puzzle}, ablating this single scalar drops the correct move's probability from $5.8\%$ to $1.0\%$, corresponding to a log-odds reduction of $1.79$. Among forgotten puzzles in which patching shows look-ahead to be present and the correct move is on top at the penultimate layer, cutting this entry alone removes the correct move from the top in $29.9\%$ of cases. Bilinear probes decode the third move with up to $79\%$ accuracy for forgotten puzzles, compared to $90\%$ for solved puzzles, though probes on randomly initialized models also show lower accuracy on forgotten puzzles. Results for additional look-ahead heads identified by \citet{cruz2025understanding} and the corresponding findings on solved puzzles are in Appendix~\ref{app:look_ahead}.

Taken together, the evidence demonstrates that the look-ahead mechanism operates normally in forgotten puzzles—future moves are represented, causally important, and linearly decodable. The model computes the correct tactical solution but does not use it. The presence of algorithmic structure does not guarantee algorithmic behavior. If the failure is not in look-ahead, what overrides it? 

\subsection{Safety priors override tactical solutions} The forgotten puzzle characterization revealed that the model systematically replaces forcing tactical solutions with safe, positional alternatives. To quantify whether this pattern reflects learned safety priors, we analyze how chess concepts prioritized by layer-wise policies evolve across network depth. Since the model's internal representation of safety is unknown, we rely on handcrafted evaluation terms as human-interpretable proxies, focusing on concepts with a direct relationship to safe play.

\paragraph{Layer-wise concept preferences}
To examine which chess concepts each layer prioritizes, we analyze how layer-wise policies weight moves by their conceptual effects. Following \citet{McGrath_2022}, who trained linear concept probes on AlphaZero's representations, we use Stockfish~8's \citep{stockfish8} handcrafted evaluation terms as human-interpretable concepts, a choice made for comparability with \citet{sadler2019gamechanger}. Each concept maps a position to a scalar value that measures an interpretable property, such as king safety or threats. Rather than probing for concepts, we measure \emph{concept preference} directly from layer-wise move probabilities.

For each move $m$ from position $s$ to resulting position $s'$, and each concept $c$, we compute $\Delta c_m = c(s') - c(s)$, representing the change in $c$ caused by $m$. All evaluations are from the perspective of the current player, so positive values indicate an improvement. At each layer $\ell$, we compute:

\begin{equation}\label{eq:concept_preference}\Delta c_\ell = \sum_{m \in \text{legal}(s)} \pi_\ell(m) \cdot \Delta c_m = \mathbb{E}_{\pi_\ell}[\Delta c_m]\end{equation} 

where $\pi_\ell(m)$ is the move probability assigned by layer $\ell$'s policy. This represents the expected concept change when sampling moves according to $\pi_\ell$. Full details and plots are provided in Appendix~\ref{app:concept_analysis}.

\begin{figure}[h]
    \centering
    \begin{subfigure}[b]{0.25\textwidth}
        \centering
        \includegraphics[width=\textwidth]{Figures/concept_legend.pdf}
        \vspace{15pt}
    \end{subfigure}%
    \hfill
    \begin{subfigure}[b]{0.3708\textwidth}
        \centering
        \includegraphics[width=\textwidth]{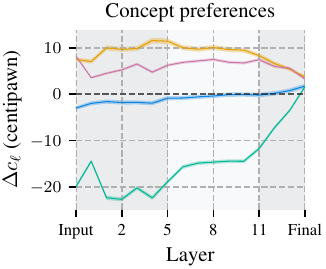}
    \end{subfigure}%
    \begin{subfigure}[b]{0.35\textwidth}
        \centering
        \includegraphics[width=\textwidth]{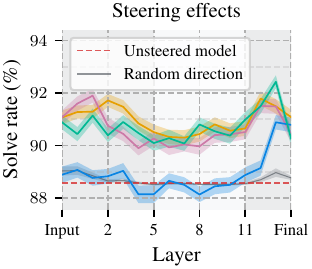}
    \end{subfigure}
    \caption{Safety-related concept analysis across layers. \textbf{Left:} Mean concept preference $\Delta c_\ell$ for four safety-related concepts, measured in centipawns with $95\%$ CI. Late layers shift from aggressive toward conservative concept preferences.    \textbf{Right:} Held-out solve rate when steering with each concept's gradient, with the strength $\alpha \in \{\pm 0.25, \pm 0.5, \ldots, \pm 1.5\}$ selected per layer on half of the puzzles and evaluated on the other half. Lines show the mean and shaded areas one standard deviation across $200$ random splits. Steering against the late-layer safety preferences substantially improves solve rates, peaking at layer $13$. The \randomdirection{gray line} shows the mean of $50$ random unit vectors under the same protocol, with $95\%$ CI across vectors, while the \baselineunsteered{dashed line} indicates the unsteered baseline.}
\label{fig:concept-steering}
\end{figure}

The left panel of Figure~\ref{fig:concept-steering} shows the evolution of four safety-related concepts across layers. Early and middle layers favor aggressive over defensive concepts, with higher $\Delta c_\ell$ for \kingsafetyopp{opponent king vulnerability} and \threatsmine{own threats}, while later layers increasingly prioritize safe play, increasing \kingsafetymine{own king safety} and reducing \threatsopp{opponent threats}, with all four concepts converging to similar values.

\paragraph{Causal validation via concept steering} As a proof of concept for the hypothesis that safety priors contribute to the forgotten puzzle phenomenon, we perform activation steering \citep{dathathri2020plugplaylanguagemodels, subramanisubramani2022extractinglatentsteeringvectorsetal-2022-extracting} using a first-order approximation in activation space with respect to the concept preference metric. For a given concept $c$ at layer $\ell$ with steering strength $\alpha$, we modify the residual stream activations $\mathbf{h}_\ell$ as: \begin{equation}\label{eq:steering}
\mathbf{h}_\ell' = \mathbf{h}_\ell \pm \alpha \cdot \mathbb{E}[\nabla_{\mathbf{h}_\ell} \Delta c_L] \cdot \frac{\mathbb{E}[\|\mathbf{h}_\ell\|]}{\|\mathbb{E}[\nabla_{\mathbf{h}_\ell} \Delta c_L]\|}\end{equation}
where $\Delta c_L$ is the concept preference of the full model and the gradient is normalized and scaled to the layer's mean activation norm. Gradients are computed on positions sampled from general play, independent of any puzzle dataset. We sweep over steering strengths $\alpha$ and intervention layers $\ell$ for each concept and measure the solve rate on the general puzzle dataset, capturing both recovery of forgotten puzzles and regressions. Using Monte Carlo cross-validation \citep{MCCV}, we select the best strength for each concept and layer on a random half of the puzzles, evaluate it on the other and report the mean and standard deviation over $200$ random splits.

The right panel of Figure~\ref{fig:concept-steering} shows steering effects across layers. They are largest when steering the model to place less emphasis on mitigating \threatsopp{opponent threats} at layer $13$, which reaches a solve rate of $92.4 \pm 0.2\%$ compared to the $88.6\%$ \baselineunsteered{baseline}. With $\alpha = -0.5$, the best strength on all puzzles, this setting solves $519$ additional puzzles with $129$ regressions. Notably, it recovers $62.4\%$ of the forgotten puzzles in the general dataset while also solving $34.4\%$ of puzzles the model never solved at any layer. Applied to the held-out forgotten dataset, the same configuration recovers $71.7\%$ of its puzzles. The optimal steering direction aligns with the concept preference analysis: for all four concepts, performance improves when placing less emphasis on safety, specifically reducing the weight on \kingsafetymine{own king safety} and \threatsopp{opponent threats against us} while increasing emphasis on \threatsmine{threats we pose} and \kingsafetyopp{opponent king vulnerability}. For three of four concepts, steering yields improvements across all layers, substantially exceeding the \randomdirection{random baseline}, with the strongest single effect at layer $13$, just before the override. 

The observed regressions suggest that not all puzzles benefit from reduced safety preferences. Nevertheless, puzzles are adversarial by construction—the winning move is frequently a sacrifice or forcing sequence that conflicts with conservative heuristics learned from the training distribution, explaining why reducing safety preferences yields a substantial net improvement on puzzles. That a simple gradient-based intervention targeting safety-related concepts suffices to recover a majority of forgotten puzzles supports the hypothesis that learned safety priors are a primary cause of the forgotten puzzle phenomenon. While other concepts also improve performance above the \randomdirection{random baseline}, the most directly safety-related concepts yield the largest and most consistent effects; complete steering analyses for all concepts are provided in Appendix~\ref{app:steering}.

\subsection{Search and the training objective}
\label{sec:search-experiments}

Leela’s policy is trained to match the outcome of search rather than to solve puzzles, so forgotten puzzles could be the product of a training objective that is misspecified on these positions rather than a genuine failure of the network. To test this, we run the self-play search of the training procedure on each puzzle with the original settings and compare the resulting training target with the policy predicted by the network. Additionally, we examine how forgotten puzzles affect the full engine in deployment, with details of both experiments in Appendix~\ref{app:search_experiments}. 

\begin{wrapfigure}{l}{0.5\textwidth}
    \centering
    \vspace{-13pt}
    \includegraphics{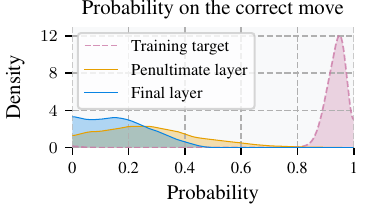}

    \vspace{-10pt}
    \includegraphics{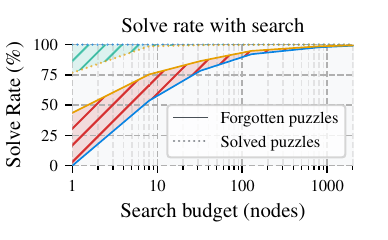}

    \vspace{-1pt}
    \caption{Training target and search. \textbf{Top:} Probability assigned to the correct move on forgotten puzzles by the \trainingtarget{training target}, the \penultimatelayer{penultimate layer} and the \finallayer{final layer}. \textbf{Bottom:} Engine solve rate by search budget, with the policy read from the \finallayer{final layer} or the \penultimatelayer{penultimate layer}. Shading shows the net effect of the final layer, \forgotten{worse} on forgotten and \mbox{\solvedgain{better}} on solved puzzles. With a single node, the engine plays the policy's top move.}
    \label{fig:search}
    \vspace{-30pt}
\end{wrapfigure}

\paragraph{Training target} The top panel of Figure~\ref{fig:search} shows that on forgotten puzzles the \trainingtarget{training target} places a mean of $93.1\%$ of its visits on the correct move, and its most visited move is the correct one in $99.4\%$ of puzzles. The correct move also has the highest value at the root in $99.5\%$ of puzzles, so more search would concentrate the target further. Under the policy's own training loss, the \penultimatelayer{penultimate layer} is closer to the target than the \finallayer{final layer} in $77.5\%$ of forgotten puzzles. These positions most likely never occurred in training, but had they occurred, the training objective would have asked for the solution. The override therefore does not come from the objective itself, but is consistent with a prior that is sensible on the training distribution and carries over to positions where it conflicts with the solution.

\paragraph{Search in deployment} We run Leela with default settings at different search budgets and count a puzzle as solved if the engine plays the correct move. The bottom panel of Figure~\ref{fig:search} shows that, given enough search, the engine recovers forgotten puzzles, but the \finallayer{final layer} makes this \forgotten{more expensive} than a policy read from the \penultimatelayer{penultimate layer}. On solved puzzles, where the final layer's top move is already correct, it \solvedgain{helps} instead.

\section{Related work}

\paragraph{Chess models and their interpretability} The success of AlphaZero \citep{silver2018general} and Leela Chess Zero \citep{leela} has motivated a growing body of interpretability work on neural chess engines, including concept probing \citep{McGrath_2022}, concept discovery \citep{schut2023bridginghumanaiknowledgegap}, and evidence of learned look-ahead \citep{jenner2024evidencelearnedlookaheadchessplaying, cruz2025understanding}. Separately, \citet{karvonen2024emergentworldmodelslatent} studied world models in language models trained on chess move sequences, and \citet{ruoss2024amortizedplanninglargescaletransformers} trained a transformer on Stockfish’s evaluations to play chess without external search. These models differ in training paradigm but are architecturally similar, suggesting the tension between learned algorithms and distributional priors we identify may extend to them as well.

\paragraph{Intermediate decoding} 
\citet{nostalgebraist2020logitlens} introduced the logit lens for decoder-only language models, with subsequent work extending it to encoder-decoder architectures \citep{langedijk2024decoderlenslayerwiseinterpretationencoderdecoder}, to anticipating future tokens \citep{Pal_2023} and aiming to improve prediction accuracy \citep{belrose2025elicitinglatentpredictionstransformers, din2024jumpconclusionsshortcuttingtransformers}, though with objectives that penalize the intermediate-to-final disagreements we study. The logit lens has been applied to Leela Zero, a superhuman Go model, though limited to divergence metrics \citep{du2023insidemindsuperhumango}, and used to identify distinct computational phases in language models \citep{lad2025remarkablerobustnessllmsstages}, a pattern we also observe in Leela. 

\paragraph{Learned algorithms in neural networks} Beyond language models, where circuit discovery has become a major research direction, recent work has mechanistically identified learned algorithms in other domains including search in synthetic tasks \citep{brinkmann2024mechanistic} and planning in Sokoban \citep{bush2025interpreting, taufeeque2026path}. These works demonstrate that identified circuits are causally relevant for the model's output. In contrast, \citet{nikankin2025arithmetic} find that neural networks can solve seemingly algorithmic tasks like arithmetic through ensembles of heuristics rather than generalizable algorithms. Our work sits between these findings: algorithmically computed solutions can be present and causally active yet still be overridden in the final output by competing heuristics.

\section{Conclusion}
We have shown that the presence of learned algorithmic structure in a neural network does not guarantee that the network's output will reflect it. In Leela Chess Zero, look-ahead computes correct tactical solutions that are systematically overridden by safety priors in the final layers, a failure of the network rather than of its training objective. Concept preference analysis shows that late layers shift toward safe play, and gradient-based steering against these priors recovers a majority of forgotten puzzles, providing causal evidence for this override. More broadly, our results suggest that circuit discovery alone is insufficient for predicting model behavior: identified algorithms may not determine the output, and incorrect outputs may not indicate absent capabilities. Mechanistic analysis can expose such failures and point to interventions that correct them.

\paragraph{Limitations} The main limitations of our work are: (1) While our results suggest safety priors as a primary cause of forgotten puzzles, we do not provide a precise mechanistic description of how these priors are implemented. (2) We verify the forgotten puzzle phenomenon across multiple Leela networks (Appendix~\ref{app:multi-model-logit-lens}), but our causal analyses are conducted on a single network. (3) Our concept analysis uses Stockfish's evaluation terms as proxies and is restricted to one-step position evaluations; the model likely encodes features that do not map cleanly onto these concepts.

\paragraph{Impact} There has been growing confidence that identifying algorithmic circuits in neural networks yields understanding of model behavior. Our results introduce a caveat: correctly identified algorithms may be overridden by competing priors, meaning that behavioral predictions based on circuit analysis alone may be incomplete. While our analysis focuses on a chess-playing model, the tension between learned algorithms and priors likely exists in any model trained on broad distributions, with or without search (Appendix~\ref{app:limitations_directions}). Indeed, related phenomena have been observed in language models, where correct answers decodable at intermediate layers are overridden in late layers by false demonstrations \citep{halawi2024overthinking} or sycophancy priors \citep{wang2026truth}. Search is also not unique to Leela: in reasoning models, the language model acts as a policy guiding a search over the space of possible chains of thought \citep{chollet2024o3}, and reinforcement learning with verifiable rewards distills successful chains back into the model \citep{lambert2025tulu3, guo2025deepseekr1}. Beyond interpretability, such overrides connect to concerns about hidden capabilities in deployed models—capabilities present in internal computations but not reflected in behavior \citep{gekhman2025insideout}, which may surface under distribution shift or adversarial inputs \citep{wei2023jailbroken}. Whether similar overrides affect algorithmic solutions in domains such as mathematical reasoning remains an open question. 

\section*{Author contributions}

Elias conceived, implemented, and conducted the study, performed all experiments, and wrote the manuscript. Sebastian and Wojciech provided supervision and feedback throughout the project.

\section*{Acknowledgments and Disclosure of Funding}
We gratefully acknowledge  \citet{jenner2024evidencelearnedlookaheadchessplaying} for the codebase and inference framework on which this work is built, and \citet{ruoss2024amortizedplanninglargescaletransformers} for the puzzle evaluation and tournament setup. Without their open-source releases, this research would not have been possible. We also thank the \citet{leela} and the \citet{stockfish} for making their engines and networks openly available, and \citet{lichess2025} for its open puzzle database and bot platform. Our gratitude also goes to Ekkehard Schnoor for pointing us to chess as a research direction, and to Marie and Enno Sandmann for always distracting us at the right moments.

This work was supported by the Federal Ministry of Research, Technology and Space (BMFTR) as grants [BIFOLD (01IS18025A, 01IS180371I), xJuRAG (16IS25015B)]; the European Union’s Horizon Europe research and innovation programme (EU Horizon Europe) as grant ACHILLES (101189689); and the German Research Foundation (DFG) as research unit DeSBi [KI-FOR 5363] (459422098). The authors declare no competing interests.

\bibliography{bibliography}
\bibliographystyle{plainnat}

\clearpage
\appendix
\etocdepthtag.toc{mtappendix}
\etocsettagdepth{mtmain}{none}
\etocsettagdepth{mtappendix}{subsection}
\renewcommand{\contentsname}{Appendix Contents}
\tableofcontents

\section{Limitations and future directions}
\label{app:limitations_directions}

\paragraph{Scope of analysis}
 Our causal analyses focus on a single chess model with a Post-LN architecture, though we verify that the forgotten puzzle phenomenon itself replicates across other Leela networks (Appendix~\ref{app:multi-model-logit-lens}). The findings function as an existence proof: algorithmically computed solutions can be overridden by competing priors in a deployed neural network. However, this single-domain analysis cannot disentangle whether the override reflects domain-specific characteristics of chess, architecture-specific properties of Post-LN transformers, or more general principles of neural network computation. Comparison across architectures and domains, particularly Pre-LN language models, would help isolate these effects. No Pre-LN Leela models exist for direct architectural comparison; the development team has chosen Post-LN over Pre-LN in their training paradigm \citep{dje2025postnorm}.

\paragraph{Models without search}
Can the override arise in models that are deployed without search? Its ingredients do not require search. Essentially every model trained today minimizes an expected loss over a data distribution with parameters shared across inputs, and so learns priors that are right in expectation. The override arises where an internal computation conflicts with such a prior and the two are reconciled late in the network. Related overrides have been reported in language models that use no search \citep{halawi2024overthinking, wang2026truth}. In Leela, the search that wraps the network at deployment corrects much of the override (Section~\ref{sec:search-experiments}); a model deployed without such a search has no correction, so the override reaches its output directly and may be more visible, not less.

\paragraph{Reasoning models and search}
Reasoning models do use a form of search, at least during training. There, the model explores the space of possible chains of thought \citep{chollet2024o3}, and its training resembles Leela's. In AlphaZero and Leela, each node of the search tree is a position. For every move in a self-play game, MCTS builds a tree from the current position, and the policy at that position is trained toward the visit counts at the root. Reinforcement learning with verifiable rewards \citep{lambert2025tulu3, guo2025deepseekr1} applies a similar idea with tokens as nodes. For each prompt, the model samples several complete chains of thought, each a single path through the tree of possible continuations, and checks the final answer. Tokens in chains that reach a correct answer are made more likely and tokens in chains that fail less likely. In both cases, the outcome of the search is distilled back into the network that guides the search as its prior, so that the network itself becomes more likely to produce what the search found. Training can therefore only learn from what the prior lets the search find. Every continuation lies in the support of a softmax policy, but the prior determines which are reached within a sampling budget in both Leela and reasoning models.

This view also gives one reading of why reinforcement learning on language models took as long as it did to work. If no sampled chain reaches the correct answer, the reward is uniform and there is nothing to upweight. The prior has to succeed occasionally before the process can bootstrap itself, which required pretrained language models strong enough to guide the search through the space of chains of thought. In Leela, the search overcomes the prior on forgotten puzzles, because MCTS backs up value estimates through the tree. Sampled chains have no such correction, so a correct chain is found only as often as the prior allows. A prior like Leela's safety prior would therefore rarely be corrected in a reasoning model, because training rarely sees a reward on the problems it gets wrong.

\paragraph{Generalization and applications}
While our analysis focuses on chess transformers, our methodology could inform interpretability research in other domains. First, concept preference analysis offers an alternative to representation probing: rather than identifying whether concepts are encoded, it tracks which concepts each layer prioritizes in driving outputs. Applied to language models, this could trace behavioral tendencies such as truthfulness or specificity across layers. Second, our observation that safety priors may override algorithmic solutions suggests that in tasks with verifiable ground truth, suppressing learned distributional biases—such as final-layer tendencies toward common tokens—might improve performance on tasks requiring precise reasoning.

\section{Compute resources and licensing}
\label{app:compute}

\paragraph{Compute resources}
All experiments were conducted on a single workstation with 2$\times$ NVIDIA RTX 5090 (32 GB) and an AMD Threadripper PRO 7965WX (24 cores). Most individual experiments complete within a few hours on a single GPU. The most compute-intensive components are the internal tournaments ($\sim$1--3 days due to sequential game play), concept delta computation via Stockfish ($\sim$16 hours parallelized across CPU cores), the full steering sweep across all concepts ($\sim$2~GPU-days), and the search experiments ($\sim$half a day). All main paper experiments can be reproduced in approximately 2--3 days on comparable hardware.

\paragraph{Licenses and data availability}
The Lichess puzzle database is available under a Creative Commons CC0 license \citep{lichess2025}. The CCRL dataset is publicly available without a stated license \citep{ccrl_dataset}. The puzzle dataset, Encyclopedia of Chess Openings positions, and tournament code from \citet{ruoss2024amortizedplanninglargescaletransformers} are released under the Apache 2.0 license, with the Encyclopedia of Chess Openings originally compiled by \citet{matanovic1978encyclopaedia}. The inference framework from \citet{jenner2024evidencelearnedlookaheadchessplaying}, Leela Chess Zero \citep{leela}, and Stockfish 8 and 17.1 \citep{stockfish} are all released under the GPLv3 license. The Lichess bots run on the lichess-bot framework \citep{lichessbot2023github}, released under the AGPLv3 license.

\section{Post-LN logit lens extension}
\label{app:post_ln_logit_lens}

The logit lens technique projects intermediate layer representations through the model's output head to examine what the network would predict at different depths. This approach relies on the assumption that representations across layers exist in a shared basis that allows meaningful projection to output space—an assumption enabled by residual connections that create additive pathways for information flow \citep{jastrzębski2018residualconnectionsencourageiterative}. In transformer architectures, architectural inductive biases further encourage information about specific tokens to remain localized to their corresponding positions throughout the network, creating a privileged basis that facilitates cross-layer interpretation \citep{jenner2024evidencelearnedlookaheadchessplaying}.

In Pre-LN transformers, this involves taking intermediate representations, applying the final layer normalization, and projecting through the unembedding matrix. This approach works because layer normalization precedes each sublayer in Pre-LN models, leaving the residual stream unchanged and maintaining representational consistency across layers.

Post-LN architectures complicate this process by applying layer normalization after each sublayer's output is added to the residual stream. Unlike Pre-LN models where only a final normalization is needed, Post-LN models have sequential normalization operations that directly transform the residual stream at each layer. These intermediate normalizations create dependencies between layers that prevent simply taking an intermediate representation and applying only the final layer normalization and output projection, as the intermediate representation has not undergone the normalization transformations it would experience in a complete forward pass.

Our goal is to develop an extension that maps intermediate layer representations to the representational basis expected by the policy head, accounting for the normalization transformations unique to Post-LN architectures.

\subsection{Pre- vs Post-LN architectures and DeepNorm}

\label{app:norm_differences}

The key difference between Pre-LN and Post-LN architectures lies in when layer normalization is applied relative to the residual connections. This placement affects how representations evolve through the network and impacts the applicability of interpretability techniques.

\paragraph{Pre-LN}
In Pre-LN transformers, layer normalization is applied before each sublayer (attention and feed-forward), with the computation following the pattern:
\begin{align}
\mathbf{h}_{\ell}' &= \mathbf{h}_{\ell-1} + \mathrm{MHA}_\ell(\mathrm{LayerNorm}(\mathbf{h}_{\ell-1})) \\
\mathbf{h}_{\ell} &= \mathbf{h}_{\ell}' + \mathrm{FFN}_\ell(\mathrm{LayerNorm}(\mathbf{h}_{\ell}'))
\end{align}
where $\mathbf{h}_{\ell}$ denotes the hidden representation at layer $\ell$, $\mathrm{MHA}_\ell$ is the multi-head attention operation, and $\mathrm{FFN}_\ell$ is the feed-forward network at layer $\ell$. This design ensures that the residual stream itself is never directly transformed by normalization operations, allowing intermediate representations to be projected through the final layer normalization and output head in a straightforward manner.

\paragraph{Post-LN}
Post-LN models apply layer normalization after adding sublayer outputs to the residual stream:
\begin{align}
\mathbf{h}_{\ell}' &= \mathrm{LayerNorm}(\mathbf{h}_{\ell-1} + \mathrm{MHA}_\ell(\mathbf{h}_{\ell-1})) \\
\mathbf{h}_{\ell} &= \mathrm{LayerNorm}(\mathbf{h}_{\ell}' + \mathrm{FFN}_\ell(\mathbf{h}_{\ell}'))
\end{align}
Each layer normalization operation directly transforms the accumulated representation, creating a sequence of transformations that intermediate representations must undergo to reach the final output space.

\paragraph{DeepNorm}
Leela employs DeepNorm \citep{deepnet} scaling to stabilize Post-LN training, modifying the computation to include residual scaling factors:
\begin{align}
\mathbf{h}_{\ell}' &= \mathrm{LayerNorm}(\alpha \cdot \mathbf{h}_{\ell-1} + \mathrm{MHA}_\ell(\mathbf{h}_{\ell-1})) \\
\mathbf{h}_{\ell} &= \mathrm{LayerNorm}(\alpha \cdot \mathbf{h}_{\ell}' + \mathrm{FFN}_\ell(\mathbf{h}_{\ell}'))
\end{align}
where $\alpha = (2N)^{1/4} \approx 2.34$ for $N=15$ layers. This scaling, combined with specialized weight initialization, enables stable training while preserving Post-LN's representational advantages.

\subsection{Zero ablation methodology for Post-LN architectures}

The standard logit lens can be understood as performing zero ablation—setting all sublayer outputs to zero for layers beyond $\ell$, then applying the final layer normalization and projection to output space. We extend this principle to Post-LN models by applying the same zero ablation of sublayer outputs while preserving the normalization operations and $\alpha$ that would transform these representations.

Specifically, for examining layer $k$, we set $\mathrm{MHA}_\ell(\cdot) = 0$ and $\mathrm{FFN}_\ell(\cdot) = 0$ for all $\ell > k$ during a forward pass, while preserving the $\alpha$ and layer normalization scaling parameters ($\gamma$). We set layer normalization biases ($\beta$) to zero for layers beyond $k$ to maintain better consistency with the standard logit lens paradigm. The rationale for these choices will be explained in Appendix~\ref{sec:implications_logit_lens}, following the decomposition analysis.

This approach ensures that representations from layers 1 through $k$ undergo the same sequence of transformations they would experience in the complete network, while removing contributions from later layers. Normalization statistics ($\mu$ and $\sigma$) are recomputed during this modified forward pass to reflect the actual distribution of the truncated representations, rather than using statistics computed on the full model output. This follows the same principle as Pre-LN logit lens implementations, which directly apply the final layer normalization to the intermediate activations being analyzed.

\subsection{Transformer encoder decomposition}

To provide intuition for why our proposed logit lens extension makes principled choices for Post-LN architectures, we present a decomposition of the transformer encoder output and reinterpret the logit lens in terms of this decomposition. Following \citet{mickus-etal-2022-dissect}, the final representation of the encoder at layer $L$ and token position $t$ can be decomposed into:

\begin{align}
    \mathbf{h}_{L,t} &= \mathbf{i}_{L,t} + {\mathbf{z}}^{\mathrm{MHA}}_{L,t} + {\mathbf{z}}^{\mathrm{FFN}}_{L,t} + \mathbf{b}_{L,t} - \mathbf{m}_{L,t} \\
    \mathbf{i}_{L,t} &= \alpha^{2L} \cdot \frac{\bigodot_{\ell=1}^L \gamma_\ell^{\mathrm{MHA}} \odot \gamma_\ell^{\mathrm{FFN}}}{\prod_{\ell=1}^L\sigma_{\ell,t}^{\mathrm{MHA}} \sigma_{\ell,t}^{\mathrm{FFN}}} \odot \mathbf{h}_{0,t} \\ {\mathbf{z}}^{\mathrm{MHA}}_{L,t} &= \sum_{\ell=1}^L \alpha^{2L-2\ell+1} \cdot \frac{\bigodot_{\ell'=\ell}^L \gamma_{\ell'}^{\mathrm{MHA}} \odot \gamma_{\ell'}^{\mathrm{FFN}}}{\prod_{\ell'=\ell}^L\sigma_{\ell',t}^{\mathrm{MHA}} \sigma_{\ell',t}^{\mathrm{FFN}}} \odot \tilde{\mathbf{z}}_{\ell,t}^{\mathrm{MHA}} \\
    {\mathbf{z}}^{\mathrm{FFN}}_{L,t} &= \sum_{\ell=1}^L \alpha^{2L-2\ell} \cdot \frac{\gamma_\ell^{\mathrm{FFN}} \bigodot_{\ell'=\ell+1}^L \gamma_{\ell'}^{\mathrm{MHA}} \odot \gamma_{\ell'}^{\mathrm{FFN}}}{\sigma_{\ell,t}^{\mathrm{FFN}}\prod_{\ell'=\ell+1}^L\sigma_{\ell',t}^{\mathrm{MHA}} \sigma_{\ell',t}^{\mathrm{FFN}}} \odot \tilde{\mathbf{z}}_{\ell,t}^{\mathrm{FFN}} \\
    \mathbf{b}_{L,t} &= \sum_{\ell=1}^L \alpha^{2L-2\ell+1} \cdot \frac{\bigodot_{\ell'=\ell}^L \gamma_{\ell'}^{\mathrm{MHA}} \odot \gamma_{\ell'}^{\mathrm{FFN}}}{\prod_{\ell'=\ell}^L\sigma_{\ell',t}^{\mathrm{MHA}} \sigma_{\ell',t}^{\mathrm{FFN}}} \odot  b_{\ell}^{\mathrm{MHA},V} W_\ell^{\mathrm{MHA},O} \\
    &+ \sum_{\ell=1}^L \alpha^{2L-2\ell+1} \cdot \frac{\bigodot_{\ell'=\ell}^L \gamma_{\ell'}^{\mathrm{MHA}} \odot \gamma_{\ell'}^{\mathrm{FFN}}}{\prod_{\ell'=\ell}^L\sigma_{\ell',t}^{\mathrm{MHA}} \sigma_{\ell',t}^{\mathrm{FFN}}} \odot  b_\ell^{\mathrm{MHA},O} \notag\\
    &+ \sum_{\ell=1}^L \alpha^{2L-2\ell} \cdot \frac{\gamma_\ell^{\mathrm{FFN}} \bigodot_{\ell'=\ell+1}^L \gamma_{\ell'}^{\mathrm{MHA}} \odot \gamma_{\ell'}^{\mathrm{FFN}}}{\sigma_{\ell,t}^{\mathrm{FFN}}\prod_{\ell'=\ell+1}^L\sigma_{\ell',t}^{\mathrm{MHA}} \sigma_{\ell',t}^{\mathrm{FFN}}} \odot  b_\ell^{\mathrm{FFN},O} \notag\\
    &+ \sum_{\ell=1}^L \alpha^{2L-2\ell+1} \cdot \frac{\gamma_\ell^{\mathrm{FFN}} \bigodot_{\ell'=\ell+1}^L \gamma_{\ell'}^{\mathrm{MHA}} \odot \gamma_{\ell'}^{\mathrm{FFN}}}{\sigma_{\ell,t}^{\mathrm{FFN}}\prod_{\ell'=\ell+1}^L\sigma_{\ell',t}^{\mathrm{MHA}} \sigma_{\ell',t}^{\mathrm{FFN}}} \odot  \beta_\ell^{\mathrm{MHA}} \notag\\
    &+ \sum_{\ell=1}^L \alpha^{2L-2\ell} \cdot \frac{ \bigodot_{\ell'=\ell+1}^L \gamma_{\ell'}^{\mathrm{MHA}} \odot \gamma_{\ell'}^{\mathrm{FFN}}}{\prod_{\ell'=\ell+1}^L\sigma_{\ell',t}^{\mathrm{MHA}} \sigma_{\ell',t}^{\mathrm{FFN}}} \odot \beta_\ell^{\mathrm{FFN}} \notag \\
    \mathbf{m}_{L,t} &=\sum_{\ell=1}^L \alpha^{2L-2\ell+1} \cdot \frac{\bigodot_{\ell'=\ell}^L \gamma_{\ell'}^{\mathrm{MHA}} \odot \gamma_{\ell'}^{\mathrm{FFN}}}{\prod_{\ell'=\ell}^L\sigma_{\ell',t}^{\mathrm{MHA}} \sigma_{\ell',t}^{\mathrm{FFN}}} \odot \mu_{\ell,t}^{\mathrm{MHA}} \mathbf{1} \\
    &+\sum_{\ell=1}^L \alpha^{2L-2\ell} \cdot \frac{\gamma_\ell^{\mathrm{FFN}} \bigodot_{\ell'=\ell+1}^L \gamma_{\ell'}^{\mathrm{MHA}} \odot \gamma_{\ell'}^{\mathrm{FFN}}}{\sigma_{\ell,t}^{\mathrm{FFN}}\prod_{\ell'=\ell+1}^L\sigma_{\ell',t}^{\mathrm{MHA}} \sigma_{\ell',t}^{\mathrm{FFN}}} \odot \mu_{\ell,t}^{\mathrm{FFN}} \mathbf{1} \notag
\end{align}

where $\mathbf{h}_{0,t}$ is the initial input embedding, $\tilde{\mathbf{z}}_{\ell,t}^{\mathrm{MHA}}$ and $\tilde{\mathbf{z}}_{\ell,t}^{\mathrm{FFN}}$ are the raw unbiased outputs from multi-head attention and feed-forward networks at layer $\ell$, and $\alpha = (2N)^{1/4}$ is the DeepNorm scaling factor. The bias terms include $b_{\ell}^{\mathrm{MHA},V}$ (value projection bias), $W_\ell^{\mathrm{MHA},O}$ (output projection matrix), $b_\ell^{\mathrm{MHA},O}$ (attention output bias), and $b_\ell^{\mathrm{FFN},O}$ (feed-forward output bias). Layer normalization parameters are $\gamma_\ell$ (learned scales), $\sigma_{\ell,t}$ (computed standard deviations), $\mu_{\ell,t}$ (computed means), and $\beta_\ell$ (learned biases), with superscripts indicating MHA or FFN sublayers. The notation $\bigodot$ denotes element-wise multiplication.

The decomposition separates the final representation into the transformed input embedding ($\mathbf{i}_{L,t}$), accumulated sublayer contributions (${\mathbf{z}}^{\mathrm{MHA}}_{L,t}$, ${\mathbf{z}}^{\mathrm{FFN}}_{L,t}$), bias terms ($\mathbf{b}_{L,t}$), and mean centering effects ($\mathbf{m}_{L,t}$). Each component is scaled by normalization parameters from subsequent layers.

\subsection{Implications for the logit lens}

\label{sec:implications_logit_lens}

The decomposition framework provides direct justification for our zero ablation approach. The Post-LN logit lens can be understood as truncating the summations in ${\mathbf{z}}^{\mathrm{MHA}}_{L,t}$, ${\mathbf{z}}^{\mathrm{FFN}}_{L,t}$, and $\mathbf{b}_{L,t}$ to include only layers $\ell \leq k$, while recomputing the normalization statistics ($\mu$ and $\sigma$) based on the modified representations. 

\paragraph{Preservation of scaling parameters}
The decomposition reveals why preserving the $\gamma$ and $\alpha$ scaling factors is essential: all terms in the truncated representation—including the input embedding $\mathbf{i}_{L,t}$ and retained sublayer contributions—are transformed by these parameters from subsequent layers. Removing these transformations would fundamentally alter how the truncated representation maps to the output space of the encoder, although under our zero-ablation configuration the $\alpha$ factor is absorbed by layer normalization and has no numerical effect.

\paragraph{Treatment of bias terms}
The bias terms $\beta$ from layer normalization appear in $\mathbf{b}_{L,t}$ alongside attention and feed-forward biases, indicating they are structurally equivalent. For consistency with the zero ablation principle of removing sublayer contributions beyond layer $k$, we ablate all bias terms including layer normalization biases. Empirically, preserving versus ablating these biases produces qualitatively similar results; for example, preserving the layer normalization biases changes the solve rate by at most $2.7$ percentage points at any layer and leaves the forgotten set nearly unchanged. We provide code to examine all configurations. Our approach uses the policy head unchanged, which contains its own bias terms that may contribute to biased outputs. Moreover, the decomposition only captures bias terms that can be linearly separated from the representation—biases that interact with activation functions cannot be decomposed into constant additive components.

\paragraph{Representational alignment versus learned priors}
The treatment of bias terms reflects broader questions about their computational role. Bias terms may serve as representational alignment mechanisms that bridge coordinate system differences between layers, or they may encode learned priors about the task domain. The Tuned Lens \citep{belrose2025elicitinglatentpredictionstransformers} learns affine transformations with bias terms to achieve representational alignment between intermediate and final layers in Pre-LN architectures, motivated by biased outputs observed with the standard logit lens. In practice, bias terms likely serve both functions simultaneously, and disentangling them remains an open problem.

\paragraph{Learned decoders}
The Tuned Lens and linear shortcuts \citep{din2024jumpconclusionsshortcuttingtransformers} are trained to reproduce the model's final layer, so their objective penalizes exactly the disagreement with the final output that defines forgotten puzzles. Our zero-ablation readout has no such target and adds no trained parameters.

\section{Model architecture and search}
\label{app:model_architecture}

We analyze the \texttt{T82-768x15x24h} transformer model from Leela Chess Zero, the network studied by \citet{jenner2024evidencelearnedlookaheadchessplaying}, using their inference framework for our experiments. This model has $15$ transformer layers, $768$-dimensional representations, $24$ attention heads, and approximately $95$ million parameters, and follows the transformer architecture developed by the Leela Chess Zero team \citep{monroe2026chessformer}. It processes chess positions through three stages: input encoding transforms board states into token representations (with each of the $64$ squares treated as a discrete token), a $15$-layer transformer encoder processes these representations, and task-specific output heads generate move probabilities, position evaluations, and game length predictions. We use the original model that incorporates historical board information for all experiments unless stated otherwise, as this represents the model deployed in practice rather than the variant fine-tuned for interpretability research by \citet{jenner2024evidencelearnedlookaheadchessplaying}. Besides the current position, the model's input encodes the seven preceding positions of the game. Since all puzzles in the Lichess database are taken from real games, this history is available. To obtain it, we download each puzzle's source game from Lichess \citep{lichess2025} and replay it up to the puzzle position. Positions sampled from CCRL \citep{ccrl_dataset} likewise come with their full games. The only inputs without a history of their own are the corrupted positions used for activation patching (Appendix~\ref{app:look_ahead}). While minor differences exist on individual positions, the fine-tuned version is generally trained to behave equivalently, and both exhibit the forgotten puzzle phenomenon (Appendix~\ref{app:multi-model-logit-lens}).

\subsection{Input encoding}

Leela's input encoding transforms chess positions through binary feature planes, chess-specific positional encodings, and learned projections into the model's embedding space.

\paragraph{Board representation}

The board state is encoded using $112$ binary feature planes of size $8\times8$, with the first $12$ planes representing current piece positions for each piece type and color, and one plane marking whether the position is a repetition. Historical context is incorporated through the $7$ previous board positions ($91 = 7 \times 13$ planes) following the same procedure, plus auxiliary planes encoding castling rights ($4$ planes), side to move ($1$ plane), fifty-move clock ($1$ plane holding the number of half-moves since the last capture or pawn move), and two constant planes ($0$s and $1$s) that are architectural remnants from CNNs without functional meaning.

\paragraph{Positional encoding}

Leela employs chess-specific positional encodings that capture movement relationships between squares in a $64 \times 64$ matrix. Each square receives a $64$-dimensional vector where position $(i,j)$ is set to 1 if any piece could legally move from square $i$ to square $j$ in one move regardless of current board state, $0$ otherwise, and $-1$ for diagonal entries $(i=j)$ to distinguish self-reference.

\paragraph{Input preparation}

The $112$ feature planes are reshaped from $112 \times 8 \times 8$ to $64 \times 112$, where each of the $64$ entries corresponds to a board square with the $112$-dimensional feature vector for that square concatenated with its corresponding $64$-dimensional positional encoding vector, forming a $64 \times 176$ tensor. This combined representation undergoes a linear transformation with Mish activation \citep{misra2020mishselfregularizednonmonotonic}, followed by elementwise scaling and shifting operations, producing the final $64 \times 768$ input for the transformer layers. This enriched preprocessing was motivated by observations that early attention layers contributed minimally to performance \citep{monroe2024transformerprogress}. While no information mixing between tokens occurs here, this involves substantially more processing than typical language models (which simply use embedding matrices plus positional encodings), potentially explaining why our logit lens mappings yield meaningful results even before the first transformer layer.

\subsection{Transformer encoder}

The transformer encoder consists of $15$ identical layers with a model dimensionality of $768$, processing $64$ tokens (one per board square) through multi-head attention with Smolgen enhancement, feed-forward networks, and Post-LN normalization with DeepNorm scaling. Unlike autoregressive models, all tokens can attend to each other bidirectionally.

\paragraph{Attention with Smolgen}
Each layer uses $24$ attention heads with $32$-dimensional head size, applying scaled dot-product attention to the $64$ board square tokens. The Smolgen module enhances standard self-attention by enabling attention scores to depend not only on individual square contents but also on the global board position. It compresses all square representations into a global vector, processes this through MLPs, then generates supplementary attention logits of shape $24 \times 64 \times 64$ that are added to the standard query-key dot products before softmax normalization.

\paragraph{Feed-forward}
Each FFN layer expands from $768$ to $1024$ dimensions with squared ReLU activation. Unlike other domains that benefit from $4 \times$ expansion ratios, chess models show little improvement from larger feed-forward networks \citep{monroe2024transformerprogress}.

\paragraph{Post-LN with DeepNorm}
The model employs a Post-LN architecture, the original design used in early transformers \citep{vaswani2017attention}, applying layer normalization after each sublayer (attention and feed-forward) within the residual stream rather than before sublayers as in modern Pre-LN variants, following empirical comparisons that demonstrated superior performance \citep{dje2025postnorm}. To mitigate vanishing gradients, the model uses DeepNorm scaling \citep{deepnet}, which applies a constant upscaling factor to residual connections and uses specialized weight initialization. See Appendix~\ref{app:norm_differences} for complete equations.

\subsection{Output heads}

The transformer encoder's final $64 \times 768$ representations are processed by three specialized heads: the policy head generates move probability distributions, the value head predicts win/draw/loss outcomes, and the moves-left head estimates remaining game length.

\paragraph{Policy head}
The policy head first processes each square's $768$-dimensional representation through a shared MLP with Mish activation. These processed representations are then transformed via two separate linear projections: one creating ``source'' representations for squares where moves originate, and another creating ``target'' representations for destination squares. The source and target representations are matrix-multiplied along the $768$-dimensional axis, producing a $64 \times 64$ matrix where each entry contains a scalar logit representing the likelihood of a move from the corresponding source square to target square. Promotion moves require additional processing through two specialized branches that handle move selection and promotion-type preferences separately. The final output combines standard move logits with promotion logits, which are then filtered to extract only legal moves for the current position before applying softmax normalization.

\paragraph{Value head}
The value head processes each square's representation through an MLP with Mish activation, reducing them to $32$ dimensions per square. These $64$ representations are flattened into a single $2048$-dimensional vector to enable global position assessment. A second MLP layer compresses this to $128$ dimensions, followed by a linear projection to $3$ logits, which are then converted to win, draw, and loss outcome probabilities via softmax normalization. Unlike AlphaZero's single scalar output, this three-way classification provides more nuanced position evaluation.

\paragraph{Moves-left head}
The moves-left head predicts the number of moves remaining until game termination. Each square's representation is processed through an MLP, then flattened into a global vector for position-level assessment. Two additional MLP layers progressively reduce the dimensionality to produce a final scalar output estimating remaining game length. The output layer uses Mish activation because ReLU would provide zero gradients when pre-activation values are negative during training, even though the target output is always positive.

\subsection{Search}
\label{app:search_mechanics}
Leela's search \citep{leela} is a variant of Monte Carlo tree search that follows AlphaZero \citep{silver2018general}. For every legal move $a$ in a position $s$ of the game tree, it stores the prior $P(s,a)$ given by the policy head, the number of visits $N(s,a)$ and the mean value $Q(s,a)$ of all evaluations below the move. Each iteration starts at the root and, at every node, follows the move that maximizes
\begin{equation}
Q(s,a) + c\, P(s,a)\, \frac{\sqrt{\sum_b N(s,b)}}{1 + N(s,a)},
\end{equation}
where $c$ grows slowly with the number of visits. The search thus favors moves with a high value and, among moves with few visits, those with a high prior. Moves that have not been visited yet enter this rule with a first-play-urgency value in place of $Q$. Unlike AlphaZero, Leela can also add a small, optional term from its moves-left head, which predicts the remaining length of the game, so that shorter wins are preferred.

When an iteration reaches a position that has not been evaluated, the network evaluates it once. The policy head gives the priors of the new position's moves, with an optional temperature applied to its logits, and the value head gives the probabilities of a win, a draw and a loss. The win probability minus the loss probability is passed back along the path and averaged into $Q$ at every node, with its sign flipped at each ply. Checkmates, stalemates and draws by rule are scored exactly, without the network. All of this holds in both self-play training and deployment. Our analyses of the policy network use the plain softmax of its logits, without the search temperature.

In self-play training games, Dirichlet noise is added to the priors at the root to encourage exploration, and the search stops once further visits barely change the distribution of visits at the root, measured by the gain in KL divergence per visit. These normalized visit counts are the training target of the policy head, and the move played in the game is sampled from them with a temperature that decreases over the first moves of the game to introduce opening diversity. 

In deployment, no noise is added, the search runs for a given budget of nodes or time, and smart pruning stops it early once the most visited move can no longer be overtaken. The engine then plays the most visited move.

\section{Intermediate policy dynamics}
\label{app:policy_dynamics}

\begin{figure}[h!]
    \centering
    \begin{subfigure}[b]{0.48\textwidth}
        \centering
        \includegraphics[width=\textwidth]{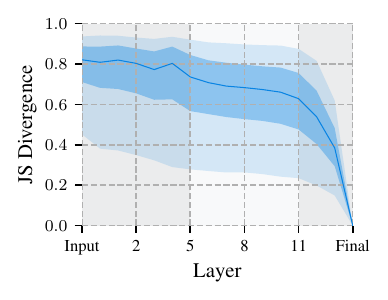}
        \caption{JS Divergence to final policy}
    \end{subfigure}
    \hfill
    \begin{subfigure}[b]{0.48\textwidth}
        \centering
        \includegraphics[width=\textwidth]{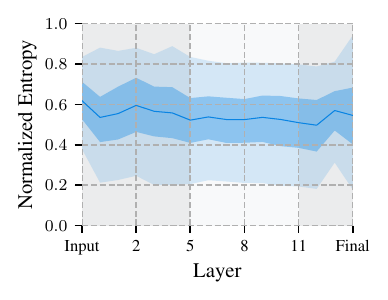}
        \caption{Policy entropy}
    \end{subfigure}
    
    \begin{subfigure}[b]{0.48\textwidth}
        \centering
        \includegraphics[width=\textwidth]{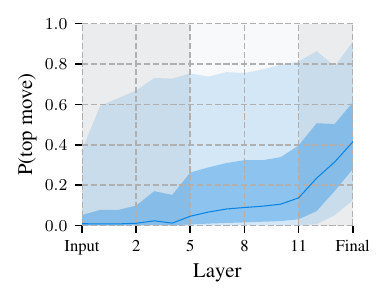}
        \caption{Probability of final top move}
    \end{subfigure}
    \hfill
    \begin{subfigure}[b]{0.48\textwidth}
        \centering
        \includegraphics[width=\textwidth]{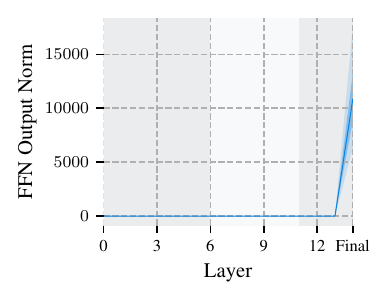}
        \caption{FFN output norm by layer}
    \end{subfigure}
    
    \begin{subfigure}[b]{0.48\textwidth}
        \centering
        \includegraphics[width=\textwidth]{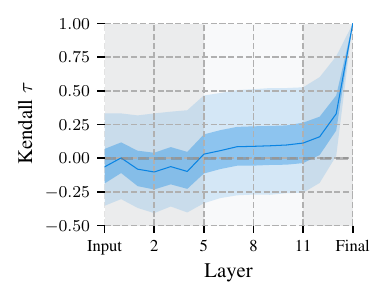}
        \caption{Kendall's $\tau$ (all moves)}
    \end{subfigure}
    \hfill
    \begin{subfigure}[b]{0.48\textwidth}
        \centering
        \includegraphics[width=\textwidth]{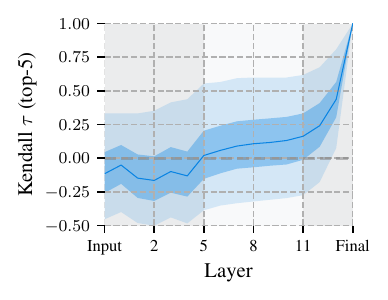}
        \caption{Kendall's $\tau$ (top-5 moves)}
    \end{subfigure}
    
    \caption{Policy dynamics across network depth on $10{,}000$ 
    CCRL positions. Lines show medians; inner and outer shaded 
    regions represent the $25$th--$75$th and $5$th--$95$th 
    percentile ranges. Background shading indicates the three 
    computational phases.}
    \label{fig:core_metrics}
\end{figure}

To provide a deeper, quantitative view of the model's inference process, we computed several policy metrics across the network's depth. The following analyses were conducted on a sample of $10{,}000$ positions from the CCRL dataset \citep{ccrl_dataset}. Figure~\ref{fig:core_metrics} presents five metrics that characterize the policy's evolution. We plot the median, the $25$th--$75$th percentile range, and the $5$th--$95$th percentile range and indicate the proposed phases as shaded background. 

\paragraph{Jensen-Shannon divergence} The Jensen-Shannon divergence shows a median that remains high until the final layers, with the $5$th percentile being much lower. This shows that a small subset of positions converge early to the final policy, while the vast majority have different move preferences than the final output throughout most layers. The phase structure is again apparent, with higher variability and erratic changes in early layers, a more stable plateau in the middle phase, and a sharp decline in the final phase. In this metric, the transition to the final phase appears to occur slightly later—around layer $12$ rather than $11$—though consistent with the overall three-phase progression described in the main text.

\paragraph{Entropy} 
The policy entropy shows a relatively stable median with modest variation across layers, indicating that overall certainty changes only slightly with depth. The $95$th and $5$th percentiles are very high and very low respectively, indicating that for any layer there are positions where the layer is either very certain or very uncertain about what moves are good. The phase structure is again apparent, with more erratic fluctuations in the early layers, a stable middle phase, and increased variability toward the end.

\paragraph{Probability of final top move} 
The median probability assigned to the final top move is near zero for the first layers, then grows approximately linearly with a steeper increase in the final layers. The outliers are very high (between $0.6$ and $0.8$), indicating that all layers correctly predict the top move for some positions with high probability, having already converged. The phase structure is again visible, with slightly more stable behavior in the middle layers and a pronounced acceleration in the final phase, which here again appears to begin around layer $12$.

\paragraph{FFN output norms} \label{app:mlp_output_norm} Following \citet{lad2025remarkablerobustnessllmsstages}, we examine the L$2$ norm of each transformer FFN sublayer's output during a standard forward pass to probe for their proposed residual sharpening mechanism. While the relatively constant policy entropy suggests no sharpening at the distributional level, the FFN output norms exhibit an extreme last-layer increase, suggesting a feature consolidation process analogous to that observed in language models. This is consistent with the final feed-forward network playing a dominant role in reshaping the policy, as observed in the forgotten puzzle override.

\paragraph{Ranking correlation (Kendall's \texorpdfstring{$\tau$}{tau})}
To analyze the stability of move preferences, we compute Kendall's $\tau$ rank correlation between intermediate and final policies. We calculate this metric in two ways: first using all legal moves, and second, to mitigate noise from low-probability moves that are never seriously considered, only over moves that appear in the top five at any layer. Both approaches yield nearly identical results, with the correlation being negative in early layers and remaining low until around layer $12$, where it increases sharply. The $5$th and $95$th percentiles in the top-$5$ moves analysis show that there are positions where intermediate layer move rankings are either very similar or very dissimilar to the final ranking until late in the network. The phase structure is again apparent, with high variability in early layers, relatively stable correlations across the middle phase up to layer $11$, and a sharp rise beginning around layer $12$, consistent with the transition to the late phase.

\clearpage

\section{Complete tables for the tournament and Lichess play}
\label{app:tournament}

\subsection{Internal tournament}

We evaluate playing strength through internal round-robin tournaments processed with BayesElo \citep{coulom208whole} using the default confidence parameter of $0.5$. Each model pairing played $200$ distinct openings from the Encyclopedia of Chess Openings~\citep{matanovic1978encyclopaedia}, with one game per side at deterministic temperature $\tau = 0$ and five games per side at stochastic temperature $\tau = 1.0$. Games are adjudicated by Stockfish at $0.01$ seconds per move, which ends a game once its evaluation exceeds $1{,}300$ centipawns in either direction. Tables~\ref{tab:tournament_results_tau0} and~\ref{tab:tournament_results_tau1} report the resulting Elo ratings, computed from all match outcomes. Each table lists the model’s Elo estimate with asymmetric confidence bounds, total score percentage (wins plus half draws), average opponent rating, and draw rate. The Lc0 anchor is evaluated under argmax in both regimes following \citet{ruoss2024amortizedplanninglargescaletransformers}; in the $\tau=1$ tournament it therefore wins nearly every game against the stochastic field, so the absolute Elo scale is meaningless and only relative ordering and gaps within the tournament are interpretable. The results show consistent improvement in playing strength with network depth, with clear early- and late-layer acceleration in both deterministic and stochastic settings, while middle layers exhibit a prolonged performance plateau.

\begin{table*}[htbp]
\centering
\caption{Playing strength evaluation through internal tournament ($\tau=0$)}
\label{tab:tournament_results_tau0}
\resizebox{\textwidth}{!}{
\begin{tabular}{clccccccc}
\toprule
\textbf{Rank} & \textbf{Model} & \textbf{Elo} & $\mathbf{+}$ & $\mathbf{-}$ & \textbf{Games} & \textbf{Score (\%)} & \textbf{Avg. Oppo.} & \textbf{Draws (\%)} \\
\midrule
1  & Lc0 Policy Net (Anchor) & 2292 & 25 & 25 & 6400 & 97 & 1043 & 2 \\
2  & Full Model                       & 2268 & 25 & 25 & 6400 & 96 & 1044 & 3 \\
3  & Logit Lens Layer 13              & 1670 & 22 & 21 & 6400 & 85 & 1081 & 2 \\
4  & Logit Lens Layer 12              & 1356 & 13 & 12 & 6400 & 74 & 1101 & 5 \\
5  & Logit Lens Layer 11              & 1117 & 10 & 10 & 6400 & 58 & 1116 & 7 \\
6  & Logit Lens Layer 10              & 1082 & 10 & 10 & 6400 & 55 & 1118 & 11 \\
7  & Logit Lens Layer 9               & 1055 & 10 & 10 & 6400 & 53 & 1120 & 12 \\
8  & Logit Lens Layer 8               & 1030 & 10 & 10 & 6400 & 51 & 1121 & 11 \\
9  & Logit Lens Layer 7               & 1028 & 10 & 10 & 6400 & 51 & 1122 & 12 \\
10 & Logit Lens Layer 6                & 997 & 10 & 10 & 6400 & 48 & 1124 & 12 \\
11 & Logit Lens Layer 5                & 993 &  9 &  9 & 6400 & 48 & 1124 & 17 \\
12 & Logit Lens Layer 4                & 908 &  9 &  9 & 6400 & 40 & 1129 & 18 \\
13 & Logit Lens Layer 3                & 832 &  9 &  9 & 6400 & 32 & 1134 & 16 \\
14 & Logit Lens Layer 2                & 730 & 10 & 10 & 6400 & 23 & 1140 & 16 \\
15 & Logit Lens Layer 1                & 655 & 11 & 10 & 6400 & 17 & 1145 & 17 \\
16 & Logit Lens Layer 0                & 609 & 11 & 12 & 6400 & 16 & 1148 & 8 \\
17 & Logit Lens Input                  & 351 & 17 & 17 & 6400 &  4 & 1164 & 4 \\
\bottomrule
\end{tabular}
}
\end{table*}

\begin{table*}[htbp]
\centering
\caption{Playing strength evaluation through internal tournament ($\tau=1$)}
\label{tab:tournament_results_tau1}
\resizebox{\textwidth}{!}{
\begin{tabular}{clccccccc}
\toprule
\textbf{Rank} & \textbf{Model} & \textbf{Elo} & $\mathbf{+}$ & $\mathbf{-}$ & \textbf{Games} & \textbf{Score (\%)} & \textbf{Avg. Oppo.} & \textbf{Draws (\%)} \\
\midrule
1  & Lc0 Policy Net (Anchor) & 2292 & -- & -- & 32000 & 100 & 749 & 0 \\
2  & Full Model                       & 1367 &  8 &  8 & 32000 &  88 & 807 & 1 \\
3  & Logit Lens Layer 13              & 1119 &  6 &  6 & 32000 &  76 & 822 & 2 \\
4  & Logit Lens Layer 12              & 1065 &  5 &  5 & 32000 &  73 & 826 & 3 \\
5  & Logit Lens Layer 11               & 876 &  5 &  5 & 32000 &  58 & 837 & 5 \\
6  & Logit Lens Layer 10               & 872 &  5 &  5 & 32000 &  57 & 838 & 5 \\
7  & Logit Lens Layer 9                & 851 &  5 &  5 & 32000 &  55 & 839 & 5 \\
8  & Logit Lens Layer 7                & 816 &  5 &  5 & 32000 &  52 & 841 & 6 \\
9  & Logit Lens Layer 8                & 813 &  5 &  5 & 32000 &  52 & 841 & 6 \\
10 & Logit Lens Layer 5                & 810 &  5 &  5 & 32000 &  52 & 842 & 6 \\
11 & Logit Lens Layer 6                & 799 &  5 &  5 & 32000 &  51 & 842 & 5 \\
12 & Logit Lens Layer 4                & 718 &  5 &  5 & 32000 &  43 & 847 & 7 \\
13 & Logit Lens Layer 3                & 585 &  5 &  5 & 32000 &  31 & 856 & 7 \\
14 & Logit Lens Layer 2                & 469 &  6 &  6 & 32000 &  22 & 863 & 7 \\
15 & Logit Lens Layer 1                & 391 &  6 &  6 & 32000 &  17 & 868 & 5 \\
16 & Logit Lens Layer 0                & 387 &  6 &  6 & 32000 &  17 & 868 & 4 \\
17 & Logit Lens Input                   & 46 &  9 & 10 & 32000 &   4 & 889 & 0 \\
\bottomrule
\end{tabular}
}
\end{table*}

\clearpage

\subsection{Lichess bot performance}

To validate our internal tournament findings in a real-world setting, we deployed layer-wise policies as bots on Lichess \citep{lichess2025} using the bot API framework \citep{lichessbot2023github}. The bots sample from the policy for the first five moves to introduce opening diversity and then play the top move. They participated in Bullet (1+0, 2+1), Blitz (3+0, 3+2, 5+0, 5+3), and Rapid (10+0, 10+5, 15+10) time controls. Ratings and game counts were recorded on 28 October 2026. At the time of writing, the bots are still playing on Lichess, so the linked profiles may show more recent numbers.

Several experimental constraints shaped the evaluation: (1) bots played exclusively against other bots per Lichess policy, (2) the limited pool of weak bots meant repeated matchups for early layers, and (3) deterministic play after the opening often led to repetitive games. Despite these limitations, the results corroborate our internal tournament findings, showing progressive improvement across layers with the most significant transitions occurring at similar points in the network depth.
Table~\ref{tab:lichess_results} presents the playing strength and performance statistics for our layer-wise Lichess bots, with bot names linked to their respective Lichess profiles.

\begin{table}[htbp]
\centering
\caption{Performance of layer-wise policies on Lichess across different time controls}
\label{tab:lichess_results}
\begin{tabular}{lccccccc}
\toprule
\textbf{Bot} & \multicolumn{3}{c}{\textbf{Rating}} & \textbf{Total} & \multicolumn{3}{c}{\textbf{Performance}} \\
\cmidrule(lr){2-4} \cmidrule(lr){6-8}
& \textbf{Bullet} & \textbf{Blitz} & \textbf{Rapid} & \textbf{Games} & \textbf{W} & \textbf{D} & \textbf{L} \\
\midrule
\href{https://lichess.org/@/LLLBot-In}{LLLBot-In} & 817 $\pm$ 45 & 545 $\pm$ 45 & 492 $\pm$ 45 & 2190 & 19 & 528 & 1643 \\
\href{https://lichess.org/@/LLLBot-0}{LLLBot-0} & 843 $\pm$ 45 & 631 $\pm$ 45 & 617 $\pm$ 45 & 2023 & 30 & 478 & 1515 \\
\href{https://lichess.org/@/LLLBot-1}{LLLBot-1} & 907 $\pm$ 45 & 611 $\pm$ 45 & 569 $\pm$ 45 & 1929 & 29 & 456 & 1444 \\
\href{https://lichess.org/@/LLLBot-2}{LLLBot-2} & 930 $\pm$ 45 & 655 $\pm$ 45 & 669 $\pm$ 45 & 1820 & 43 & 478 & 1299 \\
\href{https://lichess.org/@/LLLBot-3}{LLLBot-3} & 961 $\pm$ 45 & 740 $\pm$ 45 & 789 $\pm$ 45 & 1799 & 43 & 454 & 1302 \\
\href{https://lichess.org/@/LLLBot-4}{LLLBot-4} & 998 $\pm$ 45 & 723 $\pm$ 45 & 791 $\pm$ 45 & 1689 & 51 & 494 & 1144 \\
\href{https://lichess.org/@/LLLBot-5}{LLLBot-5} & 1071 $\pm$ 45 & 815 $\pm$ 45 & 871 $\pm$ 45 & 1634 & 105 & 388 & 1141 \\
\href{https://lichess.org/@/LLLBot-6}{LLLBot-6} & 1085 $\pm$ 45 & 815 $\pm$ 45 & 907 $\pm$ 45 & 1529 & 131 & 326 & 1072 \\
\href{https://lichess.org/@/LLLBot-7}{LLLBot-7} & 1087 $\pm$ 45 & 827 $\pm$ 45 & 954 $\pm$ 45 & 2135 & 194 & 537 & 1404 \\
\href{https://lichess.org/@/LLLBot-8}{LLLBot-8} & 1096 $\pm$ 45 & 906 $\pm$ 45 & 954 $\pm$ 45 & 2004 & 194 & 502 & 1308 \\
\href{https://lichess.org/@/LLLBot-9}{LLLBot-9} & 1094 $\pm$ 45 & 952 $\pm$ 45 & 985 $\pm$ 45 & 1849 & 208 & 446 & 1195 \\
\href{https://lichess.org/@/LLLBot-10}{LLLBot-10} & 1124 $\pm$ 45 & 860 $\pm$ 45 & 1008 $\pm$ 45 & 1805 & 210 & 408 & 1187 \\
\href{https://lichess.org/@/LLLBot-11}{LLLBot-11} & 1159 $\pm$ 45 & 946 $\pm$ 45 & 1072 $\pm$ 45 & 1652 & 262 & 297 & 1093 \\
\href{https://lichess.org/@/LLLBot-12}{LLLBot-12} & 1399 $\pm$ 45 & 1205 $\pm$ 45 & 1324 $\pm$ 45 & 1529 & 468 & 282 & 779 \\
\href{https://lichess.org/@/LLLBot-13}{LLLBot-13} & 1622 $\pm$ 45 & 1482 $\pm$ 45 & 1637 $\pm$ 45 & 1321 & 659 & 162 & 500 \\
\href{https://lichess.org/@/LLLBot-Full}{LLLBot-Full} & 2306 $\pm$ 45 & 2099 $\pm$ 45 & 2237 $\pm$ 45 & 959 & 521 & 116 & 322 \\
\bottomrule
\end{tabular}
\end{table}

\section{Forgotten puzzles across Lc0 networks}
\label{app:multi-model-logit-lens}

To verify that the forgotten puzzle phenomenon is not specific to the T82-768x15x24h network analyzed in the main paper, we apply the same logit lens to four additional Lc0 checkpoints from the publicly available \href{https://lczero.org/play/networks/bestnets/}{Lc0 network repository}, spanning three width classes ($256$, $512$, $768$), two depths ($10$ and $15$ encoder blocks), and multiple training generations.

\begin{figure}[h]
\captionsetup[subfigure]{justification=centering}
  \centering
  \begin{subfigure}[t]{0.48\textwidth}
    \centering
    \includegraphics[width=\textwidth]{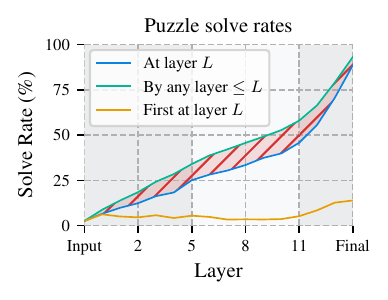}
    \caption{T82 (768x15x24h,\\the model analyzed in this paper)}
    \label{fig:msr-lc0-original}
  \end{subfigure}\hfill
  \begin{subfigure}[t]{0.48\textwidth}
    \centering
    \includegraphics[width=\textwidth]{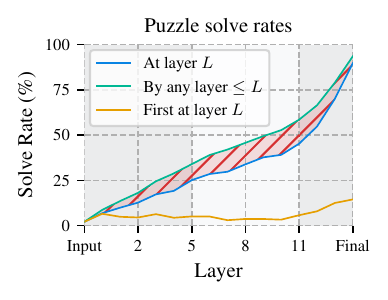}
    \caption{T82 no-history fine-tune (768x15x24h,\\from \citealp{jenner2024evidencelearnedlookaheadchessplaying})}
    \label{fig:msr-lc0}
  \end{subfigure}

  \vspace{0.5em}

  \begin{subfigure}[t]{0.48\textwidth}
    \centering
    \includegraphics[width=\textwidth]{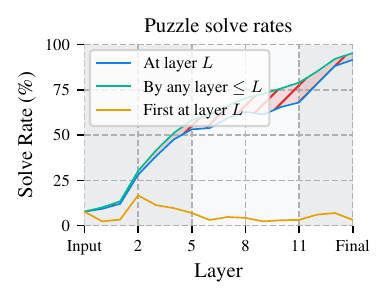}
    \caption{BT3 (768x15x24h)}
    \label{fig:msr-bt3}
  \end{subfigure}\hfill
  \begin{subfigure}[t]{0.48\textwidth}
    \centering
    \includegraphics[width=\textwidth]{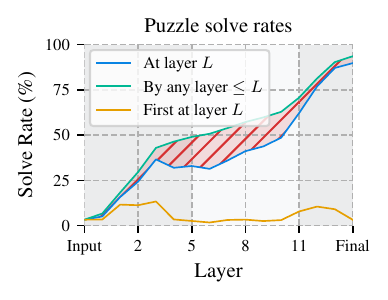}
    \caption{T3 (512x15x16h)}
    \label{fig:msr-t3}
  \end{subfigure}

  \vspace{0.5em}

  \begin{subfigure}[t]{0.48\textwidth}
    \centering
    \includegraphics[width=\textwidth]{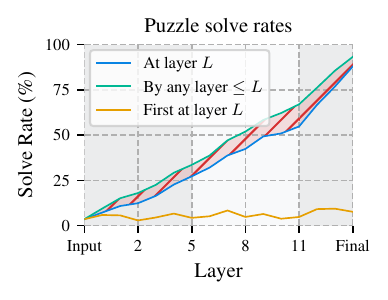}
    \caption{T1 (512x15x8h)}
    \label{fig:msr-t1-512}
  \end{subfigure}\hfill
  \begin{subfigure}[t]{0.48\textwidth}
    \centering
    \includegraphics[width=\textwidth]{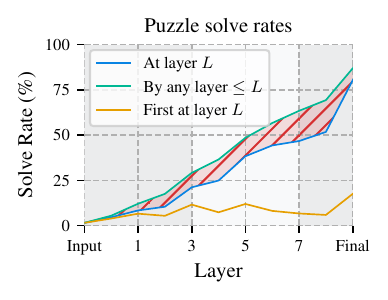}
    \caption{T1 (256x10, only 10 encoder blocks)}
    \label{fig:msr-t1-256x10}
  \end{subfigure}

  \caption{Layer-wise puzzle solve rates for six Lc0 networks of varying width, depth, and training generation. The hatched \forgotten{forgotten puzzle} band between \byanylayer{solved by any layer $\leq L$} and \atlayer{solved at $L$} is visible in every model.}
  \label{fig:multi-model-puzzle-solve-rates}
\end{figure}

Figure~\ref{fig:multi-model-puzzle-solve-rates} reports puzzle solve rates
across layers for each network, evaluated on the general dataset consisting of $10{,}000$ puzzles. The \forgotten{forgotten} puzzle band---the
gap between \byanylayer{solved by any layer $\leq L$} and \atlayer{solved at
$L$}---is visible in every network, indicating that solution forgetting
is a general property rather than a
single-checkpoint artifact.

\clearpage

\section{Complete probabilities for the solved example puzzle from Figure~\ref{fig:puzzle_example}}
\label{app:probs_example_puzzle}

This puzzle (Figure~\ref{fig:puzzle_62FeU}) features a knight sacrifice leading to mate (PV: 1.~\protect\bmove{Ng3+} 2.~\protect\wmove{hxg3} \protect\wmove{Rh6\mate}). The input layer exhibits piece-specific bias toward queen moves, with multiple captures dominating early probabilities. The winning move \protect\wmove{Ng3+} first becomes the top choice at layer $6$ ($26.7\%$) but is then briefly overtaken by \protect\wmove{Qg1+}, which dominates layers $9$--$10$. This temporary preference for \protect\wmove{Qg1+} likely reflects a learned heuristic that queen checks are tactically important, even though this particular check ultimately loses the queen without compensation. After layer $10$, \protect\wmove{Ng3+} regains its position as the preferred move, with the exception of layer $12$ where a rook lift \protect\wmove{Rh6} ($51.7\%$) briefly becomes the top candidate, potentially pinning the h-pawn and threatening the king along the h-file. The winning move's probability follows a non-monotonic trajectory, fluctuating throughout the layers before ultimately surging to $49.2\%$ at layer $13$ and $87.5\%$ in the final output. This contrasts with some of the other cases in Appendix~\ref{app:puzzles}, where the correct move is either identified immediately after early layers or remains unconsidered until the very late layers.  Full probabilities are in Tables~\ref{tab:puzzle_62FeU_probs1} and~\ref{tab:puzzle_62FeU_probs2}.

\begin{figure}[h!]
    \centering
    \includegraphics[width=0.9\textwidth]{Figures/Puzzles/Examples/puzzle_visualization_62FeU.pdf}
    \caption{Layer-wise policy evolution for puzzle \href{https://lichess.org/training/62FeU}{\texttt{62FeU}} from Figure~\ref{fig:puzzle_example}.}
    \label{fig:puzzle_62FeU}
\end{figure}

\include{Figures/Puzzles/Examples/puzzle_tables_62FeU}

\clearpage

\section{Correctly solved puzzle case studies}
\label{app:puzzles}

\subsection{Knight fork and discovered attack}
This puzzle (Figure~\ref{fig:puzzle_9JvtA}) features a tactical sequence beginning with a knight check that sets up a discovered attack (PV: 1. \protect\wmove{Nd6+} \protect\wmove{exd6} 2. \protect\wmove{Bxc6+}). After the knight is captured, the bishop takes the queen's only defender with check. Black must answer the check instead of saving the queen, and White captures the undefended queen on the next move. The input layer exhibits piece-specific bias toward knight moves, which are abandoned in early layers as the model shifts focus to queen captures in layer $1$. The queen trade \protect\wmove{Qxa5}, with the queen being protected by a knight, dominates through layers $4$--$13$ with probabilities around $50\%$. Apart from $11.55\%$ at layer $2$, the winning move \protect\wmove{Nd6+} maintains low probability until layer $12$ ($1.65\%$), then suddenly jumps to $24.37\%$ at layer $13$ before becoming the decisive top choice in the final output. This sudden increase at layer $13$—the same layer \citet{cruz2025understanding} identified as containing one of the look-ahead heads—may indicate the model requires multi-move analysis rather than simple tactical heuristics for this position. Full probabilities are in Tables~\ref{tab:puzzle_9JvtA_probs1} and~\ref{tab:puzzle_9JvtA_probs2}.

\begin{figure}[h!]
    \centering
    \includegraphics[width=0.9\textwidth]{Figures/Puzzles/Examples/puzzle_visualization_9JvtA.pdf}
    \caption{Layer-wise policy evolution for puzzle \href{https://lichess.org/training/9JvtA}{\texttt{9JvtA}}.}
    \label{fig:puzzle_9JvtA}
\end{figure}

\input{Figures/Puzzles/Examples/puzzle_tables_9JvtA.tex}

\clearpage

\subsection{Early and consistent solution lock-in}
This puzzle (Figure~\ref{fig:puzzle_70v1B}) features a back-rank mate initiated by a queen sacrifice (PV: 1.~\protect\wmove{Qxe8+} \protect\wmove{Rxe8} 2.~\protect\wmove{Rxe8\mate}). The winning move \protect\wmove{Qxe8+} becomes the top candidate immediately at layer $0$ ($50.5\%$) and maintains dominance through nearly all subsequent layers. The only significant competitor is the pawn capture \protect\wmove{bxc3}, which briefly overtakes the solution at layer $3$ ($56.3\%$ vs $40.7\%$) and layer $7$ ($52.9\%$ vs $42.9\%$), and only disappears in the final output. All other moves remain at or below $1.5\%$ from layer $1$ onwards. The solution reaches $88.9\%$ in the final output. This early convergence contrasts with the delayed emergence seen in other case studies and may reflect the winning move combining both queen-move and capture biases present in initial layers. Full probabilities are in Tables~\ref{tab:puzzle_70v1B_probs1} and~\ref{tab:puzzle_70v1B_probs2}.

\begin{figure}[h!]
    \centering
    \includegraphics[width=0.9\textwidth]{Figures/Puzzles/Examples/puzzle_visualization_70v1B.pdf}
    \caption{Layer-wise policy evolution for puzzle \href{https://lichess.org/training/70v1B}{\texttt{70v1B}}.}
    \label{fig:puzzle_70v1B}
\end{figure}

\input{Figures/Puzzles/Examples/puzzle_tables_70v1B.tex}

\clearpage

\subsection{Knight sacrifice with rook mate}

This puzzle (Figure~\ref{fig:puzzle_0KBPy}), featured in \citet{jenner2024evidencelearnedlookaheadchessplaying}, requires a knight sacrifice (PV: 1.~\protect\wmove{Ng6+} \protect\wmove{hxg6} 2.~\protect\wmove{Rh4\mate}). The input encoding favors specific pieces, while layer $0$ is dominated by the queen check \protect\wmove{Qg8+} ($40.0\%$), which reappears briefly at layer $2$. Through middle layers the model favors the rook move \protect\wmove{Rc4}, which attacks the black queen; it peaks at $69.4\%$ (layer $4$) and remains the top candidate through layer $8$. Meanwhile, the queen maneuver \protect\wmove{Qe6} threatening the opposing queen and rook while being protected by the knight is also considered in layers $7$--$11$. The correct sacrifice \protect\wmove{Ng6+} remains below $10\%$ through layer $7$, rises to $10.8\%$ at layer $8$, becomes the top candidate at layer $9$ ($14.7\%$), and climbs to a peak of $53.1\%$ at layer $12$. After a brief dip at layer $13$ ($33.2\%$), where the quiet pawn push \protect\wmove{h4} is considered, the solution recovers to dominate the final output at $70.7\%$. Full probabilities are in Tables~\ref{tab:puzzle_0KBPy_probs1} and~\ref{tab:puzzle_0KBPy_probs2}.

\begin{figure}[h!]
    \centering
    \includegraphics[width=0.9\textwidth]{Figures/Puzzles/Examples/puzzle_visualization_0KBPy.pdf}
    \caption{Layer-wise policy evolution for puzzle \href{https://lichess.org/training/0KBPy}{\texttt{0KBPy}}.}
    \label{fig:puzzle_0KBPy}
\end{figure}

\input{Figures/Puzzles/Examples/puzzle_tables_0KBPy.tex}

\clearpage

\subsection{Queen sacrifice for rook mate}
This puzzle (Figure~\ref{fig:puzzle_09SS5}) involves a queen sacrifice enabling a back-rank rook mate (PV: 1.~\protect\wmove{Qe8+} \protect\wmove{Rxe8} 2.~\protect\wmove{Rxe8\mate}). The input encoding favors quiet queen moves \protect\wmove{Qd7} ($30.6\%$) and \protect\wmove{Qc7} ($20.7\%$). From layer $0$ onwards, the model shifts toward captures and checks: the queen capture \protect\wmove{Qxf7+} ($28.9\%$--$39.3\%$, layers $0$--$4$), the material capture \protect\wmove{Qxa3} ($33.8\%$ at layer $0$, declining to $2.9\%$ by layer $12$), and the rook capture \protect\wmove{Rxa3}, which persists throughout ($10\%$--$20\%$ across most layers). The pawn check \protect\wmove{gxf7+} dominates the middle layers, peaking at $43.5\%$ (layer $6$) and remaining the top candidate through layer $12$—a forcing move that delivers check but lacks the mating continuation. The winning move \protect\wmove{Qe8+} receives minimal probability ($<2\%$) through layer $7$, then emerges gradually from layer $8$ ($3.9\%$). The solution only narrowly leads at layer $13$ ($22.3\%$ vs $20.0\%$ for \protect\wmove{gxf7+}), with the decisive separation occurring entirely in the final layer where it reaches $58.9\%$. Full probabilities are in Tables~\ref{tab:puzzle_09SS5_probs1} and~\ref{tab:puzzle_09SS5_probs2}.

\begin{figure}[h!]
    \centering
    \includegraphics[width=0.9\textwidth]{Figures/Puzzles/Examples/puzzle_visualization_09SS5.pdf}
    \caption{Layer-wise policy evolution for puzzle \href{https://lichess.org/training/09SS5}{\texttt{09SS5}}.}
    \label{fig:puzzle_09SS5}
\end{figure}

\input{Figures/Puzzles/Examples/puzzle_tables_09SS5.tex}

\clearpage

\subsection{Bishop sacrifice for discovered attack}
This puzzle (Figure~\ref{fig:puzzle_7UUlN}) features a bishop sacrifice winning material via a discovered attack (PV: 1.~\protect\bmove{Bh2+} 2.~\protect\wmove{Kxh2} \protect\wmove{Rxc6}). The input encoding strongly favors the pawn push \protect\wmove{a4} ($57.5\%$). From layer $1$, the immediate material capture \protect\wmove{Rxe3} dominates, peaking at $86.8\%$ (layer $3$) and remaining the top candidate through layer $6$. The rook centralization \protect\wmove{Re5} then emerges as a competitor, and the two alternate as the top choice through layers $7$--$11$, with \protect\wmove{Rxe3} leading at layers $8$ and $10$ and \protect\wmove{Re5} at layers $7$, $9$, and $11$. The winning move \protect\wmove{Bh2+} receives negligible probability ($<2\%$) through layer $5$, then builds steadily from layer $6$ ($14.6\%$) to $\sim$$23\%$ across layers $8$--$11$, before overtaking all competitors at layer $12$ ($51.2\%$) and reaching $92.9\%$ in the final output, demonstrating a late-layer override of a simple material-gain heuristic in favor of a deeper sacrificial sequence. Full probabilities are in Tables~\ref{tab:puzzle_7UUlN_probs1} and~\ref{tab:puzzle_7UUlN_probs2}.

\begin{figure}[h!]
    \centering
    \includegraphics[width=0.9\textwidth]{Figures/Puzzles/Examples/puzzle_visualization_7UUlN.pdf}
    \caption{Layer-wise policy evolution for puzzle \href{https://lichess.org/training/7UUlN}{\texttt{7UUlN}}.}
    \label{fig:puzzle_7UUlN}
\end{figure}

\input{Figures/Puzzles/Examples/puzzle_tables_7UUlN.tex}

\clearpage

\section{Complete probabilities for the forgotten example puzzle from Figure~\ref{fig:forgotten_puzzle}}
\label{app:forgotten_puzzles_example}

This puzzle (Figure~\ref{fig:puzzle_ANklq}) features a mate-in-two requiring a queen sacrifice (PV: 1.~\protect\bmove{Qxf1+} 2.~\protect\wmove{Kxf1} \protect\wmove{Rd1\mate}). The winning move \protect\wmove{Qxf1+} becomes the top candidate immediately at layer $0$ ($45.7\%$) and remains decisively the top choice through layer $13$, with its primary competitor being the queen check \protect\wmove{Qxh2+}, which maintains a persistent secondary role ($23\%$--$31\%$ across layers $0$--$11$ except layer $1$). The solution peaks sharply at layer $12$ ($88.3\%$) as \protect\wmove{Qxh2+} drops to $3.3\%$, but then falls to $30.6\%$ in layer $13$ and is abandoned in the final output ($5.8\%$). The losing move \protect\wmove{Qc8}—a quiet queen retreat to the back rank—receives negligible probability ($<0.2\%$) through layer $12$, then surges to $2.4\%$ (layer $13$) and $37.0\%$ in the final output. A second quiet queen move \protect\wmove{Qd7} follows the same pattern, rising from $0.3\%$ (layer $12$) to $24.7\%$ in the final output. The model replaces a winning queen sacrifice with quiet queen retreats, consistent with the safety-prior mechanism identified in the main text. The value head evaluates the current position as losing ($96.0\%$ loss) despite the mate-in-two available, yet correctly assigns near-certain victory ($99.8\%$ win) to the position after \protect\wmove{Qxf1+} while evaluating all alternatives as losing ($98.9\%$--$100\%$ loss), including the model's preferred \protect\wmove{Qc8} at $98.9\%$ loss (Table~\ref{tab:puzzle_ANklq_eval}). Full probabilities are in Tables~\ref{tab:puzzle_ANklq_probs1} and~\ref{tab:puzzle_ANklq_probs2}.

\begin{figure}[h!]
    \centering
    \includegraphics[width=0.9\textwidth]{Figures/Puzzles/Forgotten/puzzle_visualization_ANklq.pdf}
    \caption{Layer-wise policy evolution for puzzle \href{https://lichess.org/training/ANklq}{\texttt{ANklq}}.}
    \label{fig:puzzle_ANklq}
\end{figure}

\include{Figures/Puzzles/Forgotten/puzzle_evaluation_ANklq}

\include{Figures/Puzzles/Forgotten/puzzle_tables_ANklq}

\clearpage

\section{Forgotten puzzle case studies}
\label{app:forgotten_puzzles}

\subsection{Rook sacrifice for queen capture}

This puzzle (Figure~\ref{fig:puzzle_58Ib0}) presents a rook sacrifice winning the queen (PV: 1. \protect\wmove{Rxg7+} \protect\wmove{Kxg7} 2. \protect\wmove{Qxh4}). The winning move \protect\wmove{Rxg7+} dominates throughout nearly all layers, serving as the top candidate from layer $0$ onwards (excluding layer $1$) with probabilities exceeding $50\%$ from layer $2$ and peaking at $82.72\%$ at layer $10$. Its primary competitor, the queen capture \protect\wmove{Qxa7}, maintains $27\%$--$52\%$ probability through layers $0$--$7$ before declining steadily, since  this move leaves the rook hanging and gains insufficient compensation. The solution is overtaken in the final layer by \protect\wmove{Kf1}—a king move that receives negligible probability ($<1\%$) throughout layers $0$--$10$, then surges to $6.20\%$ (layer $12$), $30.42\%$ (layer $13$), and $52.98\%$ (final output). Notably, the model's value head evaluates the current position as unfavorable ($71.8\%$ loss probability), yet when applied to the position after each legal move, correctly distinguishes the forcing sequence: it assigns near-certain victory ($99.5\%$ win) to the position after \protect\wmove{Rxg7+} while evaluating all alternatives as losing positions ($91\%$--$100\%$ loss) (Table~\ref{tab:puzzle_58Ib0_eval}). Full probabilities are in Tables~\ref{tab:puzzle_58Ib0_probs1} and~\ref{tab:puzzle_58Ib0_probs2}.
\begin{figure}[h!]
    \centering
    \includegraphics[width=0.9\textwidth]{Figures/Puzzles/Forgotten/puzzle_visualization_58Ib0.pdf}
    \caption{Layer-wise policy evolution for puzzle \href{https://lichess.org/training/58Ib0}{\texttt{58Ib0}}.}
    \label{fig:puzzle_58Ib0}
    \end{figure}

\include{Figures/Puzzles/Forgotten/puzzle_evaluation_58Ib0}

\include{Figures/Puzzles/Forgotten/puzzle_tables_58Ib0}

\clearpage

\subsection{Pawn sacrifice for queen capture}

This puzzle (Figure~\ref{fig:puzzle_6PIrs}) presents a pawn sacrifice forcing a queen capture (PV: 1. \protect\bmove{g5+} 2. \protect\wmove{fxg5}  \protect\wmove{Qxe5}). The winning move \protect\wmove{g5+} exhibits highly non-monotonic behavior with substantial probability jumps between consecutive layers: $0.55\%$ to $18.46\%$ (layers $4$--$5$), $22.21\%$ to $0.38\%$ (layers $8$--$9$), and $0.77\%$ to $41.93\%$ (layers $12$--$13$). The solution briefly emerges as the top candidate at layer $13$ ($41.93\%$), but drops to fourth place in the final output ($11.17\%$). Through most middle layers ($2$--$11$), the model strongly favors the materially conservative queen trade \protect\wmove{Qxe5}, peaking at $89.57\%$ (layer $4$). The final layers instead prioritize queen retreats to safety: \protect\wmove{Qd3} ($29.14\%$), \protect\wmove{Qc4} ($25.06\%$), and \protect\wmove{Qd1} ($14.27\%$)—all receiving minimal probability through layer $11$. The sharp inter-layer transitions suggest algorithmic computation of forcing sequences at layer $13$, presumably through look-ahead, rather than gradual accumulation of tactical heuristics. The model's value head evaluates the current position as unfavorable ($75.9\%$ loss probability), yet correctly assigns near-certain victory ($98.3\%$ win) to the position after \protect\wmove{g5+} while evaluating all alternatives as losing positions ($89.1\%$--$100\%$ loss) (Table~\ref{tab:puzzle_6PIrs_eval}). Full probabilities are in Tables~\ref{tab:puzzle_6PIrs_probs1} and~\ref{tab:puzzle_6PIrs_probs2}.

\begin{figure}[h!]
    \centering
    \includegraphics[width=0.9\textwidth]{Figures/Puzzles/Forgotten/puzzle_visualization_6PIrs.pdf}
    \caption{Layer-wise policy evolution for puzzle \href{https://lichess.org/training/6PIrs}{\texttt{6PIrs}}.}
    \label{fig:puzzle_6PIrs}
\end{figure}

\include{Figures/Puzzles/Forgotten/puzzle_evaluation_6PIrs}

\include{Figures/Puzzles/Forgotten/puzzle_tables_6PIrs}

\clearpage

\subsection{Queen sacrifice to back-rank mate}

This puzzle (Figure~\ref{fig:puzzle_1Egyn}) presents a forced mate-in-two sequence requiring a queen sacrifice (PV: 1. \protect\wmove{Qxc8+} \protect\wmove{Bxc8} 2. \protect\wmove{Re8\mate}). The winning move \protect\wmove{Qxc8+} dominates from layers $3$--$13$, peaking at $72.8\%$ (layer $5$). However, this solution is abandoned in the final layer, where \protect\wmove{Qxa7} surges to $63.7\%$ in the output—a materially safe queen capture that does not lead to a sacrifice, receiving only $1.8\%$--$12.4\%$ probability in layers $4$--$12$. The reversal may reflect a safety prior against queen sacrifices overriding mid-layer tactical calculations. The model's value head evaluates the current position as unfavorable ($82.3\%$ loss) despite the mate-in-two available, yet correctly assigns near-certain victory ($100\%$ win) to the position after \protect\wmove{Qxc8+} while evaluating all alternatives as near-certain losses ($99\%$--$100\%$ loss) (Table~\ref{tab:puzzle_1Egyn_eval}). This illustrates a failure mode where sound mid-layer analysis is overwritten by conservative final-layer adjustments despite accurate position evaluation capabilities. Full probabilities are in Tables~\ref{tab:puzzle_1Egyn_probs1} and~\ref{tab:puzzle_1Egyn_probs2}.

\begin{figure}[h!]
    \centering
    \includegraphics[width=0.9\textwidth]{Figures/Puzzles/Forgotten/puzzle_visualization_1Egyn.pdf}
    \caption{Layer-wise policy evolution for puzzle \href{https://lichess.org/training/1Egyn}{\texttt{1Egyn}}.}
    \label{fig:puzzle_1Egyn}
\end{figure}

\include{Figures/Puzzles/Forgotten/puzzle_evaluation_1Egyn}

\include{Figures/Puzzles/Forgotten/puzzle_tables_1Egyn}

\clearpage

\subsection{Rook sacrifice for material gain}

This puzzle (Figure~\ref{fig:puzzle_DsaGi}) presents a rook sacrifice forcing material advantage (PV: 1. \protect\bmove{Rxg4+} 2.~\protect\wmove{Kxg4} \protect\wmove{Kxe6}). The winning move \protect\wmove{Rxg4+} dominates from layers $4$--$13$, maintaining probabilities above $61\%$ throughout and peaking at $88\%$ at layer $10$. However, the solution drops dramatically in the final layer to third place ($12.14\%$), overtaken by \protect\wmove{Rf1+} ($40.4\%$) and \protect\wmove{Rh1} ($23\%$)—two materially conservative rook moves that receive minimal probability ($<8\%$) through layers $0$--$12$. Unlike previous cases, these alternative moves are strategically sound rather than clearly losing: Stockfish evaluates \protect\wmove{Rf1+} at $-0.05$ pawns and \protect\wmove{Rh1} at $-0.15$ pawns, both near-neutral positions. The model's value head partly reflects this when applied to the position after each move: while correctly assigning near-certain victory ($97.9\%$ win) to the position after \protect\wmove{Rxg4+}, it evaluates the alternatives as likely losing (\protect\wmove{Rf1+}: $73.6\%$ loss, \protect\wmove{Rh1}: $63.3\%$ loss), rather than the near-certain losses ($89\%$+) seen in previous puzzles (Table~\ref{tab:puzzle_DsaGi_eval}). This represents the most substantial late-layer reversal  observed ($88\%$ to $12.14\%$) besides the puzzle in Figure~\ref{fig:forgotten_puzzle}, demonstrating that the safety prior against piece sacrifices can override even stronger mid-layer confidence than in previous cases. Full probabilities are in Tables~\ref{tab:puzzle_DsaGi_probs1} and~\ref{tab:puzzle_DsaGi_probs2}.

\begin{figure}[h!]
    \centering
    \includegraphics[width=0.9\textwidth]{Figures/Puzzles/Forgotten/puzzle_visualization_DsaGi.pdf}
    \caption{Layer-wise policy evolution for puzzle \href{https://lichess.org/training/DsaGi}{\texttt{DsaGi}}.}
    \label{fig:puzzle_DsaGi}
\end{figure}

\include{Figures/Puzzles/Forgotten/puzzle_evaluation_DsaGi}

\include{Figures/Puzzles/Forgotten/puzzle_tables_DsaGi}

\clearpage

\subsection{Narrowly forgotten queen sacrifice}

This puzzle (Figure~\ref{fig:puzzle_00aDl}) presents a forced mate-in-two sequence requiring a queen sacrifice (PV: 1. \protect\bmove{Qxb1+} 2. \protect\wmove{Kxb1} \protect\wmove{Ra1\mate}). The winning move \protect\wmove{Qxb1+} leads from layer $8$ through layer $13$, maintaining probabilities between $18.87\%$ and $62.90\%$ from layers $0$--$12$ and rising again to $56.96\%$ in layer $13$. However, in the final layer, the materially conservative pawn advance \protect\wmove{b5} marginally overtakes the solution ($36.12\%$ vs $35.99\%$), despite receiving only $8.85\%$ probability in the preceding layer. This final-layer reversal demonstrates another instance of a safety prior against queen sacrifices overriding established tactical analysis, though the margin is narrower than in previous cases. Notably, the model's value head evaluates the current position as unfavorable ($81.1\%$ loss probability), yet when applied to the position after each legal move, correctly assigns near-certain victory ($99.3\%$ win) to the position after \protect\wmove{Qxb1+} while evaluating all alternatives as losing positions ($99.4\%$--$100\%$ loss) (Table~\ref{tab:puzzle_00aDl_eval}). Full probabilities are in Tables~\ref{tab:puzzle_00aDl_probs1} and~\ref{tab:puzzle_00aDl_probs2}.
\begin{figure}[h!]
    \centering
    \includegraphics[width=0.9\textwidth]{Figures/Puzzles/Forgotten/puzzle_visualization_00aDl.pdf}
    \caption{Layer-wise policy evolution for puzzle \href{https://lichess.org/training/00aDl}{\texttt{00aDl}}.}
    \label{fig:puzzle_00aDl}
\end{figure}

\include{Figures/Puzzles/Forgotten/puzzle_evaluation_00aDl}

\include{Figures/Puzzles/Forgotten/puzzle_tables_00aDl}

\clearpage

\section{Look-ahead analysis details}
\label{app:look_ahead}

This appendix provides full methodological details for the three look-ahead analyses replicated from \citet{jenner2024evidencelearnedlookaheadchessplaying} on forgotten puzzles. Error bars and confidence intervals follow their methodology throughout. Model variant details for look-ahead experiments are provided in Appendix~\ref{app:model_architecture}.

\paragraph{Dataset construction}
We start from the full Lichess puzzle database of approximately $900{,}000$ puzzles and apply the same filtering criteria as \citet{jenner2024evidencelearnedlookaheadchessplaying}: puzzles must have a principal variation of at least three moves, a weaker sparring model must assign less than $5\%$ probability to the first move (ensuring difficulty), and more than $70\%$ to the opponent's response (ensuring the line is forcing). We use the same sparring model as \citet{jenner2024evidencelearnedlookaheadchessplaying}, available in their public inference framework. Mate-in-one puzzles are excluded. We then select forgotten puzzles: positions where the correct first move is the top candidate at some intermediate layer but not at the final layer. To ensure these represent genuine forgetting rather than absence of look-ahead, we additionally require the full model to assign $>50\%$ probability to all subsequent player moves in the principal variation, confirming that the model demonstrably understands the winning continuation and thus plausibly employs look-ahead. This yields $8{,}648$ forgotten puzzles. For comparison, \citet{jenner2024evidencelearnedlookaheadchessplaying} apply the same filtering but select positions the model solves correctly; repeating this selection with the original model and full game history (Appendix~\ref{app:model_architecture}) yields $23{,}601$ positions.  Neither dataset directly filters for the presence of look-ahead mechanisms; any effects observed in the analyses below are emergent rather than selected for.

\paragraph{Corruption generation}
Following \citet{jenner2024evidencelearnedlookaheadchessplaying}, we generate corrupted board states by applying small modifications to the original position, such as adding or removing pawns or moving non-pawn pieces to empty squares. Illegal board states are discarded. Corruptions are selected to substantially affect the full model while minimally affecting the sparring model, thereby targeting mechanisms beyond simple heuristics. Although the model does not select the correct move in forgotten puzzles, it still assigns nonzero probability to it—consistent with the solution being computed but overridden—and log-odds reduction remains a meaningful metric for measuring causal importance. Specifically, a corruption is retained if (1) the log-odds of the correct move under the full model decrease by at least $1.0$, (2) the corruption worsens the position (value change $<-0.1$, where position value is $p_{\text{win}} - p_{\text{loss}}$), (3) the sparring model's log-odds of the correct move decrease by less than $0.2$, and (4) the Jensen-Shannon divergence between the sparring model's clean and corrupted policies is below $0.2$. Among the corruptions that pass, we use the one with the lowest divergence. The log-odds criterion differs from \citet{jenner2024evidencelearnedlookaheadchessplaying}, who require the correct move's probability to drop below $10\%$; since the correct move already has low probability in forgotten puzzles, we use a relative decrease instead. For solved puzzles we keep their probability criterion. Of the $8{,}648$ forgotten puzzles, $7{,}460$ ($86.3\%$) yield valid corruptions; of the $23{,}601$ solved puzzles, $20{,}670$ ($87.6\%$).

Corrupted positions do not arise from an actual game and therefore have no move history. We feed each corrupted position together with the history of its clean counterpart, i.e., only the current position is modified. This makes the input slightly out of distribution, since the last position in the history no longer leads to the current one by a legal move. To check that this does not affect our conclusions, we repeated corruption generation and patching in two alternative settings: (i) filling all history slots of both the clean and the corrupted board with the current position, which is how Leela handles positions given without history, and (ii) using the fine-tuned model that ignores history (Appendix~\ref{app:model_architecture}). Both yield the same qualitative results.

\paragraph{Activation patching}
For each position-corruption pair, we run a clean forward pass and a corrupted forward pass. Token representations from the corrupted pass are patched into the clean pass one square at a time at each layer, and the effect on the log-odds of the correct move is measured. We report the mean effect across positions for four categories of squares: the first move target, the third move target together with the target squares of all later moves in the principal variation, the corrupted square(s), and all remaining squares, taking the largest effect within each category that contains several squares. Following \citet{jenner2024evidencelearnedlookaheadchessplaying}, we exclude puzzles in which these categories overlap, which happens when a corrupted square is the target of a principal-variation move or when the first move's target square is also targeted by a later move. This leaves $5{,}597$ forgotten and $12{,}080$ solved puzzles.

\begin{figure}[h]
\centering
\begin{subfigure}[b]{0.48\textwidth}
    \centering
    \includegraphics[width=\textwidth]{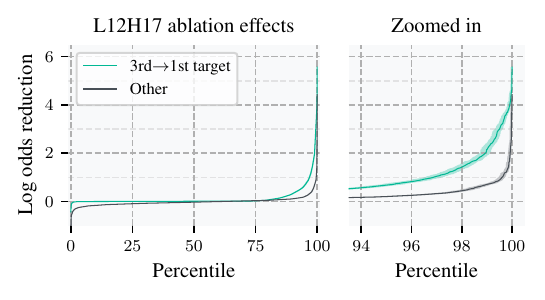}
\end{subfigure}%
\hfill
\begin{subfigure}[b]{0.48\textwidth}
    \centering
    \includegraphics[width=\textwidth]{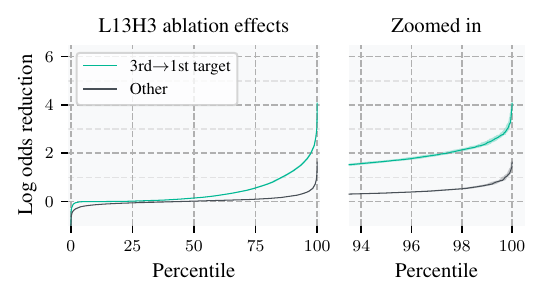}
\end{subfigure}
\begin{subfigure}[b]{0.48\textwidth}
    \centering
    \includegraphics[width=\textwidth]{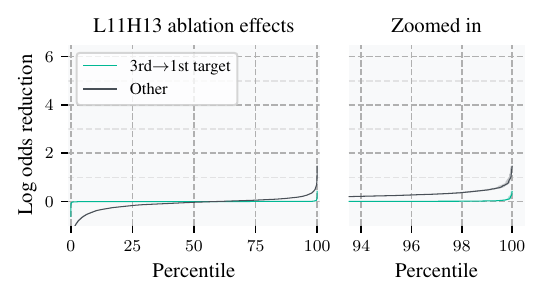}
\end{subfigure}%
\hfill
\begin{subfigure}[b]{0.48\textwidth}
    \centering
    \includegraphics[width=\textwidth]{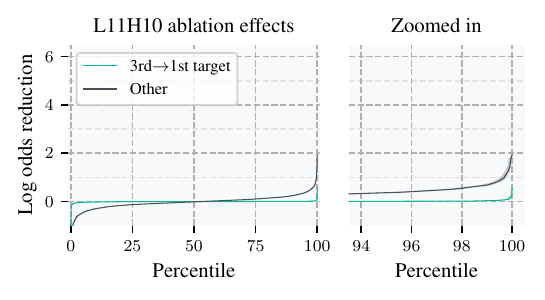}
\end{subfigure}
\caption{Attention head ablation results on forgotten puzzles for four additional look-ahead heads identified by \citet{cruz2025understanding}. Format follows Figure~\ref{fig:ablation_forgotten}: the \firsttarget{single attention entry} between first and third move target squares versus all \othersquare{other} $4{,}095$ entries, shown as percentile curves sorted by effect size.}
\label{fig:head_ablation_additional}
\end{figure}

\paragraph{Attention head ablation}
We replicate the attention ablation from \citet{jenner2024evidencelearnedlookaheadchessplaying}. For each forgotten puzzle, we zero the attention score where the first move's target square (query) attends to the third move's target square (key), preventing information flow between these squares. We measure the log-odds reduction of the correct move. As a baseline, we simultaneously zero all remaining $4{,}095$ attention entries in the same head. Results for L12H12 are presented in the main text (Figure~\ref{fig:ablation_forgotten}). Figure~\ref{fig:head_ablation_additional} shows results for four additional heads identified by \citet{cruz2025understanding}. L12H17 and L13H3 show a similar pattern to L12H12, with the single look-ahead entry producing outsized effects compared to the baseline, while L11H13 and L11H10 show no visible effects on forgotten puzzles.

\paragraph{Presence and necessity of look-ahead}
Neither dataset filters for look-ahead, so both the forgotten and the solved set may contain positions where it plays no role. We count look-ahead as present in a puzzle if patching the \thirdtarget{\third{} move target} square lowers the log-odds of the correct move at the full model by more than the $95$th percentile of the same patch applied to a random square. This holds for $85.9\%$ of the $5{,}597$ forgotten puzzles used for patching. To get a lower bound on how often look-ahead is necessary for the intermediate solution, we cut the L12H12 entry described above and read out the policy at the penultimate layer. Among the $2{,}068$ puzzles in which look-ahead is present and the correct move is on top at the penultimate layer, cutting this entry removes the correct move from the top in $29.9\%$ of cases, against $9.8\%$ when all other $4{,}095$ entries are cut and $5.8\%$ when only the most-attended other entry is cut, and lowers its log-odds by more than $1$ in $14.7\%$ of cases, against $1.6\%$ and $1.4\%$. The controls flip the top move this often because the correct move is often only narrowly on top at this layer, with a median probability of $0.39$. The corresponding entry in L13H3 partly backs up L12H12: cutting both removes the correct move from the top in $44.4\%$ of cases and lowers its log-odds by more than $1$ in $26.6\%$ of cases, against $7.5\%$ and $1.7\%$ when each head's most-attended other entry is cut. Without the presence condition, among all $3{,}781$ forgotten puzzles with the correct move on top at the penultimate layer, cutting both entries removes it from the top in $32.2\%$ of cases, against $8.4\%$ for the control, and lowers its log-odds by more than $1$ in $18.4\%$ of cases, against $1.9\%$. Since look-ahead information can also reach the \firsttarget{\first{} move target} square through other paths, these rates are lower bounds on how often look-ahead is necessary for the intermediate solution.

\paragraph{Probing}
We train bilinear probes following the two-step architecture of \citet{jenner2024evidencelearnedlookaheadchessplaying}: first predicting the target square of the third move from activations at the first move's target square, then predicting the source square conditioned on the predicted target. On forgotten puzzles, we condition on the ground-truth first move target square rather than the model's predicted move, since by definition the model does not select the correct first move. Probes are trained on a $70/30$ train/test split with five independent training runs. As a baseline, following \citet{jenner2024evidencelearnedlookaheadchessplaying}, we train probes with the same architecture on a randomly initialized model.

\begin{figure}[h]
\centering
\begin{subfigure}[b]{\textwidth}
    \centering
    \includegraphics[width=\textwidth]{Figures/solved_act_patching.pdf}
    \vspace{-20pt}
    \caption{Activation patching}
\end{subfigure}

\begin{subfigure}[b]{0.58\textwidth}
    \centering
    \includegraphics[height=4.5cm]{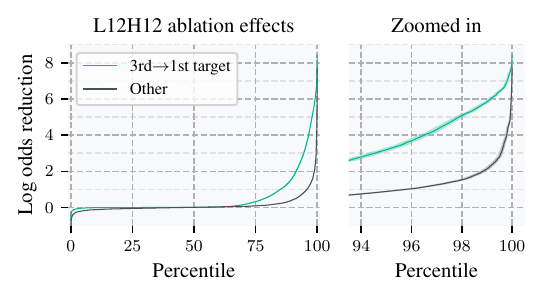}
    \vspace{-15pt}
    \caption{Attention head ablation}
\end{subfigure}%
\hfill
\begin{subfigure}[b]{0.42\textwidth}
    \centering
    \includegraphics[height=4.5cm]{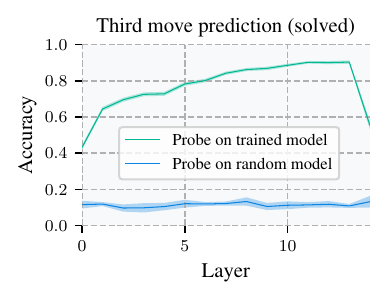}
    \vspace{-15pt}
    \caption{Probing}
\end{subfigure}

\caption{Look-ahead analyses following \citet{jenner2024evidencelearnedlookaheadchessplaying} on our solved puzzle set, for direct comparison with the forgotten puzzle results in Figure~\ref{fig:look_ahead_forgotten}.}

\label{fig:look_ahead_solved}
\end{figure}

\paragraph{Comparison with solved puzzles}
For reference, Figure~\ref{fig:look_ahead_solved} shows the same analyses, following \citet{jenner2024evidencelearnedlookaheadchessplaying}, on our solved puzzle set, enabling direct visual comparison with the forgotten puzzle results in Figure~\ref{fig:look_ahead_forgotten}.

\section{Value head evaluation}
\label{app:value_head}

The policy and value heads share the transformer encoder and are trained on the same self-play games: the policy head to predict the search outcome and the value head to predict the game outcome (Appendix~\ref{app:search_mechanics}). The value head therefore approximates the value of a position under that kind of play rather than an objective assessment of it, so the two heads should behave consistently. Yet Section~\ref{sec:limits-look-ahead} reports that the value head prefers the position after the solution move in $97.0\%$ of forgotten puzzles, which raises two questions. First, does the value head see the win at the puzzle position itself, where the policy misses it? Second, if it does not, why does it still assign some probability to a win on forgotten puzzles, and to a loss on solved ones? We answer the first by evaluating the value head at the puzzle position, and the second with the Bellman equation, which relates the value of a position to the moves the policy plays from it.

\paragraph{At the puzzle position}
The comparison in the introduction of Section~\ref{sec:limits-look-ahead} is between the positions that result from playing a move, so there is no same-input contradiction between the two heads. At the puzzle position itself, the value head does not see a win in the forgotten positions: a draw is the most likely single outcome, and loss is more likely than win (Table~\ref{tab:value_wdl}). The policy and the value head agree here. The value head evaluates forgotten positions markedly lower than solved ones, with a mean win probability of $21.6\%$ against $63.0\%$. What it evaluates accurately is the position after the move has been played, where the winning line is usually clear: after the solution move, its mean win probability on forgotten puzzles is $75.6\%$. Search reaches the solution only because this position is evaluated correctly; if it were not, the search would abandon that branch (Appendix~\ref{app:search_experiments}).

\begin{table}[h!]
\centering
\caption{Mean value-head prediction at the puzzle position (side to move) and the win probability after the solution move, from the mover's side.}
\label{tab:value_wdl}
\begin{tabular}{lrrrrr}
\toprule
 & & \multicolumn{3}{c}{\textbf{Puzzle position}} & \textbf{After solution move} \\
\cmidrule(lr){3-5}\cmidrule(lr){6-6}
\textbf{Set} & $n$ & \textbf{Win} & \textbf{Draw} & \textbf{Loss} & \textbf{Win} \\
\midrule
Forgotten puzzles & $8{,}648$ & $21.6\%$ & $42.1\%$ & $36.3\%$ & $75.6\%$ \\
Solved puzzles & $23{,}601$ & $63.0\%$ & $21.6\%$ & $15.4\%$ & $88.6\%$ \\
\bottomrule
\end{tabular}
\end{table}

\paragraph{Bellman consistency}
For any policy $\pi$, the Bellman equation \citep{sutton2018reinforcement} relates its value function at a position to the values of the positions its own moves lead to, ${V^\pi(s) = \sum_m \pi(m \mid s) \cdot V^\pi(s \to m)}$, with win and loss swapped to the perspective of the player at $s$. The value of a position is the mean of the values of the resulting positions, weighted by the probability the policy assigns to the moves leading to them. We write $\hat V(s)$ for the right-hand side, the model's own one-ply expansion, and compare it with the value head's direct reading at the puzzle position, $V(s)$. This is a theoretical property of the value function rather than something the loss explicitly imposes, so measuring it checks how well the head has learned the value of its own policy. It holds in aggregate (Table~\ref{tab:value_bellman}). The two components in question match most closely: the win probability on forgotten puzzles, $21.6\%$ against $23.0\%$, and the loss probability on solved puzzles, $15.4\%$ against $16.0\%$. Both therefore come from the moves the policy plays. Residual mass on the correct move raises the value of a forgotten puzzle without making it solved, and mass on losing moves lowers the value of a solved one without making it unsolved. The value head reflects the whole distribution; solving depends on the argmax alone. One systematic deviation remains: the direct evaluation is slightly less resolved than the same evaluation one move later, and puts more mass on a draw than its own expansion, by $6.2$ percentage points on forgotten and $8.0$ on solved puzzles.

\begin{table}[h!]
\centering
\caption{The value head's direct prediction at the puzzle position, $V(s)$, and its one-ply expansion under the model's own policy, $\hat V(s)$, averaged over puzzles. Bold marks the components discussed in the text.}
\label{tab:value_bellman}
\begin{tabular}{llrrr}
\toprule
\textbf{Set} & & \textbf{Win} & \textbf{Draw} & \textbf{Loss} \\
\midrule
Forgotten puzzles ($n = 8{,}648$) & $V(s)$ & $\mathbf{21.6\%}$ & $42.1\%$ & $36.3\%$ \\
 & $\hat V(s)$ & $\mathbf{23.0\%}$ & $36.0\%$ & $41.1\%$ \\
\midrule
Solved puzzles ($n = 23{,}601$) & $V(s)$ & $63.0\%$ & $21.6\%$ & $\mathbf{15.4\%}$ \\
 & $\hat V(s)$ & $70.4\%$ & $13.6\%$ & $\mathbf{16.0\%}$ \\
\bottomrule
\end{tabular}
\end{table}

\paragraph{Where the win mass comes from}
Table~\ref{tab:value_decomposition} decomposes the win component of the expansion on forgotten puzzles by what Stockfish says each move leads to. More than half of the win mass, $56\%$, is the term for the correct move, and most of the rest comes from drawing and near-drawing alternatives to which the value head still assigns some win probability.

\begin{table}[h!]
\centering
\caption{Win component of the one-ply expansion on forgotten puzzles, split by the Stockfish 17.1 evaluation of each legal move at depth 12, in pawns from the mover's side. Policy mass and contribution are means over puzzles; the win probability after the move is the policy-weighted mean within each band, so that policy mass times win probability gives the contribution.}
\label{tab:value_decomposition}
\resizebox{\textwidth}{!}{%
\begin{tabular}{lrrrr}
\toprule
\textbf{Moves} & \textbf{Policy mass} & \textbf{Win after move} & \textbf{Contribution} & \textbf{Share} \\
\midrule
The solution move & $16.9\%$ & $75.7\%$ & $12.8\%$ & $56\%$ \\
Drawing ($-0.5$ to $+0.5$) & $28.3\%$ & $15.3\%$ & $4.3\%$ & $19\%$ \\
Slight edge for the player ($+0.5$ to $+2$) & $8.4\%$ & $36.4\%$ & $3.1\%$ & $13\%$ \\
Slight edge for the opponent ($-2$ to $-0.5$) & $18.7\%$ & $8.4\%$ & $1.6\%$ & $7\%$ \\
Losing (below $-2$) & $27.0\%$ & $3.3\%$ & $0.9\%$ & $4\%$ \\
Another winning move (above $+2$ or mate) & $0.7\%$ & $47.0\%$ & $0.3\%$ & $1\%$ \\
\midrule
Total & $100.0\%$ & & $23.0\%$ & $100\%$ \\
\bottomrule
\end{tabular}%
}
\end{table}

\clearpage

\section{Stockfish concept preference analysis}
\label{app:concept_analysis}

We analyze how different network layers prioritize chess concepts by measuring the expected change in concept values induced by each layer's move distribution. Unlike \citet{McGrath_2022}, who probed for concept \emph{representations} using linear classifiers on intermediate activations, we directly measure concept \emph{preferences} from policy outputs. This approach reveals what concepts each layer prioritizes when selecting moves, complementing representation-based analyses.

\paragraph{Chess concepts}

We use Stockfish 8's \citep{stockfish8} handcrafted evaluation function as our source of chess concepts, following \citet{McGrath_2022} to enable direct comparison with prior work from \citet{sadler2019gamechanger}.  Stockfish decomposes position evaluation into interpretable components including material balance, piece-specific features, king safety, threats, mobility, passed pawns, and spatial control. Each concept is evaluated separately for the midgame (mg) and the endgame (eg). A phase-interpolated value (ph) combines the two as a weighted sum based on the game phase. All concepts are represented as continuous values. Table~\ref{tab:concepts} summarizes the main concept categories we analyze.

\begin{table}[h]
\centering
\caption{Summary of chess concepts from Stockfish 8's evaluation function taken from \citet{McGrath_2022}. Concepts are enumerated as \texttt{<concept\_name> <side> <game\_phase>}, where side is \texttt{[mine|opponent|t]} for current player, opponent, or total (difference), and game phase is \texttt{[mg|eg|ph]} for midgame, endgame, or phase-interpolated value.}
\label{tab:concepts}
\resizebox{0.82\textwidth}{!}{%
\begin{tabular}{p{3cm}|p{10cm}}
\toprule
\textbf{Concept} & \textbf{Description} \\
\midrule
\texttt{material t [mg|eg|ph]} & Material score, where each piece on the board has a predefined value that changes depending on the phase of the game. \\
\midrule
\texttt{imbalance t [mg|eg|ph]} & Imbalance score that compares the piece count of each piece type for both colors. E.g., it awards having a pair of bishops vs a bishop and a knight. \\
\midrule
\texttt{pawns t [mg|eg|ph]} & Evaluation of the pawn structure. E.g., the evaluation considers isolated, doubled, connected, backward, blocked, weak, etc. pawns. \\
\midrule
\texttt{knights [mine|opponent|t] [mg|eg|ph]} & Evaluation of knights. E.g., extra points are given to knights that occupy outposts protected by pawns. \\
\midrule
\texttt{bishops [mine|opponent|t] [mg|eg|ph]} & Evaluation of bishops. E.g., bishops that occupy the same color squares as pawns are penalized. \\
\midrule
\texttt{rooks [mine|opponent|t] [mg|eg|ph]} & Evaluation of rooks. E.g., rooks that occupy open or semi-open files have higher valuation.  \\
\midrule
\texttt{queens [mine|opponent|t] [mg|eg|ph]} &  Evaluation of queens. E.g., queens that have relative pin or discovered attack against them are penalized. \\
\midrule
\texttt{mobility [mine|opponent|t] [mg|eg|ph]} & Evaluation of piece mobility score. It depends on the number of squares attacked by the pieces. \\
\midrule
\texttt{king safety [mine|opponent|t] [mg|eg|ph]} & A complex concept related to king safety. It depends on the number and type of pieces that attack squares around the king, shelter strength, number of pawns around the king, penalties for being on a pawnless flank, etc. \\
\midrule
\texttt{threats [mine|opponent|t] [mg|eg|ph]} & Evaluation of threats to pieces, such as whether a pawn can safely advance and attack an opponent's higher value piece, hanging pieces, possible x-ray attacks by rooks, etc. \\
\midrule
\texttt{passed pawns [mine|opponent|t] [mg|eg|ph]} & Evaluates bonuses for passed pawns. The closer a pawn is to the promotion rank, the higher the bonus. \\
\midrule
\texttt{space [mine|opponent|t] [mg|eg|ph]} &Evaluation of the space. It depends on the number of safe squares available for minor pieces on the central four files on ranks 2 to 4. \\
\midrule
\texttt{total t [mg|eg|ph]} & The total evaluation of a given position. It encapsulates all the above concepts. \\
\bottomrule
\end{tabular}
}
\end{table}

\paragraph{Stockfish version}
Stockfish evaluated positions with a handcrafted function from its first release in 2008 until version 12 in 2020, which made Efficiently Updatable Neural Networks, NNUE \citep{nasu2018nnue}, the main evaluation \citep{stockfish2020nnue}. NNUE is a small neural network that runs on a CPU fast enough to evaluate every position in Stockfish's alpha-beta search, whereas Leela's much larger network runs on a GPU and evaluates far fewer positions in its Monte Carlo tree search. The handcrafted evaluation remained in the engine for some time after that until it was removed entirely in 2023, so no newer interpretable evaluation exists. Stockfish 8, released in 2016, was AlphaZero's opponent in its evaluation matches \citep{silver2018general}, and \citet{sadler2019gamechanger} study AlphaZero's play using games from these matches in a book widely read in the chess community. \citet{McGrath_2022} use Stockfish 8's terms for this reason, and following them keeps our concept measurements comparable with theirs and legible to the chess community. We use the terms as interpretable functions of the position rather than as an evaluation oracle, so the engine's playing strength does not matter here.

\paragraph{Perspective normalization}
Stockfish evaluates all positions from White's perspective, with positive values favoring White. To ensure consistency with Leela's current-player perspective, we convert Stockfish's evaluations to player-relative coordinates following \citet{McGrath_2022}. For concepts with side-specific values (knights, bishops, rooks, queens, mobility, king safety, threats, passed pawns, space), we relabel White/Black as Mine/Opponent based on the side to move. \textbf{When White is to move}, White concepts become Mine and Black concepts become Opponent, with values unchanged. \textbf{When Black is to move}, Black concepts become Mine and White concepts become Opponent, with all values negated to reflect the perspective change. For aggregate concepts (material, imbalance, pawns, total), we negate values when Black is to move. This ensures that positive concept values always represent advantages for the player to move.

\paragraph{Concept delta calculation}

For each position $s$ with player $p$ to move, we compute concept preferences as follows:

\begin{enumerate}
    \item \textbf{Evaluate initial position}: Use Stockfish 8 to obtain concept values $c(s)$ for all concepts $c$, normalized to player $p$'s perspective.
    
    \item \textbf{Generate and evaluate legal moves}: For each legal move $m$ leading to position $s'$, evaluate $c(s')$ using Stockfish and normalize to player $p$'s perspective (accounting for the perspective flip after the move).
    
    \item \textbf{Calculate concept deltas}: For each move-concept pair, compute:
    \begin{equation}
    \Delta c_m = c(s') - c(s)
    \end{equation}
    where both $c(s')$ and $c(s)$ are evaluated from player $p$'s perspective, ensuring $\Delta c_m$ consistently represents concept change from the moving player's viewpoint.
    
    \item \textbf{Compute layer-wise preferences}: For each layer $\ell$, obtain the move probability distribution $\pi_\ell$ using the logit lens and calculate the expected concept delta:
    \begin{equation}
    \Delta c_\ell = \sum_{m \in \text{legal}(s)} \pi_\ell(m) \cdot \Delta c_m = \mathbb{E}_{\pi_\ell}[\Delta c_m]
    \end{equation}
\end{enumerate}

This weighted average represents the expected change in concept $c$ when sampling moves according to layer $\ell$'s policy.

\paragraph{Dataset and sampling}

We sample $10{,}000$ positions from the CCRL dataset \citep{ccrl_dataset}, a standard benchmark consisting of $2.5$ million computer games from CCRL $40/40$ and $40/4$ tournaments. The dataset contains games between strong chess engines and provides diverse positions across different game phases and strategic themes.

Positions are sampled using a two-stage hierarchical process until $10{,}000$ unique positions are collected:
\begin{enumerate}
    \item Sample 10\% of games uniformly from the dataset
    \item From each selected game, sample 10\% of positions uniformly
    \item Filter duplicate positions to ensure uniqueness
\end{enumerate}

\paragraph{Implementation details}

For each sampled position, we:
\begin{itemize}
    \item Extract move policies $\pi_\ell$ for the input embedding layer plus all $15$ transformer layers, where the last layer corresponds to the final model output
    \item Evaluate the initial position and all legal move outcomes (typically $30$--$40$ moves per position) using Stockfish 8
    \item Calculate concept deltas and probability-weighted averages for all $93$ Stockfish concepts across midgame, endgame, and phase-interpolated variants
    \item Store results for analysis and visualization
\end{itemize}

Figures in the main paper present phase-interpolated (ph) concept values, as these represent Stockfish's actual evaluation for a given position. Comprehensive results showing all concept variants (mg, eg, ph) across all layers are provided in this appendix (Figures~\ref{fig:concepts-part1}--\ref{fig:concepts-part7}).

\paragraph{Statistical analysis}

We report mean $\Delta c_\ell$ values across all $10{,}000$ positions with $95\%$ confidence intervals computed via the $t$-distribution. Many concepts have values of zero—and therefore zero deltas—in positions lacking relevant pieces (e.g., bishop concepts when no bishops are present, or passed-pawn bonuses without passed pawns) or where the concept is undefined. The endgame variants of the queen and space terms are zero in every position, so their concept preference is zero at every layer and they yield no steering direction. This produces zero-inflated distributions that bias means toward zero. We do not control for this effect, which makes absolute $\Delta c_\ell$ values across concepts skewed and not directly comparable. Nevertheless, relative trends across layers remain informative for understanding how concept preferences evolve with depth.

\paragraph{Interpretation of concept evolution}

Analysis of concept preferences across layers reveals three distinct patterns corresponding to the computational phases identified in playing strength progression. In the early phase (layers \texttt{Input} to $5$), most concepts exhibit erratic, volatile shifts with substantial fluctuations in preference values. During the middle phase (layers $6$ to $10$), concept preferences stabilize significantly, with most concepts maintaining relatively constant values across this range, mirroring the performance plateau observed in tournament play and puzzle solving. In the late phase (layers $11$ to \texttt{Final}), concept preferences show consistent, smooth trends—typically monotonic increases or decreases for each concept, with minimal erratic behavior. This systematic evolution coincides with the sharp capability improvements and emergence of look-ahead mechanisms in late layers. The shift from aggressive to safety-oriented concepts occurs primarily in this phase: the model increasingly prefers moves that raise \kingsafetymine{own king safety} and reduce \threatsopp{opponent threats} (prioritizing defensive considerations), while its preference for moves that raise \kingsafetyopp{opponent king vulnerability} and \threatsmine{own threats} decreases (de-emphasizing aggressive play). These patterns hold across most concepts, though some show different dynamics requiring deeper chess expertise to interpret. The consistency of trends in middle and late phases is particularly striking given the volatility of early layers, suggesting fundamentally different computational regimes across network depth.

\input{Figures/concepts_individual/concepts_figures.tex}

\clearpage

\section{Concept steering details}
\label{app:steering}

This appendix provides complete details for the gradient-based concept steering experiments described in the main text. We group threats and king safety as safety-related concepts based on their intuitive connection to conservative play; this categorization is ours rather than intrinsic to Stockfish's evaluation terms. The consistency of the effect across four independently defined safety-related concepts suggests the result reflects a genuine property of the model rather than an artifact of any single proxy. All steering experiments use the original model with position history.

\paragraph{Gradient computation}
For each concept $c$ and layer $\ell$, we compute the gradient of the concept preference of the complete model $\Delta c_L$ (Equation~\ref{eq:concept_preference}) with respect to the residual stream activations after layer $\ell$:
\begin{equation}
\hat{\mathbf{v}}_{c,\ell} = \frac{\mathbb{E}[\nabla_{\mathbf{h}_\ell} \Delta c_L]}{\|\mathbb{E}[\nabla_{\mathbf{h}_\ell} \Delta c_L]\|}
\end{equation}
Gradients are computed over $10{,}000$ positions sampled from the CCRL~\citep{ccrl_dataset} dataset using the same hierarchical sampling procedure described in Appendix~\ref{app:concept_analysis}. Gradients are averaged across all $64$ squares and positions to obtain a single $768$-dimensional steering vector $\hat{\mathbf{v}}_{c,\ell}$ per concept per layer based on $640{,}000$ activations.

\paragraph{Steering strength}
The steering strengths $\alpha \in \{\pm 0.25, \pm 0.5, \ldots, \pm 1.5\}$ reported in the main text are scaled by the per-layer mean activation norm $\mathbb{E}[\|\mathbf{h}_\ell\|_2]$, computed over the CCRL gradient positions, to ensure comparable intervention magnitude across layers, which have different activation scales. The applied perturbation is:
\begin{equation}
\mathbf{h}_\ell' = \mathbf{h}_\ell + \alpha \cdot \mathbb{E}[\|\mathbf{h}_\ell\|_2] \cdot \hat{\mathbf{v}}_{c,\ell}
\end{equation}
where $\hat{\mathbf{v}}_{c,\ell}$ is the normalized gradient direction. Positive $\alpha$ steers toward increasing the concept preference; negative $\alpha$ steers toward decreasing it. The steering vector is applied uniformly to all $64$ square positions. We sweep over all strengths and all layers, evaluating each configuration on the full $10{,}000$ general puzzle dataset using full principal variation correctness.

\paragraph{Held-out evaluation}
Selecting a setting and reporting its solve rate on the same puzzles overstates the effect of steering, so we use Monte Carlo cross-validation. We split the $10{,}000$ puzzles randomly into two halves, select the setting on one half and measure the solve rate on the other, and repeat this for $200$ splits with a fixed seed. For the per-layer curves in Figure~\ref{fig:concept-steering} and Figures~\ref{fig:steering-part1}--\ref{fig:steering-part7}, the strength is selected separately for each concept and layer. Lines show the mean over splits and shaded areas one standard deviation, which describes the variation across splits rather than a confidence interval.

The highest curve in Figure~\ref{fig:concept-steering}, \threatsopp{opponent threats} at layer~$13$, reaches a held-out solve rate of $92.43 \pm 0.25\%$, the value reported in the main text. Picking the highest curve is itself a selection over concepts and layers, so we also select concept, layer and strength jointly by cross-validation, among all concept terms. This gives nearly the same held-out solve rate of $92.31 \pm 0.29\%$, with both values given as mean and standard deviation over splits. In every split, the selection lands on a safety-related concept, and on \threatsopp{opponent threats} at layer~$13$ in $180$ of the $200$ splits.

\paragraph{Fixed setting and transfer}
The counts of gained and lost puzzles in the main text and the transfer test use one fixed setting, \threatsopp{opponent threats} at layer~$13$ with $\alpha = -0.5$, the best strength on all $10{,}000$ puzzles. It reaches $92.46\%$ there, only $0.03$ points above its held-out value. Because this setting is selected on the general dataset only, the forgotten dataset, which played no part in the selection, serves as a held-out test set. After removing the $53$ puzzles it shares with the general dataset, steering recovers $71.7\%$ of the remaining $8{,}595$ puzzles.

\paragraph{Random baseline}
To establish that the steering effects are concept-specific rather than artifacts of generic activation perturbation, we compare against $50$ random steering vectors drawn from an isotropic Gaussian distribution, each normalized to unit length. Each random vector is evaluated like a concept, with the strength selected per layer by the same cross-validation. The \randomdirection{random baseline} shows the mean held-out solve rate across the $50$ vectors, with $95\%$ confidence intervals across vectors.

\paragraph{Per-concept steering results}
Figures~\ref{fig:steering-part1}--\ref{fig:steering-part7} show the held-out steering results for all Stockfish concepts. Each plot displays the held-out solve rate across layers with one standard deviation over splits, with point color indicating the most frequently selected strength. The \forgotten{unsteered baseline} and \randomdirection{random baseline} are shown for reference. Three of the four safety-related concepts---\threatsopp{opponent threats}, \threatsmine{own threats} and \kingsafetyopp{opponent king vulnerability}---consistently produce the largest improvements, reaching $91.8$ to $92.4\%$ at their best layers, while \kingsafetymine{own king safety} improves the solve rate only in the last layers, reaching $90.9\%$ at layer $13$. While other concepts also improve performance above the random baseline, likely reflecting correlations between piece-specific concepts and tactical move types common in puzzles, the effects are smaller and less consistent across layers. Steering in middle layers is typically least effective, again mirroring the computational phases identified in the main text. The most directly safety-related concepts yield the largest effects, supporting the interpretation that learned safety priors are a primary cause of the forgotten puzzle phenomenon.

\paragraph{Concepts versus move types}
Safe play, as we use the term, includes preferences over move types such as avoiding sacrifices, preserving material and preferring quiet moves, so safety-related concepts are expected to correlate with move types. A stronger concern is that a concept merely marks specific move types, such as captures or checks, so that steering it amounts to a preference for these moves regardless of their underlying risk. In $98.3\%$ of forgotten puzzles, the first move of the solution is a capture, check or sacrifice, while the model replaces it with a quiet move in $68.3\%$. For each concept and each of these four move types, we measure as an AUC how well the concept's change $\Delta c$ separates moves of that type from the other candidates in the same position, namely the solution's first move and the model's eight most probable moves, and take the largest deviation from $0.5$ as the concept's encoding strength. As a positive control, material separates captures almost perfectly. We use each of the $31$ concepts in its phase-interpolated form, so that each concept counts once. Consistent with this expectation, stronger move-type encoding goes with moderately stronger steering, with Spearman correlations of $0.45$ for forgotten recovery at layer~$13$ and $0.37$ for the peak held-out solve rate. Move-type encoding alone does not explain the effect, however. To test whether the safety-related concepts steer better than their move-type encoding predicts, we fit a linear relation between encoding strength and steering efficacy across the $27$ other concepts. \threatsopp{Opponent threats}, \kingsafetyopp{opponent king vulnerability} and \threatsmine{own threats} steer better than this fit predicts by $1.7$ to $3.2$ times the typical deviation of the other concepts from it. \threatsopp{Opponent threats} is the largest outlier of all $31$ concepts, while \kingsafetymine{own king safety}, which is effective only in the last layers, lies only slightly above the fit.

\paragraph{Early-layer steering}
Steering toward aggression in early layers is nearly as effective as the best late setting. On all puzzles, steering \kingsafetyopp{opponent king vulnerability} at layer~$2$ reaches $92.1\%$, compared with $92.5\%$ for \threatsopp{opponent threats} at layer~$13$, but it also loses more previously solved puzzles, $186$ against $129$. To see how an intervention this early recovers forgotten puzzles, we read out the probability of the correct move with the logit lens at every layer after the steering point. For the $240$ forgotten puzzles that steering at layer~$2$ recovers, the trajectory stays close to the unsteered one through the intermediate layers, and about three quarters of the gain in log-odds of the correct move arises in the final layer, which lowers the correct move in only $32\%$ of these puzzles with steering. Early steering therefore acts mainly by weakening the final-layer override rather than by producing a stronger solution in earlier layers.

\input{Figures/steering/steering_figures}

\clearpage

\section{Search experiments}
\label{app:search_experiments}

\paragraph{Setup}
We run both experiments of Section~\ref{sec:search-experiments} on the $8{,}648$ forgotten and $23{,}601$ solved puzzles, starting each search from the puzzle position with the full move history of its game, as in all other analyses, and scoring only the first move of the solution. We compare two networks: the final-layer network is the unmodified network analyzed throughout the paper, and the penultimate-layer network is identical except that its policy head reads the logit lens at the penultimate layer, while its value and moves-left heads still use the full model. All searches run in Leela's engine \citep{leela}, version 0.33 at commit d8ce482, with the final-layer network on the CUDA backend and the penultimate-layer network on the ONNX backend. To check that the engine uses the policy we analyze, we compare its prior at a single node, with temperature one and smart pruning off, against our own implementation. The top move agrees in at least $99.58\%$ of puzzles for each network and dataset, and move probabilities differ by at most $0.011$. The remaining differences are numerical, since the engine's two backends and our implementation compute the same network with different kernels. Every disagreement on the top move is a near-tie, in which the two most likely moves differ by at most $0.0034$ in our implementation.

\paragraph{Training target}
For the training target, we run the self-play search described in Appendix~\ref{app:search_mechanics} with the settings of the Leela run that generated the network's training data, listed in Table~\ref{tab:search_settings}, to obtain the target the training procedure would assign to each puzzle. Each puzzle is searched once until the KL-divergence stop fires, which happens after a median of $861$ visits on forgotten and $809$ on solved puzzles, and always before the cap of $10{,}000$. Since the engine has no seed for the root noise, each puzzle gets a single noise draw. Because unvisited root moves enter the selection rule with the value of a certain win, every root move is evaluated at least once in practice.

\paragraph{Deployment}
For deployment, we use the engine's default settings and limit each search to $1$, $8$, $32$, $128$, $800$ or $2{,}048$ nodes, running each budget separately. Smart pruning is on, as in normal use, so a search can stop before reaching its budget. Here we count a puzzle as solved if the engine plays the first move of the solution.

\begin{table}[h]
\centering
\caption{Search settings for the training target, taken from the self-play run that generated the network's training data, and for deployment, which uses the engine's defaults. In both profiles, $c_{\text{puct}}$ grows with the visit count with the engine's default factor $3.894$ and base $38{,}739$, and a twofold repetition counts as a draw within the search.}
\label{tab:search_settings}
\resizebox{\textwidth}{!}{%

\begin{tabular}{p{4.0cm}p{4.9cm}p{3.8cm}}
\toprule
Setting & Training target & Deployment \\
\midrule
Policy temperature & $1.45$ & $1.359$ \\
$c_{\text{puct}}$, tree / root & $1.32$ / $2.0$ & $1.745$ / $1.745$ \\
First-play urgency, tree / root & reduction $0.26$ / absolute $1.0$ & reduction $0.33$ / reduction $0.33$ \\
Moves-left utility & max.\ effect $0.2$, threshold $0$, slope $0.007$, quadratic $0.85$, scaled $0.15$, constant $0$ & max.\ effect $0.0345$, threshold $0.8$, slope $0.0027$, quadratic $-0.6521$, scaled $1.6521$, constant $0$ \\
Root noise & Dirichlet, $\varepsilon = 0.1$, $\alpha = 0.12$ & none \\
Smart pruning & off & factor $1.33$ \\
Stopping & KL gain below $5 \cdot 10^{-5}$ per visit, checked every $100$ visits, at most $10{,}000$ visits & budget of $1$ to $2{,}048$ nodes \\
\bottomrule
\end{tabular}
}
\end{table}

\newpage

\paragraph{Results}
Table~\ref{tab:search_target} gives the training target and Table~\ref{tab:search_deployment} the deployment solve rates on both datasets. Stockfish confirms that the solution wins in all $49$ forgotten puzzles, $0.6\%$, in which the training target misses it. In most of them, the value head rates the position after the solution as losing, so the search spends few visits on it. Even at $2{,}048$ nodes, the penultimate-layer network solves slightly more forgotten puzzles, $99.3\%$ against $99.1\%$, and only $38$ forgotten puzzles are never solved by it at any budget, against $71$ for the final-layer network. On solved puzzles, the penultimate-layer network catches up quickly: from $76.7\%$ at a single node, it reaches $98.6\%$ at $8$ nodes and $99.6\%$ at $32$, so the final layer's advantage there is limited to the smallest budgets.

\begin{table}[h]
\centering
\caption{Training target on forgotten and solved puzzles, compared with the policies read from the final and the penultimate layer.}
\label{tab:search_target}
\begin{tabular}{lrr}
\toprule
 & Forgotten & Solved \\
\midrule
Puzzles & $8{,}648$ & $23{,}601$ \\
Visits when the search stops, median & $861$ & $809$ \\
Target probability on the solution, mean & $0.931$ & $0.953$ \\
Target's top move is the solution & $99.4\%$ & $100.0\%$ \\
Solution has the highest value at the root & $99.5\%$ & $>99.99\%$ \\
Final-layer probability on the solution, median & $0.159$ & $0.795$ \\
Penultimate-layer probability on the solution, median & $0.262$ & $0.469$ \\
Penultimate-layer top move agrees with target & $43.9\%$ & $76.7\%$ \\
Cross-entropy to target, final layer (nats) & $2.223$ & $0.523$ \\
Cross-entropy to target, penultimate layer (nats) & $1.716$ & $1.026$ \\
Penultimate layer closer to target than final layer & $77.5\%$ & $7.3\%$ \\
\bottomrule
\end{tabular}
\end{table}

\begin{table}[h]
\centering
\caption{Share of puzzles in which the engine plays the solution's first move, by search budget in nodes, with engine defaults.}
\label{tab:search_deployment}
\resizebox{\textwidth}{!}{%

\begin{tabular}{llrrrrrr}
\toprule
Puzzles & Network & $1$ & $8$ & $32$ & $128$ & $800$ & $2{,}048$ \\
\midrule
Forgotten & Final layer & $0.0\%$ & $53.7\%$ & $78.3\%$ & $92.0\%$ & $97.8\%$ & $99.1\%$ \\
 & Penultimate layer & $43.7\%$ & $75.3\%$ & $86.4\%$ & $94.8\%$ & $98.6\%$ & $99.3\%$ \\
\addlinespace
Solved & Final layer & $100.0\%$ & $99.98\%$ & $>99.99\%$ & $>99.99\%$ & $100.0\%$ & $100.0\%$ \\
 & Penultimate layer & $76.7\%$ & $98.6\%$ & $99.6\%$ & $99.9\%$ & $99.97\%$ & $99.99\%$ \\
\bottomrule
\end{tabular}%
}
\end{table}

\clearpage

\end{document}

%% file: Figures/Puzzles/Examples/puzzle_tables_62FeU.tex
\begin{table*}[h!]
\centering
\caption{Move probabilities by layer for puzzle \href{https://lichess.org/training/62FeU}{\texttt{62FeU}} (Part 1: Input to Layer 6)}
\label{tab:puzzle_62FeU_probs1}
\resizebox{\textwidth}{!}{%
\begin{tabular}{lrrrrrrrr}
\toprule
\textbf{Move} & \textbf{Input} & \textbf{Layer 0} & \textbf{Layer 1} & \textbf{Layer 2} & \textbf{Layer 3} & \textbf{Layer 4} & \textbf{Layer 5} & \textbf{Layer 6} \\
\midrule
\wmove{Ng3+} & \cellcolor{heatmapcolor!7!white}6.54\% & 0.09\% & 0.14\% & \cellcolor{heatmapcolor!2!white}1.79\% & \cellcolor{heatmapcolor!8!white}7.59\% & 0.41\% & \cellcolor{heatmapcolor!15!white}15.23\% & \cellcolor{heatmapcolor!27!white}26.71\% \\
\wmove{Rh6} & 0.01\% & 0.03\% & 0.04\% & 0.13\% & 0.12\% & 0.15\% & 0.02\% & \cellcolor{heatmapcolor!3!white}2.56\% \\
\wmove{Qg1+} & \cellcolor{heatmapcolor!9!white}9.16\% & \cellcolor{heatmapcolor!1!white}1.11\% & \cellcolor{heatmapcolor!4!white}4.36\% & \cellcolor{heatmapcolor!8!white}7.52\% & \cellcolor{heatmapcolor!7!white}6.52\% & 0.43\% & \cellcolor{heatmapcolor!1!white}1.30\% & \cellcolor{heatmapcolor!23!white}22.74\% \\
\wmove{Qxb2} & \cellcolor{heatmapcolor!3!white}2.80\% & \cellcolor{heatmapcolor!18!white}17.91\% & \cellcolor{heatmapcolor!20!white}19.80\% & \cellcolor{heatmapcolor!24!white}24.22\% & \cellcolor{heatmapcolor!24!white}24.28\% & \cellcolor{heatmapcolor!28!white}27.55\% & \cellcolor{heatmapcolor!13!white}13.09\% & \cellcolor{heatmapcolor!1!white}0.96\% \\
\wmove{Qxd2} & \cellcolor{heatmapcolor!7!white}6.94\% & \cellcolor{heatmapcolor!16!white}15.90\% & \cellcolor{heatmapcolor!19!white}19.44\% & \cellcolor{heatmapcolor!21!white}21.33\% & \cellcolor{heatmapcolor!26!white}26.00\% & \cellcolor{heatmapcolor!22!white}21.59\% & \cellcolor{heatmapcolor!10!white}9.50\% & \cellcolor{heatmapcolor!1!white}0.66\% \\
\wmove{Rxe5} & 0.07\% & \cellcolor{heatmapcolor!17!white}16.82\% & \cellcolor{heatmapcolor!21!white}21.35\% & \cellcolor{heatmapcolor!12!white}12.49\% & \cellcolor{heatmapcolor!6!white}5.78\% & \cellcolor{heatmapcolor!10!white}9.91\% & \cellcolor{heatmapcolor!24!white}24.47\% & \cellcolor{heatmapcolor!10!white}10.04\% \\
\wmove{Qf2} & \cellcolor{heatmapcolor!24!white}24.04\% & 0.17\% & 0.33\% & 0.12\% & 0.14\% & 0.26\% & 0.09\% & 0.05\% \\
\wmove{Qxe5} & 0.03\% & \cellcolor{heatmapcolor!23!white}23.35\% & \cellcolor{heatmapcolor!19!white}18.84\% & \cellcolor{heatmapcolor!2!white}1.76\% & \cellcolor{heatmapcolor!1!white}0.63\% & \cellcolor{heatmapcolor!1!white}0.52\% & \cellcolor{heatmapcolor!1!white}0.72\% & 0.26\% \\
\wmove{Qxf4} & 0.49\% & \cellcolor{heatmapcolor!6!white}6.44\% & \cellcolor{heatmapcolor!7!white}6.63\% & \cellcolor{heatmapcolor!14!white}13.56\% & \cellcolor{heatmapcolor!11!white}11.31\% & \cellcolor{heatmapcolor!14!white}13.92\% & \cellcolor{heatmapcolor!14!white}13.76\% & \cellcolor{heatmapcolor!20!white}20.34\% \\
\wmove{Qxc4} & \cellcolor{heatmapcolor!1!white}0.80\% & \cellcolor{heatmapcolor!10!white}9.96\% & \cellcolor{heatmapcolor!6!white}6.49\% & \cellcolor{heatmapcolor!10!white}9.72\% & \cellcolor{heatmapcolor!12!white}11.84\% & \cellcolor{heatmapcolor!12!white}11.95\% & \cellcolor{heatmapcolor!16!white}16.31\% & \cellcolor{heatmapcolor!10!white}10.29\% \\
\wmove{h6} & 0.01\% & \cellcolor{heatmapcolor!1!white}1.04\% & 0.02\% & 0.01\% & 0.00\% & 0.02\% & 0.01\% & 0.00\% \\
\wmove{Qd3} & \cellcolor{heatmapcolor!12!white}12.24\% & 0.07\% & 0.12\% & 0.04\% & 0.09\% & 0.04\% & 0.08\% & 0.08\% \\
\wmove{Qe3} & \cellcolor{heatmapcolor!11!white}11.37\% & 0.05\% & 0.07\% & \cellcolor{heatmapcolor!1!white}0.58\% & 0.16\% & 0.31\% & 0.26\% & \cellcolor{heatmapcolor!1!white}0.56\% \\
\wmove{Qc3} & \cellcolor{heatmapcolor!11!white}10.54\% & 0.04\% & 0.09\% & \cellcolor{heatmapcolor!2!white}2.37\% & \cellcolor{heatmapcolor!1!white}1.38\% & \cellcolor{heatmapcolor!1!white}1.16\% & \cellcolor{heatmapcolor!1!white}0.97\% & \cellcolor{heatmapcolor!1!white}0.70\% \\
\wmove{Ne3} & \cellcolor{heatmapcolor!10!white}10.26\% & 0.09\% & 0.18\% & \cellcolor{heatmapcolor!1!white}1.02\% & 0.40\% & 0.44\% & 0.21\% & 0.07\% \\
\wmove{Nh6} & 0.13\% & 0.24\% & 0.03\% & 0.08\% & 0.01\% & 0.07\% & 0.01\% & \cellcolor{heatmapcolor!1!white}1.01\% \\
\wmove{Rc6} & 0.12\% & \cellcolor{heatmapcolor!2!white}1.51\% & \cellcolor{heatmapcolor!1!white}0.77\% & \cellcolor{heatmapcolor!1!white}0.66\% & \cellcolor{heatmapcolor!1!white}0.67\% & \cellcolor{heatmapcolor!4!white}4.08\% & \cellcolor{heatmapcolor!1!white}1.47\% & \cellcolor{heatmapcolor!1!white}0.87\% \\
\wmove{h5} & 0.05\% & \cellcolor{heatmapcolor!3!white}2.57\% & 0.11\% & 0.04\% & 0.02\% & 0.03\% & 0.03\% & 0.02\% \\
\wmove{Kh8} & 0.15\% & 0.05\% & 0.02\% & 0.22\% & \cellcolor{heatmapcolor!1!white}0.81\% & \cellcolor{heatmapcolor!2!white}2.20\% & 0.34\% & 0.24\% \\
\wmove{Nd6} & 0.08\% & 0.01\% & 0.02\% & 0.07\% & 0.14\% & 0.08\% & 0.09\% & 0.10\% \\
\wmove{Qe4} & \cellcolor{heatmapcolor!2!white}1.62\% & 0.09\% & 0.13\% & 0.06\% & 0.06\% & 0.07\% & 0.10\% & 0.13\% \\
\wmove{Rb6} & 0.07\% & 0.01\% & 0.01\% & 0.08\% & 0.32\% & \cellcolor{heatmapcolor!1!white}1.36\% & 0.30\% & 0.18\% \\
\wmove{Re8} & 0.06\% & \cellcolor{heatmapcolor!1!white}0.60\% & 0.22\% & 0.17\% & 0.10\% & 0.06\% & 0.02\% & 0.03\% \\
\wmove{f6} & 0.07\% & 0.06\% & 0.00\% & 0.01\% & 0.01\% & 0.02\% & 0.05\% & 0.05\% \\
\wmove{g6} & 0.07\% & 0.11\% & 0.01\% & 0.07\% & 0.05\% & 0.04\% & 0.03\% & 0.02\% \\
\wmove{a5} & 0.04\% & \cellcolor{heatmapcolor!1!white}0.91\% & 0.13\% & 0.17\% & 0.06\% & 0.07\% & 0.09\% & 0.02\% \\
\wmove{Ne7} & 0.25\% & 0.01\% & 0.01\% & 0.05\% & 0.06\% & 0.10\% & 0.12\% & 0.09\% \\
\wmove{Qd8} & 0.05\% & 0.13\% & 0.04\% & \cellcolor{heatmapcolor!1!white}0.76\% & 0.24\% & 0.22\% & 0.12\% & 0.06\% \\
\wmove{Re7} & 0.32\% & 0.01\% & 0.05\% & 0.05\% & 0.12\% & \cellcolor{heatmapcolor!1!white}0.68\% & 0.17\% & 0.09\% \\
\wmove{Kf8} & 0.06\% & 0.04\% & 0.02\% & 0.04\% & 0.25\% & \cellcolor{heatmapcolor!1!white}0.68\% & 0.17\% & 0.22\% \\
\wmove{Nh4} & \cellcolor{heatmapcolor!1!white}0.65\% & 0.04\% & 0.01\% & 0.04\% & 0.05\% & 0.03\% & 0.03\% & 0.05\% \\
\wmove{g5} & \cellcolor{heatmapcolor!1!white}0.54\% & 0.38\% & 0.23\% & 0.07\% & 0.04\% & 0.03\% & 0.01\% & 0.01\% \\
\wmove{Qd5} & 0.02\% & 0.05\% & 0.06\% & 0.31\% & 0.22\% & 0.14\% & \cellcolor{heatmapcolor!1!white}0.52\% & 0.38\% \\
\wmove{Qd7} & 0.04\% & 0.04\% & 0.02\% & 0.09\% & 0.22\% & 0.41\% & 0.11\% & 0.16\% \\
\wmove{Rg6} & 0.07\% & 0.01\% & 0.02\% & 0.09\% & 0.16\% & 0.41\% & 0.04\% & 0.05\% \\
\wmove{Qd6} & 0.02\% & 0.02\% & 0.03\% & 0.08\% & 0.06\% & 0.19\% & 0.07\% & 0.07\% \\
\wmove{Rd6} & 0.10\% & 0.01\% & 0.03\% & 0.06\% & 0.06\% & 0.25\% & 0.07\% & 0.06\% \\
\wmove{Rf6} & 0.10\% & 0.01\% & 0.12\% & 0.07\% & 0.03\% & 0.11\% & 0.03\% & 0.03\% \\
\bottomrule
\end{tabular}%
}
\end{table*}

\begin{table*}[h!]
\centering
\caption{Move probabilities by layer for puzzle \href{https://lichess.org/training/62FeU}{\texttt{62FeU}} (Part 2: Layer 7 to Final)}
\label{tab:puzzle_62FeU_probs2}
\resizebox{\textwidth}{!}{%
\begin{tabular}{lrrrrrrrr}
\toprule
\textbf{Move} & \textbf{Layer 7} & \textbf{Layer 8} & \textbf{Layer 9} & \textbf{Layer 10} & \textbf{Layer 11} & \textbf{Layer 12} & \textbf{Layer 13} & \textbf{Final} \\
\midrule
\wmove{Ng3+} & \cellcolor{heatmapcolor!27!white}27.02\% & \cellcolor{heatmapcolor!20!white}20.06\% & \cellcolor{heatmapcolor!19!white}19.00\% & \cellcolor{heatmapcolor!25!white}25.38\% & \cellcolor{heatmapcolor!39!white}39.13\% & \cellcolor{heatmapcolor!28!white}28.03\% & \cellcolor{heatmapcolor!49!white}49.15\% & \cellcolor{heatmapcolor!88!white}87.55\% \\
\wmove{Rh6} & \cellcolor{heatmapcolor!9!white}8.66\% & \cellcolor{heatmapcolor!16!white}15.63\% & \cellcolor{heatmapcolor!9!white}9.35\% & \cellcolor{heatmapcolor!10!white}10.04\% & \cellcolor{heatmapcolor!28!white}28.26\% & \cellcolor{heatmapcolor!52!white}51.73\% & \cellcolor{heatmapcolor!10!white}9.61\% & 0.24\% \\
\wmove{Qg1+} & \cellcolor{heatmapcolor!19!white}18.86\% & \cellcolor{heatmapcolor!15!white}15.30\% & \cellcolor{heatmapcolor!49!white}49.02\% & \cellcolor{heatmapcolor!47!white}47.22\% & \cellcolor{heatmapcolor!22!white}22.42\% & \cellcolor{heatmapcolor!3!white}2.68\% & \cellcolor{heatmapcolor!2!white}2.08\% & 0.28\% \\
\wmove{Qxb2} & 0.27\% & \cellcolor{heatmapcolor!3!white}2.92\% & 0.42\% & 0.35\% & 0.20\% & 0.10\% & \cellcolor{heatmapcolor!1!white}0.86\% & 0.25\% \\
\wmove{Qxd2} & 0.19\% & \cellcolor{heatmapcolor!2!white}1.72\% & 0.34\% & 0.27\% & 0.13\% & 0.08\% & \cellcolor{heatmapcolor!1!white}0.63\% & 0.24\% \\
\wmove{Rxe5} & \cellcolor{heatmapcolor!10!white}9.74\% & \cellcolor{heatmapcolor!7!white}6.51\% & \cellcolor{heatmapcolor!6!white}5.83\% & \cellcolor{heatmapcolor!4!white}3.66\% & \cellcolor{heatmapcolor!3!white}2.71\% & \cellcolor{heatmapcolor!1!white}1.44\% & \cellcolor{heatmapcolor!1!white}0.71\% & 0.37\% \\
\wmove{Qf2} & 0.07\% & 0.21\% & 0.16\% & \cellcolor{heatmapcolor!1!white}0.55\% & 0.08\% & 0.04\% & 0.34\% & 0.27\% \\
\wmove{Qxe5} & 0.17\% & 0.25\% & 0.37\% & 0.13\% & 0.09\% & 0.06\% & \cellcolor{heatmapcolor!1!white}0.67\% & 0.37\% \\
\wmove{Qxf4} & \cellcolor{heatmapcolor!13!white}13.50\% & \cellcolor{heatmapcolor!18!white}18.19\% & \cellcolor{heatmapcolor!5!white}5.50\% & \cellcolor{heatmapcolor!7!white}6.82\% & \cellcolor{heatmapcolor!1!white}0.50\% & 0.17\% & \cellcolor{heatmapcolor!1!white}0.55\% & 0.27\% \\
\wmove{Qxc4} & \cellcolor{heatmapcolor!12!white}12.17\% & \cellcolor{heatmapcolor!13!white}13.07\% & \cellcolor{heatmapcolor!5!white}4.70\% & \cellcolor{heatmapcolor!1!white}1.31\% & 0.37\% & 0.20\% & \cellcolor{heatmapcolor!1!white}0.75\% & 0.29\% \\
\wmove{h6} & 0.02\% & 0.14\% & 0.12\% & \cellcolor{heatmapcolor!1!white}0.61\% & \cellcolor{heatmapcolor!1!white}1.38\% & \cellcolor{heatmapcolor!4!white}3.87\% & \cellcolor{heatmapcolor!15!white}14.66\% & 0.25\% \\
\wmove{Qd3} & 0.07\% & 0.15\% & 0.22\% & 0.16\% & 0.33\% & 0.15\% & \cellcolor{heatmapcolor!1!white}0.59\% & 0.26\% \\
\wmove{Qe3} & 0.37\% & 0.15\% & 0.06\% & 0.03\% & 0.08\% & 0.04\% & 0.37\% & 0.25\% \\
\wmove{Qc3} & \cellcolor{heatmapcolor!1!white}1.42\% & 0.38\% & 0.22\% & 0.08\% & 0.12\% & 0.03\% & 0.50\% & 0.22\% \\
\wmove{Ne3} & 0.04\% & 0.04\% & 0.03\% & 0.02\% & 0.03\% & 0.03\% & 0.24\% & 0.21\% \\
\wmove{Nh6} & \cellcolor{heatmapcolor!4!white}4.29\% & \cellcolor{heatmapcolor!3!white}3.37\% & \cellcolor{heatmapcolor!2!white}2.22\% & \cellcolor{heatmapcolor!2!white}1.66\% & \cellcolor{heatmapcolor!2!white}2.45\% & \cellcolor{heatmapcolor!9!white}9.23\% & \cellcolor{heatmapcolor!6!white}6.39\% & 0.23\% \\
\wmove{Rc6} & \cellcolor{heatmapcolor!1!white}0.58\% & 0.22\% & 0.24\% & 0.06\% & 0.06\% & 0.05\% & 0.21\% & 0.34\% \\
\wmove{h5} & 0.20\% & 0.07\% & 0.19\% & 0.32\% & 0.24\% & 0.14\% & \cellcolor{heatmapcolor!1!white}0.88\% & 0.25\% \\
\wmove{Kh8} & 0.33\% & 0.25\% & 0.48\% & 0.32\% & 0.34\% & 0.48\% & 0.27\% & 0.27\% \\
\wmove{Nd6} & 0.17\% & 0.12\% & 0.34\% & 0.15\% & 0.09\% & 0.12\% & \cellcolor{heatmapcolor!2!white}1.97\% & \cellcolor{heatmapcolor!1!white}0.82\% \\
\wmove{Qe4} & 0.11\% & 0.07\% & 0.07\% & 0.03\% & 0.04\% & 0.02\% & 0.31\% & 0.26\% \\
\wmove{Rb6} & 0.12\% & 0.05\% & 0.05\% & 0.02\% & 0.03\% & 0.05\% & 0.28\% & 0.28\% \\
\wmove{Re8} & 0.10\% & 0.13\% & 0.12\% & 0.07\% & 0.06\% & 0.20\% & \cellcolor{heatmapcolor!1!white}0.70\% & \cellcolor{heatmapcolor!1!white}1.35\% \\
\wmove{f6} & 0.33\% & 0.13\% & 0.11\% & 0.14\% & 0.15\% & 0.27\% & \cellcolor{heatmapcolor!1!white}1.28\% & 0.29\% \\
\wmove{g6} & 0.04\% & 0.03\% & 0.05\% & 0.08\% & 0.07\% & 0.12\% & \cellcolor{heatmapcolor!1!white}1.05\% & 0.31\% \\
\wmove{a5} & 0.04\% & 0.01\% & 0.02\% & 0.01\% & 0.01\% & 0.02\% & 0.18\% & 0.23\% \\
\wmove{Ne7} & 0.07\% & 0.05\% & 0.04\% & 0.02\% & 0.02\% & 0.01\% & \cellcolor{heatmapcolor!1!white}0.84\% & 0.39\% \\
\wmove{Qd8} & 0.08\% & 0.08\% & 0.07\% & 0.05\% & 0.06\% & 0.05\% & 0.48\% & \cellcolor{heatmapcolor!1!white}0.81\% \\
\wmove{Re7} & 0.08\% & 0.04\% & 0.03\% & 0.01\% & 0.02\% & 0.03\% & 0.33\% & 0.26\% \\
\wmove{Kf8} & 0.21\% & 0.22\% & 0.21\% & 0.10\% & 0.05\% & 0.08\% & 0.22\% & 0.35\% \\
\wmove{Nh4} & 0.05\% & 0.04\% & 0.03\% & 0.04\% & 0.05\% & 0.07\% & 0.39\% & 0.20\% \\
\wmove{g5} & 0.02\% & 0.02\% & 0.05\% & 0.04\% & 0.03\% & 0.04\% & \cellcolor{heatmapcolor!1!white}0.61\% & 0.29\% \\
\wmove{Qd5} & 0.35\% & 0.13\% & 0.14\% & 0.05\% & 0.07\% & 0.05\% & 0.41\% & 0.27\% \\
\wmove{Qd7} & 0.04\% & 0.09\% & 0.10\% & 0.07\% & 0.08\% & 0.06\% & 0.22\% & 0.35\% \\
\wmove{Rg6} & 0.05\% & 0.04\% & 0.05\% & 0.05\% & 0.10\% & 0.09\% & 0.23\% & 0.28\% \\
\wmove{Qd6} & 0.07\% & 0.04\% & 0.03\% & 0.03\% & 0.05\% & 0.06\% & 0.41\% & 0.28\% \\
\wmove{Rd6} & 0.06\% & 0.04\% & 0.02\% & 0.01\% & 0.04\% & 0.04\% & 0.34\% & 0.29\% \\
\wmove{Rf6} & 0.05\% & 0.05\% & 0.03\% & 0.02\% & 0.04\% & 0.06\% & 0.29\% & 0.28\% \\
\bottomrule
\end{tabular}%
}
\end{table*}

%% file: Figures/Puzzles/Examples/puzzle_tables_9JvtA.tex
\begin{table*}[h!]
\centering
\caption{Move probabilities by layer for puzzle \href{https://lichess.org/training/9JvtA}{\texttt{9JvtA}} (Part 1: Input to Layer 6)}
\label{tab:puzzle_9JvtA_probs1}
\resizebox{\textwidth}{!}{%
\begin{tabular}{lrrrrrrrr}
\toprule
\textbf{Move} & \textbf{Input} & \textbf{Layer 0} & \textbf{Layer 1} & \textbf{Layer 2} & \textbf{Layer 3} & \textbf{Layer 4} & \textbf{Layer 5} & \textbf{Layer 6} \\
\midrule
\wmove{Nd6+} & \cellcolor{heatmapcolor!32!white}32.13\% & 0.05\% & \cellcolor{heatmapcolor!4!white}3.60\% & \cellcolor{heatmapcolor!12!white}11.55\% & 0.14\% & 0.10\% & \cellcolor{heatmapcolor!1!white}0.75\% & 0.50\% \\
\wmove{Qxa5} & 0.20\% & \cellcolor{heatmapcolor!4!white}4.42\% & \cellcolor{heatmapcolor!6!white}6.22\% & \cellcolor{heatmapcolor!9!white}9.20\% & \cellcolor{heatmapcolor!27!white}26.54\% & \cellcolor{heatmapcolor!51!white}50.71\% & \cellcolor{heatmapcolor!48!white}47.82\% & \cellcolor{heatmapcolor!58!white}58.34\% \\
\wmove{Qxc6+} & \cellcolor{heatmapcolor!1!white}0.90\% & \cellcolor{heatmapcolor!27!white}26.71\% & \cellcolor{heatmapcolor!56!white}56.31\% & \cellcolor{heatmapcolor!20!white}19.67\% & \cellcolor{heatmapcolor!21!white}20.88\% & \cellcolor{heatmapcolor!16!white}16.16\% & \cellcolor{heatmapcolor!8!white}7.71\% & \cellcolor{heatmapcolor!6!white}5.78\% \\
\wmove{b4} & \cellcolor{heatmapcolor!1!white}1.44\% & \cellcolor{heatmapcolor!47!white}46.77\% & 0.36\% & \cellcolor{heatmapcolor!1!white}0.88\% & \cellcolor{heatmapcolor!1!white}0.58\% & 0.01\% & 0.01\% & 0.16\% \\
\wmove{Qxd4} & \cellcolor{heatmapcolor!1!white}1.50\% & \cellcolor{heatmapcolor!7!white}7.14\% & \cellcolor{heatmapcolor!23!white}22.75\% & \cellcolor{heatmapcolor!40!white}39.84\% & \cellcolor{heatmapcolor!34!white}34.48\% & \cellcolor{heatmapcolor!24!white}23.90\% & \cellcolor{heatmapcolor!30!white}29.99\% & \cellcolor{heatmapcolor!23!white}23.33\% \\
\wmove{Nf6+} & \cellcolor{heatmapcolor!36!white}36.04\% & 0.01\% & 0.17\% & \cellcolor{heatmapcolor!2!white}2.27\% & 0.05\% & 0.13\% & 0.01\% & 0.01\% \\
\wmove{Nc5} & \cellcolor{heatmapcolor!17!white}17.13\% & 0.04\% & 0.27\% & \cellcolor{heatmapcolor!1!white}0.60\% & 0.17\% & 0.09\% & 0.13\% & 0.09\% \\
\wmove{a3} & 0.03\% & \cellcolor{heatmapcolor!8!white}7.50\% & \cellcolor{heatmapcolor!3!white}2.85\% & \cellcolor{heatmapcolor!6!white}6.10\% & 0.13\% & \cellcolor{heatmapcolor!1!white}0.57\% & 0.16\% & 0.20\% \\
\wmove{Nf3} & 0.07\% & 0.01\% & 0.03\% & 0.20\% & \cellcolor{heatmapcolor!4!white}4.38\% & 0.12\% & \cellcolor{heatmapcolor!5!white}4.86\% & \cellcolor{heatmapcolor!3!white}3.17\% \\
\wmove{f4} & \cellcolor{heatmapcolor!3!white}3.27\% & 0.27\% & 0.03\% & 0.04\% & 0.04\% & 0.15\% & 0.05\% & 0.11\% \\
\wmove{Bf3} & 0.29\% & 0.02\% & \cellcolor{heatmapcolor!1!white}0.51\% & 0.37\% & \cellcolor{heatmapcolor!2!white}1.85\% & \cellcolor{heatmapcolor!1!white}0.86\% & \cellcolor{heatmapcolor!3!white}3.02\% & \cellcolor{heatmapcolor!2!white}1.93\% \\
\wmove{Qb4} & 0.10\% & 0.44\% & \cellcolor{heatmapcolor!1!white}0.51\% & \cellcolor{heatmapcolor!1!white}0.92\% & \cellcolor{heatmapcolor!3!white}2.59\% & 0.20\% & 0.11\% & 0.27\% \\
\wmove{Ng5} & \cellcolor{heatmapcolor!3!white}2.51\% & 0.05\% & 0.02\% & 0.31\% & 0.17\% & 0.29\% & 0.02\% & 0.05\% \\
\wmove{Qb3} & 0.09\% & 0.10\% & 0.20\% & \cellcolor{heatmapcolor!1!white}1.28\% & \cellcolor{heatmapcolor!2!white}2.22\% & 0.42\% & \cellcolor{heatmapcolor!1!white}0.98\% & \cellcolor{heatmapcolor!1!white}1.08\% \\
\wmove{Qa3} & 0.04\% & \cellcolor{heatmapcolor!2!white}1.51\% & \cellcolor{heatmapcolor!1!white}0.93\% & \cellcolor{heatmapcolor!1!white}1.41\% & \cellcolor{heatmapcolor!2!white}1.87\% & 0.50\% & 0.50\% & \cellcolor{heatmapcolor!1!white}0.87\% \\
\wmove{Nc3} & 0.36\% & 0.02\% & \cellcolor{heatmapcolor!2!white}1.86\% & \cellcolor{heatmapcolor!1!white}1.02\% & 0.21\% & 0.19\% & 0.04\% & 0.20\% \\
\wmove{g4} & \cellcolor{heatmapcolor!2!white}1.75\% & 0.30\% & 0.24\% & \cellcolor{heatmapcolor!1!white}0.71\% & 0.38\% & 0.25\% & 0.03\% & 0.02\% \\
\wmove{Rb1} & 0.16\% & 0.10\% & 0.09\% & 0.17\% & 0.21\% & \cellcolor{heatmapcolor!1!white}1.25\% & \cellcolor{heatmapcolor!1!white}1.09\% & \cellcolor{heatmapcolor!2!white}1.68\% \\
\wmove{f3} & 0.20\% & 0.00\% & 0.00\% & 0.00\% & 0.00\% & 0.11\% & 0.04\% & 0.08\% \\
\wmove{h3} & 0.03\% & \cellcolor{heatmapcolor!1!white}1.44\% & 0.12\% & 0.41\% & 0.03\% & 0.12\% & 0.22\% & 0.08\% \\
\wmove{Bh3} & 0.02\% & 0.06\% & 0.11\% & 0.15\% & \cellcolor{heatmapcolor!1!white}0.98\% & \cellcolor{heatmapcolor!1!white}0.96\% & 0.43\% & 0.24\% \\
\wmove{Kf1} & 0.11\% & 0.37\% & \cellcolor{heatmapcolor!1!white}0.96\% & 0.25\% & 0.25\% & 0.19\% & 0.17\% & 0.09\% \\
\wmove{b3} & 0.17\% & 0.21\% & 0.08\% & \cellcolor{heatmapcolor!1!white}0.92\% & 0.01\% & 0.03\% & 0.03\% & 0.01\% \\
\wmove{Bf1} & 0.01\% & 0.08\% & \cellcolor{heatmapcolor!1!white}0.87\% & 0.35\% & \cellcolor{heatmapcolor!1!white}0.51\% & 0.29\% & 0.17\% & 0.05\% \\
\wmove{Nh3} & 0.15\% & \cellcolor{heatmapcolor!1!white}0.83\% & 0.12\% & 0.21\% & 0.04\% & 0.13\% & 0.23\% & 0.22\% \\
\wmove{Kd1} & 0.12\% & 0.11\% & 0.05\% & 0.09\% & 0.41\% & \cellcolor{heatmapcolor!1!white}0.75\% & 0.40\% & \cellcolor{heatmapcolor!1!white}0.56\% \\
\wmove{e3} & 0.42\% & 0.49\% & 0.03\% & 0.23\% & 0.16\% & \cellcolor{heatmapcolor!1!white}0.75\% & 0.24\% & 0.25\% \\
\wmove{Qc4} & 0.25\% & 0.11\% & 0.03\% & 0.08\% & 0.16\% & 0.11\% & 0.18\% & 0.16\% \\
\wmove{h4} & 0.09\% & \cellcolor{heatmapcolor!1!white}0.52\% & \cellcolor{heatmapcolor!1!white}0.56\% & 0.45\% & 0.15\% & 0.21\% & 0.16\% & 0.14\% \\
\wmove{Qd1} & 0.11\% & 0.09\% & 0.03\% & 0.08\% & 0.07\% & 0.17\% & 0.16\% & 0.08\% \\
\wmove{Qb5} & 0.12\% & 0.12\% & 0.06\% & 0.11\% & 0.30\% & 0.13\% & 0.16\% & 0.15\% \\
\wmove{Qc2} & 0.16\% & 0.10\% & 0.04\% & 0.12\% & 0.06\% & 0.16\% & 0.14\% & 0.10\% \\
\bottomrule
\end{tabular}%
}
\end{table*}

\begin{table*}[h!]
\centering
\caption{Move probabilities by layer for puzzle \href{https://lichess.org/training/9JvtA}{\texttt{9JvtA}} (Part 2: Layer 7 to Final)}
\label{tab:puzzle_9JvtA_probs2}
\resizebox{\textwidth}{!}{%
\begin{tabular}{lrrrrrrrr}
\toprule
\textbf{Move} & \textbf{Layer 7} & \textbf{Layer 8} & \textbf{Layer 9} & \textbf{Layer 10} & \textbf{Layer 11} & \textbf{Layer 12} & \textbf{Layer 13} & \textbf{Final} \\
\midrule
\wmove{Nd6+} & \cellcolor{heatmapcolor!1!white}1.22\% & \cellcolor{heatmapcolor!1!white}1.26\% & \cellcolor{heatmapcolor!2!white}2.22\% & \cellcolor{heatmapcolor!3!white}2.73\% & \cellcolor{heatmapcolor!2!white}1.72\% & \cellcolor{heatmapcolor!2!white}1.65\% & \cellcolor{heatmapcolor!24!white}24.37\% & \cellcolor{heatmapcolor!65!white}65.34\% \\
\wmove{Qxa5} & \cellcolor{heatmapcolor!58!white}58.41\% & \cellcolor{heatmapcolor!53!white}52.60\% & \cellcolor{heatmapcolor!51!white}50.88\% & \cellcolor{heatmapcolor!61!white}61.05\% & \cellcolor{heatmapcolor!57!white}56.84\% & \cellcolor{heatmapcolor!57!white}56.52\% & \cellcolor{heatmapcolor!46!white}46.15\% & \cellcolor{heatmapcolor!12!white}12.18\% \\
\wmove{Qxc6+} & \cellcolor{heatmapcolor!7!white}6.60\% & \cellcolor{heatmapcolor!15!white}14.67\% & \cellcolor{heatmapcolor!25!white}25.41\% & \cellcolor{heatmapcolor!22!white}22.02\% & \cellcolor{heatmapcolor!30!white}30.43\% & \cellcolor{heatmapcolor!29!white}28.79\% & \cellcolor{heatmapcolor!3!white}3.26\% & \cellcolor{heatmapcolor!1!white}0.61\% \\
\wmove{b4} & 0.44\% & \cellcolor{heatmapcolor!2!white}1.59\% & 0.40\% & \cellcolor{heatmapcolor!1!white}0.70\% & \cellcolor{heatmapcolor!2!white}1.71\% & \cellcolor{heatmapcolor!3!white}3.37\% & \cellcolor{heatmapcolor!3!white}3.06\% & 0.29\% \\
\wmove{Qxd4} & \cellcolor{heatmapcolor!23!white}22.78\% & \cellcolor{heatmapcolor!20!white}19.53\% & \cellcolor{heatmapcolor!15!white}14.91\% & \cellcolor{heatmapcolor!9!white}8.88\% & \cellcolor{heatmapcolor!4!white}4.27\% & \cellcolor{heatmapcolor!5!white}4.65\% & \cellcolor{heatmapcolor!1!white}0.79\% & 0.29\% \\
\wmove{Nf6+} & 0.01\% & 0.02\% & 0.02\% & 0.02\% & 0.02\% & 0.02\% & \cellcolor{heatmapcolor!11!white}11.00\% & \cellcolor{heatmapcolor!14!white}13.67\% \\
\wmove{Nc5} & 0.08\% & 0.12\% & 0.20\% & 0.13\% & 0.14\% & 0.09\% & \cellcolor{heatmapcolor!1!white}0.69\% & 0.29\% \\
\wmove{a3} & \cellcolor{heatmapcolor!1!white}0.93\% & \cellcolor{heatmapcolor!1!white}0.78\% & 0.28\% & \cellcolor{heatmapcolor!1!white}0.55\% & 0.22\% & 0.19\% & \cellcolor{heatmapcolor!2!white}1.90\% & 0.28\% \\
\wmove{Nf3} & \cellcolor{heatmapcolor!2!white}1.91\% & \cellcolor{heatmapcolor!2!white}1.70\% & \cellcolor{heatmapcolor!1!white}0.89\% & \cellcolor{heatmapcolor!1!white}0.83\% & 0.41\% & 0.14\% & \cellcolor{heatmapcolor!1!white}0.73\% & 0.28\% \\
\wmove{f4} & 0.40\% & 0.48\% & 0.31\% & \cellcolor{heatmapcolor!1!white}0.57\% & \cellcolor{heatmapcolor!2!white}1.95\% & \cellcolor{heatmapcolor!1!white}1.26\% & \cellcolor{heatmapcolor!1!white}1.22\% & 0.26\% \\
\wmove{Bf3} & \cellcolor{heatmapcolor!1!white}0.72\% & \cellcolor{heatmapcolor!1!white}0.55\% & 0.35\% & 0.16\% & 0.11\% & \cellcolor{heatmapcolor!1!white}0.51\% & 0.38\% & 0.29\% \\
\wmove{Qb4} & \cellcolor{heatmapcolor!1!white}0.68\% & \cellcolor{heatmapcolor!1!white}0.60\% & 0.26\% & 0.12\% & 0.08\% & 0.06\% & 0.05\% & 0.27\% \\
\wmove{Ng5} & 0.22\% & 0.45\% & 0.03\% & 0.03\% & 0.02\% & 0.19\% & 0.34\% & 0.27\% \\
\wmove{Qb3} & \cellcolor{heatmapcolor!2!white}1.94\% & \cellcolor{heatmapcolor!2!white}2.00\% & \cellcolor{heatmapcolor!1!white}0.67\% & 0.28\% & 0.12\% & 0.32\% & 0.26\% & 0.29\% \\
\wmove{Qa3} & \cellcolor{heatmapcolor!1!white}0.78\% & \cellcolor{heatmapcolor!1!white}1.30\% & \cellcolor{heatmapcolor!1!white}0.72\% & 0.37\% & 0.21\% & 0.14\% & 0.12\% & 0.29\% \\
\wmove{Nc3} & 0.25\% & 0.13\% & 0.07\% & 0.11\% & 0.06\% & 0.17\% & \cellcolor{heatmapcolor!1!white}0.66\% & 0.29\% \\
\wmove{g4} & 0.02\% & 0.02\% & 0.02\% & 0.04\% & 0.04\% & 0.04\% & 0.16\% & 0.29\% \\
\wmove{Rb1} & \cellcolor{heatmapcolor!1!white}0.55\% & \cellcolor{heatmapcolor!1!white}0.51\% & \cellcolor{heatmapcolor!1!white}0.77\% & 0.10\% & 0.10\% & 0.12\% & 0.38\% & 0.31\% \\
\wmove{f3} & 0.03\% & 0.06\% & 0.02\% & 0.02\% & 0.02\% & 0.13\% & \cellcolor{heatmapcolor!1!white}1.47\% & 0.26\% \\
\wmove{h3} & 0.06\% & 0.04\% & 0.04\% & 0.03\% & 0.05\% & 0.04\% & 0.33\% & 0.30\% \\
\wmove{Bh3} & 0.20\% & 0.22\% & 0.21\% & 0.05\% & 0.07\% & 0.05\% & 0.16\% & 0.32\% \\
\wmove{Kf1} & 0.07\% & 0.05\% & 0.08\% & 0.07\% & 0.04\% & 0.02\% & 0.13\% & 0.33\% \\
\wmove{b3} & 0.08\% & 0.05\% & 0.03\% & 0.04\% & 0.02\% & 0.01\% & 0.15\% & 0.25\% \\
\wmove{Bf1} & 0.04\% & 0.02\% & 0.03\% & 0.03\% & 0.05\% & 0.05\% & 0.15\% & 0.34\% \\
\wmove{Nh3} & 0.18\% & 0.17\% & 0.08\% & 0.03\% & 0.08\% & 0.19\% & 0.36\% & 0.29\% \\
\wmove{Kd1} & \cellcolor{heatmapcolor!1!white}0.54\% & 0.29\% & 0.37\% & 0.32\% & 0.40\% & 0.28\% & 0.21\% & 0.32\% \\
\wmove{e3} & 0.26\% & 0.25\% & 0.19\% & 0.18\% & 0.19\% & 0.10\% & \cellcolor{heatmapcolor!1!white}0.55\% & 0.28\% \\
\wmove{Qc4} & 0.14\% & 0.21\% & 0.19\% & 0.19\% & 0.25\% & \cellcolor{heatmapcolor!1!white}0.56\% & 0.32\% & 0.28\% \\
\wmove{h4} & 0.17\% & 0.11\% & 0.06\% & 0.06\% & 0.08\% & 0.06\% & 0.20\% & 0.33\% \\
\wmove{Qd1} & 0.08\% & 0.04\% & 0.05\% & 0.04\% & 0.04\% & 0.04\% & 0.15\% & 0.32\% \\
\wmove{Qb5} & 0.10\% & 0.13\% & 0.17\% & 0.17\% & 0.11\% & 0.13\% & 0.10\% & 0.29\% \\
\wmove{Qc2} & 0.11\% & 0.06\% & 0.07\% & 0.09\% & 0.14\% & 0.13\% & 0.18\% & 0.29\% \\
\bottomrule
\end{tabular}%
}
\end{table*}

%% file: Figures/Puzzles/Examples/puzzle_tables_70v1B.tex
\begin{table*}[h!]
\centering
\caption{Move probabilities by layer for puzzle \href{https://lichess.org/training/70v1B}{\texttt{70v1B}} (Part 1: Input to Layer 6)}
\label{tab:puzzle_70v1B_probs1}
\resizebox{\textwidth}{!}{%
\begin{tabular}{lrrrrrrrr}
\toprule
\textbf{Move} & \textbf{Input} & \textbf{Layer 0} & \textbf{Layer 1} & \textbf{Layer 2} & \textbf{Layer 3} & \textbf{Layer 4} & \textbf{Layer 5} & \textbf{Layer 6} \\
\midrule
\wmove{Qxe8+} & \cellcolor{heatmapcolor!11!white}11.14\% & \cellcolor{heatmapcolor!51!white}50.54\% & \cellcolor{heatmapcolor!72!white}71.54\% & \cellcolor{heatmapcolor!81!white}80.98\% & \cellcolor{heatmapcolor!41!white}40.69\% & \cellcolor{heatmapcolor!65!white}64.80\% & \cellcolor{heatmapcolor!67!white}66.88\% & \cellcolor{heatmapcolor!68!white}68.29\% \\
\wmove{Qe7} & \cellcolor{heatmapcolor!57!white}56.93\% & \cellcolor{heatmapcolor!1!white}0.68\% & 0.16\% & 0.06\% & 0.08\% & 0.09\% & 0.06\% & 0.09\% \\
\wmove{bxc3} & 0.18\% & \cellcolor{heatmapcolor!20!white}20.37\% & \cellcolor{heatmapcolor!22!white}21.51\% & \cellcolor{heatmapcolor!14!white}13.64\% & \cellcolor{heatmapcolor!56!white}56.27\% & \cellcolor{heatmapcolor!31!white}30.75\% & \cellcolor{heatmapcolor!31!white}30.98\% & \cellcolor{heatmapcolor!27!white}27.02\% \\
\wmove{Qe6} & \cellcolor{heatmapcolor!21!white}20.58\% & \cellcolor{heatmapcolor!1!white}1.07\% & \cellcolor{heatmapcolor!1!white}1.00\% & 0.16\% & 0.12\% & 0.10\% & 0.10\% & 0.16\% \\
\wmove{h4} & 0.08\% & \cellcolor{heatmapcolor!7!white}6.79\% & \cellcolor{heatmapcolor!1!white}0.81\% & 0.44\% & 0.37\% & 0.11\% & 0.07\% & 0.29\% \\
\wmove{b4} & \cellcolor{heatmapcolor!1!white}0.72\% & \cellcolor{heatmapcolor!6!white}6.44\% & \cellcolor{heatmapcolor!1!white}1.49\% & 0.42\% & 0.06\% & 0.05\% & 0.02\% & 0.04\% \\
\wmove{a4} & 0.07\% & \cellcolor{heatmapcolor!5!white}4.65\% & 0.20\% & 0.08\% & 0.02\% & 0.02\% & 0.09\% & 0.14\% \\
\wmove{Qe5} & \cellcolor{heatmapcolor!1!white}0.69\% & \cellcolor{heatmapcolor!2!white}1.88\% & 0.49\% & 0.30\% & 0.22\% & 0.13\% & 0.06\% & 0.10\% \\
\wmove{Qf2} & \cellcolor{heatmapcolor!2!white}1.59\% & 0.15\% & 0.03\% & 0.12\% & 0.09\% & 0.03\% & 0.04\% & 0.10\% \\
\wmove{Qc4} & \cellcolor{heatmapcolor!1!white}0.62\% & \cellcolor{heatmapcolor!1!white}0.56\% & \cellcolor{heatmapcolor!1!white}1.50\% & \cellcolor{heatmapcolor!1!white}1.15\% & 0.20\% & 0.13\% & 0.12\% & 0.23\% \\
\wmove{f5} & \cellcolor{heatmapcolor!1!white}0.63\% & \cellcolor{heatmapcolor!1!white}1.44\% & 0.01\% & 0.01\% & 0.01\% & 0.01\% & 0.04\% & 0.14\% \\
\wmove{Rb1} & 0.11\% & 0.04\% & 0.02\% & 0.12\% & 0.19\% & \cellcolor{heatmapcolor!1!white}1.31\% & 0.06\% & 0.29\% \\
\wmove{Qg2} & \cellcolor{heatmapcolor!1!white}1.14\% & 0.10\% & 0.02\% & 0.07\% & 0.03\% & 0.04\% & 0.03\% & 0.08\% \\
\wmove{a3} & 0.02\% & \cellcolor{heatmapcolor!1!white}0.94\% & 0.05\% & 0.04\% & 0.04\% & 0.07\% & 0.06\% & 0.25\% \\
\wmove{g4} & \cellcolor{heatmapcolor!1!white}0.82\% & \cellcolor{heatmapcolor!1!white}0.83\% & 0.02\% & 0.04\% & 0.09\% & 0.04\% & 0.05\% & 0.12\% \\
\wmove{Qb5} & \cellcolor{heatmapcolor!1!white}0.62\% & 0.10\% & 0.06\% & 0.09\% & 0.05\% & 0.07\% & 0.08\% & 0.17\% \\
\wmove{Qg4} & \cellcolor{heatmapcolor!1!white}0.61\% & 0.34\% & 0.09\% & 0.23\% & 0.16\% & 0.12\% & 0.10\% & 0.19\% \\
\wmove{Rd1} & 0.13\% & 0.17\% & 0.06\% & 0.05\% & 0.04\% & \cellcolor{heatmapcolor!1!white}0.55\% & 0.07\% & 0.09\% \\
\wmove{Qd1} & 0.08\% & 0.49\% & 0.16\% & 0.26\% & 0.09\% & 0.14\% & 0.07\% & 0.14\% \\
\wmove{Be3} & 0.05\% & 0.01\% & 0.02\% & 0.13\% & 0.02\% & 0.05\% & 0.08\% & 0.23\% \\
\wmove{Qf3} & 0.48\% & 0.17\% & 0.02\% & 0.05\% & 0.04\% & 0.07\% & 0.07\% & 0.11\% \\
\wmove{Qe3} & 0.42\% & 0.11\% & 0.03\% & 0.07\% & 0.03\% & 0.03\% & 0.09\% & 0.10\% \\
\wmove{b3} & 0.11\% & 0.34\% & 0.04\% & 0.06\% & 0.05\% & 0.07\% & 0.05\% & 0.30\% \\
\wmove{Qe4} & 0.40\% & 0.14\% & 0.15\% & 0.16\% & 0.11\% & 0.08\% & 0.06\% & 0.16\% \\
\wmove{Qa6} & 0.05\% & 0.31\% & 0.07\% & 0.06\% & 0.04\% & 0.05\% & 0.05\% & 0.08\% \\
\wmove{Qc2} & 0.37\% & 0.13\% & 0.04\% & 0.13\% & 0.06\% & 0.12\% & 0.07\% & 0.17\% \\
\wmove{Qh5} & 0.13\% & 0.30\% & 0.05\% & 0.11\% & 0.10\% & 0.09\% & 0.08\% & 0.18\% \\
\wmove{h3} & 0.02\% & 0.29\% & 0.05\% & 0.07\% & 0.04\% & 0.01\% & 0.03\% & 0.04\% \\
\wmove{Kh1} & 0.26\% & 0.06\% & 0.04\% & 0.06\% & 0.04\% & 0.10\% & 0.06\% & 0.13\% \\
\wmove{Kf2} & 0.10\% & 0.12\% & 0.07\% & 0.30\% & 0.28\% & 0.10\% & 0.03\% & 0.08\% \\
\wmove{Qd3} & 0.22\% & 0.07\% & 0.04\% & 0.09\% & 0.05\% & 0.10\% & 0.09\% & 0.15\% \\
\wmove{Kg2} & 0.11\% & 0.03\% & 0.03\% & 0.16\% & 0.11\% & 0.14\% & 0.06\% & 0.12\% \\
\wmove{Qd2} & 0.07\% & 0.09\% & 0.03\% & 0.04\% & 0.04\% & 0.05\% & 0.03\% & 0.04\% \\
\wmove{Bd2} & 0.10\% & 0.01\% & 0.01\% & 0.02\% & 0.02\% & 0.06\% & 0.05\% & 0.07\% \\
\wmove{Kf1} & 0.10\% & 0.04\% & 0.03\% & 0.10\% & 0.07\% & 0.12\% & 0.04\% & 0.04\% \\
\wmove{Qf1} & 0.10\% & 0.16\% & 0.03\% & 0.08\% & 0.04\% & 0.05\% & 0.03\% & 0.05\% \\
\wmove{Rf1} & 0.16\% & 0.04\% & 0.02\% & 0.07\% & 0.03\% & 0.13\% & 0.03\% & 0.03\% \\
\bottomrule
\end{tabular}%
}
\end{table*}

\begin{table*}[h!]
\centering
\caption{Move probabilities by layer for puzzle \href{https://lichess.org/training/70v1B}{\texttt{70v1B}} (Part 2: Layer 7 to Final)}
\label{tab:puzzle_70v1B_probs2}
\resizebox{\textwidth}{!}{%
\begin{tabular}{lrrrrrrrr}
\toprule
\textbf{Move} & \textbf{Layer 7} & \textbf{Layer 8} & \textbf{Layer 9} & \textbf{Layer 10} & \textbf{Layer 11} & \textbf{Layer 12} & \textbf{Layer 13} & \textbf{Final} \\
\midrule
\wmove{Qxe8+} & \cellcolor{heatmapcolor!43!white}42.88\% & \cellcolor{heatmapcolor!60!white}60.35\% & \cellcolor{heatmapcolor!75!white}74.76\% & \cellcolor{heatmapcolor!69!white}69.45\% & \cellcolor{heatmapcolor!78!white}78.19\% & \cellcolor{heatmapcolor!75!white}75.30\% & \cellcolor{heatmapcolor!54!white}53.67\% & \cellcolor{heatmapcolor!89!white}88.91\% \\
\wmove{Qe7} & 0.11\% & 0.11\% & 0.12\% & 0.30\% & 0.43\% & 0.21\% & 0.26\% & 0.33\% \\
\wmove{bxc3} & \cellcolor{heatmapcolor!53!white}52.88\% & \cellcolor{heatmapcolor!36!white}35.65\% & \cellcolor{heatmapcolor!21!white}20.83\% & \cellcolor{heatmapcolor!25!white}25.08\% & \cellcolor{heatmapcolor!17!white}17.09\% & \cellcolor{heatmapcolor!20!white}20.41\% & \cellcolor{heatmapcolor!36!white}35.91\% & 0.28\% \\
\wmove{Qe6} & 0.13\% & 0.07\% & 0.08\% & 0.08\% & 0.15\% & 0.12\% & 0.25\% & 0.41\% \\
\wmove{h4} & 0.38\% & 0.25\% & 0.22\% & \cellcolor{heatmapcolor!1!white}0.86\% & 0.28\% & 0.14\% & 0.31\% & 0.34\% \\
\wmove{b4} & 0.05\% & 0.11\% & 0.47\% & \cellcolor{heatmapcolor!1!white}0.56\% & 0.10\% & 0.22\% & \cellcolor{heatmapcolor!1!white}1.02\% & 0.33\% \\
\wmove{a4} & 0.14\% & 0.10\% & 0.04\% & 0.09\% & 0.10\% & 0.17\% & \cellcolor{heatmapcolor!1!white}0.97\% & 0.29\% \\
\wmove{Qe5} & 0.11\% & 0.09\% & 0.13\% & 0.14\% & 0.24\% & 0.23\% & 0.22\% & 0.37\% \\
\wmove{Qf2} & 0.05\% & 0.04\% & 0.04\% & 0.04\% & 0.04\% & 0.07\% & 0.15\% & 0.24\% \\
\wmove{Qc4} & 0.17\% & 0.09\% & 0.16\% & 0.14\% & 0.12\% & 0.10\% & 0.33\% & 0.45\% \\
\wmove{f5} & 0.29\% & \cellcolor{heatmapcolor!1!white}1.17\% & 0.44\% & 0.36\% & 0.45\% & 0.22\% & \cellcolor{heatmapcolor!1!white}0.56\% & 0.30\% \\
\wmove{Rb1} & 0.08\% & 0.05\% & 0.13\% & 0.19\% & 0.17\% & 0.12\% & 0.37\% & 0.37\% \\
\wmove{Qg2} & 0.09\% & 0.04\% & 0.04\% & 0.04\% & 0.06\% & 0.10\% & 0.16\% & 0.27\% \\
\wmove{a3} & 0.18\% & 0.11\% & 0.10\% & 0.12\% & 0.07\% & 0.07\% & \cellcolor{heatmapcolor!1!white}0.50\% & 0.32\% \\
\wmove{g4} & 0.13\% & 0.14\% & 0.25\% & 0.44\% & 0.23\% & 0.22\% & \cellcolor{heatmapcolor!1!white}0.75\% & 0.30\% \\
\wmove{Qb5} & 0.13\% & 0.12\% & 0.10\% & 0.08\% & 0.08\% & 0.10\% & 0.21\% & 0.21\% \\
\wmove{Qg4} & 0.13\% & 0.08\% & 0.10\% & 0.11\% & 0.10\% & 0.06\% & 0.20\% & 0.35\% \\
\wmove{Rd1} & 0.09\% & 0.06\% & 0.10\% & 0.06\% & 0.06\% & 0.06\% & 0.26\% & 0.27\% \\
\wmove{Qd1} & 0.11\% & 0.05\% & 0.08\% & 0.12\% & 0.12\% & 0.06\% & 0.17\% & 0.29\% \\
\wmove{Be3} & 0.15\% & 0.12\% & 0.20\% & 0.10\% & 0.06\% & 0.17\% & 0.49\% & 0.21\% \\
\wmove{Qf3} & 0.12\% & 0.05\% & 0.09\% & 0.09\% & 0.10\% & 0.07\% & 0.19\% & 0.34\% \\
\wmove{Qe3} & 0.07\% & 0.16\% & 0.12\% & 0.06\% & 0.12\% & 0.26\% & 0.23\% & 0.21\% \\
\wmove{b3} & 0.33\% & 0.15\% & 0.33\% & 0.25\% & 0.19\% & 0.09\% & 0.39\% & 0.41\% \\
\wmove{Qe4} & 0.13\% & 0.08\% & 0.13\% & 0.18\% & 0.27\% & 0.21\% & 0.23\% & 0.40\% \\
\wmove{Qa6} & 0.07\% & 0.09\% & 0.09\% & 0.10\% & 0.11\% & 0.06\% & 0.15\% & 0.38\% \\
\wmove{Qc2} & 0.20\% & 0.09\% & 0.14\% & 0.16\% & 0.15\% & 0.10\% & 0.21\% & 0.34\% \\
\wmove{Qh5} & 0.14\% & 0.10\% & 0.10\% & 0.14\% & 0.12\% & 0.07\% & 0.23\% & 0.37\% \\
\wmove{h3} & 0.10\% & 0.05\% & 0.06\% & 0.14\% & 0.07\% & 0.04\% & 0.27\% & 0.33\% \\
\wmove{Kh1} & 0.08\% & 0.06\% & 0.05\% & 0.03\% & 0.05\% & 0.07\% & 0.09\% & 0.33\% \\
\wmove{Kf2} & 0.04\% & 0.03\% & 0.04\% & 0.07\% & 0.09\% & 0.15\% & 0.10\% & 0.24\% \\
\wmove{Qd3} & 0.14\% & 0.12\% & 0.12\% & 0.08\% & 0.11\% & 0.11\% & 0.22\% & 0.28\% \\
\wmove{Kg2} & 0.07\% & 0.06\% & 0.05\% & 0.09\% & 0.19\% & 0.20\% & 0.11\% & 0.27\% \\
\wmove{Qd2} & 0.02\% & 0.03\% & 0.05\% & 0.02\% & 0.02\% & 0.04\% & 0.08\% & 0.27\% \\
\wmove{Bd2} & 0.10\% & 0.08\% & 0.16\% & 0.10\% & 0.07\% & 0.12\% & 0.27\% & 0.23\% \\
\wmove{Kf1} & 0.02\% & 0.02\% & 0.03\% & 0.05\% & 0.10\% & 0.11\% & 0.08\% & 0.26\% \\
\wmove{Qf1} & 0.05\% & 0.02\% & 0.03\% & 0.05\% & 0.10\% & 0.09\% & 0.17\% & 0.26\% \\
\wmove{Rf1} & 0.03\% & 0.01\% & 0.02\% & 0.02\% & 0.03\% & 0.05\% & 0.23\% & 0.25\% \\
\bottomrule
\end{tabular}%
}
\end{table*}

%% file: Figures/Puzzles/Examples/puzzle_tables_0KBPy.tex
\begin{table*}[h!]
\centering
\caption{Move probabilities by layer for puzzle \href{https://lichess.org/training/0KBPy}{\texttt{0KBPy}} (Part 1: Input to Layer 6)}
\label{tab:puzzle_0KBPy_probs1}
\resizebox{\textwidth}{!}{%
\begin{tabular}{lrrrrrrrr}
\toprule
\textbf{Move} & \textbf{Input} & \textbf{Layer 0} & \textbf{Layer 1} & \textbf{Layer 2} & \textbf{Layer 3} & \textbf{Layer 4} & \textbf{Layer 5} & \textbf{Layer 6} \\
\midrule
\wmove{Ng6+} & \cellcolor{heatmapcolor!10!white}9.76\% & 0.46\% & \cellcolor{heatmapcolor!1!white}0.88\% & 0.35\% & \cellcolor{heatmapcolor!6!white}6.34\% & 0.12\% & \cellcolor{heatmapcolor!1!white}1.38\% & \cellcolor{heatmapcolor!1!white}0.77\% \\
\wmove{Rc4} & 0.15\% & 0.32\% & 0.36\% & \cellcolor{heatmapcolor!1!white}1.20\% & \cellcolor{heatmapcolor!15!white}14.79\% & \cellcolor{heatmapcolor!69!white}69.39\% & \cellcolor{heatmapcolor!49!white}48.78\% & \cellcolor{heatmapcolor!42!white}41.74\% \\
\wmove{Qg8+} & 0.07\% & \cellcolor{heatmapcolor!40!white}40.00\% & \cellcolor{heatmapcolor!7!white}7.11\% & \cellcolor{heatmapcolor!18!white}18.17\% & \cellcolor{heatmapcolor!5!white}4.87\% & 0.36\% & \cellcolor{heatmapcolor!1!white}0.60\% & \cellcolor{heatmapcolor!1!white}0.77\% \\
\wmove{Ne6} & \cellcolor{heatmapcolor!19!white}19.23\% & \cellcolor{heatmapcolor!6!white}6.12\% & \cellcolor{heatmapcolor!30!white}30.11\% & \cellcolor{heatmapcolor!7!white}7.38\% & \cellcolor{heatmapcolor!9!white}8.51\% & \cellcolor{heatmapcolor!1!white}0.56\% & \cellcolor{heatmapcolor!11!white}10.91\% & \cellcolor{heatmapcolor!11!white}10.97\% \\
\wmove{h4} & 0.09\% & \cellcolor{heatmapcolor!7!white}6.94\% & 0.16\% & 0.36\% & 0.28\% & 0.07\% & 0.23\% & 0.04\% \\
\wmove{Qa4} & 0.05\% & \cellcolor{heatmapcolor!3!white}2.77\% & \cellcolor{heatmapcolor!6!white}5.95\% & \cellcolor{heatmapcolor!10!white}10.38\% & \cellcolor{heatmapcolor!16!white}16.30\% & \cellcolor{heatmapcolor!3!white}2.73\% & \cellcolor{heatmapcolor!6!white}5.80\% & \cellcolor{heatmapcolor!3!white}3.49\% \\
\wmove{Qxb5} & \cellcolor{heatmapcolor!16!white}16.21\% & \cellcolor{heatmapcolor!8!white}8.15\% & \cellcolor{heatmapcolor!8!white}7.62\% & \cellcolor{heatmapcolor!6!white}6.20\% & \cellcolor{heatmapcolor!3!white}2.83\% & \cellcolor{heatmapcolor!3!white}3.07\% & \cellcolor{heatmapcolor!3!white}2.87\% & \cellcolor{heatmapcolor!1!white}1.41\% \\
\wmove{Rd7} & \cellcolor{heatmapcolor!15!white}15.49\% & 0.31\% & 0.19\% & 0.42\% & \cellcolor{heatmapcolor!1!white}1.07\% & \cellcolor{heatmapcolor!1!white}0.75\% & 0.35\% & \cellcolor{heatmapcolor!1!white}0.89\% \\
\wmove{Qc2} & 0.38\% & \cellcolor{heatmapcolor!1!white}0.71\% & \cellcolor{heatmapcolor!2!white}1.63\% & \cellcolor{heatmapcolor!14!white}13.84\% & \cellcolor{heatmapcolor!5!white}4.56\% & \cellcolor{heatmapcolor!1!white}0.78\% & 0.24\% & \cellcolor{heatmapcolor!1!white}0.88\% \\
\wmove{Qe6} & \cellcolor{heatmapcolor!2!white}1.77\% & \cellcolor{heatmapcolor!4!white}4.49\% & \cellcolor{heatmapcolor!1!white}1.31\% & 0.20\% & \cellcolor{heatmapcolor!1!white}0.70\% & 0.16\% & \cellcolor{heatmapcolor!1!white}1.50\% & \cellcolor{heatmapcolor!5!white}5.27\% \\
\wmove{Ra4} & 0.12\% & \cellcolor{heatmapcolor!1!white}0.51\% & \cellcolor{heatmapcolor!1!white}1.20\% & \cellcolor{heatmapcolor!2!white}1.83\% & \cellcolor{heatmapcolor!4!white}3.50\% & \cellcolor{heatmapcolor!2!white}2.12\% & \cellcolor{heatmapcolor!13!white}12.92\% & \cellcolor{heatmapcolor!7!white}6.79\% \\
\wmove{Qb4} & 0.19\% & \cellcolor{heatmapcolor!12!white}12.14\% & \cellcolor{heatmapcolor!11!white}11.15\% & \cellcolor{heatmapcolor!3!white}3.00\% & \cellcolor{heatmapcolor!3!white}2.59\% & 0.49\% & 0.36\% & 0.37\% \\
\wmove{Rd6} & \cellcolor{heatmapcolor!10!white}10.35\% & 0.26\% & 0.23\% & \cellcolor{heatmapcolor!1!white}0.53\% & \cellcolor{heatmapcolor!2!white}1.96\% & \cellcolor{heatmapcolor!1!white}1.40\% & 0.44\% & \cellcolor{heatmapcolor!1!white}0.74\% \\
\wmove{Qd3} & 0.18\% & 0.28\% & \cellcolor{heatmapcolor!1!white}0.59\% & 0.45\% & 0.44\% & 0.22\% & 0.19\% & 0.31\% \\
\wmove{Nd3} & 0.33\% & 0.18\% & 0.06\% & \cellcolor{heatmapcolor!1!white}0.62\% & \cellcolor{heatmapcolor!1!white}1.27\% & 0.14\% & \cellcolor{heatmapcolor!2!white}1.53\% & \cellcolor{heatmapcolor!1!white}1.31\% \\
\wmove{Rb4} & 0.09\% & \cellcolor{heatmapcolor!3!white}2.99\% & \cellcolor{heatmapcolor!9!white}9.44\% & \cellcolor{heatmapcolor!1!white}0.75\% & \cellcolor{heatmapcolor!1!white}0.50\% & 0.25\% & 0.04\% & 0.04\% \\
\wmove{Nd5} & \cellcolor{heatmapcolor!9!white}9.23\% & 0.10\% & 0.21\% & \cellcolor{heatmapcolor!2!white}2.13\% & \cellcolor{heatmapcolor!3!white}3.00\% & 0.24\% & \cellcolor{heatmapcolor!1!white}0.92\% & \cellcolor{heatmapcolor!1!white}0.73\% \\
\wmove{Qc4} & 0.31\% & \cellcolor{heatmapcolor!2!white}1.70\% & \cellcolor{heatmapcolor!5!white}5.11\% & \cellcolor{heatmapcolor!8!white}8.20\% & \cellcolor{heatmapcolor!4!white}4.18\% & \cellcolor{heatmapcolor!1!white}1.06\% & \cellcolor{heatmapcolor!2!white}2.18\% & \cellcolor{heatmapcolor!3!white}2.54\% \\
\wmove{Re4} & 0.15\% & 0.29\% & \cellcolor{heatmapcolor!1!white}1.43\% & \cellcolor{heatmapcolor!2!white}2.45\% & \cellcolor{heatmapcolor!3!white}2.82\% & \cellcolor{heatmapcolor!1!white}0.97\% & \cellcolor{heatmapcolor!1!white}1.38\% & \cellcolor{heatmapcolor!7!white}7.30\% \\
\wmove{a4} & 0.08\% & \cellcolor{heatmapcolor!3!white}3.03\% & 0.08\% & 0.24\% & 0.26\% & 0.11\% & 0.22\% & 0.07\% \\
\wmove{Kh2} & 0.20\% & \cellcolor{heatmapcolor!1!white}0.93\% & 0.20\% & 0.38\% & 0.18\% & 0.19\% & \cellcolor{heatmapcolor!1!white}0.66\% & \cellcolor{heatmapcolor!2!white}2.47\% \\
\wmove{Rd8} & \cellcolor{heatmapcolor!3!white}2.55\% & 0.12\% & 0.06\% & 0.31\% & \cellcolor{heatmapcolor!1!white}0.67\% & \cellcolor{heatmapcolor!2!white}1.80\% & 0.30\% & \cellcolor{heatmapcolor!3!white}3.34\% \\
\wmove{Rd2} & 0.18\% & 0.20\% & 0.24\% & \cellcolor{heatmapcolor!1!white}0.73\% & \cellcolor{heatmapcolor!2!white}2.22\% & \cellcolor{heatmapcolor!5!white}4.69\% & \cellcolor{heatmapcolor!1!white}1.23\% & \cellcolor{heatmapcolor!1!white}1.35\% \\
\wmove{Qf7} & \cellcolor{heatmapcolor!2!white}1.78\% & \cellcolor{heatmapcolor!1!white}0.73\% & \cellcolor{heatmapcolor!4!white}3.87\% & \cellcolor{heatmapcolor!4!white}4.31\% & \cellcolor{heatmapcolor!2!white}2.14\% & 0.47\% & 0.19\% & 0.20\% \\
\wmove{Kf1} & 0.13\% & 0.38\% & 0.12\% & \cellcolor{heatmapcolor!1!white}0.63\% & \cellcolor{heatmapcolor!2!white}1.72\% & \cellcolor{heatmapcolor!1!white}1.18\% & 0.41\% & \cellcolor{heatmapcolor!1!white}0.87\% \\
\wmove{Rd5} & \cellcolor{heatmapcolor!4!white}3.77\% & 0.27\% & \cellcolor{heatmapcolor!2!white}2.25\% & \cellcolor{heatmapcolor!1!white}1.07\% & 0.19\% & 0.07\% & 0.14\% & \cellcolor{heatmapcolor!1!white}0.65\% \\
\wmove{Qd1} & 0.12\% & 0.38\% & 0.27\% & \cellcolor{heatmapcolor!1!white}0.52\% & \cellcolor{heatmapcolor!1!white}1.35\% & \cellcolor{heatmapcolor!1!white}0.80\% & 0.28\% & 0.28\% \\
\wmove{Qc3} & 0.10\% & \cellcolor{heatmapcolor!1!white}0.51\% & \cellcolor{heatmapcolor!1!white}0.70\% & \cellcolor{heatmapcolor!3!white}3.10\% & \cellcolor{heatmapcolor!2!white}2.00\% & \cellcolor{heatmapcolor!1!white}1.10\% & 0.45\% & \cellcolor{heatmapcolor!1!white}0.82\% \\
\wmove{e4} & \cellcolor{heatmapcolor!1!white}0.54\% & 0.49\% & \cellcolor{heatmapcolor!1!white}0.91\% & \cellcolor{heatmapcolor!3!white}3.02\% & \cellcolor{heatmapcolor!1!white}0.92\% & 0.08\% & 0.10\% & 0.22\% \\
\wmove{Qa2} & 0.03\% & \cellcolor{heatmapcolor!1!white}1.39\% & \cellcolor{heatmapcolor!2!white}1.96\% & \cellcolor{heatmapcolor!3!white}2.76\% & \cellcolor{heatmapcolor!3!white}2.71\% & \cellcolor{heatmapcolor!1!white}1.38\% & \cellcolor{heatmapcolor!1!white}1.00\% & \cellcolor{heatmapcolor!1!white}1.24\% \\
\wmove{Nh5} & \cellcolor{heatmapcolor!2!white}2.21\% & 0.49\% & 0.05\% & 0.18\% & \cellcolor{heatmapcolor!1!white}0.51\% & 0.39\% & 0.16\% & 0.08\% \\
\wmove{f3} & 0.17\% & 0.12\% & 0.20\% & \cellcolor{heatmapcolor!1!white}0.60\% & \cellcolor{heatmapcolor!2!white}1.83\% & 0.13\% & 0.11\% & 0.08\% \\
\wmove{Ne2} & \cellcolor{heatmapcolor!1!white}1.09\% & 0.13\% & \cellcolor{heatmapcolor!1!white}0.90\% & 0.13\% & 0.05\% & 0.07\% & 0.27\% & 0.21\% \\
\wmove{Rd3} & 0.11\% & 0.20\% & \cellcolor{heatmapcolor!1!white}1.19\% & \cellcolor{heatmapcolor!1!white}0.63\% & 0.16\% & 0.08\% & 0.17\% & 0.08\% \\
\wmove{Kh1} & 0.34\% & \cellcolor{heatmapcolor!1!white}0.59\% & \cellcolor{heatmapcolor!1!white}0.71\% & \cellcolor{heatmapcolor!1!white}1.13\% & 0.27\% & \cellcolor{heatmapcolor!1!white}1.10\% & \cellcolor{heatmapcolor!1!white}0.59\% & 0.41\% \\
\wmove{g4} & \cellcolor{heatmapcolor!1!white}1.09\% & 0.48\% & 0.13\% & \cellcolor{heatmapcolor!1!white}0.69\% & \cellcolor{heatmapcolor!1!white}0.69\% & 0.11\% & 0.25\% & 0.25\% \\
\wmove{Rd1} & 0.16\% & 0.17\% & 0.26\% & 0.32\% & 0.41\% & \cellcolor{heatmapcolor!1!white}1.06\% & \cellcolor{heatmapcolor!1!white}0.53\% & \cellcolor{heatmapcolor!1!white}0.54\% \\
\wmove{Qd5} & \cellcolor{heatmapcolor!1!white}1.04\% & 0.23\% & \cellcolor{heatmapcolor!1!white}1.04\% & 0.50\% & 0.40\% & 0.22\% & 0.22\% & 0.26\% \\
\wmove{g3} & 0.17\% & 0.41\% & 0.09\% & 0.30\% & \cellcolor{heatmapcolor!1!white}0.80\% & 0.10\% & 0.12\% & 0.24\% \\
\bottomrule
\end{tabular}%
}
\end{table*}

\begin{table*}[h!]
\centering
\caption{Move probabilities by layer for puzzle \href{https://lichess.org/training/0KBPy}{\texttt{0KBPy}} (Part 2: Layer 7 to Final)}
\label{tab:puzzle_0KBPy_probs2}
\resizebox{\textwidth}{!}{%
\begin{tabular}{lrrrrrrrr}
\toprule
\textbf{Move} & \textbf{Layer 7} & \textbf{Layer 8} & \textbf{Layer 9} & \textbf{Layer 10} & \textbf{Layer 11} & \textbf{Layer 12} & \textbf{Layer 13} & \textbf{Final} \\
\midrule
\wmove{Ng6+} & \cellcolor{heatmapcolor!3!white}3.14\% & \cellcolor{heatmapcolor!11!white}10.83\% & \cellcolor{heatmapcolor!15!white}14.71\% & \cellcolor{heatmapcolor!16!white}16.18\% & \cellcolor{heatmapcolor!35!white}35.31\% & \cellcolor{heatmapcolor!53!white}53.13\% & \cellcolor{heatmapcolor!33!white}33.17\% & \cellcolor{heatmapcolor!71!white}70.65\% \\
\wmove{Rc4} & \cellcolor{heatmapcolor!33!white}33.25\% & \cellcolor{heatmapcolor!15!white}14.94\% & \cellcolor{heatmapcolor!13!white}13.26\% & \cellcolor{heatmapcolor!5!white}5.06\% & \cellcolor{heatmapcolor!1!white}1.01\% & 0.19\% & 0.28\% & 0.18\% \\
\wmove{Qg8+} & \cellcolor{heatmapcolor!2!white}1.51\% & \cellcolor{heatmapcolor!2!white}1.85\% & \cellcolor{heatmapcolor!6!white}5.92\% & \cellcolor{heatmapcolor!10!white}9.55\% & \cellcolor{heatmapcolor!1!white}1.02\% & \cellcolor{heatmapcolor!1!white}0.73\% & \cellcolor{heatmapcolor!1!white}0.52\% & 0.16\% \\
\wmove{Ne6} & \cellcolor{heatmapcolor!8!white}8.39\% & \cellcolor{heatmapcolor!12!white}12.37\% & \cellcolor{heatmapcolor!11!white}10.91\% & \cellcolor{heatmapcolor!10!white}10.09\% & \cellcolor{heatmapcolor!10!white}10.47\% & \cellcolor{heatmapcolor!9!white}9.26\% & \cellcolor{heatmapcolor!12!white}12.17\% & \cellcolor{heatmapcolor!1!white}1.46\% \\
\wmove{h4} & 0.04\% & 0.04\% & 0.01\% & 0.04\% & \cellcolor{heatmapcolor!1!white}1.05\% & \cellcolor{heatmapcolor!2!white}1.58\% & \cellcolor{heatmapcolor!18!white}17.72\% & \cellcolor{heatmapcolor!5!white}4.76\% \\
\wmove{Qa4} & \cellcolor{heatmapcolor!2!white}2.40\% & \cellcolor{heatmapcolor!3!white}3.49\% & \cellcolor{heatmapcolor!1!white}1.38\% & \cellcolor{heatmapcolor!3!white}2.76\% & \cellcolor{heatmapcolor!3!white}2.78\% & 0.47\% & 0.31\% & 0.14\% \\
\wmove{Qxb5} & \cellcolor{heatmapcolor!2!white}1.99\% & \cellcolor{heatmapcolor!5!white}4.93\% & \cellcolor{heatmapcolor!3!white}2.81\% & \cellcolor{heatmapcolor!2!white}2.12\% & \cellcolor{heatmapcolor!1!white}0.64\% & \cellcolor{heatmapcolor!1!white}0.62\% & \cellcolor{heatmapcolor!1!white}1.30\% & 0.16\% \\
\wmove{Rd7} & \cellcolor{heatmapcolor!1!white}1.08\% & \cellcolor{heatmapcolor!1!white}0.99\% & \cellcolor{heatmapcolor!1!white}1.22\% & \cellcolor{heatmapcolor!2!white}2.37\% & \cellcolor{heatmapcolor!4!white}3.61\% & \cellcolor{heatmapcolor!1!white}1.34\% & 0.26\% & 0.21\% \\
\wmove{Qc2} & \cellcolor{heatmapcolor!1!white}1.09\% & \cellcolor{heatmapcolor!3!white}2.72\% & \cellcolor{heatmapcolor!6!white}5.97\% & \cellcolor{heatmapcolor!7!white}6.75\% & \cellcolor{heatmapcolor!1!white}1.36\% & 0.28\% & 0.21\% & 0.16\% \\
\wmove{Qe6} & \cellcolor{heatmapcolor!9!white}8.67\% & \cellcolor{heatmapcolor!13!white}13.05\% & \cellcolor{heatmapcolor!14!white}13.50\% & \cellcolor{heatmapcolor!14!white}13.51\% & \cellcolor{heatmapcolor!9!white}8.71\% & \cellcolor{heatmapcolor!3!white}2.56\% & \cellcolor{heatmapcolor!3!white}3.14\% & 0.48\% \\
\wmove{Ra4} & \cellcolor{heatmapcolor!6!white}6.37\% & \cellcolor{heatmapcolor!5!white}4.91\% & 0.40\% & \cellcolor{heatmapcolor!2!white}2.20\% & \cellcolor{heatmapcolor!2!white}1.72\% & 0.23\% & 0.12\% & 0.15\% \\
\wmove{Qb4} & 0.48\% & \cellcolor{heatmapcolor!1!white}0.85\% & \cellcolor{heatmapcolor!1!white}0.70\% & \cellcolor{heatmapcolor!1!white}0.98\% & \cellcolor{heatmapcolor!1!white}0.74\% & \cellcolor{heatmapcolor!1!white}0.77\% & \cellcolor{heatmapcolor!1!white}0.58\% & \cellcolor{heatmapcolor!1!white}0.71\% \\
\wmove{Rd6} & 0.46\% & 0.31\% & 0.45\% & 0.50\% & \cellcolor{heatmapcolor!1!white}0.67\% & 0.48\% & 0.14\% & 0.18\% \\
\wmove{Qd3} & \cellcolor{heatmapcolor!1!white}1.20\% & \cellcolor{heatmapcolor!1!white}1.38\% & \cellcolor{heatmapcolor!3!white}2.80\% & \cellcolor{heatmapcolor!4!white}4.34\% & \cellcolor{heatmapcolor!7!white}7.01\% & \cellcolor{heatmapcolor!10!white}9.79\% & \cellcolor{heatmapcolor!6!white}6.29\% & \cellcolor{heatmapcolor!4!white}4.04\% \\
\wmove{Nd3} & \cellcolor{heatmapcolor!2!white}2.31\% & \cellcolor{heatmapcolor!3!white}2.88\% & \cellcolor{heatmapcolor!4!white}4.05\% & \cellcolor{heatmapcolor!2!white}2.28\% & \cellcolor{heatmapcolor!4!white}3.70\% & \cellcolor{heatmapcolor!3!white}3.28\% & \cellcolor{heatmapcolor!10!white}9.63\% & 0.49\% \\
\wmove{Rb4} & 0.04\% & 0.06\% & 0.04\% & 0.09\% & 0.10\% & 0.09\% & 0.05\% & 0.15\% \\
\wmove{Nd5} & \cellcolor{heatmapcolor!1!white}0.95\% & \cellcolor{heatmapcolor!1!white}0.92\% & 0.48\% & 0.29\% & 0.24\% & 0.24\% & \cellcolor{heatmapcolor!1!white}0.50\% & 0.15\% \\
\wmove{Qc4} & \cellcolor{heatmapcolor!1!white}1.41\% & \cellcolor{heatmapcolor!1!white}1.06\% & \cellcolor{heatmapcolor!1!white}0.93\% & \cellcolor{heatmapcolor!1!white}0.54\% & 0.30\% & 0.24\% & 0.22\% & 0.17\% \\
\wmove{Re4} & \cellcolor{heatmapcolor!7!white}6.63\% & \cellcolor{heatmapcolor!3!white}3.24\% & \cellcolor{heatmapcolor!4!white}4.44\% & \cellcolor{heatmapcolor!5!white}4.90\% & \cellcolor{heatmapcolor!3!white}3.39\% & \cellcolor{heatmapcolor!2!white}1.89\% & \cellcolor{heatmapcolor!1!white}0.50\% & 0.19\% \\
\wmove{a4} & 0.18\% & 0.08\% & 0.01\% & 0.03\% & 0.09\% & 0.03\% & \cellcolor{heatmapcolor!1!white}1.02\% & \cellcolor{heatmapcolor!6!white}5.80\% \\
\wmove{Kh2} & \cellcolor{heatmapcolor!5!white}5.48\% & \cellcolor{heatmapcolor!4!white}3.97\% & \cellcolor{heatmapcolor!3!white}2.81\% & \cellcolor{heatmapcolor!3!white}2.75\% & \cellcolor{heatmapcolor!1!white}0.81\% & \cellcolor{heatmapcolor!3!white}2.87\% & \cellcolor{heatmapcolor!3!white}3.23\% & \cellcolor{heatmapcolor!6!white}5.54\% \\
\wmove{Rd8} & \cellcolor{heatmapcolor!2!white}2.37\% & \cellcolor{heatmapcolor!5!white}5.16\% & \cellcolor{heatmapcolor!2!white}1.86\% & \cellcolor{heatmapcolor!2!white}1.57\% & \cellcolor{heatmapcolor!2!white}2.43\% & \cellcolor{heatmapcolor!1!white}0.85\% & 0.37\% & 0.17\% \\
\wmove{Rd2} & \cellcolor{heatmapcolor!1!white}1.06\% & \cellcolor{heatmapcolor!1!white}0.74\% & \cellcolor{heatmapcolor!1!white}1.34\% & 0.25\% & 0.14\% & 0.19\% & 0.42\% & 0.25\% \\
\wmove{Qf7} & 0.41\% & \cellcolor{heatmapcolor!1!white}0.76\% & \cellcolor{heatmapcolor!1!white}0.87\% & \cellcolor{heatmapcolor!2!white}1.69\% & \cellcolor{heatmapcolor!1!white}1.04\% & 0.30\% & 0.31\% & 0.20\% \\
\wmove{Kf1} & \cellcolor{heatmapcolor!2!white}1.68\% & \cellcolor{heatmapcolor!2!white}2.08\% & \cellcolor{heatmapcolor!2!white}2.11\% & \cellcolor{heatmapcolor!2!white}1.84\% & \cellcolor{heatmapcolor!4!white}4.25\% & \cellcolor{heatmapcolor!3!white}3.32\% & 0.32\% & 0.15\% \\
\wmove{Rd5} & \cellcolor{heatmapcolor!1!white}0.94\% & \cellcolor{heatmapcolor!1!white}0.62\% & 0.33\% & 0.35\% & \cellcolor{heatmapcolor!1!white}0.64\% & \cellcolor{heatmapcolor!1!white}0.62\% & 0.42\% & 0.22\% \\
\wmove{Qd1} & \cellcolor{heatmapcolor!1!white}1.01\% & \cellcolor{heatmapcolor!1!white}0.93\% & \cellcolor{heatmapcolor!2!white}1.71\% & \cellcolor{heatmapcolor!2!white}1.89\% & \cellcolor{heatmapcolor!3!white}3.13\% & \cellcolor{heatmapcolor!1!white}1.05\% & \cellcolor{heatmapcolor!1!white}0.62\% & 0.46\% \\
\wmove{Qc3} & \cellcolor{heatmapcolor!1!white}0.63\% & \cellcolor{heatmapcolor!1!white}0.55\% & \cellcolor{heatmapcolor!1!white}0.65\% & \cellcolor{heatmapcolor!1!white}0.87\% & 0.45\% & 0.29\% & 0.17\% & 0.15\% \\
\wmove{e4} & 0.31\% & 0.30\% & 0.50\% & \cellcolor{heatmapcolor!1!white}0.64\% & 0.46\% & 0.48\% & \cellcolor{heatmapcolor!1!white}0.61\% & 0.18\% \\
\wmove{Qa2} & \cellcolor{heatmapcolor!2!white}1.54\% & \cellcolor{heatmapcolor!1!white}1.18\% & \cellcolor{heatmapcolor!1!white}1.05\% & \cellcolor{heatmapcolor!1!white}0.78\% & 0.45\% & 0.15\% & 0.34\% & 0.27\% \\
\wmove{Nh5} & 0.17\% & 0.17\% & 0.05\% & 0.19\% & 0.11\% & 0.16\% & 0.37\% & 0.18\% \\
\wmove{f3} & 0.30\% & 0.17\% & 0.21\% & 0.21\% & 0.17\% & 0.12\% & \cellcolor{heatmapcolor!1!white}0.57\% & 0.18\% \\
\wmove{Ne2} & 0.47\% & 0.30\% & \cellcolor{heatmapcolor!1!white}0.50\% & 0.38\% & \cellcolor{heatmapcolor!1!white}0.50\% & \cellcolor{heatmapcolor!1!white}0.67\% & \cellcolor{heatmapcolor!1!white}1.46\% & 0.18\% \\
\wmove{Rd3} & 0.10\% & 0.09\% & 0.05\% & 0.08\% & 0.06\% & 0.08\% & 0.24\% & 0.19\% \\
\wmove{Kh1} & \cellcolor{heatmapcolor!1!white}0.54\% & \cellcolor{heatmapcolor!1!white}0.79\% & \cellcolor{heatmapcolor!1!white}0.83\% & \cellcolor{heatmapcolor!1!white}0.95\% & 0.48\% & \cellcolor{heatmapcolor!1!white}0.69\% & 0.45\% & 0.19\% \\
\wmove{g4} & 0.29\% & 0.24\% & 0.11\% & 0.17\% & 0.23\% & 0.13\% & 0.25\% & 0.40\% \\
\wmove{Rd1} & \cellcolor{heatmapcolor!1!white}0.57\% & 0.36\% & 0.47\% & 0.15\% & 0.09\% & 0.30\% & \cellcolor{heatmapcolor!1!white}0.76\% & 0.25\% \\
\wmove{Qd5} & 0.45\% & \cellcolor{heatmapcolor!1!white}0.64\% & \cellcolor{heatmapcolor!1!white}0.55\% & \cellcolor{heatmapcolor!1!white}0.63\% & \cellcolor{heatmapcolor!1!white}0.63\% & 0.47\% & \cellcolor{heatmapcolor!1!white}0.68\% & 0.16\% \\
\wmove{g3} & 0.11\% & 0.05\% & 0.03\% & 0.03\% & 0.01\% & 0.06\% & 0.26\% & 0.29\% \\
\bottomrule
\end{tabular}%
}
\end{table*}

%% file: Figures/Puzzles/Examples/puzzle_tables_09SS5.tex
\begin{table*}[h!]
\centering
\caption{Move probabilities by layer for puzzle \href{https://lichess.org/training/09SS5}{\texttt{09SS5}} (Part 1: Input to Layer 6)}
\label{tab:puzzle_09SS5_probs1}
\resizebox{\textwidth}{!}{%
\begin{tabular}{lrrrrrrrr}
\toprule
\textbf{Move} & \textbf{Input} & \textbf{Layer 0} & \textbf{Layer 1} & \textbf{Layer 2} & \textbf{Layer 3} & \textbf{Layer 4} & \textbf{Layer 5} & \textbf{Layer 6} \\
\midrule
\wmove{Qe8+} & \cellcolor{heatmapcolor!2!white}1.73\% & 0.11\% & 0.16\% & 0.24\% & \cellcolor{heatmapcolor!1!white}1.28\% & 0.02\% & 0.05\% & \cellcolor{heatmapcolor!1!white}1.11\% \\
\wmove{gxf7+} & \cellcolor{heatmapcolor!14!white}13.74\% & \cellcolor{heatmapcolor!3!white}2.54\% & \cellcolor{heatmapcolor!30!white}29.80\% & \cellcolor{heatmapcolor!33!white}33.00\% & \cellcolor{heatmapcolor!16!white}15.57\% & \cellcolor{heatmapcolor!37!white}37.26\% & \cellcolor{heatmapcolor!36!white}36.43\% & \cellcolor{heatmapcolor!44!white}43.50\% \\
\wmove{Qxf7+} & \cellcolor{heatmapcolor!4!white}4.30\% & \cellcolor{heatmapcolor!31!white}31.00\% & \cellcolor{heatmapcolor!39!white}39.29\% & \cellcolor{heatmapcolor!29!white}28.88\% & \cellcolor{heatmapcolor!32!white}32.40\% & \cellcolor{heatmapcolor!35!white}35.43\% & \cellcolor{heatmapcolor!33!white}32.63\% & \cellcolor{heatmapcolor!31!white}30.50\% \\
\wmove{Qxa3} & 0.43\% & \cellcolor{heatmapcolor!34!white}33.82\% & \cellcolor{heatmapcolor!16!white}15.83\% & \cellcolor{heatmapcolor!24!white}23.85\% & \cellcolor{heatmapcolor!25!white}25.18\% & \cellcolor{heatmapcolor!13!white}13.23\% & \cellcolor{heatmapcolor!13!white}13.23\% & \cellcolor{heatmapcolor!10!white}10.04\% \\
\wmove{Qd7} & \cellcolor{heatmapcolor!31!white}30.62\% & 0.06\% & 0.18\% & 0.05\% & 0.03\% & 0.03\% & 0.02\% & 0.02\% \\
\wmove{Rxa3} & 0.06\% & \cellcolor{heatmapcolor!27!white}26.89\% & \cellcolor{heatmapcolor!10!white}10.22\% & \cellcolor{heatmapcolor!9!white}8.50\% & \cellcolor{heatmapcolor!21!white}21.20\% & \cellcolor{heatmapcolor!12!white}12.43\% & \cellcolor{heatmapcolor!16!white}15.65\% & \cellcolor{heatmapcolor!12!white}11.68\% \\
\wmove{Qc7} & \cellcolor{heatmapcolor!21!white}20.69\% & 0.06\% & 0.04\% & 0.11\% & 0.06\% & 0.04\% & 0.02\% & 0.02\% \\
\wmove{Qe4} & 0.01\% & 0.05\% & 0.06\% & 0.12\% & 0.06\% & 0.05\% & 0.06\% & 0.03\% \\
\wmove{Rh3} & 0.01\% & 0.13\% & 0.08\% & 0.16\% & 0.22\% & 0.03\% & 0.38\% & \cellcolor{heatmapcolor!1!white}1.42\% \\
\wmove{Qf6} & \cellcolor{heatmapcolor!7!white}6.82\% & 0.05\% & 0.14\% & 0.15\% & 0.10\% & 0.06\% & 0.15\% & 0.11\% \\
\wmove{Rf3} & 0.04\% & 0.08\% & 0.23\% & 0.15\% & 0.09\% & 0.02\% & 0.18\% & 0.29\% \\
\wmove{Qb7} & \cellcolor{heatmapcolor!1!white}1.29\% & 0.10\% & 0.05\% & 0.04\% & 0.02\% & 0.02\% & 0.02\% & 0.02\% \\
\wmove{Qd6} & \cellcolor{heatmapcolor!4!white}4.19\% & 0.05\% & 0.18\% & 0.05\% & 0.04\% & 0.03\% & 0.03\% & 0.02\% \\
\wmove{Qe6} & \cellcolor{heatmapcolor!4!white}3.74\% & 0.06\% & 0.08\% & 0.33\% & 0.08\% & 0.03\% & 0.05\% & 0.03\% \\
\wmove{Re4} & 0.05\% & 0.07\% & 0.06\% & 0.07\% & 0.12\% & 0.03\% & 0.12\% & 0.15\% \\
\wmove{Qg5} & 0.43\% & \cellcolor{heatmapcolor!3!white}2.70\% & \cellcolor{heatmapcolor!1!white}0.81\% & 0.21\% & 0.21\% & 0.09\% & 0.11\% & 0.07\% \\
\wmove{Qb4} & \cellcolor{heatmapcolor!2!white}2.24\% & 0.20\% & 0.20\% & \cellcolor{heatmapcolor!1!white}1.01\% & 0.38\% & 0.08\% & 0.03\% & 0.03\% \\
\wmove{Qf8+} & \cellcolor{heatmapcolor!2!white}1.79\% & 0.10\% & \cellcolor{heatmapcolor!1!white}0.74\% & 0.36\% & \cellcolor{heatmapcolor!2!white}1.50\% & 0.08\% & 0.10\% & 0.23\% \\
\wmove{Qd8+} & \cellcolor{heatmapcolor!2!white}1.67\% & 0.10\% & 0.21\% & 0.06\% & 0.09\% & 0.02\% & 0.02\% & 0.03\% \\
\wmove{Qh4} & \cellcolor{heatmapcolor!2!white}1.51\% & 0.27\% & 0.21\% & \cellcolor{heatmapcolor!1!white}0.82\% & 0.10\% & 0.05\% & 0.03\% & 0.02\% \\
\wmove{f4} & 0.42\% & 0.20\% & 0.07\% & 0.09\% & 0.07\% & 0.08\% & 0.04\% & 0.06\% \\
\wmove{Re6} & \cellcolor{heatmapcolor!1!white}1.18\% & 0.07\% & 0.06\% & 0.06\% & 0.06\% & 0.03\% & 0.03\% & 0.03\% \\
\wmove{Rg3} & 0.03\% & 0.11\% & 0.10\% & 0.11\% & 0.09\% & 0.02\% & 0.18\% & 0.18\% \\
\wmove{Qa7} & \cellcolor{heatmapcolor!1!white}0.59\% & 0.08\% & 0.06\% & 0.05\% & 0.05\% & 0.04\% & 0.02\% & 0.01\% \\
\wmove{Qc5} & \cellcolor{heatmapcolor!1!white}0.75\% & 0.07\% & 0.04\% & 0.43\% & 0.14\% & 0.06\% & 0.03\% & 0.03\% \\
\wmove{Re5} & \cellcolor{heatmapcolor!1!white}0.60\% & 0.08\% & 0.16\% & 0.07\% & 0.05\% & 0.04\% & 0.05\% & 0.05\% \\
\wmove{Qe5} & \cellcolor{heatmapcolor!1!white}0.59\% & 0.06\% & 0.11\% & 0.22\% & 0.09\% & 0.06\% & 0.04\% & 0.03\% \\
\wmove{Kg2} & 0.03\% & 0.06\% & 0.24\% & 0.39\% & 0.28\% & 0.06\% & 0.09\% & 0.08\% \\
\wmove{Rc3} & 0.03\% & 0.13\% & 0.08\% & 0.05\% & 0.12\% & 0.15\% & 0.04\% & 0.04\% \\
\wmove{Rb3} & 0.03\% & 0.20\% & 0.11\% & 0.09\% & 0.12\% & 0.12\% & 0.03\% & 0.03\% \\
\wmove{Kh1} & 0.08\% & 0.09\% & 0.08\% & 0.06\% & 0.04\% & 0.03\% & 0.03\% & 0.02\% \\
\wmove{Kh2} & 0.05\% & 0.09\% & 0.09\% & 0.11\% & 0.09\% & 0.05\% & 0.05\% & 0.05\% \\
\wmove{Rd3} & 0.04\% & 0.09\% & 0.06\% & 0.04\% & 0.03\% & 0.05\% & 0.03\% & 0.03\% \\
\wmove{Re2} & 0.11\% & 0.06\% & 0.06\% & 0.03\% & 0.03\% & 0.05\% & 0.02\% & 0.02\% \\
\wmove{Re1} & 0.05\% & 0.11\% & 0.07\% & 0.03\% & 0.02\% & 0.12\% & 0.02\% & 0.02\% \\
\wmove{f3} & 0.04\% & 0.06\% & 0.01\% & 0.01\% & 0.00\% & 0.00\% & 0.00\% & 0.00\% \\
\bottomrule
\end{tabular}%
}
\end{table*}

\begin{table*}[h!]
\centering
\caption{Move probabilities by layer for puzzle \href{https://lichess.org/training/09SS5}{\texttt{09SS5}} (Part 2: Layer 7 to Final)}
\label{tab:puzzle_09SS5_probs2}
\resizebox{\textwidth}{!}{%
\begin{tabular}{lrrrrrrrr}
\toprule
\textbf{Move} & \textbf{Layer 7} & \textbf{Layer 8} & \textbf{Layer 9} & \textbf{Layer 10} & \textbf{Layer 11} & \textbf{Layer 12} & \textbf{Layer 13} & \textbf{Final} \\
\midrule
\wmove{Qe8+} & 0.46\% & \cellcolor{heatmapcolor!4!white}3.87\% & \cellcolor{heatmapcolor!4!white}4.40\% & \cellcolor{heatmapcolor!8!white}7.59\% & \cellcolor{heatmapcolor!16!white}16.43\% & \cellcolor{heatmapcolor!16!white}15.84\% & \cellcolor{heatmapcolor!22!white}22.31\% & \cellcolor{heatmapcolor!59!white}58.86\% \\
\wmove{gxf7+} & \cellcolor{heatmapcolor!42!white}41.53\% & \cellcolor{heatmapcolor!37!white}37.46\% & \cellcolor{heatmapcolor!42!white}42.34\% & \cellcolor{heatmapcolor!36!white}36.43\% & \cellcolor{heatmapcolor!27!white}27.24\% & \cellcolor{heatmapcolor!36!white}36.29\% & \cellcolor{heatmapcolor!20!white}20.04\% & \cellcolor{heatmapcolor!1!white}1.46\% \\
\wmove{Qxf7+} & \cellcolor{heatmapcolor!30!white}29.73\% & \cellcolor{heatmapcolor!27!white}27.14\% & \cellcolor{heatmapcolor!28!white}28.31\% & \cellcolor{heatmapcolor!20!white}19.52\% & \cellcolor{heatmapcolor!15!white}15.49\% & \cellcolor{heatmapcolor!6!white}5.55\% & \cellcolor{heatmapcolor!2!white}1.83\% & 0.27\% \\
\wmove{Qxa3} & \cellcolor{heatmapcolor!10!white}10.11\% & \cellcolor{heatmapcolor!11!white}11.00\% & \cellcolor{heatmapcolor!8!white}8.21\% & \cellcolor{heatmapcolor!8!white}7.69\% & \cellcolor{heatmapcolor!4!white}4.35\% & \cellcolor{heatmapcolor!3!white}2.92\% & \cellcolor{heatmapcolor!7!white}6.87\% & 0.45\% \\
\wmove{Qd7} & 0.02\% & 0.01\% & 0.01\% & 0.01\% & 0.02\% & 0.02\% & 0.17\% & 0.24\% \\
\wmove{Rxa3} & \cellcolor{heatmapcolor!15!white}15.22\% & \cellcolor{heatmapcolor!16!white}16.47\% & \cellcolor{heatmapcolor!13!white}13.12\% & \cellcolor{heatmapcolor!20!white}19.88\% & \cellcolor{heatmapcolor!19!white}19.22\% & \cellcolor{heatmapcolor!20!white}20.38\% & \cellcolor{heatmapcolor!13!white}12.69\% & \cellcolor{heatmapcolor!9!white}9.17\% \\
\wmove{Qc7} & 0.03\% & 0.02\% & 0.04\% & 0.02\% & 0.02\% & 0.01\% & 0.26\% & 0.21\% \\
\wmove{Qe4} & 0.03\% & 0.07\% & 0.19\% & 0.30\% & \cellcolor{heatmapcolor!1!white}0.54\% & \cellcolor{heatmapcolor!1!white}0.87\% & \cellcolor{heatmapcolor!8!white}8.36\% & \cellcolor{heatmapcolor!10!white}9.86\% \\
\wmove{Rh3} & \cellcolor{heatmapcolor!1!white}0.93\% & \cellcolor{heatmapcolor!2!white}1.98\% & \cellcolor{heatmapcolor!1!white}1.26\% & \cellcolor{heatmapcolor!4!white}3.77\% & \cellcolor{heatmapcolor!7!white}6.90\% & \cellcolor{heatmapcolor!7!white}6.86\% & \cellcolor{heatmapcolor!6!white}6.19\% & \cellcolor{heatmapcolor!6!white}6.35\% \\
\wmove{Qf6} & 0.15\% & 0.06\% & 0.06\% & 0.07\% & 0.08\% & 0.04\% & 0.14\% & 0.18\% \\
\wmove{Rf3} & 0.21\% & 0.27\% & 0.19\% & \cellcolor{heatmapcolor!2!white}1.53\% & \cellcolor{heatmapcolor!6!white}5.85\% & \cellcolor{heatmapcolor!7!white}6.66\% & \cellcolor{heatmapcolor!6!white}5.66\% & \cellcolor{heatmapcolor!2!white}1.98\% \\
\wmove{Qb7} & 0.02\% & 0.01\% & 0.02\% & 0.02\% & 0.04\% & 0.20\% & \cellcolor{heatmapcolor!6!white}6.48\% & \cellcolor{heatmapcolor!4!white}4.20\% \\
\wmove{Qd6} & 0.03\% & 0.02\% & 0.03\% & 0.02\% & 0.02\% & 0.01\% & 0.08\% & 0.18\% \\
\wmove{Qe6} & 0.03\% & 0.02\% & 0.05\% & 0.03\% & 0.03\% & 0.04\% & 0.08\% & 0.18\% \\
\wmove{Re4} & 0.11\% & 0.15\% & 0.07\% & 0.29\% & \cellcolor{heatmapcolor!2!white}1.69\% & \cellcolor{heatmapcolor!2!white}1.86\% & \cellcolor{heatmapcolor!3!white}2.77\% & \cellcolor{heatmapcolor!1!white}0.75\% \\
\wmove{Qg5} & 0.11\% & 0.06\% & 0.05\% & 0.05\% & 0.03\% & 0.02\% & 0.08\% & 0.20\% \\
\wmove{Qb4} & 0.07\% & 0.05\% & 0.10\% & 0.07\% & 0.05\% & 0.01\% & 0.09\% & 0.17\% \\
\wmove{Qf8+} & 0.19\% & 0.24\% & 0.27\% & \cellcolor{heatmapcolor!1!white}1.02\% & 0.21\% & 0.03\% & 0.41\% & 0.22\% \\
\wmove{Qd8+} & 0.02\% & 0.04\% & 0.04\% & 0.07\% & 0.21\% & 0.02\% & 0.07\% & 0.25\% \\
\wmove{Qh4} & 0.03\% & 0.05\% & 0.06\% & 0.06\% & 0.07\% & 0.06\% & 0.46\% & \cellcolor{heatmapcolor!1!white}0.71\% \\
\wmove{f4} & 0.15\% & 0.30\% & 0.38\% & \cellcolor{heatmapcolor!1!white}0.75\% & 0.40\% & \cellcolor{heatmapcolor!1!white}0.56\% & \cellcolor{heatmapcolor!1!white}1.42\% & 0.16\% \\
\wmove{Re6} & 0.04\% & 0.03\% & 0.04\% & 0.05\% & 0.06\% & 0.04\% & 0.11\% & 0.24\% \\
\wmove{Rg3} & 0.14\% & 0.19\% & 0.10\% & 0.27\% & 0.42\% & \cellcolor{heatmapcolor!1!white}1.08\% & \cellcolor{heatmapcolor!1!white}1.05\% & \cellcolor{heatmapcolor!1!white}0.54\% \\
\wmove{Qa7} & 0.02\% & 0.01\% & 0.02\% & 0.02\% & 0.01\% & 0.01\% & \cellcolor{heatmapcolor!1!white}0.94\% & \cellcolor{heatmapcolor!1!white}0.77\% \\
\wmove{Qc5} & 0.04\% & 0.03\% & 0.06\% & 0.05\% & 0.04\% & 0.02\% & 0.07\% & 0.22\% \\
\wmove{Re5} & 0.04\% & 0.05\% & 0.04\% & 0.05\% & 0.14\% & 0.18\% & 0.16\% & 0.21\% \\
\wmove{Qe5} & 0.03\% & 0.02\% & 0.02\% & 0.02\% & 0.02\% & 0.01\% & 0.08\% & 0.18\% \\
\wmove{Kg2} & 0.10\% & 0.07\% & 0.10\% & 0.06\% & 0.02\% & 0.01\% & 0.10\% & 0.18\% \\
\wmove{Rc3} & 0.12\% & 0.05\% & 0.10\% & 0.04\% & 0.09\% & 0.08\% & 0.16\% & 0.35\% \\
\wmove{Rb3} & 0.07\% & 0.05\% & 0.08\% & 0.06\% & 0.08\% & 0.07\% & 0.26\% & 0.19\% \\
\wmove{Kh1} & 0.02\% & 0.02\% & 0.03\% & 0.03\% & 0.09\% & 0.11\% & 0.13\% & 0.21\% \\
\wmove{Kh2} & 0.04\% & 0.05\% & 0.04\% & 0.05\% & 0.02\% & 0.01\% & 0.11\% & 0.19\% \\
\wmove{Rd3} & 0.06\% & 0.07\% & 0.11\% & 0.10\% & 0.09\% & 0.10\% & 0.16\% & 0.18\% \\
\wmove{Re2} & 0.04\% & 0.03\% & 0.05\% & 0.02\% & 0.03\% & 0.02\% & 0.10\% & 0.17\% \\
\wmove{Re1} & 0.03\% & 0.02\% & 0.02\% & 0.01\% & 0.01\% & 0.01\% & 0.12\% & 0.17\% \\
\wmove{f3} & 0.02\% & 0.00\% & 0.01\% & 0.00\% & 0.00\% & 0.00\% & 0.01\% & 0.15\% \\
\bottomrule
\end{tabular}%
}
\end{table*}

%% file: Figures/Puzzles/Examples/puzzle_tables_7UUlN.tex
\begin{table*}[h!]
\centering
\caption{Move probabilities by layer for puzzle \href{https://lichess.org/training/7UUlN}{\texttt{7UUlN}} (Part 1: Input to Layer 6)}
\label{tab:puzzle_7UUlN_probs1}
\resizebox{\textwidth}{!}{%
\begin{tabular}{lrrrrrrrr}
\toprule
\textbf{Move} & \textbf{Input} & \textbf{Layer 0} & \textbf{Layer 1} & \textbf{Layer 2} & \textbf{Layer 3} & \textbf{Layer 4} & \textbf{Layer 5} & \textbf{Layer 6} \\
\midrule
\wmove{Bh2+} & 0.08\% & 0.02\% & \cellcolor{heatmapcolor!1!white}0.54\% & 0.16\% & 0.28\% & 0.29\% & \cellcolor{heatmapcolor!2!white}1.79\% & \cellcolor{heatmapcolor!15!white}14.62\% \\
\wmove{Rxe3} & \cellcolor{heatmapcolor!15!white}15.44\% & \cellcolor{heatmapcolor!31!white}30.90\% & \cellcolor{heatmapcolor!83!white}82.76\% & \cellcolor{heatmapcolor!83!white}82.54\% & \cellcolor{heatmapcolor!87!white}86.84\% & \cellcolor{heatmapcolor!87!white}86.62\% & \cellcolor{heatmapcolor!76!white}76.43\% & \cellcolor{heatmapcolor!44!white}43.60\% \\
\wmove{a4} & \cellcolor{heatmapcolor!58!white}57.52\% & \cellcolor{heatmapcolor!44!white}43.79\% & 0.40\% & 0.10\% & \cellcolor{heatmapcolor!3!white}3.44\% & 0.02\% & \cellcolor{heatmapcolor!5!white}5.46\% & \cellcolor{heatmapcolor!4!white}3.59\% \\
\wmove{Re5} & 0.49\% & 0.01\% & 0.07\% & 0.13\% & 0.44\% & \cellcolor{heatmapcolor!3!white}2.98\% & \cellcolor{heatmapcolor!9!white}9.25\% & \cellcolor{heatmapcolor!32!white}31.57\% \\
\wmove{h5} & 0.20\% & \cellcolor{heatmapcolor!14!white}13.90\% & \cellcolor{heatmapcolor!2!white}1.77\% & 0.11\% & 0.02\% & 0.00\% & 0.01\% & 0.01\% \\
\wmove{b5} & \cellcolor{heatmapcolor!2!white}2.34\% & \cellcolor{heatmapcolor!8!white}8.17\% & 0.40\% & 0.36\% & 0.01\% & 0.02\% & 0.08\% & 0.05\% \\
\wmove{Re4} & \cellcolor{heatmapcolor!6!white}5.54\% & 0.02\% & \cellcolor{heatmapcolor!2!white}1.78\% & \cellcolor{heatmapcolor!1!white}0.61\% & \cellcolor{heatmapcolor!2!white}1.55\% & 0.30\% & 0.40\% & 0.25\% \\
\wmove{Bc5} & \cellcolor{heatmapcolor!1!white}1.33\% & 0.02\% & \cellcolor{heatmapcolor!1!white}0.99\% & \cellcolor{heatmapcolor!4!white}4.38\% & \cellcolor{heatmapcolor!1!white}0.59\% & \cellcolor{heatmapcolor!1!white}1.23\% & 0.27\% & 0.12\% \\
\wmove{f5} & \cellcolor{heatmapcolor!3!white}3.35\% & \cellcolor{heatmapcolor!1!white}0.57\% & 0.13\% & 0.02\% & 0.02\% & 0.00\% & 0.01\% & 0.04\% \\
\wmove{Kh8} & \cellcolor{heatmapcolor!1!white}0.63\% & 0.11\% & \cellcolor{heatmapcolor!1!white}1.07\% & \cellcolor{heatmapcolor!3!white}2.96\% & \cellcolor{heatmapcolor!2!white}1.91\% & \cellcolor{heatmapcolor!2!white}2.12\% & 0.38\% & 0.18\% \\
\wmove{Rf6} & 0.37\% & 0.02\% & \cellcolor{heatmapcolor!2!white}2.26\% & 0.32\% & 0.16\% & 0.37\% & 0.33\% & \cellcolor{heatmapcolor!1!white}1.45\% \\
\wmove{g5} & \cellcolor{heatmapcolor!2!white}2.25\% & 0.46\% & 0.22\% & 0.05\% & 0.01\% & 0.01\% & 0.05\% & 0.08\% \\
\wmove{f6} & 0.18\% & 0.13\% & 0.03\% & 0.06\% & 0.00\% & 0.01\% & 0.02\% & 0.13\% \\
\wmove{h6} & 0.05\% & \cellcolor{heatmapcolor!1!white}0.73\% & 0.13\% & 0.08\% & 0.02\% & 0.03\% & 0.07\% & 0.11\% \\
\wmove{Bg3} & \cellcolor{heatmapcolor!1!white}0.75\% & 0.03\% & 0.34\% & 0.26\% & 0.16\% & 0.30\% & 0.18\% & 0.16\% \\
\wmove{Ra8} & \cellcolor{heatmapcolor!1!white}0.74\% & 0.12\% & 0.17\% & \cellcolor{heatmapcolor!1!white}1.38\% & 0.30\% & 0.38\% & 0.30\% & 0.27\% \\
\wmove{Rg6} & 0.31\% & 0.02\% & \cellcolor{heatmapcolor!1!white}1.38\% & 0.28\% & 0.33\% & 0.18\% & 0.13\% & 0.11\% \\
\wmove{Rh6} & 0.10\% & 0.02\% & \cellcolor{heatmapcolor!1!white}1.29\% & 0.36\% & 0.36\% & \cellcolor{heatmapcolor!1!white}0.57\% & 0.40\% & \cellcolor{heatmapcolor!1!white}0.67\% \\
\wmove{Be7} & \cellcolor{heatmapcolor!1!white}1.22\% & 0.01\% & 0.08\% & 0.15\% & 0.16\% & 0.18\% & 0.14\% & 0.04\% \\
\wmove{Ree8} & 0.39\% & 0.18\% & \cellcolor{heatmapcolor!1!white}1.14\% & \cellcolor{heatmapcolor!1!white}0.88\% & 0.30\% & 0.09\% & 0.27\% & 0.09\% \\
\wmove{Be5} & \cellcolor{heatmapcolor!1!white}1.10\% & 0.01\% & 0.21\% & \cellcolor{heatmapcolor!1!white}1.13\% & 0.47\% & \cellcolor{heatmapcolor!1!white}0.91\% & 0.37\% & \cellcolor{heatmapcolor!1!white}0.57\% \\
\wmove{Ba3} & \cellcolor{heatmapcolor!1!white}1.10\% & 0.02\% & 0.21\% & 0.12\% & \cellcolor{heatmapcolor!1!white}0.61\% & 0.24\% & \cellcolor{heatmapcolor!1!white}0.75\% & 0.26\% \\
\wmove{Bf4} & \cellcolor{heatmapcolor!1!white}0.99\% & 0.02\% & 0.22\% & \cellcolor{heatmapcolor!1!white}0.58\% & 0.17\% & 0.41\% & 0.39\% & 0.43\% \\
\wmove{Rce8} & 0.28\% & 0.26\% & \cellcolor{heatmapcolor!1!white}0.99\% & 0.48\% & 0.45\% & 0.10\% & 0.45\% & 0.13\% \\
\wmove{g6} & 0.31\% & 0.22\% & 0.03\% & 0.04\% & 0.00\% & 0.01\% & 0.01\% & 0.01\% \\
\wmove{Re7} & \cellcolor{heatmapcolor!1!white}0.79\% & 0.01\% & 0.44\% & 0.10\% & 0.14\% & 0.31\% & 0.25\% & 0.19\% \\
\wmove{Bb4} & \cellcolor{heatmapcolor!1!white}0.60\% & 0.02\% & 0.12\% & 0.31\% & 0.34\% & 0.32\% & \cellcolor{heatmapcolor!1!white}0.62\% & 0.19\% \\
\wmove{Rb8} & \cellcolor{heatmapcolor!1!white}0.53\% & 0.04\% & 0.09\% & \cellcolor{heatmapcolor!1!white}0.69\% & 0.19\% & \cellcolor{heatmapcolor!1!white}0.74\% & 0.44\% & \cellcolor{heatmapcolor!1!white}0.57\% \\
\wmove{Kf8} & 0.26\% & 0.06\% & 0.23\% & 0.39\% & 0.31\% & \cellcolor{heatmapcolor!1!white}0.57\% & 0.29\% & 0.16\% \\
\wmove{Rd8} & 0.34\% & 0.05\% & 0.21\% & 0.50\% & 0.17\% & 0.37\% & 0.25\% & 0.24\% \\
\wmove{Rf8} & 0.39\% & 0.05\% & 0.23\% & 0.27\% & 0.13\% & 0.17\% & 0.12\% & 0.07\% \\
\wmove{Bf8} & 0.04\% & 0.02\% & 0.09\% & 0.18\% & 0.11\% & 0.15\% & 0.11\% & 0.05\% \\
\bottomrule
\end{tabular}%
}
\end{table*}

\begin{table*}[h!]
\centering
\caption{Move probabilities by layer for puzzle \href{https://lichess.org/training/7UUlN}{\texttt{7UUlN}} (Part 2: Layer 7 to Final)}
\label{tab:puzzle_7UUlN_probs2}
\resizebox{\textwidth}{!}{%
\begin{tabular}{lrrrrrrrr}
\toprule
\textbf{Move} & \textbf{Layer 7} & \textbf{Layer 8} & \textbf{Layer 9} & \textbf{Layer 10} & \textbf{Layer 11} & \textbf{Layer 12} & \textbf{Layer 13} & \textbf{Final} \\
\midrule
\wmove{Bh2+} & \cellcolor{heatmapcolor!17!white}16.54\% & \cellcolor{heatmapcolor!23!white}23.41\% & \cellcolor{heatmapcolor!22!white}22.23\% & \cellcolor{heatmapcolor!23!white}23.31\% & \cellcolor{heatmapcolor!23!white}23.15\% & \cellcolor{heatmapcolor!51!white}51.16\% & \cellcolor{heatmapcolor!67!white}67.02\% & \cellcolor{heatmapcolor!93!white}92.87\% \\
\wmove{Rxe3} & \cellcolor{heatmapcolor!31!white}30.72\% & \cellcolor{heatmapcolor!51!white}50.54\% & \cellcolor{heatmapcolor!35!white}35.04\% & \cellcolor{heatmapcolor!43!white}42.78\% & \cellcolor{heatmapcolor!32!white}31.82\% & \cellcolor{heatmapcolor!6!white}5.67\% & \cellcolor{heatmapcolor!4!white}3.56\% & 0.17\% \\
\wmove{a4} & \cellcolor{heatmapcolor!1!white}1.12\% & 0.35\% & 0.10\% & 0.27\% & \cellcolor{heatmapcolor!1!white}0.90\% & \cellcolor{heatmapcolor!2!white}1.69\% & \cellcolor{heatmapcolor!6!white}5.71\% & \cellcolor{heatmapcolor!1!white}1.08\% \\
\wmove{Re5} & \cellcolor{heatmapcolor!49!white}48.65\% & \cellcolor{heatmapcolor!22!white}21.68\% & \cellcolor{heatmapcolor!39!white}38.76\% & \cellcolor{heatmapcolor!30!white}29.97\% & \cellcolor{heatmapcolor!35!white}35.14\% & \cellcolor{heatmapcolor!36!white}35.87\% & \cellcolor{heatmapcolor!3!white}3.34\% & 0.17\% \\
\wmove{h5} & 0.01\% & 0.01\% & 0.01\% & 0.01\% & 0.02\% & 0.02\% & 0.37\% & 0.26\% \\
\wmove{b5} & 0.10\% & 0.12\% & 0.05\% & 0.07\% & 0.37\% & 0.03\% & \cellcolor{heatmapcolor!1!white}1.04\% & 0.15\% \\
\wmove{Re4} & 0.06\% & 0.12\% & 0.06\% & 0.12\% & 0.29\% & 0.14\% & \cellcolor{heatmapcolor!1!white}0.75\% & 0.16\% \\
\wmove{Bc5} & 0.07\% & 0.20\% & 0.15\% & 0.13\% & 0.15\% & 0.18\% & \cellcolor{heatmapcolor!1!white}0.72\% & 0.16\% \\
\wmove{f5} & 0.15\% & 0.44\% & 0.06\% & 0.06\% & 0.30\% & 0.17\% & \cellcolor{heatmapcolor!1!white}0.73\% & \cellcolor{heatmapcolor!1!white}0.71\% \\
\wmove{Kh8} & 0.12\% & 0.19\% & 0.29\% & 0.35\% & \cellcolor{heatmapcolor!1!white}0.77\% & \cellcolor{heatmapcolor!1!white}0.78\% & \cellcolor{heatmapcolor!1!white}1.39\% & 0.19\% \\
\wmove{Rf6} & \cellcolor{heatmapcolor!1!white}0.61\% & 0.29\% & \cellcolor{heatmapcolor!1!white}0.53\% & 0.34\% & \cellcolor{heatmapcolor!2!white}1.86\% & 0.40\% & 0.37\% & 0.16\% \\
\wmove{g5} & 0.07\% & 0.08\% & 0.07\% & 0.06\% & 0.24\% & 0.11\% & \cellcolor{heatmapcolor!1!white}0.63\% & 0.16\% \\
\wmove{f6} & 0.09\% & 0.06\% & 0.03\% & 0.05\% & 0.10\% & 0.06\% & \cellcolor{heatmapcolor!2!white}1.98\% & 0.18\% \\
\wmove{h6} & 0.04\% & 0.01\% & 0.02\% & 0.02\% & 0.18\% & 0.06\% & \cellcolor{heatmapcolor!2!white}1.74\% & 0.14\% \\
\wmove{Bg3} & 0.07\% & 0.14\% & 0.21\% & 0.27\% & \cellcolor{heatmapcolor!1!white}0.64\% & \cellcolor{heatmapcolor!1!white}0.69\% & \cellcolor{heatmapcolor!1!white}1.46\% & 0.19\% \\
\wmove{Ra8} & 0.08\% & 0.07\% & 0.07\% & 0.06\% & 0.07\% & 0.06\% & 0.26\% & 0.30\% \\
\wmove{Rg6} & 0.03\% & 0.04\% & 0.03\% & 0.04\% & 0.31\% & 0.09\% & \cellcolor{heatmapcolor!1!white}0.50\% & 0.15\% \\
\wmove{Rh6} & 0.24\% & 0.12\% & 0.14\% & 0.13\% & \cellcolor{heatmapcolor!1!white}0.80\% & 0.18\% & 0.49\% & 0.15\% \\
\wmove{Be7} & 0.02\% & 0.08\% & 0.07\% & 0.08\% & 0.09\% & 0.23\% & \cellcolor{heatmapcolor!1!white}0.53\% & 0.16\% \\
\wmove{Ree8} & 0.03\% & 0.03\% & 0.02\% & 0.01\% & 0.13\% & 0.09\% & 0.47\% & 0.21\% \\
\wmove{Be5} & 0.19\% & 0.42\% & 0.27\% & 0.23\% & \cellcolor{heatmapcolor!1!white}0.80\% & \cellcolor{heatmapcolor!1!white}0.52\% & \cellcolor{heatmapcolor!1!white}0.71\% & 0.16\% \\
\wmove{Ba3} & 0.15\% & 0.25\% & 0.26\% & 0.16\% & 0.23\% & 0.13\% & \cellcolor{heatmapcolor!1!white}0.89\% & 0.16\% \\
\wmove{Bf4} & 0.25\% & 0.49\% & \cellcolor{heatmapcolor!1!white}0.59\% & \cellcolor{heatmapcolor!1!white}0.63\% & 0.40\% & \cellcolor{heatmapcolor!1!white}0.62\% & \cellcolor{heatmapcolor!1!white}1.07\% & 0.18\% \\
\wmove{Rce8} & 0.08\% & 0.08\% & 0.10\% & 0.11\% & 0.17\% & 0.16\% & 0.27\% & 0.39\% \\
\wmove{g6} & 0.01\% & 0.00\% & 0.00\% & 0.01\% & 0.04\% & 0.01\% & \cellcolor{heatmapcolor!1!white}0.81\% & 0.14\% \\
\wmove{Re7} & 0.04\% & 0.10\% & 0.09\% & 0.08\% & 0.21\% & 0.10\% & 0.49\% & 0.17\% \\
\wmove{Bb4} & 0.08\% & 0.18\% & 0.25\% & 0.13\% & 0.14\% & 0.28\% & \cellcolor{heatmapcolor!1!white}0.78\% & 0.15\% \\
\wmove{Rb8} & 0.13\% & 0.11\% & 0.15\% & 0.09\% & 0.15\% & 0.08\% & 0.40\% & 0.17\% \\
\wmove{Kf8} & 0.11\% & 0.15\% & 0.10\% & 0.14\% & 0.19\% & 0.14\% & 0.44\% & 0.17\% \\
\wmove{Rd8} & 0.06\% & 0.07\% & 0.11\% & 0.09\% & 0.12\% & 0.08\% & 0.28\% & 0.27\% \\
\wmove{Rf8} & 0.04\% & 0.06\% & 0.08\% & 0.12\% & 0.18\% & 0.15\% & 0.44\% & 0.17\% \\
\wmove{Bf8} & 0.03\% & 0.10\% & 0.07\% & 0.06\% & 0.03\% & 0.07\% & 0.36\% & 0.14\% \\
\bottomrule
\end{tabular}%
}
\end{table*}

%% file: Figures/Puzzles/Forgotten/puzzle_evaluation_ANklq.tex
\begin{table*}[h!]
\centering
\caption{Move evaluation for puzzle \href{https://lichess.org/training/ANklq}{\texttt{ANklq}}: Stockfish evaluation at depth 20 and model WDL prediction for resulting positions}
\label{tab:puzzle_ANklq_eval}
\resizebox{\textwidth}{!}{%
\begin{tabular}{llrrrrr}
\toprule
\textbf{Move} & \textbf{Stockfish} & \textbf{$\Delta$ (cp)} & \textbf{Win} & \textbf{Draw} & \textbf{Loss} & \textbf{$\Delta$ Win} \\
\midrule
\textit{Current position} & $+\infty$ & --- & $1.4\%$ & $2.6\%$ & $96.0\%$ & --- \\
\midrule
\rowcolor{heatmapcolor!30}
\wmove{Qxf1+} $\star$ & $+\infty$ & --- & $99.8\%$ & $0.1\%$ & $0.0\%$ & $+98.4\%$ \\
\wmove{Ba8} & $-4.34$ & $-\infty$ & $0.1\%$ & $0.2\%$ & $99.7\%$ & $-1.3\%$ \\
\wmove{Rd7} & $-4.36$ & $-\infty$ & $0.3\%$ & $0.6\%$ & $99.1\%$ & $-1.1\%$ \\
\wmove{Bc8} & $-4.51$ & $-\infty$ & $0.3\%$ & $0.6\%$ & $99.1\%$ & $-1.1\%$ \\
\wmove{Qc8} & $-4.87$ & $-\infty$ & $0.3\%$ & $0.7\%$ & $98.9\%$ & $-1.1\%$ \\
\wmove{Qd7} & $-4.91$ & $-\infty$ & $0.2\%$ & $0.5\%$ & $99.3\%$ & $-1.2\%$ \\
\wmove{Rb8} & $-4.95$ & $-\infty$ & $0.2\%$ & $0.3\%$ & $99.5\%$ & $-1.2\%$ \\
\wmove{Nc5} & $-4.97$ & $-\infty$ & $0.1\%$ & $0.1\%$ & $99.8\%$ & $-1.3\%$ \\
\wmove{Bh1} & $-5.23$ & $-\infty$ & $0.1\%$ & $0.2\%$ & $99.7\%$ & $-1.3\%$ \\
\wmove{Be4} & $-5.34$ & $-\infty$ & $0.1\%$ & $0.2\%$ & $99.7\%$ & $-1.3\%$ \\
\wmove{Bd5} & $-5.37$ & $-\infty$ & $0.1\%$ & $0.2\%$ & $99.6\%$ & $-1.3\%$ \\
\wmove{Nxb4} & $-5.37$ & $-\infty$ & $0.0\%$ & $0.1\%$ & $99.9\%$ & $-1.4\%$ \\
\wmove{Nc7} & $-5.83$ & $-\infty$ & $0.0\%$ & $0.1\%$ & $99.8\%$ & $-1.4\%$ \\
\wmove{Nb8} & $-6.19$ & $-\infty$ & $0.0\%$ & $0.1\%$ & $99.9\%$ & $-1.4\%$ \\
\wmove{Bf3} & $-6.21$ & $-\infty$ & $0.0\%$ & $0.1\%$ & $99.9\%$ & $-1.4\%$ \\
\wmove{Bg2} & $-6.44$ & $-\infty$ & $0.0\%$ & $0.0\%$ & $99.9\%$ & $-1.4\%$ \\
\wmove{Qg4} & $-6.53$ & $-\infty$ & $0.0\%$ & $0.0\%$ & $99.9\%$ & $-1.4\%$ \\
\wmove{Rf8} & $-6.78$ & $-\infty$ & $0.0\%$ & $0.0\%$ & $99.9\%$ & $-1.4\%$ \\
\wmove{Rc8} & $-6.78$ & $-\infty$ & $0.0\%$ & $0.0\%$ & $100.0\%$ & $-1.4\%$ \\
\wmove{Qe6} & $-6.82$ & $-\infty$ & $0.0\%$ & $0.1\%$ & $99.8\%$ & $-1.4\%$ \\
\wmove{Rd6} & $-6.87$ & $-\infty$ & $0.0\%$ & $0.0\%$ & $100.0\%$ & $-1.4\%$ \\
\wmove{Bc6} & $-6.90$ & $-\infty$ & $0.0\%$ & $0.2\%$ & $99.8\%$ & $-1.4\%$ \\
\wmove{f6} & $-6.92$ & $-\infty$ & $0.0\%$ & $0.1\%$ & $99.9\%$ & $-1.4\%$ \\
\wmove{Kh8} & $-6.93$ & $-\infty$ & $0.0\%$ & $0.0\%$ & $100.0\%$ & $-1.4\%$ \\
\wmove{Kf8} & $-6.95$ & $-\infty$ & $0.0\%$ & $0.0\%$ & $100.0\%$ & $-1.4\%$ \\
\wmove{g6} & $-6.95$ & $-\infty$ & $0.0\%$ & $0.0\%$ & $99.9\%$ & $-1.4\%$ \\
\wmove{Kh7} & $-6.97$ & $-\infty$ & $0.0\%$ & $0.0\%$ & $99.9\%$ & $-1.4\%$ \\
\wmove{Rd3} & $-7.06$ & $-\infty$ & $0.0\%$ & $0.0\%$ & $100.0\%$ & $-1.4\%$ \\
\wmove{Rd5} & $-7.07$ & $-\infty$ & $0.0\%$ & $0.1\%$ & $99.9\%$ & $-1.4\%$ \\
\wmove{Ra8} & $-7.21$ & $-\infty$ & $0.0\%$ & $0.0\%$ & $100.0\%$ & $-1.4\%$ \\
\wmove{Rd2} & $-7.30$ & $-\infty$ & $0.0\%$ & $0.0\%$ & $100.0\%$ & $-1.4\%$ \\
\wmove{f5} & $-7.32$ & $-\infty$ & $0.0\%$ & $0.0\%$ & $100.0\%$ & $-1.4\%$ \\
\wmove{Qxh4} & $-7.72$ & $-\infty$ & $0.0\%$ & $0.1\%$ & $99.9\%$ & $-1.4\%$ \\
\wmove{Rd4} & $-7.94$ & $-\infty$ & $0.0\%$ & $0.1\%$ & $99.9\%$ & $-1.4\%$ \\
\wmove{Rd1} & $-8.33$ & $-\infty$ & $0.0\%$ & $0.4\%$ & $99.6\%$ & $-1.4\%$ \\
\wmove{Qxg3+} & $-8.48$ & $-\infty$ & $0.0\%$ & $0.0\%$ & $100.0\%$ & $-1.4\%$ \\
\wmove{Qg2+} & $-8.48$ & $-\infty$ & $0.0\%$ & $0.0\%$ & $100.0\%$ & $-1.4\%$ \\
\wmove{Qxh2+} & $-8.71$ & $-\infty$ & $0.0\%$ & $0.0\%$ & $100.0\%$ & $-1.4\%$ \\
\wmove{Qf5} & $-9.18$ & $-\infty$ & $0.0\%$ & $0.0\%$ & $100.0\%$ & $-1.4\%$ \\
\wmove{Re8} & $-9.33$ & $-\infty$ & $0.0\%$ & $0.0\%$ & $100.0\%$ & $-1.4\%$ \\
\wmove{g5} & $-9.38$ & $-\infty$ & $0.0\%$ & $0.0\%$ & $100.0\%$ & $-1.4\%$ \\
\bottomrule
\end{tabular}%
}
\end{table*}

%% file: Figures/Puzzles/Forgotten/puzzle_tables_ANklq.tex
\begin{table*}[h!]
\centering
\caption{Move probabilities by layer for puzzle \href{https://lichess.org/training/ANklq}{\texttt{ANklq}} (Part 1: Input to Layer 6)}
\label{tab:puzzle_ANklq_probs1}
\resizebox{\textwidth}{!}{%
\begin{tabular}{lrrrrrrrr}
\toprule
\textbf{Move} & \textbf{Input} & \textbf{Layer 0} & \textbf{Layer 1} & \textbf{Layer 2} & \textbf{Layer 3} & \textbf{Layer 4} & \textbf{Layer 5} & \textbf{Layer 6} \\
\midrule
\wmove{Qxf1+} & 0.46\% & \cellcolor{heatmapcolor!46!white}45.73\% & \cellcolor{heatmapcolor!49!white}49.14\% & \cellcolor{heatmapcolor!47!white}46.74\% & \cellcolor{heatmapcolor!45!white}44.85\% & \cellcolor{heatmapcolor!48!white}47.54\% & \cellcolor{heatmapcolor!66!white}66.46\% & \cellcolor{heatmapcolor!48!white}47.93\% \\
\wmove{Rd2} & \cellcolor{heatmapcolor!41!white}41.09\% & 0.07\% & 0.04\% & 0.15\% & 0.44\% & \cellcolor{heatmapcolor!1!white}0.81\% & 0.08\% & 0.11\% \\
\wmove{Qc8} & 0.10\% & 0.02\% & 0.04\% & 0.03\% & 0.03\% & 0.13\% & 0.01\% & 0.02\% \\
\wmove{Qxh2+} & \cellcolor{heatmapcolor!1!white}0.51\% & \cellcolor{heatmapcolor!24!white}23.84\% & \cellcolor{heatmapcolor!13!white}13.22\% & \cellcolor{heatmapcolor!26!white}25.78\% & \cellcolor{heatmapcolor!30!white}30.20\% & \cellcolor{heatmapcolor!31!white}31.29\% & \cellcolor{heatmapcolor!24!white}23.55\% & \cellcolor{heatmapcolor!31!white}31.01\% \\
\wmove{Qg2+} & \cellcolor{heatmapcolor!26!white}26.47\% & 0.03\% & 0.37\% & 0.34\% & \cellcolor{heatmapcolor!1!white}0.67\% & 0.04\% & 0.08\% & \cellcolor{heatmapcolor!1!white}0.57\% \\
\wmove{Qd7} & 0.01\% & 0.01\% & 0.03\% & 0.02\% & 0.04\% & 0.05\% & 0.02\% & 0.02\% \\
\wmove{Qxh4} & 0.18\% & \cellcolor{heatmapcolor!13!white}13.43\% & \cellcolor{heatmapcolor!24!white}24.41\% & \cellcolor{heatmapcolor!4!white}3.60\% & \cellcolor{heatmapcolor!1!white}1.43\% & \cellcolor{heatmapcolor!1!white}0.72\% & \cellcolor{heatmapcolor!1!white}0.68\% & \cellcolor{heatmapcolor!1!white}0.59\% \\
\wmove{Qxg3+} & \cellcolor{heatmapcolor!1!white}1.06\% & \cellcolor{heatmapcolor!15!white}15.02\% & \cellcolor{heatmapcolor!9!white}9.23\% & \cellcolor{heatmapcolor!17!white}17.08\% & \cellcolor{heatmapcolor!16!white}15.67\% & \cellcolor{heatmapcolor!10!white}9.74\% & \cellcolor{heatmapcolor!5!white}5.29\% & \cellcolor{heatmapcolor!12!white}12.33\% \\
\wmove{Rd7} & 0.04\% & 0.02\% & 0.03\% & 0.04\% & 0.04\% & 0.10\% & 0.02\% & 0.04\% \\
\wmove{Rd3} & \cellcolor{heatmapcolor!13!white}12.98\% & 0.03\% & 0.02\% & 0.11\% & \cellcolor{heatmapcolor!1!white}0.63\% & \cellcolor{heatmapcolor!1!white}1.50\% & 0.27\% & 0.41\% \\
\wmove{Nc7} & 0.18\% & 0.19\% & 0.01\% & 0.02\% & 0.14\% & 0.05\% & 0.45\% & 0.37\% \\
\wmove{Rd1} & \cellcolor{heatmapcolor!6!white}5.56\% & 0.06\% & 0.06\% & 0.08\% & \cellcolor{heatmapcolor!1!white}1.16\% & 0.07\% & 0.16\% & \cellcolor{heatmapcolor!1!white}0.88\% \\
\wmove{Bc8} & 0.10\% & 0.01\% & 0.01\% & 0.01\% & 0.01\% & 0.02\% & 0.01\% & 0.01\% \\
\wmove{Nxb4} & \cellcolor{heatmapcolor!4!white}4.36\% & 0.04\% & \cellcolor{heatmapcolor!2!white}1.58\% & \cellcolor{heatmapcolor!4!white}3.76\% & \cellcolor{heatmapcolor!2!white}1.96\% & \cellcolor{heatmapcolor!1!white}1.45\% & \cellcolor{heatmapcolor!1!white}0.97\% & 0.26\% \\
\wmove{Bg2} & 0.11\% & 0.03\% & 0.18\% & 0.22\% & 0.27\% & 0.04\% & 0.27\% & \cellcolor{heatmapcolor!2!white}2.32\% \\
\wmove{Rd4} & \cellcolor{heatmapcolor!2!white}2.29\% & 0.06\% & 0.04\% & 0.11\% & 0.26\% & \cellcolor{heatmapcolor!1!white}0.83\% & 0.11\% & 0.23\% \\
\wmove{f5} & \cellcolor{heatmapcolor!1!white}1.00\% & 0.06\% & 0.01\% & 0.02\% & 0.01\% & 0.01\% & 0.01\% & 0.01\% \\
\wmove{Bd5} & 0.13\% & 0.01\% & 0.01\% & 0.02\% & 0.05\% & 0.03\% & 0.06\% & 0.06\% \\
\wmove{Bh1} & 0.04\% & 0.05\% & 0.08\% & 0.16\% & 0.41\% & 0.04\% & \cellcolor{heatmapcolor!1!white}0.57\% & \cellcolor{heatmapcolor!1!white}1.02\% \\
\wmove{Bf3} & 0.29\% & 0.16\% & 0.34\% & 0.09\% & 0.27\% & 0.07\% & 0.05\% & 0.14\% \\
\wmove{Rb8} & 0.17\% & 0.03\% & 0.01\% & 0.03\% & 0.05\% & \cellcolor{heatmapcolor!1!white}1.18\% & 0.08\% & 0.29\% \\
\wmove{Qg4} & 0.48\% & 0.02\% & 0.08\% & 0.05\% & 0.08\% & 0.10\% & 0.03\% & 0.04\% \\
\wmove{Rd5} & 0.01\% & 0.02\% & 0.05\% & 0.05\% & 0.04\% & 0.10\% & 0.03\% & 0.02\% \\
\wmove{Nc5} & 0.09\% & 0.20\% & 0.37\% & \cellcolor{heatmapcolor!1!white}0.76\% & 0.44\% & 0.46\% & 0.16\% & 0.28\% \\
\wmove{Ba8} & 0.02\% & 0.02\% & 0.02\% & 0.01\% & 0.01\% & 0.02\% & 0.01\% & 0.01\% \\
\wmove{Qe6} & 0.00\% & 0.02\% & 0.08\% & 0.05\% & 0.05\% & 0.05\% & 0.01\% & 0.01\% \\
\wmove{Ra8} & 0.23\% & 0.05\% & 0.01\% & 0.04\% & 0.07\% & 0.49\% & 0.04\% & 0.05\% \\
\wmove{Rd6} & 0.01\% & 0.02\% & 0.03\% & 0.08\% & 0.08\% & \cellcolor{heatmapcolor!1!white}0.80\% & 0.08\% & 0.24\% \\
\wmove{g6} & 0.09\% & 0.05\% & 0.01\% & 0.01\% & 0.01\% & 0.01\% & 0.00\% & 0.00\% \\
\wmove{Be4} & 0.33\% & 0.01\% & 0.01\% & 0.02\% & 0.03\% & 0.04\% & 0.04\% & 0.03\% \\
\wmove{Bc6} & 0.05\% & 0.01\% & 0.01\% & 0.01\% & 0.02\% & 0.03\% & 0.03\% & 0.02\% \\
\wmove{Rc8} & 0.11\% & 0.02\% & 0.05\% & 0.03\% & 0.04\% & \cellcolor{heatmapcolor!1!white}0.73\% & 0.02\% & 0.03\% \\
\wmove{Qf5} & 0.01\% & 0.02\% & 0.04\% & 0.05\% & 0.04\% & 0.04\% & 0.02\% & 0.03\% \\
\wmove{g5} & \cellcolor{heatmapcolor!1!white}0.68\% & 0.09\% & 0.11\% & 0.20\% & 0.15\% & 0.02\% & 0.04\% & 0.16\% \\
\wmove{f6} & 0.09\% & 0.03\% & 0.01\% & 0.00\% & 0.00\% & 0.00\% & 0.00\% & 0.00\% \\
\wmove{Re8} & 0.10\% & 0.02\% & 0.03\% & 0.02\% & 0.04\% & 0.33\% & 0.06\% & 0.12\% \\
\wmove{Nb8} & 0.07\% & 0.38\% & 0.07\% & 0.05\% & 0.05\% & 0.45\% & 0.13\% & 0.23\% \\
\wmove{Rf8} & 0.12\% & 0.02\% & 0.04\% & 0.03\% & 0.04\% & 0.27\% & 0.03\% & 0.05\% \\
\wmove{Kh7} & 0.11\% & 0.02\% & 0.04\% & 0.04\% & 0.08\% & 0.13\% & 0.02\% & 0.01\% \\
\wmove{Kh8} & 0.19\% & 0.05\% & 0.05\% & 0.03\% & 0.05\% & 0.10\% & 0.02\% & 0.02\% \\
\wmove{Kf8} & 0.08\% & 0.02\% & 0.04\% & 0.03\% & 0.08\% & 0.15\% & 0.02\% & 0.03\% \\
\bottomrule
\end{tabular}%
}
\end{table*}

\begin{table*}[h!]
\centering
\caption{Move probabilities by layer for puzzle \href{https://lichess.org/training/ANklq}{\texttt{ANklq}} (Part 2: Layer 7 to Final)}
\label{tab:puzzle_ANklq_probs2}
\resizebox{\textwidth}{!}{%
\begin{tabular}{lrrrrrrrr}
\toprule
\textbf{Move} & \textbf{Layer 7} & \textbf{Layer 8} & \textbf{Layer 9} & \textbf{Layer 10} & \textbf{Layer 11} & \textbf{Layer 12} & \textbf{Layer 13} & \textbf{Final} \\
\midrule
\wmove{Qxf1+} & \cellcolor{heatmapcolor!53!white}53.31\% & \cellcolor{heatmapcolor!52!white}51.84\% & \cellcolor{heatmapcolor!56!white}56.08\% & \cellcolor{heatmapcolor!59!white}59.27\% & \cellcolor{heatmapcolor!60!white}60.25\% & \cellcolor{heatmapcolor!88!white}88.28\% & \cellcolor{heatmapcolor!31!white}30.59\% & \cellcolor{heatmapcolor!6!white}5.79\% \\
\wmove{Rd2} & 0.10\% & 0.06\% & 0.12\% & 0.04\% & 0.08\% & 0.05\% & \cellcolor{heatmapcolor!1!white}0.60\% & 0.22\% \\
\wmove{Qc8} & 0.02\% & 0.02\% & 0.03\% & 0.04\% & 0.06\% & 0.07\% & \cellcolor{heatmapcolor!2!white}2.40\% & \cellcolor{heatmapcolor!37!white}37.04\% \\
\wmove{Qxh2+} & \cellcolor{heatmapcolor!31!white}30.50\% & \cellcolor{heatmapcolor!28!white}27.80\% & \cellcolor{heatmapcolor!26!white}26.32\% & \cellcolor{heatmapcolor!26!white}25.86\% & \cellcolor{heatmapcolor!23!white}23.28\% & \cellcolor{heatmapcolor!3!white}3.29\% & \cellcolor{heatmapcolor!3!white}3.05\% & 0.18\% \\
\wmove{Qg2+} & \cellcolor{heatmapcolor!1!white}0.50\% & \cellcolor{heatmapcolor!2!white}2.37\% & \cellcolor{heatmapcolor!2!white}1.78\% & 0.47\% & \cellcolor{heatmapcolor!1!white}0.58\% & 0.16\% & \cellcolor{heatmapcolor!1!white}1.41\% & 0.21\% \\
\wmove{Qd7} & 0.03\% & 0.05\% & 0.14\% & 0.30\% & \cellcolor{heatmapcolor!1!white}0.59\% & 0.31\% & \cellcolor{heatmapcolor!7!white}7.38\% & \cellcolor{heatmapcolor!25!white}24.66\% \\
\wmove{Qxh4} & 0.28\% & 0.19\% & 0.11\% & 0.10\% & 0.22\% & 0.25\% & \cellcolor{heatmapcolor!2!white}2.00\% & 0.26\% \\
\wmove{Qxg3+} & \cellcolor{heatmapcolor!7!white}6.65\% & \cellcolor{heatmapcolor!9!white}9.07\% & \cellcolor{heatmapcolor!9!white}8.91\% & \cellcolor{heatmapcolor!8!white}8.23\% & \cellcolor{heatmapcolor!6!white}6.30\% & \cellcolor{heatmapcolor!2!white}2.32\% & \cellcolor{heatmapcolor!11!white}11.04\% & 0.17\% \\
\wmove{Rd7} & 0.01\% & 0.01\% & 0.01\% & 0.00\% & 0.00\% & 0.02\% & \cellcolor{heatmapcolor!1!white}0.94\% & \cellcolor{heatmapcolor!13!white}13.27\% \\
\wmove{Rd3} & 0.47\% & 0.37\% & \cellcolor{heatmapcolor!1!white}0.55\% & 0.34\% & \cellcolor{heatmapcolor!1!white}0.52\% & 0.12\% & 0.36\% & 0.21\% \\
\wmove{Nc7} & 0.23\% & 0.23\% & 0.32\% & 0.20\% & 0.38\% & 0.35\% & \cellcolor{heatmapcolor!12!white}12.47\% & \cellcolor{heatmapcolor!2!white}2.19\% \\
\wmove{Rd1} & \cellcolor{heatmapcolor!1!white}1.45\% & \cellcolor{heatmapcolor!2!white}2.10\% & \cellcolor{heatmapcolor!1!white}0.87\% & \cellcolor{heatmapcolor!1!white}1.02\% & \cellcolor{heatmapcolor!3!white}3.03\% & 0.38\% & \cellcolor{heatmapcolor!1!white}0.80\% & 0.32\% \\
\wmove{Bc8} & 0.01\% & 0.02\% & 0.02\% & 0.07\% & 0.14\% & 0.35\% & \cellcolor{heatmapcolor!3!white}2.62\% & \cellcolor{heatmapcolor!4!white}4.44\% \\
\wmove{Nxb4} & 0.39\% & 0.35\% & 0.50\% & 0.29\% & 0.17\% & 0.23\% & \cellcolor{heatmapcolor!1!white}1.36\% & 0.30\% \\
\wmove{Bg2} & \cellcolor{heatmapcolor!3!white}2.99\% & \cellcolor{heatmapcolor!3!white}2.86\% & \cellcolor{heatmapcolor!1!white}1.29\% & \cellcolor{heatmapcolor!1!white}0.80\% & 0.30\% & 0.22\% & \cellcolor{heatmapcolor!1!white}1.28\% & 0.23\% \\
\wmove{Rd4} & 0.17\% & 0.19\% & 0.46\% & 0.15\% & 0.31\% & 0.24\% & 0.40\% & 0.20\% \\
\wmove{f5} & 0.01\% & 0.01\% & 0.02\% & 0.07\% & 0.11\% & 0.14\% & \cellcolor{heatmapcolor!2!white}2.19\% & 0.21\% \\
\wmove{Bd5} & 0.12\% & 0.16\% & 0.19\% & 0.23\% & 0.45\% & \cellcolor{heatmapcolor!1!white}0.60\% & \cellcolor{heatmapcolor!2!white}2.13\% & \cellcolor{heatmapcolor!1!white}0.88\% \\
\wmove{Bh1} & \cellcolor{heatmapcolor!1!white}0.68\% & 0.28\% & 0.35\% & \cellcolor{heatmapcolor!1!white}0.94\% & 0.44\% & 0.18\% & \cellcolor{heatmapcolor!2!white}1.80\% & \cellcolor{heatmapcolor!1!white}1.07\% \\
\wmove{Bf3} & 0.17\% & 0.16\% & 0.25\% & 0.31\% & 0.33\% & 0.10\% & \cellcolor{heatmapcolor!1!white}1.48\% & 0.20\% \\
\wmove{Rb8} & 0.40\% & 0.14\% & 0.07\% & 0.01\% & 0.02\% & 0.01\% & 0.27\% & \cellcolor{heatmapcolor!1!white}1.23\% \\
\wmove{Qg4} & 0.04\% & 0.05\% & 0.06\% & 0.04\% & 0.09\% & 0.06\% & \cellcolor{heatmapcolor!1!white}1.13\% & 0.24\% \\
\wmove{Rd5} & 0.02\% & 0.01\% & 0.01\% & 0.00\% & 0.00\% & 0.02\% & \cellcolor{heatmapcolor!1!white}1.06\% & 0.42\% \\
\wmove{Nc5} & 0.27\% & 0.45\% & 0.17\% & 0.24\% & 0.32\% & 0.45\% & \cellcolor{heatmapcolor!1!white}0.70\% & \cellcolor{heatmapcolor!1!white}1.05\% \\
\wmove{Ba8} & 0.01\% & 0.01\% & 0.01\% & 0.01\% & 0.01\% & 0.02\% & 0.33\% & \cellcolor{heatmapcolor!1!white}1.02\% \\
\wmove{Qe6} & 0.02\% & 0.05\% & 0.05\% & 0.13\% & 0.28\% & \cellcolor{heatmapcolor!1!white}0.55\% & \cellcolor{heatmapcolor!1!white}1.01\% & 0.31\% \\
\wmove{Ra8} & 0.06\% & 0.05\% & 0.12\% & 0.09\% & 0.15\% & 0.16\% & \cellcolor{heatmapcolor!1!white}0.97\% & 0.20\% \\
\wmove{Rd6} & 0.18\% & 0.07\% & 0.16\% & 0.05\% & 0.10\% & 0.04\% & \cellcolor{heatmapcolor!1!white}0.87\% & 0.21\% \\
\wmove{g6} & 0.01\% & 0.01\% & 0.01\% & 0.01\% & 0.02\% & 0.04\% & \cellcolor{heatmapcolor!1!white}0.86\% & 0.20\% \\
\wmove{Be4} & 0.04\% & 0.04\% & 0.05\% & 0.06\% & 0.11\% & 0.06\% & \cellcolor{heatmapcolor!1!white}0.83\% & \cellcolor{heatmapcolor!1!white}0.59\% \\
\wmove{Bc6} & 0.03\% & 0.03\% & 0.04\% & 0.07\% & 0.11\% & 0.09\% & \cellcolor{heatmapcolor!1!white}0.76\% & 0.24\% \\
\wmove{Rc8} & 0.04\% & 0.06\% & 0.07\% & 0.02\% & 0.02\% & 0.03\% & 0.24\% & 0.23\% \\
\wmove{Qf5} & 0.02\% & 0.02\% & 0.03\% & 0.04\% & 0.09\% & 0.13\% & \cellcolor{heatmapcolor!1!white}0.71\% & 0.19\% \\
\wmove{g5} & 0.05\% & 0.06\% & 0.02\% & 0.04\% & 0.05\% & 0.04\% & \cellcolor{heatmapcolor!1!white}0.50\% & 0.27\% \\
\wmove{f6} & 0.01\% & 0.01\% & 0.01\% & 0.00\% & 0.01\% & 0.01\% & \cellcolor{heatmapcolor!1!white}0.67\% & 0.21\% \\
\wmove{Re8} & 0.37\% & \cellcolor{heatmapcolor!1!white}0.59\% & \cellcolor{heatmapcolor!1!white}0.56\% & 0.21\% & \cellcolor{heatmapcolor!1!white}0.64\% & 0.28\% & 0.39\% & 0.17\% \\
\wmove{Nb8} & 0.19\% & 0.12\% & 0.07\% & 0.03\% & 0.05\% & 0.04\% & \cellcolor{heatmapcolor!1!white}0.62\% & 0.36\% \\
\wmove{Rf8} & 0.12\% & 0.05\% & 0.10\% & 0.05\% & 0.10\% & 0.06\% & \cellcolor{heatmapcolor!1!white}0.53\% & 0.21\% \\
\wmove{Kh7} & 0.02\% & 0.02\% & 0.02\% & 0.03\% & 0.05\% & 0.05\% & 0.49\% & 0.21\% \\
\wmove{Kh8} & 0.02\% & 0.02\% & 0.03\% & 0.08\% & 0.10\% & 0.09\% & 0.45\% & 0.22\% \\
\wmove{Kf8} & 0.03\% & 0.03\% & 0.04\% & 0.04\% & 0.12\% & 0.09\% & 0.30\% & 0.19\% \\
\bottomrule
\end{tabular}%
}
\end{table*}

%% file: Figures/Puzzles/Forgotten/puzzle_evaluation_58Ib0.tex
\begin{table*}[h!]
\centering
\caption{Move evaluation for puzzle \href{https://lichess.org/training/58Ib0}{\texttt{58Ib0}}: Stockfish evaluation at depth 20 and model WDL prediction for resulting positions}
\label{tab:puzzle_58Ib0_eval}
\resizebox{\textwidth}{!}{%
\begin{tabular}{llrrrrr}
\toprule
\textbf{Move} & \textbf{Stockfish} & \textbf{$\Delta$ (cp)} & \textbf{Win} & \textbf{Draw} & \textbf{Loss} & \textbf{$\Delta$ Win} \\
\midrule
\textit{Current position} & $+6.18$ & --- & $20.0\%$ & $8.2\%$ & $71.8\%$ & --- \\
\midrule
\rowcolor{heatmapcolor!30}
\wmove{Rxg7+} $\star$ & $+6.15$ & $-3$ & $99.5\%$ & $0.4\%$ & $0.1\%$ & $+79.5\%$ \\
\wmove{Kf1} & $-2.47$ & $-865$ & $0.9\%$ & $7.9\%$ & $91.1\%$ & $-19.1\%$ \\
\wmove{Qe3} & $-3.01$ & $-919$ & $0.4\%$ & $2.5\%$ & $97.1\%$ & $-19.6\%$ \\
\wmove{Qh2} & $-3.29$ & $-947$ & $0.5\%$ & $2.9\%$ & $96.6\%$ & $-19.5\%$ \\
\wmove{Rg2} & $-3.64$ & $-982$ & $0.3\%$ & $1.5\%$ & $98.2\%$ & $-19.7\%$ \\
\wmove{Qe1} & $-4.53$ & $-1071$ & $0.2\%$ & $0.7\%$ & $99.1\%$ & $-19.8\%$ \\
\wmove{Rf3} & $-7.32$ & $-1350$ & $0.1\%$ & $0.5\%$ & $99.4\%$ & $-19.9\%$ \\
\wmove{Qxa7} & $-7.41$ & $-1359$ & $0.0\%$ & $0.0\%$ & $100.0\%$ & $-20.0\%$ \\
\wmove{Rh3} & $-8.01$ & $-1419$ & $0.0\%$ & $0.1\%$ & $99.9\%$ & $-20.0\%$ \\
\wmove{Qg2} & $-8.17$ & $-1435$ & $0.0\%$ & $0.1\%$ & $99.9\%$ & $-20.0\%$ \\
\wmove{Qc5} & $-8.60$ & $-1478$ & $0.0\%$ & $0.0\%$ & $100.0\%$ & $-20.0\%$ \\
\wmove{Qf3} & $-8.60$ & $-1478$ & $0.0\%$ & $0.0\%$ & $100.0\%$ & $-20.0\%$ \\
\wmove{Qd4} & $-8.64$ & $-1482$ & $0.0\%$ & $0.0\%$ & $100.0\%$ & $-20.0\%$ \\
\wmove{Qd2} & $-8.87$ & $-1505$ & $0.0\%$ & $0.0\%$ & $100.0\%$ & $-20.0\%$ \\
\wmove{Qe2} & $-9.30$ & $-1548$ & $0.0\%$ & $0.0\%$ & $100.0\%$ & $-20.0\%$ \\
\wmove{Rg6} & $-\infty$ & $-\infty$ & $0.0\%$ & $0.0\%$ & $100.0\%$ & $-20.0\%$ \\
\wmove{Rg5} & $-\infty$ & $-\infty$ & $0.0\%$ & $0.0\%$ & $100.0\%$ & $-20.0\%$ \\
\wmove{Rg4} & $-\infty$ & $-\infty$ & $0.0\%$ & $0.0\%$ & $100.0\%$ & $-20.0\%$ \\
\wmove{Re3} & $-\infty$ & $-\infty$ & $0.0\%$ & $0.2\%$ & $99.8\%$ & $-20.0\%$ \\
\wmove{Rd3} & $-\infty$ & $-\infty$ & $0.0\%$ & $0.0\%$ & $100.0\%$ & $-20.0\%$ \\
\wmove{Rc3} & $-\infty$ & $-\infty$ & $0.0\%$ & $0.3\%$ & $99.6\%$ & $-19.9\%$ \\
\wmove{Qb6} & $-\infty$ & $-\infty$ & $0.0\%$ & $0.0\%$ & $100.0\%$ & $-20.0\%$ \\
\wmove{Qf1} & $-\infty$ & $-\infty$ & $0.0\%$ & $0.0\%$ & $100.0\%$ & $-20.0\%$ \\
\wmove{f5} & $-\infty$ & $-\infty$ & $0.0\%$ & $0.3\%$ & $99.6\%$ & $-19.9\%$ \\
\wmove{a5} & $-\infty$ & $-\infty$ & $0.1\%$ & $0.4\%$ & $99.6\%$ & $-19.9\%$ \\
\wmove{b4} & $-\infty$ & $-\infty$ & $0.0\%$ & $0.3\%$ & $99.6\%$ & $-19.9\%$ \\
\wmove{c3} & $-\infty$ & $-\infty$ & $0.1\%$ & $0.8\%$ & $99.1\%$ & $-19.9\%$ \\
\wmove{c4} & $-\infty$ & $-\infty$ & $0.0\%$ & $0.2\%$ & $99.8\%$ & $-20.0\%$ \\
\bottomrule
\end{tabular}%
}
\end{table*}

%% file: Figures/Puzzles/Forgotten/puzzle_tables_58Ib0.tex
\begin{table*}[h!]
\centering
\caption{Move probabilities by layer for puzzle \href{https://lichess.org/training/58Ib0}{\texttt{58Ib0}} (Part 1: Input to Layer 6)}
\label{tab:puzzle_58Ib0_probs1}
\resizebox{\textwidth}{!}{%
\begin{tabular}{lrrrrrrrr}
\toprule
\textbf{Move} & \textbf{Input} & \textbf{Layer 0} & \textbf{Layer 1} & \textbf{Layer 2} & \textbf{Layer 3} & \textbf{Layer 4} & \textbf{Layer 5} & \textbf{Layer 6} \\
\midrule
\wmove{Rxg7+} & \cellcolor{heatmapcolor!3!white}2.93\% & \cellcolor{heatmapcolor!53!white}52.97\% & \cellcolor{heatmapcolor!35!white}34.97\% & \cellcolor{heatmapcolor!56!white}56.12\% & \cellcolor{heatmapcolor!52!white}51.65\% & \cellcolor{heatmapcolor!54!white}54.48\% & \cellcolor{heatmapcolor!64!white}63.68\% & \cellcolor{heatmapcolor!65!white}64.57\% \\
\wmove{a5} & \cellcolor{heatmapcolor!59!white}59.08\% & \cellcolor{heatmapcolor!3!white}3.06\% & 0.04\% & 0.01\% & 0.02\% & 0.01\% & 0.02\% & 0.05\% \\
\wmove{Kf1} & 0.31\% & 0.06\% & 0.34\% & 0.29\% & 0.13\% & 0.17\% & 0.26\% & 0.45\% \\
\wmove{Qxa7} & \cellcolor{heatmapcolor!11!white}11.45\% & \cellcolor{heatmapcolor!27!white}27.21\% & \cellcolor{heatmapcolor!52!white}52.41\% & \cellcolor{heatmapcolor!32!white}31.89\% & \cellcolor{heatmapcolor!28!white}27.95\% & \cellcolor{heatmapcolor!31!white}31.24\% & \cellcolor{heatmapcolor!33!white}33.11\% & \cellcolor{heatmapcolor!30!white}29.93\% \\
\wmove{Qg2} & 0.39\% & \cellcolor{heatmapcolor!10!white}10.35\% & \cellcolor{heatmapcolor!2!white}2.08\% & 0.34\% & 0.19\% & 0.18\% & 0.04\% & 0.05\% \\
\wmove{Qb6} & \cellcolor{heatmapcolor!6!white}6.17\% & 0.13\% & 0.19\% & \cellcolor{heatmapcolor!2!white}2.31\% & \cellcolor{heatmapcolor!9!white}8.67\% & \cellcolor{heatmapcolor!2!white}2.10\% & 0.45\% & 0.37\% \\
\wmove{f5} & \cellcolor{heatmapcolor!1!white}0.69\% & \cellcolor{heatmapcolor!1!white}1.39\% & \cellcolor{heatmapcolor!3!white}2.83\% & \cellcolor{heatmapcolor!3!white}3.29\% & \cellcolor{heatmapcolor!6!white}5.62\% & 0.12\% & 0.44\% & \cellcolor{heatmapcolor!1!white}1.04\% \\
\wmove{Rg4} & 0.41\% & 0.06\% & 0.09\% & \cellcolor{heatmapcolor!1!white}0.63\% & \cellcolor{heatmapcolor!2!white}2.00\% & \cellcolor{heatmapcolor!5!white}5.14\% & \cellcolor{heatmapcolor!1!white}0.71\% & \cellcolor{heatmapcolor!1!white}1.27\% \\
\wmove{Qc5} & \cellcolor{heatmapcolor!5!white}4.78\% & 0.10\% & \cellcolor{heatmapcolor!1!white}0.85\% & 0.12\% & 0.02\% & 0.06\% & 0.04\% & 0.11\% \\
\wmove{Qe3} & \cellcolor{heatmapcolor!1!white}0.92\% & 0.07\% & 0.21\% & 0.22\% & 0.07\% & 0.31\% & 0.03\% & 0.04\% \\
\wmove{Rg2} & 0.15\% & \cellcolor{heatmapcolor!3!white}2.88\% & \cellcolor{heatmapcolor!3!white}3.08\% & 0.20\% & 0.15\% & 0.24\% & 0.05\% & 0.08\% \\
\wmove{b4} & \cellcolor{heatmapcolor!3!white}2.64\% & \cellcolor{heatmapcolor!1!white}0.58\% & 0.12\% & 0.14\% & 0.01\% & 0.01\% & 0.01\% & 0.02\% \\
\wmove{Qh2} & 0.48\% & 0.08\% & 0.15\% & 0.07\% & 0.02\% & 0.05\% & 0.02\% & 0.02\% \\
\wmove{c4} & \cellcolor{heatmapcolor!2!white}2.27\% & 0.23\% & 0.11\% & 0.10\% & 0.00\% & 0.04\% & 0.07\% & 0.26\% \\
\wmove{Rg6} & \cellcolor{heatmapcolor!2!white}1.81\% & 0.08\% & 0.50\% & \cellcolor{heatmapcolor!1!white}0.91\% & \cellcolor{heatmapcolor!1!white}0.52\% & 0.21\% & 0.29\% & 0.24\% \\
\wmove{c3} & 0.40\% & 0.06\% & 0.05\% & 0.02\% & 0.00\% & 0.00\% & 0.01\% & 0.05\% \\
\wmove{Rg5} & \cellcolor{heatmapcolor!1!white}0.92\% & 0.05\% & 0.19\% & \cellcolor{heatmapcolor!1!white}1.21\% & \cellcolor{heatmapcolor!1!white}1.45\% & \cellcolor{heatmapcolor!1!white}1.40\% & 0.15\% & \cellcolor{heatmapcolor!1!white}0.51\% \\
\wmove{Rh3} & 0.07\% & 0.05\% & 0.09\% & 0.34\% & 0.31\% & \cellcolor{heatmapcolor!1!white}1.19\% & 0.15\% & 0.22\% \\
\wmove{Qf1} & 0.02\% & 0.07\% & 0.28\% & 0.09\% & 0.03\% & 0.08\% & 0.03\% & 0.02\% \\
\wmove{Rd3} & 0.26\% & 0.06\% & 0.09\% & 0.48\% & 0.34\% & \cellcolor{heatmapcolor!1!white}0.97\% & 0.06\% & 0.11\% \\
\wmove{Qd4} & \cellcolor{heatmapcolor!1!white}0.85\% & 0.09\% & 0.19\% & 0.22\% & 0.09\% & 0.15\% & 0.04\% & 0.10\% \\
\wmove{Qe2} & \cellcolor{heatmapcolor!1!white}0.81\% & 0.06\% & 0.20\% & 0.10\% & 0.06\% & 0.13\% & 0.02\% & 0.02\% \\
\wmove{Qf3} & \cellcolor{heatmapcolor!1!white}0.75\% & 0.06\% & 0.13\% & 0.22\% & 0.25\% & 0.23\% & 0.06\% & 0.13\% \\
\wmove{Qd2} & \cellcolor{heatmapcolor!1!white}0.66\% & 0.06\% & 0.16\% & 0.11\% & 0.08\% & 0.33\% & 0.03\% & 0.04\% \\
\wmove{Rc3} & 0.21\% & 0.06\% & 0.22\% & 0.13\% & 0.03\% & 0.05\% & 0.03\% & 0.03\% \\
\wmove{Rf3} & 0.21\% & 0.05\% & 0.09\% & 0.19\% & 0.20\% & \cellcolor{heatmapcolor!1!white}0.53\% & 0.12\% & 0.21\% \\
\wmove{Re3} & 0.23\% & 0.05\% & 0.22\% & 0.13\% & 0.07\% & 0.40\% & 0.05\% & 0.06\% \\
\wmove{Qe1} & 0.15\% & 0.07\% & 0.12\% & 0.14\% & 0.07\% & 0.16\% & 0.02\% & 0.02\% \\
\bottomrule
\end{tabular}%
}
\end{table*}

\begin{table*}[h!]
\centering
\caption{Move probabilities by layer for puzzle \href{https://lichess.org/training/58Ib0}{\texttt{58Ib0}} (Part 2: Layer 7 to Final)}
\label{tab:puzzle_58Ib0_probs2}
\resizebox{\textwidth}{!}{%
\begin{tabular}{lrrrrrrrr}
\toprule
\textbf{Move} & \textbf{Layer 7} & \textbf{Layer 8} & \textbf{Layer 9} & \textbf{Layer 10} & \textbf{Layer 11} & \textbf{Layer 12} & \textbf{Layer 13} & \textbf{Final} \\
\midrule
\wmove{Rxg7+} & \cellcolor{heatmapcolor!63!white}62.81\% & \cellcolor{heatmapcolor!69!white}68.93\% & \cellcolor{heatmapcolor!82!white}82.50\% & \cellcolor{heatmapcolor!83!white}82.72\% & \cellcolor{heatmapcolor!80!white}80.06\% & \cellcolor{heatmapcolor!81!white}80.68\% & \cellcolor{heatmapcolor!51!white}50.84\% & \cellcolor{heatmapcolor!32!white}31.61\% \\
\wmove{a5} & 0.10\% & 0.09\% & 0.14\% & 0.08\% & 0.34\% & 0.07\% & \cellcolor{heatmapcolor!1!white}0.61\% & 0.26\% \\
\wmove{Kf1} & 0.31\% & \cellcolor{heatmapcolor!1!white}0.51\% & \cellcolor{heatmapcolor!1!white}0.55\% & 0.33\% & \cellcolor{heatmapcolor!1!white}1.22\% & \cellcolor{heatmapcolor!6!white}6.20\% & \cellcolor{heatmapcolor!30!white}30.42\% & \cellcolor{heatmapcolor!53!white}52.98\% \\
\wmove{Qxa7} & \cellcolor{heatmapcolor!33!white}32.68\% & \cellcolor{heatmapcolor!28!white}27.58\% & \cellcolor{heatmapcolor!14!white}14.43\% & \cellcolor{heatmapcolor!15!white}14.94\% & \cellcolor{heatmapcolor!13!white}13.42\% & \cellcolor{heatmapcolor!8!white}8.13\% & \cellcolor{heatmapcolor!2!white}2.23\% & 0.29\% \\
\wmove{Qg2} & 0.05\% & 0.05\% & 0.08\% & 0.08\% & 0.25\% & 0.35\% & \cellcolor{heatmapcolor!1!white}1.01\% & 0.33\% \\
\wmove{Qb6} & \cellcolor{heatmapcolor!1!white}0.60\% & 0.31\% & 0.17\% & 0.13\% & 0.13\% & 0.05\% & 0.27\% & 0.29\% \\
\wmove{f5} & 0.39\% & 0.28\% & 0.04\% & 0.05\% & 0.31\% & 0.07\% & \cellcolor{heatmapcolor!1!white}0.79\% & 0.23\% \\
\wmove{Rg4} & \cellcolor{heatmapcolor!1!white}0.76\% & 0.34\% & 0.29\% & 0.11\% & 0.11\% & 0.11\% & 0.29\% & 0.28\% \\
\wmove{Qc5} & 0.17\% & 0.15\% & 0.18\% & 0.11\% & \cellcolor{heatmapcolor!1!white}0.60\% & 0.23\% & \cellcolor{heatmapcolor!1!white}0.76\% & 0.30\% \\
\wmove{Qe3} & 0.07\% & 0.06\% & 0.06\% & 0.06\% & 0.16\% & 0.29\% & \cellcolor{heatmapcolor!1!white}1.40\% & \cellcolor{heatmapcolor!4!white}4.46\% \\
\wmove{Rg2} & 0.06\% & 0.07\% & 0.08\% & 0.07\% & 0.18\% & 0.14\% & \cellcolor{heatmapcolor!1!white}0.88\% & \cellcolor{heatmapcolor!2!white}1.53\% \\
\wmove{b4} & 0.08\% & 0.07\% & 0.02\% & 0.02\% & 0.08\% & 0.03\% & \cellcolor{heatmapcolor!1!white}0.77\% & 0.24\% \\
\wmove{Qh2} & 0.04\% & 0.04\% & 0.04\% & 0.03\% & 0.10\% & 0.17\% & \cellcolor{heatmapcolor!2!white}1.58\% & \cellcolor{heatmapcolor!3!white}2.60\% \\
\wmove{c4} & 0.35\% & 0.22\% & 0.06\% & 0.05\% & 0.18\% & 0.13\% & \cellcolor{heatmapcolor!1!white}0.93\% & 0.24\% \\
\wmove{Rg6} & 0.28\% & 0.19\% & 0.19\% & 0.24\% & 0.44\% & 0.12\% & 0.31\% & 0.34\% \\
\wmove{c3} & 0.10\% & 0.05\% & 0.02\% & 0.03\% & 0.15\% & 0.23\% & \cellcolor{heatmapcolor!2!white}1.75\% & 0.19\% \\
\wmove{Rg5} & 0.35\% & 0.22\% & 0.30\% & 0.34\% & 0.36\% & 0.42\% & 0.45\% & 0.29\% \\
\wmove{Rh3} & 0.20\% & 0.09\% & 0.11\% & 0.07\% & 0.12\% & 0.21\% & 0.31\% & 0.28\% \\
\wmove{Qf1} & 0.04\% & 0.04\% & 0.05\% & 0.06\% & 0.23\% & \cellcolor{heatmapcolor!1!white}0.79\% & \cellcolor{heatmapcolor!1!white}0.97\% & 0.37\% \\
\wmove{Rd3} & 0.08\% & 0.05\% & 0.06\% & 0.04\% & 0.14\% & 0.19\% & 0.34\% & 0.26\% \\
\wmove{Qd4} & 0.14\% & 0.30\% & 0.18\% & 0.14\% & 0.47\% & 0.42\% & \cellcolor{heatmapcolor!1!white}0.71\% & 0.30\% \\
\wmove{Qe2} & 0.03\% & 0.03\% & 0.03\% & 0.03\% & 0.06\% & 0.12\% & 0.33\% & 0.29\% \\
\wmove{Qf3} & 0.07\% & 0.08\% & 0.10\% & 0.08\% & 0.30\% & 0.31\% & 0.30\% & 0.26\% \\
\wmove{Qd2} & 0.03\% & 0.03\% & 0.04\% & 0.02\% & 0.05\% & 0.07\% & 0.26\% & 0.29\% \\
\wmove{Rc3} & 0.03\% & 0.08\% & 0.05\% & 0.03\% & 0.18\% & 0.10\% & \cellcolor{heatmapcolor!1!white}0.56\% & 0.27\% \\
\wmove{Rf3} & 0.10\% & 0.06\% & 0.09\% & 0.07\% & 0.19\% & 0.20\% & 0.38\% & 0.45\% \\
\wmove{Re3} & 0.06\% & 0.07\% & 0.10\% & 0.06\% & 0.10\% & 0.05\% & 0.16\% & 0.37\% \\
\wmove{Qe1} & 0.03\% & 0.03\% & 0.03\% & 0.02\% & 0.05\% & 0.12\% & 0.39\% & 0.37\% \\
\bottomrule
\end{tabular}%
}
\end{table*}

%% file: Figures/Puzzles/Forgotten/puzzle_evaluation_6PIrs.tex
\begin{table*}[h!]
\centering
\caption{Move evaluation for puzzle \href{https://lichess.org/training/6PIrs}{\texttt{6PIrs}}: Stockfish evaluation at depth 20 and model WDL prediction for resulting positions}
\label{tab:puzzle_6PIrs_eval}
\resizebox{\textwidth}{!}{%
\begin{tabular}{llrrrrr}
\toprule
\textbf{Move} & \textbf{Stockfish} & \textbf{$\Delta$ (cp)} & \textbf{Win} & \textbf{Draw} & \textbf{Loss} & \textbf{$\Delta$ Win} \\
\midrule
\textit{Current position} & $+7.17$ & --- & $1.8\%$ & $22.3\%$ & $75.9\%$ & --- \\
\midrule
\rowcolor{heatmapcolor!30}
\wmove{g5+} $\star$ & $+8.54$ & $+137$ & $98.3\%$ & $1.5\%$ & $0.2\%$ & $+96.5\%$ \\
\wmove{Qd3} & $-2.70$ & $-987$ & $0.8\%$ & $9.7\%$ & $89.5\%$ & $-1.0\%$ \\
\wmove{Qc4} & $-3.09$ & $-1026$ & $0.8\%$ & $10.1\%$ & $89.1\%$ & $-1.0\%$ \\
\wmove{Qa2} & $-3.09$ & $-1026$ & $0.5\%$ & $3.6\%$ & $95.9\%$ & $-1.3\%$ \\
\wmove{Qf3} & $-3.19$ & $-1036$ & $0.5\%$ & $4.9\%$ & $94.6\%$ & $-1.3\%$ \\
\wmove{Qd2} & $-3.24$ & $-1041$ & $0.8\%$ & $8.3\%$ & $90.9\%$ & $-1.0\%$ \\
\wmove{Qa6} & $-3.26$ & $-1043$ & $0.5\%$ & $3.8\%$ & $95.7\%$ & $-1.3\%$ \\
\wmove{Qc2} & $-3.45$ & $-1062$ & $0.6\%$ & $5.1\%$ & $94.3\%$ & $-1.2\%$ \\
\wmove{Qd1} & $-3.70$ & $-1087$ & $0.7\%$ & $8.0\%$ & $91.4\%$ & $-1.1\%$ \\
\wmove{Qh2} & $-3.72$ & $-1089$ & $0.2\%$ & $0.9\%$ & $98.8\%$ & $-1.6\%$ \\
\wmove{Qf1} & $-3.84$ & $-1101$ & $0.3\%$ & $2.0\%$ & $97.6\%$ & $-1.5\%$ \\
\wmove{Qf2} & $-4.05$ & $-1122$ & $0.3\%$ & $1.5\%$ & $98.3\%$ & $-1.5\%$ \\
\wmove{Qg2} & $-4.09$ & $-1126$ & $0.4\%$ & $2.8\%$ & $96.8\%$ & $-1.4\%$ \\
\wmove{Qxe5} & $-6.70$ & $-1387$ & $0.1\%$ & $0.1\%$ & $99.8\%$ & $-1.7\%$ \\
\wmove{Kf8} & $-12.22$ & $-1939$ & $0.0\%$ & $0.0\%$ & $100.0\%$ & $-1.8\%$ \\
\wmove{Qb2} & $-12.29$ & $-1946$ & $0.0\%$ & $0.0\%$ & $100.0\%$ & $-1.8\%$ \\
\wmove{g6} & $-12.97$ & $-2014$ & $0.0\%$ & $0.0\%$ & $100.0\%$ & $-1.8\%$ \\
\wmove{Kh8} & $-\infty$ & $-\infty$ & $0.0\%$ & $0.1\%$ & $99.9\%$ & $-1.8\%$ \\
\wmove{Kh7} & $-\infty$ & $-\infty$ & $0.0\%$ & $0.2\%$ & $99.7\%$ & $-1.8\%$ \\
\wmove{Kf7} & $-\infty$ & $-\infty$ & $0.0\%$ & $0.0\%$ & $100.0\%$ & $-1.8\%$ \\
\wmove{Qh5+} & $-\infty$ & $-\infty$ & $0.0\%$ & $0.0\%$ & $100.0\%$ & $-1.8\%$ \\
\wmove{Qb5} & $-\infty$ & $-\infty$ & $0.0\%$ & $0.0\%$ & $100.0\%$ & $-1.8\%$ \\
\wmove{Qg4+} & $-\infty$ & $-\infty$ & $0.0\%$ & $0.0\%$ & $100.0\%$ & $-1.8\%$ \\
\wmove{Qe4} & $-\infty$ & $-\infty$ & $0.0\%$ & $0.1\%$ & $99.9\%$ & $-1.8\%$ \\
\wmove{Qe3} & $-\infty$ & $-\infty$ & $0.0\%$ & $0.0\%$ & $100.0\%$ & $-1.8\%$ \\
\wmove{Qe1} & $-\infty$ & $-\infty$ & $0.0\%$ & $0.1\%$ & $99.9\%$ & $-1.8\%$ \\
\wmove{b6} & $-\infty$ & $-\infty$ & $0.0\%$ & $0.1\%$ & $99.9\%$ & $-1.8\%$ \\
\wmove{h5} & $-\infty$ & $-\infty$ & $0.0\%$ & $0.0\%$ & $100.0\%$ & $-1.8\%$ \\
\wmove{b5} & $-\infty$ & $-\infty$ & $0.0\%$ & $0.1\%$ & $99.9\%$ & $-1.8\%$ \\
\bottomrule
\end{tabular}%
}
\end{table*}

%% file: Figures/Puzzles/Forgotten/puzzle_tables_6PIrs.tex
\begin{table*}[h!]
\centering
\caption{Move probabilities by layer for puzzle \href{https://lichess.org/training/6PIrs}{\texttt{6PIrs}} (Part 1: Input to Layer 6)}
\label{tab:puzzle_6PIrs_probs1}
\resizebox{\textwidth}{!}{%
\begin{tabular}{lrrrrrrrr}
\toprule
\textbf{Move} & \textbf{Input} & \textbf{Layer 0} & \textbf{Layer 1} & \textbf{Layer 2} & \textbf{Layer 3} & \textbf{Layer 4} & \textbf{Layer 5} & \textbf{Layer 6} \\
\midrule
\wmove{Qxe5} & 0.03\% & \cellcolor{heatmapcolor!19!white}19.22\% & \cellcolor{heatmapcolor!17!white}16.74\% & \cellcolor{heatmapcolor!41!white}41.20\% & \cellcolor{heatmapcolor!47!white}46.85\% & \cellcolor{heatmapcolor!90!white}89.57\% & \cellcolor{heatmapcolor!57!white}57.18\% & \cellcolor{heatmapcolor!50!white}49.90\% \\
\wmove{Qg2} & \cellcolor{heatmapcolor!5!white}5.38\% & \cellcolor{heatmapcolor!51!white}51.44\% & \cellcolor{heatmapcolor!66!white}65.55\% & \cellcolor{heatmapcolor!3!white}2.96\% & \cellcolor{heatmapcolor!2!white}1.62\% & 0.13\% & 0.24\% & 0.19\% \\
\wmove{g5+} & 0.22\% & \cellcolor{heatmapcolor!1!white}1.01\% & 0.11\% & \cellcolor{heatmapcolor!6!white}6.18\% & \cellcolor{heatmapcolor!6!white}6.27\% & \cellcolor{heatmapcolor!1!white}0.55\% & \cellcolor{heatmapcolor!18!white}18.46\% & \cellcolor{heatmapcolor!15!white}14.62\% \\
\wmove{Qd3} & \cellcolor{heatmapcolor!6!white}6.21\% & 0.13\% & 0.13\% & 0.39\% & 0.22\% & 0.15\% & 0.37\% & 0.50\% \\
\wmove{Qd2} & \cellcolor{heatmapcolor!25!white}25.48\% & 0.14\% & 0.36\% & 0.48\% & 0.17\% & 0.11\% & 0.12\% & 0.17\% \\
\wmove{Qc4} & \cellcolor{heatmapcolor!1!white}1.15\% & 0.17\% & 0.16\% & \cellcolor{heatmapcolor!1!white}0.55\% & 0.24\% & 0.14\% & 0.40\% & \cellcolor{heatmapcolor!1!white}0.59\% \\
\wmove{Qg4+} & 0.45\% & 0.40\% & \cellcolor{heatmapcolor!1!white}1.46\% & \cellcolor{heatmapcolor!4!white}4.21\% & \cellcolor{heatmapcolor!21!white}21.21\% & \cellcolor{heatmapcolor!1!white}0.84\% & \cellcolor{heatmapcolor!3!white}3.49\% & \cellcolor{heatmapcolor!10!white}10.09\% \\
\wmove{Qd1} & \cellcolor{heatmapcolor!5!white}5.23\% & 0.17\% & 0.33\% & 0.50\% & 0.35\% & 0.15\% & \cellcolor{heatmapcolor!1!white}0.61\% & \cellcolor{heatmapcolor!1!white}1.19\% \\
\wmove{Qc2} & \cellcolor{heatmapcolor!14!white}14.22\% & 0.16\% & 0.14\% & 0.36\% & 0.24\% & 0.14\% & 0.29\% & 0.29\% \\
\wmove{Qb5} & \cellcolor{heatmapcolor!2!white}2.00\% & 0.34\% & 0.46\% & \cellcolor{heatmapcolor!14!white}14.07\% & \cellcolor{heatmapcolor!3!white}2.75\% & 0.40\% & \cellcolor{heatmapcolor!1!white}0.67\% & \cellcolor{heatmapcolor!1!white}0.65\% \\
\wmove{Qf3} & \cellcolor{heatmapcolor!8!white}7.58\% & 0.24\% & \cellcolor{heatmapcolor!3!white}2.69\% & \cellcolor{heatmapcolor!3!white}3.25\% & \cellcolor{heatmapcolor!1!white}0.89\% & 0.18\% & \cellcolor{heatmapcolor!1!white}0.93\% & \cellcolor{heatmapcolor!1!white}1.16\% \\
\wmove{h5} & 0.02\% & \cellcolor{heatmapcolor!10!white}10.08\% & \cellcolor{heatmapcolor!1!white}0.55\% & \cellcolor{heatmapcolor!6!white}6.22\% & \cellcolor{heatmapcolor!9!white}9.38\% & \cellcolor{heatmapcolor!3!white}2.67\% & \cellcolor{heatmapcolor!9!white}8.66\% & \cellcolor{heatmapcolor!4!white}4.21\% \\
\wmove{Qf2} & \cellcolor{heatmapcolor!9!white}8.87\% & 0.17\% & 0.43\% & \cellcolor{heatmapcolor!1!white}1.03\% & 0.21\% & 0.12\% & 0.22\% & 0.21\% \\
\wmove{Qh5+} & \cellcolor{heatmapcolor!1!white}0.59\% & \cellcolor{heatmapcolor!1!white}1.09\% & \cellcolor{heatmapcolor!2!white}1.88\% & \cellcolor{heatmapcolor!3!white}3.04\% & \cellcolor{heatmapcolor!2!white}2.39\% & \cellcolor{heatmapcolor!1!white}0.81\% & \cellcolor{heatmapcolor!3!white}3.19\% & \cellcolor{heatmapcolor!6!white}6.33\% \\
\wmove{Qh2} & \cellcolor{heatmapcolor!2!white}1.73\% & \cellcolor{heatmapcolor!6!white}6.20\% & \cellcolor{heatmapcolor!4!white}4.46\% & \cellcolor{heatmapcolor!1!white}1.02\% & \cellcolor{heatmapcolor!1!white}0.61\% & 0.17\% & 0.14\% & 0.13\% \\
\wmove{Qe3} & \cellcolor{heatmapcolor!6!white}5.99\% & 0.16\% & 0.14\% & \cellcolor{heatmapcolor!1!white}0.71\% & 0.23\% & 0.15\% & 0.11\% & 0.12\% \\
\wmove{Qe1} & \cellcolor{heatmapcolor!6!white}5.84\% & 0.26\% & 0.27\% & 0.34\% & 0.23\% & 0.12\% & 0.17\% & 0.15\% \\
\wmove{Qf1} & \cellcolor{heatmapcolor!5!white}5.49\% & \cellcolor{heatmapcolor!2!white}1.69\% & \cellcolor{heatmapcolor!2!white}2.47\% & \cellcolor{heatmapcolor!1!white}1.32\% & 0.30\% & 0.14\% & 0.18\% & 0.17\% \\
\wmove{Kh7} & 0.03\% & 0.12\% & 0.11\% & \cellcolor{heatmapcolor!2!white}1.70\% & \cellcolor{heatmapcolor!1!white}1.30\% & 0.41\% & \cellcolor{heatmapcolor!1!white}0.99\% & \cellcolor{heatmapcolor!2!white}2.21\% \\
\wmove{g6} & 0.03\% & \cellcolor{heatmapcolor!1!white}0.57\% & 0.06\% & \cellcolor{heatmapcolor!1!white}0.69\% & 0.11\% & 0.20\% & \cellcolor{heatmapcolor!2!white}1.66\% & \cellcolor{heatmapcolor!4!white}4.30\% \\
\wmove{b5} & 0.19\% & \cellcolor{heatmapcolor!4!white}3.90\% & 0.28\% & \cellcolor{heatmapcolor!4!white}4.28\% & \cellcolor{heatmapcolor!2!white}1.88\% & 0.19\% & 0.37\% & \cellcolor{heatmapcolor!1!white}0.74\% \\
\wmove{Kf7} & 0.01\% & 0.09\% & 0.25\% & \cellcolor{heatmapcolor!2!white}1.68\% & \cellcolor{heatmapcolor!1!white}0.81\% & \cellcolor{heatmapcolor!1!white}0.88\% & 0.43\% & \cellcolor{heatmapcolor!1!white}0.66\% \\
\wmove{Qb2} & \cellcolor{heatmapcolor!2!white}1.57\% & 0.23\% & 0.14\% & 0.36\% & 0.16\% & 0.13\% & 0.11\% & 0.11\% \\
\wmove{Qe4} & \cellcolor{heatmapcolor!1!white}1.21\% & 0.17\% & 0.11\% & \cellcolor{heatmapcolor!1!white}0.75\% & 0.24\% & 0.16\% & 0.16\% & 0.49\% \\
\wmove{b6} & 0.03\% & \cellcolor{heatmapcolor!1!white}0.99\% & 0.03\% & 0.07\% & 0.01\% & 0.00\% & 0.01\% & 0.01\% \\
\wmove{Qa6} & 0.10\% & 0.49\% & 0.28\% & \cellcolor{heatmapcolor!1!white}0.94\% & 0.18\% & 0.18\% & 0.26\% & 0.19\% \\
\wmove{Kf8} & 0.02\% & 0.10\% & 0.15\% & \cellcolor{heatmapcolor!1!white}0.87\% & \cellcolor{heatmapcolor!1!white}0.64\% & \cellcolor{heatmapcolor!1!white}0.69\% & 0.26\% & 0.28\% \\
\wmove{Kh8} & 0.06\% & 0.15\% & 0.09\% & \cellcolor{heatmapcolor!1!white}0.51\% & 0.35\% & \cellcolor{heatmapcolor!1!white}0.51\% & 0.16\% & 0.14\% \\
\wmove{Qa2} & 0.26\% & 0.11\% & 0.16\% & 0.32\% & 0.17\% & 0.12\% & 0.16\% & 0.18\% \\
\bottomrule
\end{tabular}%
}
\end{table*}

\begin{table*}[h!]
\centering
\caption{Move probabilities by layer for puzzle \href{https://lichess.org/training/6PIrs}{\texttt{6PIrs}} (Part 2: Layer 7 to Final)}
\label{tab:puzzle_6PIrs_probs2}
\resizebox{\textwidth}{!}{%
\begin{tabular}{lrrrrrrrr}
\toprule
\textbf{Move} & \textbf{Layer 7} & \textbf{Layer 8} & \textbf{Layer 9} & \textbf{Layer 10} & \textbf{Layer 11} & \textbf{Layer 12} & \textbf{Layer 13} & \textbf{Final} \\
\midrule
\wmove{Qxe5} & \cellcolor{heatmapcolor!53!white}53.49\% & \cellcolor{heatmapcolor!59!white}59.50\% & \cellcolor{heatmapcolor!78!white}77.55\% & \cellcolor{heatmapcolor!56!white}55.82\% & \cellcolor{heatmapcolor!56!white}56.34\% & \cellcolor{heatmapcolor!11!white}11.45\% & \cellcolor{heatmapcolor!3!white}2.52\% & 0.49\% \\
\wmove{Qg2} & 0.13\% & 0.07\% & 0.08\% & 0.12\% & 0.10\% & 0.12\% & 0.18\% & 0.36\% \\
\wmove{g5+} & \cellcolor{heatmapcolor!23!white}22.91\% & \cellcolor{heatmapcolor!22!white}22.21\% & 0.38\% & 0.34\% & \cellcolor{heatmapcolor!6!white}6.39\% & \cellcolor{heatmapcolor!1!white}0.77\% & \cellcolor{heatmapcolor!42!white}41.93\% & \cellcolor{heatmapcolor!11!white}11.17\% \\
\wmove{Qd3} & \cellcolor{heatmapcolor!1!white}0.88\% & \cellcolor{heatmapcolor!1!white}0.94\% & \cellcolor{heatmapcolor!3!white}2.92\% & \cellcolor{heatmapcolor!10!white}10.24\% & \cellcolor{heatmapcolor!10!white}9.81\% & \cellcolor{heatmapcolor!29!white}29.23\% & \cellcolor{heatmapcolor!14!white}13.54\% & \cellcolor{heatmapcolor!29!white}29.14\% \\
\wmove{Qd2} & 0.16\% & 0.15\% & 0.34\% & \cellcolor{heatmapcolor!2!white}1.66\% & \cellcolor{heatmapcolor!1!white}1.47\% & \cellcolor{heatmapcolor!4!white}4.29\% & \cellcolor{heatmapcolor!6!white}5.51\% & \cellcolor{heatmapcolor!10!white}9.92\% \\
\wmove{Qc4} & \cellcolor{heatmapcolor!1!white}1.00\% & \cellcolor{heatmapcolor!2!white}1.53\% & \cellcolor{heatmapcolor!1!white}1.49\% & \cellcolor{heatmapcolor!5!white}4.79\% & \cellcolor{heatmapcolor!5!white}5.07\% & \cellcolor{heatmapcolor!20!white}19.61\% & \cellcolor{heatmapcolor!14!white}14.48\% & \cellcolor{heatmapcolor!25!white}25.06\% \\
\wmove{Qg4+} & \cellcolor{heatmapcolor!4!white}3.88\% & \cellcolor{heatmapcolor!2!white}1.70\% & \cellcolor{heatmapcolor!1!white}0.76\% & \cellcolor{heatmapcolor!1!white}1.28\% & \cellcolor{heatmapcolor!1!white}1.32\% & \cellcolor{heatmapcolor!1!white}0.56\% & 0.19\% & 0.17\% \\
\wmove{Qd1} & \cellcolor{heatmapcolor!1!white}1.12\% & \cellcolor{heatmapcolor!1!white}0.94\% & \cellcolor{heatmapcolor!2!white}1.70\% & \cellcolor{heatmapcolor!6!white}5.72\% & \cellcolor{heatmapcolor!5!white}5.08\% & \cellcolor{heatmapcolor!11!white}11.34\% & \cellcolor{heatmapcolor!8!white}7.67\% & \cellcolor{heatmapcolor!14!white}14.27\% \\
\wmove{Qc2} & 0.27\% & 0.16\% & 0.20\% & 0.46\% & 0.43\% & \cellcolor{heatmapcolor!1!white}1.20\% & \cellcolor{heatmapcolor!1!white}1.15\% & \cellcolor{heatmapcolor!2!white}1.99\% \\
\wmove{Qb5} & 0.43\% & \cellcolor{heatmapcolor!1!white}0.88\% & \cellcolor{heatmapcolor!1!white}1.05\% & \cellcolor{heatmapcolor!1!white}1.15\% & \cellcolor{heatmapcolor!1!white}0.53\% & 0.47\% & 0.29\% & 0.17\% \\
\wmove{Qf3} & \cellcolor{heatmapcolor!1!white}1.21\% & \cellcolor{heatmapcolor!1!white}0.92\% & \cellcolor{heatmapcolor!1!white}0.90\% & \cellcolor{heatmapcolor!3!white}2.81\% & \cellcolor{heatmapcolor!3!white}2.88\% & \cellcolor{heatmapcolor!11!white}10.71\% & \cellcolor{heatmapcolor!3!white}3.29\% & \cellcolor{heatmapcolor!3!white}2.98\% \\
\wmove{h5} & \cellcolor{heatmapcolor!1!white}1.38\% & \cellcolor{heatmapcolor!1!white}0.99\% & \cellcolor{heatmapcolor!1!white}1.02\% & \cellcolor{heatmapcolor!1!white}0.59\% & 0.17\% & 0.16\% & 0.21\% & 0.18\% \\
\wmove{Qf2} & 0.13\% & 0.07\% & 0.08\% & 0.14\% & 0.14\% & 0.13\% & 0.23\% & 0.21\% \\
\wmove{Qh5+} & \cellcolor{heatmapcolor!2!white}2.19\% & \cellcolor{heatmapcolor!2!white}1.67\% & \cellcolor{heatmapcolor!2!white}1.97\% & \cellcolor{heatmapcolor!4!white}4.22\% & \cellcolor{heatmapcolor!2!white}2.31\% & \cellcolor{heatmapcolor!1!white}0.60\% & 0.17\% & 0.17\% \\
\wmove{Qh2} & 0.12\% & 0.09\% & 0.11\% & 0.15\% & 0.12\% & 0.11\% & 0.21\% & 0.20\% \\
\wmove{Qe3} & 0.09\% & 0.07\% & 0.09\% & 0.18\% & 0.25\% & 0.20\% & 0.30\% & 0.18\% \\
\wmove{Qe1} & 0.13\% & 0.08\% & 0.11\% & 0.21\% & 0.23\% & 0.17\% & 0.17\% & 0.18\% \\
\wmove{Qf1} & 0.14\% & 0.08\% & 0.10\% & 0.16\% & 0.11\% & 0.11\% & 0.21\% & 0.25\% \\
\wmove{Kh7} & \cellcolor{heatmapcolor!3!white}3.22\% & \cellcolor{heatmapcolor!3!white}2.93\% & \cellcolor{heatmapcolor!4!white}4.00\% & \cellcolor{heatmapcolor!4!white}4.05\% & \cellcolor{heatmapcolor!2!white}2.22\% & \cellcolor{heatmapcolor!5!white}4.85\% & \cellcolor{heatmapcolor!3!white}2.57\% & 0.18\% \\
\wmove{g6} & \cellcolor{heatmapcolor!4!white}4.15\% & \cellcolor{heatmapcolor!2!white}2.10\% & \cellcolor{heatmapcolor!1!white}1.11\% & \cellcolor{heatmapcolor!1!white}1.13\% & 0.15\% & 0.36\% & \cellcolor{heatmapcolor!1!white}1.07\% & 0.20\% \\
\wmove{b5} & \cellcolor{heatmapcolor!1!white}0.96\% & \cellcolor{heatmapcolor!1!white}1.30\% & \cellcolor{heatmapcolor!2!white}2.17\% & \cellcolor{heatmapcolor!2!white}1.52\% & \cellcolor{heatmapcolor!2!white}1.93\% & \cellcolor{heatmapcolor!1!white}0.61\% & \cellcolor{heatmapcolor!1!white}1.09\% & 0.19\% \\
\wmove{Kf7} & \cellcolor{heatmapcolor!1!white}0.70\% & 0.43\% & 0.43\% & \cellcolor{heatmapcolor!1!white}0.62\% & \cellcolor{heatmapcolor!1!white}0.73\% & \cellcolor{heatmapcolor!1!white}0.91\% & 0.30\% & 0.18\% \\
\wmove{Qb2} & 0.10\% & 0.07\% & 0.10\% & 0.16\% & 0.15\% & 0.13\% & 0.26\% & 0.18\% \\
\wmove{Qe4} & 0.46\% & 0.41\% & 0.49\% & \cellcolor{heatmapcolor!1!white}1.19\% & \cellcolor{heatmapcolor!1!white}0.89\% & 0.25\% & 0.38\% & 0.18\% \\
\wmove{b6} & 0.02\% & 0.02\% & 0.01\% & 0.01\% & 0.02\% & 0.01\% & \cellcolor{heatmapcolor!1!white}0.71\% & 0.17\% \\
\wmove{Qa6} & 0.17\% & 0.20\% & 0.20\% & 0.30\% & 0.22\% & 0.40\% & \cellcolor{heatmapcolor!1!white}0.60\% & \cellcolor{heatmapcolor!1!white}0.81\% \\
\wmove{Kf8} & 0.27\% & 0.21\% & 0.33\% & 0.42\% & \cellcolor{heatmapcolor!1!white}0.64\% & \cellcolor{heatmapcolor!1!white}0.80\% & 0.20\% & 0.17\% \\
\wmove{Kh8} & 0.16\% & 0.19\% & 0.22\% & 0.41\% & 0.17\% & 0.28\% & 0.28\% & 0.18\% \\
\wmove{Qa2} & 0.13\% & 0.09\% & 0.10\% & 0.16\% & 0.12\% & 0.16\% & 0.31\% & 0.48\% \\
\bottomrule
\end{tabular}%
}
\end{table*}

%% file: Figures/Puzzles/Forgotten/puzzle_evaluation_1Egyn.tex
\begin{table*}[h!]
\centering
\caption{Move evaluation for puzzle \href{https://lichess.org/training/1Egyn}{\texttt{1Egyn}}: Stockfish evaluation at depth 20 and model WDL prediction for resulting positions}
\label{tab:puzzle_1Egyn_eval}
\resizebox{\textwidth}{!}{%
\begin{tabular}{llrrrrr}
\toprule
\textbf{Move} & \textbf{Stockfish} & \textbf{$\Delta$ (cp)} & \textbf{Win} & \textbf{Draw} & \textbf{Loss} & \textbf{$\Delta$ Win} \\
\midrule
\textit{Current position} & $+\infty$ & --- & $12.3\%$ & $5.4\%$ & $82.3\%$ & --- \\
\midrule
\rowcolor{heatmapcolor!30}
\wmove{Qxc8+} $\star$ & $+\infty$ & --- & $100.0\%$ & $0.0\%$ & $0.0\%$ & $+87.7\%$ \\
\wmove{Qxa7} & $-4.62$ & $-\infty$ & $0.3\%$ & $0.5\%$ & $99.2\%$ & $-12.0\%$ \\
\wmove{Qe4} & $-7.11$ & $-\infty$ & $0.0\%$ & $0.0\%$ & $100.0\%$ & $-12.3\%$ \\
\wmove{Qb4} & $-7.23$ & $-\infty$ & $0.0\%$ & $0.0\%$ & $99.9\%$ & $-12.3\%$ \\
\wmove{Qc6} & $-7.39$ & $-\infty$ & $0.0\%$ & $0.0\%$ & $100.0\%$ & $-12.3\%$ \\
\wmove{Qb6} & $-7.64$ & $-\infty$ & $0.0\%$ & $0.0\%$ & $99.9\%$ & $-12.3\%$ \\
\wmove{Qf3} & $-7.67$ & $-\infty$ & $0.0\%$ & $0.0\%$ & $99.9\%$ & $-12.3\%$ \\
\wmove{Qb5} & $-7.70$ & $-\infty$ & $0.0\%$ & $0.0\%$ & $100.0\%$ & $-12.3\%$ \\
\wmove{h4} & $-8.96$ & $-\infty$ & $0.0\%$ & $0.0\%$ & $100.0\%$ & $-12.3\%$ \\
\wmove{h3} & $-9.11$ & $-\infty$ & $0.0\%$ & $0.0\%$ & $100.0\%$ & $-12.3\%$ \\
\wmove{Qe7} & $-9.17$ & $-\infty$ & $0.0\%$ & $0.0\%$ & $100.0\%$ & $-12.3\%$ \\
\wmove{d4} & $-9.18$ & $-\infty$ & $0.0\%$ & $0.0\%$ & $100.0\%$ & $-12.3\%$ \\
\wmove{c4} & $-9.22$ & $-\infty$ & $0.0\%$ & $0.0\%$ & $100.0\%$ & $-12.3\%$ \\
\wmove{b4} & $-9.33$ & $-\infty$ & $0.0\%$ & $0.0\%$ & $100.0\%$ & $-12.3\%$ \\
\wmove{g3} & $-9.33$ & $-\infty$ & $0.0\%$ & $0.0\%$ & $100.0\%$ & $-12.3\%$ \\
\wmove{Qc7} & $-9.35$ & $-\infty$ & $0.0\%$ & $0.0\%$ & $100.0\%$ & $-12.3\%$ \\
\wmove{Rf1} & $-9.44$ & $-\infty$ & $0.0\%$ & $0.0\%$ & $100.0\%$ & $-12.3\%$ \\
\wmove{g4} & $-9.46$ & $-\infty$ & $0.0\%$ & $0.0\%$ & $100.0\%$ & $-12.3\%$ \\
\wmove{Kh1} & $-9.47$ & $-\infty$ & $0.0\%$ & $0.0\%$ & $100.0\%$ & $-12.3\%$ \\
\wmove{c3} & $-9.52$ & $-\infty$ & $0.0\%$ & $0.0\%$ & $100.0\%$ & $-12.3\%$ \\
\wmove{f3} & $-9.55$ & $-\infty$ & $0.0\%$ & $0.0\%$ & $100.0\%$ & $-12.3\%$ \\
\wmove{Rd1} & $-9.66$ & $-\infty$ & $0.0\%$ & $0.0\%$ & $100.0\%$ & $-12.3\%$ \\
\wmove{Kf1} & $-9.68$ & $-\infty$ & $0.0\%$ & $0.0\%$ & $100.0\%$ & $-12.3\%$ \\
\wmove{f4} & $-10.00$ & $-\infty$ & $0.0\%$ & $0.0\%$ & $100.0\%$ & $-12.3\%$ \\
\wmove{Qa8} & $-10.26$ & $-\infty$ & $0.0\%$ & $0.0\%$ & $100.0\%$ & $-12.3\%$ \\
\wmove{Qb8} & $-10.29$ & $-\infty$ & $0.0\%$ & $0.0\%$ & $100.0\%$ & $-12.3\%$ \\
\wmove{Qxf7+} & $-11.09$ & $-\infty$ & $0.0\%$ & $0.0\%$ & $100.0\%$ & $-12.3\%$ \\
\wmove{Qa6} & $-11.11$ & $-\infty$ & $0.0\%$ & $0.0\%$ & $100.0\%$ & $-12.3\%$ \\
\wmove{Qd7} & $-12.92$ & $-\infty$ & $0.0\%$ & $0.0\%$ & $100.0\%$ & $-12.3\%$ \\
\wmove{Qd5} & $-\infty$ & $-\infty$ & $0.0\%$ & $0.0\%$ & $100.0\%$ & $-12.3\%$ \\
\wmove{Rxe6} & $-\infty$ & $-\infty$ & $0.0\%$ & $0.0\%$ & $100.0\%$ & $-12.3\%$ \\
\wmove{Re5} & $-\infty$ & $-\infty$ & $0.0\%$ & $0.0\%$ & $100.0\%$ & $-12.3\%$ \\
\wmove{Re4} & $-\infty$ & $-\infty$ & $0.0\%$ & $0.0\%$ & $100.0\%$ & $-12.3\%$ \\
\wmove{Re3} & $-\infty$ & $-\infty$ & $0.0\%$ & $0.0\%$ & $100.0\%$ & $-12.3\%$ \\
\wmove{Re2} & $-\infty$ & $-\infty$ & $0.0\%$ & $0.0\%$ & $100.0\%$ & $-12.3\%$ \\
\wmove{Rc1} & $-\infty$ & $-\infty$ & $0.0\%$ & $0.0\%$ & $100.0\%$ & $-12.3\%$ \\
\wmove{Rb1} & $-\infty$ & $-\infty$ & $0.0\%$ & $0.0\%$ & $100.0\%$ & $-12.3\%$ \\
\wmove{Ra1} & $-\infty$ & $-\infty$ & $0.0\%$ & $0.0\%$ & $100.0\%$ & $-12.3\%$ \\
\bottomrule
\end{tabular}%
}
\end{table*}

%% file: Figures/Puzzles/Forgotten/puzzle_tables_1Egyn.tex
\begin{table*}[h!]
\centering
\caption{Move probabilities by layer for puzzle \href{https://lichess.org/training/1Egyn}{\texttt{1Egyn}} (Part 1: Input to Layer 6)}
\label{tab:puzzle_1Egyn_probs1}
\resizebox{\textwidth}{!}{%
\begin{tabular}{lrrrrrrrr}
\toprule
\textbf{Move} & \textbf{Input} & \textbf{Layer 0} & \textbf{Layer 1} & \textbf{Layer 2} & \textbf{Layer 3} & \textbf{Layer 4} & \textbf{Layer 5} & \textbf{Layer 6} \\
\midrule
\wmove{Qxa7} & 0.13\% & \cellcolor{heatmapcolor!58!white}58.43\% & \cellcolor{heatmapcolor!73!white}72.93\% & \cellcolor{heatmapcolor!39!white}38.65\% & \cellcolor{heatmapcolor!12!white}12.37\% & \cellcolor{heatmapcolor!4!white}3.65\% & \cellcolor{heatmapcolor!3!white}3.45\% & \cellcolor{heatmapcolor!3!white}2.67\% \\
\wmove{Qxc8+} & \cellcolor{heatmapcolor!5!white}5.06\% & \cellcolor{heatmapcolor!35!white}35.05\% & \cellcolor{heatmapcolor!10!white}9.71\% & \cellcolor{heatmapcolor!37!white}36.67\% & \cellcolor{heatmapcolor!49!white}49.39\% & \cellcolor{heatmapcolor!55!white}54.67\% & \cellcolor{heatmapcolor!73!white}72.84\% & \cellcolor{heatmapcolor!49!white}48.69\% \\
\wmove{Rxe6} & \cellcolor{heatmapcolor!2!white}1.67\% & 0.17\% & 0.20\% & 0.40\% & \cellcolor{heatmapcolor!4!white}3.85\% & \cellcolor{heatmapcolor!14!white}14.24\% & \cellcolor{heatmapcolor!8!white}8.29\% & \cellcolor{heatmapcolor!13!white}13.42\% \\
\wmove{Qd7} & \cellcolor{heatmapcolor!27!white}27.16\% & 0.05\% & 0.33\% & \cellcolor{heatmapcolor!1!white}0.68\% & 0.40\% & 0.11\% & 0.10\% & 0.09\% \\
\wmove{Qe7} & \cellcolor{heatmapcolor!27!white}26.77\% & 0.05\% & 0.49\% & 0.36\% & \cellcolor{heatmapcolor!1!white}0.62\% & 0.20\% & 0.09\% & 0.20\% \\
\wmove{c4} & 0.33\% & 0.04\% & 0.38\% & 0.20\% & 0.31\% & 0.32\% & \cellcolor{heatmapcolor!1!white}0.84\% & \cellcolor{heatmapcolor!3!white}2.51\% \\
\wmove{Qc7} & \cellcolor{heatmapcolor!11!white}11.15\% & 0.16\% & 0.10\% & 0.08\% & 0.16\% & 0.15\% & 0.11\% & 0.15\% \\
\wmove{Qb8} & \cellcolor{heatmapcolor!8!white}7.99\% & 0.11\% & 0.37\% & \cellcolor{heatmapcolor!1!white}0.53\% & \cellcolor{heatmapcolor!1!white}1.13\% & \cellcolor{heatmapcolor!1!white}0.67\% & 0.40\% & \cellcolor{heatmapcolor!1!white}0.88\% \\
\wmove{Rc1} & 0.07\% & 0.04\% & 0.29\% & \cellcolor{heatmapcolor!1!white}1.46\% & \cellcolor{heatmapcolor!5!white}5.14\% & \cellcolor{heatmapcolor!5!white}4.71\% & \cellcolor{heatmapcolor!4!white}3.90\% & \cellcolor{heatmapcolor!8!white}7.98\% \\
\wmove{Rb1} & 0.10\% & 0.05\% & 0.34\% & \cellcolor{heatmapcolor!4!white}3.68\% & \cellcolor{heatmapcolor!7!white}6.77\% & \cellcolor{heatmapcolor!4!white}3.81\% & \cellcolor{heatmapcolor!2!white}2.34\% & \cellcolor{heatmapcolor!6!white}5.70\% \\
\wmove{h3} & 0.01\% & 0.05\% & 0.20\% & \cellcolor{heatmapcolor!1!white}0.57\% & \cellcolor{heatmapcolor!1!white}1.23\% & 0.19\% & 0.38\% & \cellcolor{heatmapcolor!5!white}4.53\% \\
\wmove{Qxf7+} & \cellcolor{heatmapcolor!2!white}2.32\% & \cellcolor{heatmapcolor!1!white}1.37\% & \cellcolor{heatmapcolor!3!white}2.62\% & \cellcolor{heatmapcolor!3!white}2.94\% & \cellcolor{heatmapcolor!2!white}2.18\% & \cellcolor{heatmapcolor!3!white}3.33\% & \cellcolor{heatmapcolor!2!white}1.76\% & \cellcolor{heatmapcolor!1!white}0.84\% \\
\wmove{Qc6} & \cellcolor{heatmapcolor!5!white}4.90\% & 0.05\% & 0.15\% & 0.36\% & \cellcolor{heatmapcolor!1!white}0.54\% & 0.21\% & 0.15\% & 0.12\% \\
\wmove{Qa8} & \cellcolor{heatmapcolor!4!white}4.40\% & \cellcolor{heatmapcolor!3!white}2.71\% & \cellcolor{heatmapcolor!2!white}1.75\% & \cellcolor{heatmapcolor!1!white}1.08\% & \cellcolor{heatmapcolor!1!white}0.67\% & 0.42\% & 0.37\% & \cellcolor{heatmapcolor!1!white}1.33\% \\
\wmove{Ra1} & 0.10\% & 0.15\% & \cellcolor{heatmapcolor!1!white}0.60\% & \cellcolor{heatmapcolor!4!white}4.11\% & \cellcolor{heatmapcolor!4!white}3.70\% & \cellcolor{heatmapcolor!4!white}3.51\% & \cellcolor{heatmapcolor!1!white}1.24\% & \cellcolor{heatmapcolor!2!white}2.08\% \\
\wmove{Re2} & 0.03\% & 0.04\% & 0.24\% & 0.44\% & \cellcolor{heatmapcolor!1!white}1.20\% & \cellcolor{heatmapcolor!2!white}1.75\% & 0.41\% & \cellcolor{heatmapcolor!3!white}3.42\% \\
\wmove{Rd1} & 0.07\% & 0.05\% & 0.32\% & 0.49\% & \cellcolor{heatmapcolor!2!white}2.01\% & \cellcolor{heatmapcolor!2!white}2.44\% & \cellcolor{heatmapcolor!1!white}1.32\% & \cellcolor{heatmapcolor!2!white}1.62\% \\
\wmove{f3} & 0.05\% & 0.05\% & 0.37\% & 0.15\% & 0.17\% & 0.06\% & 0.04\% & 0.09\% \\
\wmove{Qb6} & \cellcolor{heatmapcolor!2!white}1.74\% & 0.04\% & 0.17\% & 0.28\% & 0.37\% & 0.24\% & 0.27\% & 0.25\% \\
\wmove{Re5} & \cellcolor{heatmapcolor!2!white}1.63\% & 0.04\% & 0.20\% & \cellcolor{heatmapcolor!1!white}0.74\% & \cellcolor{heatmapcolor!1!white}1.06\% & \cellcolor{heatmapcolor!1!white}1.35\% & 0.31\% & \cellcolor{heatmapcolor!2!white}1.64\% \\
\wmove{Qd5} & \cellcolor{heatmapcolor!2!white}1.61\% & 0.05\% & \cellcolor{heatmapcolor!1!white}1.44\% & 0.27\% & 0.42\% & 0.11\% & 0.06\% & 0.03\% \\
\wmove{Re4} & 0.01\% & 0.04\% & 0.41\% & 0.20\% & 0.27\% & \cellcolor{heatmapcolor!1!white}1.05\% & 0.11\% & 0.29\% \\
\wmove{d4} & 0.15\% & 0.06\% & \cellcolor{heatmapcolor!1!white}0.73\% & \cellcolor{heatmapcolor!1!white}1.03\% & \cellcolor{heatmapcolor!1!white}0.57\% & 0.08\% & 0.08\% & 0.08\% \\
\wmove{c3} & 0.06\% & 0.09\% & \cellcolor{heatmapcolor!1!white}0.96\% & 0.39\% & 0.27\% & 0.06\% & 0.02\% & 0.02\% \\
\wmove{h4} & 0.03\% & 0.12\% & 0.17\% & 0.27\% & 0.35\% & 0.08\% & 0.08\% & 0.18\% \\
\wmove{Kh1} & 0.12\% & 0.21\% & \cellcolor{heatmapcolor!1!white}0.87\% & 0.29\% & 0.49\% & 0.39\% & 0.08\% & 0.11\% \\
\wmove{Qb5} & 0.37\% & 0.04\% & 0.24\% & 0.37\% & \cellcolor{heatmapcolor!1!white}0.69\% & 0.22\% & 0.14\% & 0.10\% \\
\wmove{Re3} & 0.00\% & 0.04\% & 0.20\% & 0.39\% & 0.36\% & \cellcolor{heatmapcolor!1!white}0.67\% & 0.05\% & 0.07\% \\
\wmove{Qb4} & 0.37\% & 0.07\% & 0.47\% & \cellcolor{heatmapcolor!1!white}0.51\% & \cellcolor{heatmapcolor!1!white}0.60\% & 0.14\% & 0.10\% & 0.05\% \\
\wmove{Qa6} & 0.03\% & 0.05\% & 0.15\% & 0.27\% & \cellcolor{heatmapcolor!1!white}0.59\% & 0.34\% & 0.25\% & 0.30\% \\
\wmove{b4} & 0.38\% & 0.09\% & 0.36\% & 0.46\% & 0.34\% & 0.08\% & 0.07\% & 0.11\% \\
\wmove{g4} & 0.38\% & 0.09\% & \cellcolor{heatmapcolor!1!white}0.57\% & 0.17\% & 0.11\% & 0.03\% & 0.01\% & 0.01\% \\
\wmove{f4} & \cellcolor{heatmapcolor!1!white}0.57\% & 0.05\% & 0.29\% & 0.17\% & 0.13\% & 0.03\% & 0.02\% & 0.04\% \\
\wmove{Qf3} & 0.03\% & 0.05\% & 0.37\% & 0.34\% & 0.44\% & 0.13\% & 0.07\% & 0.07\% \\
\wmove{Kf1} & 0.05\% & 0.09\% & 0.24\% & 0.31\% & 0.42\% & 0.27\% & 0.05\% & 0.07\% \\
\wmove{g3} & 0.06\% & 0.04\% & 0.25\% & 0.15\% & 0.23\% & 0.07\% & 0.07\% & 0.10\% \\
\wmove{Qe4} & 0.03\% & 0.04\% & 0.33\% & 0.36\% & 0.27\% & 0.14\% & 0.08\% & 0.14\% \\
\wmove{Rf1} & 0.07\% & 0.05\% & 0.19\% & 0.19\% & 0.15\% & 0.05\% & 0.03\% & 0.03\% \\
\bottomrule
\end{tabular}%
}
\end{table*}

\begin{table*}[h!]
\centering
\caption{Move probabilities by layer for puzzle \href{https://lichess.org/training/1Egyn}{\texttt{1Egyn}} (Part 2: Layer 7 to Final)}
\label{tab:puzzle_1Egyn_probs2}
\resizebox{\textwidth}{!}{%
\begin{tabular}{lrrrrrrrr}
\toprule
\textbf{Move} & \textbf{Layer 7} & \textbf{Layer 8} & \textbf{Layer 9} & \textbf{Layer 10} & \textbf{Layer 11} & \textbf{Layer 12} & \textbf{Layer 13} & \textbf{Final} \\
\midrule
\wmove{Qxa7} & \cellcolor{heatmapcolor!2!white}1.84\% & \cellcolor{heatmapcolor!7!white}6.79\% & \cellcolor{heatmapcolor!5!white}5.08\% & \cellcolor{heatmapcolor!4!white}3.98\% & \cellcolor{heatmapcolor!5!white}5.37\% & \cellcolor{heatmapcolor!12!white}12.37\% & \cellcolor{heatmapcolor!23!white}23.09\% & \cellcolor{heatmapcolor!64!white}63.67\% \\
\wmove{Qxc8+} & \cellcolor{heatmapcolor!47!white}47.17\% & \cellcolor{heatmapcolor!30!white}30.34\% & \cellcolor{heatmapcolor!32!white}31.97\% & \cellcolor{heatmapcolor!43!white}42.62\% & \cellcolor{heatmapcolor!58!white}58.43\% & \cellcolor{heatmapcolor!61!white}61.50\% & \cellcolor{heatmapcolor!54!white}54.05\% & \cellcolor{heatmapcolor!29!white}28.66\% \\
\wmove{Rxe6} & \cellcolor{heatmapcolor!13!white}13.02\% & \cellcolor{heatmapcolor!26!white}25.77\% & \cellcolor{heatmapcolor!29!white}29.43\% & \cellcolor{heatmapcolor!29!white}29.42\% & \cellcolor{heatmapcolor!20!white}20.49\% & \cellcolor{heatmapcolor!11!white}10.78\% & \cellcolor{heatmapcolor!3!white}3.06\% & 0.16\% \\
\wmove{Qd7} & 0.06\% & 0.07\% & 0.20\% & 0.12\% & 0.09\% & 0.06\% & 0.12\% & 0.21\% \\
\wmove{Qe7} & 0.10\% & 0.11\% & 0.19\% & 0.08\% & 0.07\% & 0.06\% & 0.23\% & 0.22\% \\
\wmove{c4} & \cellcolor{heatmapcolor!4!white}3.99\% & \cellcolor{heatmapcolor!9!white}9.02\% & \cellcolor{heatmapcolor!12!white}11.81\% & \cellcolor{heatmapcolor!16!white}16.30\% & \cellcolor{heatmapcolor!10!white}10.39\% & \cellcolor{heatmapcolor!9!white}8.67\% & \cellcolor{heatmapcolor!2!white}1.58\% & 0.23\% \\
\wmove{Qc7} & 0.10\% & 0.09\% & 0.21\% & 0.13\% & 0.08\% & 0.04\% & 0.13\% & 0.20\% \\
\wmove{Qb8} & \cellcolor{heatmapcolor!1!white}1.38\% & \cellcolor{heatmapcolor!1!white}0.85\% & \cellcolor{heatmapcolor!1!white}1.00\% & 0.39\% & 0.43\% & 0.23\% & 0.16\% & 0.19\% \\
\wmove{Rc1} & \cellcolor{heatmapcolor!7!white}6.78\% & 0.34\% & \cellcolor{heatmapcolor!1!white}1.03\% & \cellcolor{heatmapcolor!1!white}0.51\% & 0.47\% & \cellcolor{heatmapcolor!1!white}0.54\% & 0.14\% & 0.22\% \\
\wmove{Rb1} & \cellcolor{heatmapcolor!8!white}7.56\% & \cellcolor{heatmapcolor!4!white}4.31\% & \cellcolor{heatmapcolor!4!white}3.81\% & \cellcolor{heatmapcolor!1!white}0.66\% & 0.49\% & 0.39\% & 0.25\% & 0.20\% \\
\wmove{h3} & \cellcolor{heatmapcolor!7!white}7.37\% & \cellcolor{heatmapcolor!7!white}7.22\% & \cellcolor{heatmapcolor!2!white}2.36\% & \cellcolor{heatmapcolor!2!white}1.75\% & \cellcolor{heatmapcolor!1!white}1.06\% & \cellcolor{heatmapcolor!3!white}3.46\% & \cellcolor{heatmapcolor!7!white}6.99\% & 0.23\% \\
\wmove{Qxf7+} & \cellcolor{heatmapcolor!1!white}1.35\% & \cellcolor{heatmapcolor!6!white}5.57\% & \cellcolor{heatmapcolor!2!white}2.49\% & \cellcolor{heatmapcolor!1!white}1.07\% & 0.35\% & 0.28\% & \cellcolor{heatmapcolor!2!white}1.52\% & 0.19\% \\
\wmove{Qc6} & 0.07\% & 0.08\% & 0.16\% & 0.09\% & 0.07\% & 0.05\% & 0.45\% & 0.21\% \\
\wmove{Qa8} & \cellcolor{heatmapcolor!1!white}1.16\% & 0.45\% & \cellcolor{heatmapcolor!1!white}1.08\% & 0.41\% & 0.33\% & 0.16\% & 0.10\% & 0.21\% \\
\wmove{Ra1} & \cellcolor{heatmapcolor!3!white}2.89\% & \cellcolor{heatmapcolor!3!white}3.44\% & \cellcolor{heatmapcolor!2!white}1.64\% & 0.32\% & 0.23\% & 0.24\% & 0.18\% & 0.22\% \\
\wmove{Re2} & \cellcolor{heatmapcolor!2!white}1.88\% & 0.45\% & \cellcolor{heatmapcolor!1!white}0.79\% & 0.13\% & 0.06\% & 0.09\% & 0.27\% & 0.22\% \\
\wmove{Rd1} & \cellcolor{heatmapcolor!1!white}0.76\% & \cellcolor{heatmapcolor!2!white}1.58\% & \cellcolor{heatmapcolor!3!white}2.81\% & 0.33\% & 0.19\% & 0.06\% & 0.13\% & 0.21\% \\
\wmove{f3} & 0.35\% & 0.31\% & 0.24\% & 0.02\% & 0.03\% & 0.04\% & \cellcolor{heatmapcolor!2!white}1.76\% & 0.20\% \\
\wmove{Qb6} & 0.21\% & 0.19\% & 0.28\% & 0.11\% & 0.07\% & 0.04\% & 0.19\% & 0.23\% \\
\wmove{Re5} & \cellcolor{heatmapcolor!1!white}0.58\% & \cellcolor{heatmapcolor!1!white}0.98\% & \cellcolor{heatmapcolor!1!white}1.41\% & \cellcolor{heatmapcolor!1!white}0.63\% & \cellcolor{heatmapcolor!1!white}0.77\% & 0.26\% & 0.37\% & 0.21\% \\
\wmove{Qd5} & 0.02\% & 0.03\% & 0.05\% & 0.04\% & 0.02\% & 0.04\% & 0.17\% & 0.21\% \\
\wmove{Re4} & 0.11\% & 0.07\% & 0.14\% & 0.03\% & 0.03\% & 0.05\% & \cellcolor{heatmapcolor!1!white}0.62\% & 0.21\% \\
\wmove{d4} & 0.06\% & 0.23\% & 0.07\% & 0.03\% & 0.02\% & 0.02\% & 0.32\% & 0.21\% \\
\wmove{c3} & 0.02\% & 0.04\% & 0.08\% & 0.06\% & 0.03\% & 0.02\% & 0.12\% & 0.24\% \\
\wmove{h4} & 0.24\% & 0.33\% & 0.23\% & 0.06\% & 0.08\% & 0.15\% & \cellcolor{heatmapcolor!1!white}0.94\% & 0.22\% \\
\wmove{Kh1} & 0.09\% & 0.13\% & 0.33\% & 0.36\% & 0.09\% & 0.08\% & 0.25\% & 0.18\% \\
\wmove{Qb5} & 0.07\% & 0.09\% & 0.14\% & 0.05\% & 0.03\% & 0.03\% & 0.16\% & 0.23\% \\
\wmove{Re3} & 0.03\% & 0.04\% & 0.04\% & 0.01\% & 0.00\% & 0.02\% & 0.12\% & 0.22\% \\
\wmove{Qb4} & 0.03\% & 0.07\% & 0.05\% & 0.01\% & 0.01\% & 0.01\% & 0.15\% & 0.22\% \\
\wmove{Qa6} & 0.19\% & 0.11\% & 0.25\% & 0.04\% & 0.03\% & 0.02\% & 0.21\% & 0.23\% \\
\wmove{b4} & 0.13\% & 0.39\% & 0.10\% & 0.02\% & 0.01\% & 0.01\% & \cellcolor{heatmapcolor!1!white}0.58\% & 0.22\% \\
\wmove{g4} & 0.01\% & 0.01\% & 0.02\% & 0.01\% & 0.01\% & 0.02\% & 0.13\% & 0.19\% \\
\wmove{f4} & 0.04\% & 0.08\% & 0.18\% & 0.08\% & 0.05\% & 0.08\% & 0.35\% & 0.18\% \\
\wmove{Qf3} & 0.09\% & 0.19\% & 0.12\% & 0.03\% & 0.03\% & 0.02\% & 0.24\% & 0.23\% \\
\wmove{Kf1} & 0.05\% & 0.05\% & 0.05\% & 0.03\% & 0.01\% & 0.02\% & 0.15\% & 0.23\% \\
\wmove{g3} & 0.08\% & 0.07\% & 0.07\% & 0.01\% & 0.03\% & 0.05\% & 0.39\% & 0.19\% \\
\wmove{Qe4} & 0.07\% & 0.09\% & 0.08\% & 0.04\% & 0.04\% & 0.04\% & 0.26\% & 0.24\% \\
\wmove{Rf1} & 0.05\% & 0.02\% & 0.02\% & 0.01\% & 0.00\% & 0.02\% & 0.08\% & 0.25\% \\
\bottomrule
\end{tabular}%
}
\end{table*}

%% file: Figures/Puzzles/Forgotten/puzzle_evaluation_DsaGi.tex
\begin{table*}[h!]
\centering
\caption{Move evaluation for puzzle \href{https://lichess.org/training/DsaGi}{\texttt{DsaGi}}: Stockfish evaluation at depth 20 and model WDL prediction for resulting positions}
\label{tab:puzzle_DsaGi_eval}
\resizebox{\textwidth}{!}{%
\begin{tabular}{llrrrrr}
\toprule
\textbf{Move} & \textbf{Stockfish} & \textbf{$\Delta$ (cp)} & \textbf{Win} & \textbf{Draw} & \textbf{Loss} & \textbf{$\Delta$ Win} \\
\midrule
\textit{Current position} & $+5.76$ & --- & $4.5\%$ & $23.1\%$ & $72.4\%$ & --- \\
\midrule
\rowcolor{heatmapcolor!30}
\wmove{Rxg4+} $\star$ & $+6.10$ & $+34$ & $97.9\%$ & $1.9\%$ & $0.2\%$ & $+93.4\%$ \\
\wmove{Rf1+} & $-0.05$ & $-581$ & $0.8\%$ & $25.6\%$ & $73.6\%$ & $-3.7\%$ \\
\wmove{Rh1} & $-0.15$ & $-591$ & $0.9\%$ & $35.8\%$ & $63.3\%$ & $-3.7\%$ \\
\wmove{Kg7} & $-2.46$ & $-822$ & $0.6\%$ & $8.9\%$ & $90.5\%$ & $-4.0\%$ \\
\wmove{Rc1} & $-2.75$ & $-851$ & $0.3\%$ & $3.7\%$ & $96.0\%$ & $-4.2\%$ \\
\wmove{Rd1} & $-2.77$ & $-853$ & $0.1\%$ & $0.9\%$ & $99.0\%$ & $-4.4\%$ \\
\wmove{Ra1} & $-2.81$ & $-857$ & $0.2\%$ & $1.9\%$ & $97.9\%$ & $-4.3\%$ \\
\wmove{Rb1} & $-2.86$ & $-862$ & $0.2\%$ & $1.3\%$ & $98.5\%$ & $-4.3\%$ \\
\wmove{Bc3} & $-3.28$ & $-904$ & $0.2\%$ & $2.4\%$ & $97.3\%$ & $-4.3\%$ \\
\wmove{Bb2} & $-3.48$ & $-924$ & $0.3\%$ & $4.1\%$ & $95.6\%$ & $-4.2\%$ \\
\wmove{Bg7} & $-3.50$ & $-926$ & $0.2\%$ & $1.1\%$ & $98.7\%$ & $-4.4\%$ \\
\wmove{Bh8} & $-3.56$ & $-932$ & $0.1\%$ & $0.9\%$ & $98.9\%$ & $-4.4\%$ \\
\wmove{Kf8} & $-3.63$ & $-939$ & $0.1\%$ & $0.7\%$ & $99.1\%$ & $-4.4\%$ \\
\wmove{Rg2} & $-3.64$ & $-940$ & $0.3\%$ & $5.0\%$ & $94.7\%$ & $-4.2\%$ \\
\wmove{Ba1} & $-3.75$ & $-951$ & $0.2\%$ & $1.1\%$ & $98.8\%$ & $-4.4\%$ \\
\wmove{a4} & $-3.75$ & $-951$ & $0.5\%$ & $13.1\%$ & $86.4\%$ & $-4.1\%$ \\
\wmove{c4} & $-3.88$ & $-964$ & $0.2\%$ & $1.4\%$ & $98.4\%$ & $-4.4\%$ \\
\wmove{Be5+} & $-7.19$ & $-1295$ & $0.0\%$ & $0.0\%$ & $99.9\%$ & $-4.5\%$ \\
\wmove{Rg3} & $-7.42$ & $-1318$ & $0.0\%$ & $0.1\%$ & $99.9\%$ & $-4.5\%$ \\
\wmove{Re1} & $-7.42$ & $-1318$ & $0.0\%$ & $0.1\%$ & $99.9\%$ & $-4.5\%$ \\
\wmove{Bf6} & $-7.49$ & $-1325$ & $0.0\%$ & $0.1\%$ & $99.8\%$ & $-4.5\%$ \\
\wmove{Be3+} & $-7.88$ & $-1364$ & $0.0\%$ & $0.1\%$ & $99.9\%$ & $-4.5\%$ \\
\wmove{Bf2} & $-9.17$ & $-1493$ & $0.0\%$ & $0.1\%$ & $99.9\%$ & $-4.5\%$ \\
\bottomrule
\end{tabular}%
}
\end{table*}

%% file: Figures/Puzzles/Forgotten/puzzle_tables_DsaGi.tex
\begin{table*}[h!]
\centering
\caption{Move probabilities by layer for puzzle \href{https://lichess.org/training/DsaGi}{\texttt{DsaGi}} (Part 1: Input to Layer 6)}
\label{tab:puzzle_DsaGi_probs1}
\resizebox{\textwidth}{!}{%
\begin{tabular}{lrrrrrrrr}
\toprule
\textbf{Move} & \textbf{Input} & \textbf{Layer 0} & \textbf{Layer 1} & \textbf{Layer 2} & \textbf{Layer 3} & \textbf{Layer 4} & \textbf{Layer 5} & \textbf{Layer 6} \\
\midrule
\wmove{a4} & \cellcolor{heatmapcolor!15!white}15.39\% & \cellcolor{heatmapcolor!89!white}88.66\% & \cellcolor{heatmapcolor!15!white}14.83\% & \cellcolor{heatmapcolor!16!white}15.77\% & \cellcolor{heatmapcolor!10!white}10.39\% & 0.24\% & \cellcolor{heatmapcolor!1!white}1.31\% & \cellcolor{heatmapcolor!1!white}0.93\% \\
\wmove{Rxg4+} & 0.38\% & \cellcolor{heatmapcolor!1!white}0.66\% & \cellcolor{heatmapcolor!12!white}11.54\% & \cellcolor{heatmapcolor!22!white}22.23\% & \cellcolor{heatmapcolor!29!white}29.10\% & \cellcolor{heatmapcolor!72!white}71.66\% & \cellcolor{heatmapcolor!81!white}80.50\% & \cellcolor{heatmapcolor!75!white}75.47\% \\
\wmove{Rf1+} & \cellcolor{heatmapcolor!3!white}3.45\% & 0.12\% & \cellcolor{heatmapcolor!2!white}1.81\% & \cellcolor{heatmapcolor!7!white}7.09\% & \cellcolor{heatmapcolor!4!white}4.04\% & 0.47\% & 0.29\% & \cellcolor{heatmapcolor!1!white}0.60\% \\
\wmove{Kg7} & 0.12\% & 0.31\% & \cellcolor{heatmapcolor!26!white}25.84\% & \cellcolor{heatmapcolor!11!white}10.54\% & \cellcolor{heatmapcolor!5!white}5.47\% & \cellcolor{heatmapcolor!1!white}1.05\% & \cellcolor{heatmapcolor!2!white}2.15\% & \cellcolor{heatmapcolor!1!white}1.43\% \\
\wmove{Rh1} & \cellcolor{heatmapcolor!5!white}5.32\% & 0.13\% & \cellcolor{heatmapcolor!3!white}2.62\% & \cellcolor{heatmapcolor!2!white}2.47\% & \cellcolor{heatmapcolor!4!white}3.66\% & 0.35\% & \cellcolor{heatmapcolor!1!white}0.57\% & \cellcolor{heatmapcolor!1!white}1.02\% \\
\wmove{Bh8} & 0.13\% & 0.06\% & \cellcolor{heatmapcolor!1!white}0.54\% & \cellcolor{heatmapcolor!1!white}0.51\% & \cellcolor{heatmapcolor!19!white}18.57\% & \cellcolor{heatmapcolor!2!white}1.73\% & \cellcolor{heatmapcolor!8!white}8.11\% & \cellcolor{heatmapcolor!17!white}17.35\% \\
\wmove{Rg3} & \cellcolor{heatmapcolor!7!white}7.11\% & 0.10\% & \cellcolor{heatmapcolor!5!white}4.71\% & \cellcolor{heatmapcolor!10!white}10.33\% & \cellcolor{heatmapcolor!14!white}14.05\% & \cellcolor{heatmapcolor!6!white}6.37\% & \cellcolor{heatmapcolor!1!white}0.68\% & 0.31\% \\
\wmove{Rg2} & \cellcolor{heatmapcolor!14!white}13.65\% & 0.19\% & \cellcolor{heatmapcolor!2!white}2.17\% & \cellcolor{heatmapcolor!4!white}3.87\% & \cellcolor{heatmapcolor!3!white}3.22\% & \cellcolor{heatmapcolor!4!white}4.06\% & 0.38\% & 0.28\% \\
\wmove{Bf2} & \cellcolor{heatmapcolor!12!white}12.36\% & 0.08\% & \cellcolor{heatmapcolor!1!white}1.08\% & \cellcolor{heatmapcolor!4!white}4.40\% & \cellcolor{heatmapcolor!1!white}0.77\% & \cellcolor{heatmapcolor!1!white}1.30\% & 0.27\% & 0.21\% \\
\wmove{c4} & \cellcolor{heatmapcolor!1!white}1.25\% & \cellcolor{heatmapcolor!9!white}8.55\% & \cellcolor{heatmapcolor!11!white}10.83\% & \cellcolor{heatmapcolor!3!white}2.58\% & \cellcolor{heatmapcolor!1!white}0.51\% & 0.05\% & \cellcolor{heatmapcolor!1!white}0.51\% & 0.43\% \\
\wmove{Be3+} & \cellcolor{heatmapcolor!10!white}9.81\% & 0.05\% & \cellcolor{heatmapcolor!1!white}1.11\% & \cellcolor{heatmapcolor!3!white}2.74\% & \cellcolor{heatmapcolor!1!white}0.73\% & \cellcolor{heatmapcolor!1!white}0.76\% & 0.12\% & 0.05\% \\
\wmove{Bc3} & \cellcolor{heatmapcolor!10!white}9.74\% & 0.07\% & \cellcolor{heatmapcolor!1!white}1.04\% & 0.38\% & 0.49\% & 0.37\% & \cellcolor{heatmapcolor!2!white}1.61\% & 0.20\% \\
\wmove{Ra1} & \cellcolor{heatmapcolor!3!white}3.41\% & 0.07\% & \cellcolor{heatmapcolor!6!white}5.73\% & \cellcolor{heatmapcolor!3!white}2.58\% & \cellcolor{heatmapcolor!1!white}0.84\% & \cellcolor{heatmapcolor!1!white}0.75\% & 0.18\% & 0.11\% \\
\wmove{Rb1} & \cellcolor{heatmapcolor!5!white}4.89\% & 0.14\% & \cellcolor{heatmapcolor!3!white}2.50\% & \cellcolor{heatmapcolor!2!white}1.50\% & \cellcolor{heatmapcolor!2!white}1.51\% & \cellcolor{heatmapcolor!2!white}2.38\% & \cellcolor{heatmapcolor!1!white}0.56\% & 0.18\% \\
\wmove{Be5+} & 0.42\% & 0.05\% & \cellcolor{heatmapcolor!2!white}2.37\% & \cellcolor{heatmapcolor!4!white}4.46\% & \cellcolor{heatmapcolor!1!white}0.68\% & \cellcolor{heatmapcolor!1!white}1.05\% & 0.15\% & 0.06\% \\
\wmove{Bg7} & 0.45\% & 0.29\% & \cellcolor{heatmapcolor!4!white}3.97\% & 0.22\% & \cellcolor{heatmapcolor!1!white}0.58\% & 0.22\% & \cellcolor{heatmapcolor!1!white}0.98\% & 0.37\% \\
\wmove{Re1} & \cellcolor{heatmapcolor!3!white}3.29\% & 0.08\% & \cellcolor{heatmapcolor!1!white}0.60\% & \cellcolor{heatmapcolor!1!white}1.26\% & \cellcolor{heatmapcolor!1!white}1.12\% & \cellcolor{heatmapcolor!2!white}1.86\% & 0.27\% & 0.22\% \\
\wmove{Rd1} & \cellcolor{heatmapcolor!3!white}3.17\% & 0.11\% & \cellcolor{heatmapcolor!1!white}0.97\% & \cellcolor{heatmapcolor!1!white}1.12\% & \cellcolor{heatmapcolor!1!white}0.92\% & \cellcolor{heatmapcolor!2!white}1.85\% & 0.31\% & 0.16\% \\
\wmove{Rc1} & \cellcolor{heatmapcolor!1!white}1.29\% & 0.08\% & \cellcolor{heatmapcolor!2!white}2.23\% & \cellcolor{heatmapcolor!1!white}1.43\% & \cellcolor{heatmapcolor!1!white}0.90\% & \cellcolor{heatmapcolor!1!white}1.35\% & 0.28\% & 0.19\% \\
\wmove{Ba1} & \cellcolor{heatmapcolor!2!white}2.02\% & 0.04\% & \cellcolor{heatmapcolor!1!white}1.18\% & \cellcolor{heatmapcolor!1!white}0.96\% & \cellcolor{heatmapcolor!1!white}1.16\% & \cellcolor{heatmapcolor!1!white}0.74\% & 0.29\% & 0.15\% \\
\wmove{Bb2} & \cellcolor{heatmapcolor!2!white}1.62\% & 0.06\% & \cellcolor{heatmapcolor!1!white}0.74\% & \cellcolor{heatmapcolor!1!white}1.25\% & 0.42\% & \cellcolor{heatmapcolor!1!white}0.65\% & 0.09\% & 0.03\% \\
\wmove{Kf8} & 0.00\% & 0.07\% & \cellcolor{heatmapcolor!1!white}1.25\% & \cellcolor{heatmapcolor!2!white}1.61\% & \cellcolor{heatmapcolor!1!white}0.70\% & 0.48\% & 0.28\% & 0.19\% \\
\wmove{Bf6} & \cellcolor{heatmapcolor!1!white}0.73\% & 0.04\% & 0.34\% & \cellcolor{heatmapcolor!1!white}0.69\% & 0.16\% & 0.24\% & 0.11\% & 0.03\% \\
\bottomrule
\end{tabular}%
}
\end{table*}

\begin{table*}[h!]
\centering
\caption{Move probabilities by layer for puzzle \href{https://lichess.org/training/DsaGi}{\texttt{DsaGi}} (Part 2: Layer 7 to Final)}
\label{tab:puzzle_DsaGi_probs2}
\resizebox{\textwidth}{!}{%
\begin{tabular}{lrrrrrrrr}
\toprule
\textbf{Move} & \textbf{Layer 7} & \textbf{Layer 8} & \textbf{Layer 9} & \textbf{Layer 10} & \textbf{Layer 11} & \textbf{Layer 12} & \textbf{Layer 13} & \textbf{Final} \\
\midrule
\wmove{a4} & \cellcolor{heatmapcolor!1!white}0.96\% & 0.48\% & 0.04\% & 0.20\% & \cellcolor{heatmapcolor!1!white}0.86\% & \cellcolor{heatmapcolor!1!white}0.64\% & \cellcolor{heatmapcolor!1!white}1.17\% & \cellcolor{heatmapcolor!1!white}0.72\% \\
\wmove{Rxg4+} & \cellcolor{heatmapcolor!72!white}72.17\% & \cellcolor{heatmapcolor!73!white}73.29\% & \cellcolor{heatmapcolor!83!white}83.14\% & \cellcolor{heatmapcolor!88!white}88.00\% & \cellcolor{heatmapcolor!78!white}78.25\% & \cellcolor{heatmapcolor!87!white}86.89\% & \cellcolor{heatmapcolor!61!white}61.03\% & \cellcolor{heatmapcolor!12!white}12.14\% \\
\wmove{Rf1+} & \cellcolor{heatmapcolor!1!white}1.23\% & \cellcolor{heatmapcolor!2!white}2.08\% & \cellcolor{heatmapcolor!1!white}0.56\% & \cellcolor{heatmapcolor!2!white}2.04\% & \cellcolor{heatmapcolor!4!white}4.41\% & \cellcolor{heatmapcolor!3!white}3.45\% & \cellcolor{heatmapcolor!17!white}16.89\% & \cellcolor{heatmapcolor!40!white}40.43\% \\
\wmove{Kg7} & \cellcolor{heatmapcolor!4!white}3.84\% & \cellcolor{heatmapcolor!4!white}3.85\% & \cellcolor{heatmapcolor!3!white}2.77\% & \cellcolor{heatmapcolor!2!white}2.18\% & \cellcolor{heatmapcolor!3!white}3.21\% & \cellcolor{heatmapcolor!3!white}3.16\% & \cellcolor{heatmapcolor!5!white}5.34\% & \cellcolor{heatmapcolor!8!white}8.01\% \\
\wmove{Rh1} & \cellcolor{heatmapcolor!1!white}0.84\% & \cellcolor{heatmapcolor!1!white}1.16\% & 0.30\% & \cellcolor{heatmapcolor!1!white}0.87\% & \cellcolor{heatmapcolor!2!white}1.89\% & \cellcolor{heatmapcolor!1!white}1.41\% & \cellcolor{heatmapcolor!5!white}4.99\% & \cellcolor{heatmapcolor!23!white}23.00\% \\
\wmove{Bh8} & \cellcolor{heatmapcolor!18!white}17.62\% & \cellcolor{heatmapcolor!15!white}14.62\% & \cellcolor{heatmapcolor!9!white}9.44\% & \cellcolor{heatmapcolor!3!white}3.12\% & \cellcolor{heatmapcolor!2!white}1.55\% & 0.36\% & \cellcolor{heatmapcolor!1!white}0.52\% & \cellcolor{heatmapcolor!1!white}0.50\% \\
\wmove{Rg3} & 0.40\% & 0.47\% & \cellcolor{heatmapcolor!1!white}0.85\% & \cellcolor{heatmapcolor!1!white}0.85\% & \cellcolor{heatmapcolor!2!white}1.88\% & 0.49\% & 0.26\% & 0.19\% \\
\wmove{Rg2} & 0.32\% & 0.37\% & 0.27\% & 0.35\% & \cellcolor{heatmapcolor!1!white}0.51\% & 0.20\% & \cellcolor{heatmapcolor!1!white}0.52\% & \cellcolor{heatmapcolor!5!white}5.38\% \\
\wmove{Bf2} & 0.24\% & \cellcolor{heatmapcolor!1!white}0.53\% & \cellcolor{heatmapcolor!1!white}0.78\% & \cellcolor{heatmapcolor!1!white}1.01\% & \cellcolor{heatmapcolor!5!white}4.98\% & \cellcolor{heatmapcolor!1!white}1.25\% & \cellcolor{heatmapcolor!1!white}0.60\% & 0.23\% \\
\wmove{c4} & 0.35\% & 0.06\% & 0.02\% & 0.04\% & 0.16\% & 0.20\% & \cellcolor{heatmapcolor!1!white}0.59\% & 0.31\% \\
\wmove{Be3+} & 0.05\% & 0.06\% & 0.04\% & 0.03\% & 0.02\% & 0.00\% & 0.15\% & 0.27\% \\
\wmove{Bc3} & 0.05\% & 0.13\% & 0.05\% & 0.02\% & 0.02\% & 0.03\% & 0.12\% & \cellcolor{heatmapcolor!1!white}1.32\% \\
\wmove{Ra1} & 0.08\% & 0.09\% & 0.05\% & 0.04\% & 0.08\% & 0.03\% & 0.12\% & \cellcolor{heatmapcolor!1!white}1.05\% \\
\wmove{Rb1} & 0.24\% & 0.29\% & 0.07\% & 0.24\% & 0.43\% & 0.25\% & \cellcolor{heatmapcolor!3!white}2.95\% & 0.36\% \\
\wmove{Be5+} & 0.06\% & 0.10\% & 0.05\% & 0.03\% & 0.04\% & 0.01\% & 0.09\% & 0.24\% \\
\wmove{Bg7} & \cellcolor{heatmapcolor!1!white}0.52\% & \cellcolor{heatmapcolor!1!white}1.22\% & \cellcolor{heatmapcolor!1!white}0.67\% & 0.26\% & 0.24\% & 0.28\% & \cellcolor{heatmapcolor!1!white}1.36\% & \cellcolor{heatmapcolor!1!white}0.50\% \\
\wmove{Re1} & 0.21\% & 0.27\% & 0.39\% & 0.22\% & \cellcolor{heatmapcolor!1!white}0.56\% & \cellcolor{heatmapcolor!1!white}0.61\% & 0.17\% & 0.20\% \\
\wmove{Rd1} & 0.16\% & 0.19\% & 0.08\% & 0.11\% & 0.16\% & 0.12\% & \cellcolor{heatmapcolor!2!white}2.18\% & 0.32\% \\
\wmove{Rc1} & 0.21\% & 0.20\% & 0.13\% & 0.10\% & 0.23\% & 0.15\% & 0.29\% & \cellcolor{heatmapcolor!2!white}2.40\% \\
\wmove{Ba1} & 0.08\% & 0.20\% & 0.13\% & 0.05\% & 0.14\% & 0.13\% & 0.13\% & 0.28\% \\
\wmove{Bb2} & 0.02\% & 0.03\% & 0.02\% & 0.02\% & 0.03\% & 0.04\% & 0.13\% & \cellcolor{heatmapcolor!2!white}1.62\% \\
\wmove{Kf8} & 0.31\% & 0.28\% & 0.15\% & 0.22\% & 0.35\% & 0.30\% & 0.27\% & 0.27\% \\
\wmove{Bf6} & 0.04\% & 0.04\% & 0.02\% & 0.01\% & 0.01\% & 0.02\% & 0.14\% & 0.27\% \\
\bottomrule
\end{tabular}%
}
\end{table*}

%% file: Figures/Puzzles/Forgotten/puzzle_evaluation_00aDl.tex
\begin{table*}[h!]
\centering
\caption{Move evaluation for puzzle \href{https://lichess.org/training/00aDl}{\texttt{00aDl}}: Stockfish evaluation at depth 20 and model WDL prediction for resulting positions}
\label{tab:puzzle_00aDl_eval}
\resizebox{\textwidth}{!}{%
\begin{tabular}{llrrrrr}
\toprule
\textbf{Move} & \textbf{Stockfish} & \textbf{$\Delta$ (cp)} & \textbf{Win} & \textbf{Draw} & \textbf{Loss} & \textbf{$\Delta$ Win} \\
\midrule
\textit{Current position} & $+\infty$ & --- & $8.3\%$ & $10.6\%$ & $81.1\%$ & --- \\
\midrule
\rowcolor{heatmapcolor!30}
\wmove{Qxb1+} $\star$ & $+\infty$ & --- & $99.3\%$ & $0.5\%$ & $0.2\%$ & $+91.0\%$ \\
\wmove{b5} & $-4.12$ & $-\infty$ & $0.2\%$ & $0.4\%$ & $99.4\%$ & $-8.1\%$ \\
\wmove{R2a4} & $-4.57$ & $-\infty$ & $0.2\%$ & $0.3\%$ & $99.5\%$ & $-8.1\%$ \\
\wmove{R2a7} & $-4.72$ & $-\infty$ & $0.1\%$ & $0.1\%$ & $99.8\%$ & $-8.2\%$ \\
\wmove{Na5} & $-4.74$ & $-\infty$ & $0.0\%$ & $0.1\%$ & $99.9\%$ & $-8.3\%$ \\
\wmove{R2a6} & $-4.77$ & $-\infty$ & $0.1\%$ & $0.1\%$ & $99.8\%$ & $-8.2\%$ \\
\wmove{c5} & $-4.78$ & $-\infty$ & $0.0\%$ & $0.1\%$ & $99.9\%$ & $-8.3\%$ \\
\wmove{Nd6} & $-5.07$ & $-\infty$ & $0.1\%$ & $0.2\%$ & $99.7\%$ & $-8.2\%$ \\
\wmove{Kg7} & $-5.40$ & $-\infty$ & $0.0\%$ & $0.1\%$ & $99.9\%$ & $-8.3\%$ \\
\wmove{b6} & $-5.42$ & $-\infty$ & $0.0\%$ & $0.1\%$ & $99.9\%$ & $-8.3\%$ \\
\wmove{e5} & $-5.45$ & $-\infty$ & $0.0\%$ & $0.1\%$ & $99.9\%$ & $-8.3\%$ \\
\wmove{Kh8} & $-5.49$ & $-\infty$ & $0.0\%$ & $0.1\%$ & $99.9\%$ & $-8.3\%$ \\
\wmove{f5} & $-5.49$ & $-\infty$ & $0.0\%$ & $0.1\%$ & $99.9\%$ & $-8.3\%$ \\
\wmove{g5} & $-5.53$ & $-\infty$ & $0.0\%$ & $0.1\%$ & $99.9\%$ & $-8.3\%$ \\
\wmove{R2a5} & $-5.55$ & $-\infty$ & $0.0\%$ & $0.1\%$ & $99.9\%$ & $-8.3\%$ \\
\wmove{R8a4} & $-5.56$ & $-\infty$ & $0.1\%$ & $0.2\%$ & $99.8\%$ & $-8.2\%$ \\
\wmove{R8a6} & $-5.60$ & $-\infty$ & $0.0\%$ & $0.1\%$ & $99.9\%$ & $-8.3\%$ \\
\wmove{R8a5} & $-5.60$ & $-\infty$ & $0.0\%$ & $0.1\%$ & $99.9\%$ & $-8.3\%$ \\
\wmove{Rc8} & $-5.63$ & $-\infty$ & $0.0\%$ & $0.1\%$ & $99.9\%$ & $-8.3\%$ \\
\wmove{Rd8} & $-5.75$ & $-\infty$ & $0.0\%$ & $0.1\%$ & $99.9\%$ & $-8.3\%$ \\
\wmove{R8a7} & $-5.75$ & $-\infty$ & $0.0\%$ & $0.1\%$ & $99.9\%$ & $-8.3\%$ \\
\wmove{Rb8} & $-5.76$ & $-\infty$ & $0.0\%$ & $0.1\%$ & $99.9\%$ & $-8.3\%$ \\
\wmove{Kf8} & $-5.82$ & $-\infty$ & $0.0\%$ & $0.1\%$ & $99.9\%$ & $-8.3\%$ \\
\wmove{Kh7} & $-5.87$ & $-\infty$ & $0.0\%$ & $0.1\%$ & $99.9\%$ & $-8.3\%$ \\
\wmove{Ne5} & $-5.96$ & $-\infty$ & $0.0\%$ & $0.0\%$ & $99.9\%$ & $-8.3\%$ \\
\wmove{Rf8} & $-5.98$ & $-\infty$ & $0.0\%$ & $0.0\%$ & $99.9\%$ & $-8.3\%$ \\
\wmove{Re8} & $-6.17$ & $-\infty$ & $0.0\%$ & $0.1\%$ & $99.9\%$ & $-8.3\%$ \\
\wmove{Ne3} & $-6.29$ & $-\infty$ & $0.0\%$ & $0.0\%$ & $99.9\%$ & $-8.3\%$ \\
\wmove{Nb6} & $-6.78$ & $-\infty$ & $0.0\%$ & $0.0\%$ & $99.9\%$ & $-8.3\%$ \\
\wmove{Nxb2} & $-7.09$ & $-\infty$ & $0.0\%$ & $0.1\%$ & $99.9\%$ & $-8.3\%$ \\
\wmove{Qxb2+} & $-7.11$ & $-\infty$ & $0.0\%$ & $0.0\%$ & $100.0\%$ & $-8.3\%$ \\
\wmove{Rxb2} & $-7.23$ & $-\infty$ & $0.0\%$ & $0.1\%$ & $99.8\%$ & $-8.3\%$ \\
\wmove{Na3} & $-7.65$ & $-\infty$ & $0.0\%$ & $0.0\%$ & $99.9\%$ & $-8.3\%$ \\
\wmove{Nd2} & $-7.77$ & $-\infty$ & $0.0\%$ & $0.0\%$ & $100.0\%$ & $-8.3\%$ \\
\wmove{R2a3} & $-8.16$ & $-\infty$ & $0.0\%$ & $0.0\%$ & $100.0\%$ & $-8.3\%$ \\
\wmove{R8a3} & $-8.34$ & $-\infty$ & $0.0\%$ & $0.1\%$ & $99.9\%$ & $-8.3\%$ \\
\bottomrule
\end{tabular}%
}
\end{table*}

%% file: Figures/Puzzles/Forgotten/puzzle_tables_00aDl.tex
\begin{table*}[h!]
\centering
\caption{Move probabilities by layer for puzzle \href{https://lichess.org/training/00aDl}{\texttt{00aDl}} (Part 1: Input to Layer 6)}
\label{tab:puzzle_00aDl_probs1}
\resizebox{\textwidth}{!}{%
\begin{tabular}{lrrrrrrrr}
\toprule
\textbf{Move} & \textbf{Input} & \textbf{Layer 0} & \textbf{Layer 1} & \textbf{Layer 2} & \textbf{Layer 3} & \textbf{Layer 4} & \textbf{Layer 5} & \textbf{Layer 6} \\
\midrule
\wmove{Qxb1+} & \cellcolor{heatmapcolor!2!white}1.80\% & \cellcolor{heatmapcolor!63!white}62.90\% & \cellcolor{heatmapcolor!48!white}47.54\% & \cellcolor{heatmapcolor!30!white}29.77\% & \cellcolor{heatmapcolor!23!white}23.48\% & \cellcolor{heatmapcolor!19!white}18.87\% & \cellcolor{heatmapcolor!28!white}27.94\% & \cellcolor{heatmapcolor!27!white}27.42\% \\
\wmove{Nd2} & \cellcolor{heatmapcolor!57!white}56.64\% & 0.41\% & \cellcolor{heatmapcolor!1!white}0.72\% & 0.18\% & \cellcolor{heatmapcolor!3!white}2.89\% & 0.03\% & \cellcolor{heatmapcolor!1!white}1.47\% & \cellcolor{heatmapcolor!1!white}0.89\% \\
\wmove{b5} & 0.43\% & 0.31\% & 0.18\% & 0.03\% & 0.37\% & 0.13\% & \cellcolor{heatmapcolor!3!white}2.56\% & \cellcolor{heatmapcolor!13!white}12.77\% \\
\wmove{Nxb2} & \cellcolor{heatmapcolor!1!white}1.28\% & \cellcolor{heatmapcolor!1!white}0.66\% & \cellcolor{heatmapcolor!1!white}0.69\% & \cellcolor{heatmapcolor!24!white}24.30\% & \cellcolor{heatmapcolor!30!white}29.59\% & \cellcolor{heatmapcolor!36!white}35.56\% & \cellcolor{heatmapcolor!31!white}31.40\% & \cellcolor{heatmapcolor!29!white}28.96\% \\
\wmove{Ne3} & \cellcolor{heatmapcolor!26!white}26.36\% & 0.10\% & 0.37\% & 0.02\% & 0.04\% & 0.06\% & 0.02\% & 0.02\% \\
\wmove{Qxb2+} & \cellcolor{heatmapcolor!9!white}8.75\% & \cellcolor{heatmapcolor!11!white}11.39\% & \cellcolor{heatmapcolor!24!white}23.80\% & \cellcolor{heatmapcolor!24!white}24.03\% & \cellcolor{heatmapcolor!21!white}20.94\% & \cellcolor{heatmapcolor!23!white}22.84\% & \cellcolor{heatmapcolor!20!white}19.88\% & \cellcolor{heatmapcolor!14!white}13.99\% \\
\wmove{Rxb2} & 0.35\% & \cellcolor{heatmapcolor!17!white}17.28\% & \cellcolor{heatmapcolor!18!white}18.29\% & \cellcolor{heatmapcolor!20!white}20.38\% & \cellcolor{heatmapcolor!21!white}20.92\% & \cellcolor{heatmapcolor!21!white}20.66\% & \cellcolor{heatmapcolor!14!white}13.81\% & \cellcolor{heatmapcolor!14!white}13.55\% \\
\wmove{Nd6} & 0.09\% & 0.11\% & 0.09\% & 0.05\% & 0.17\% & 0.08\% & \cellcolor{heatmapcolor!1!white}0.57\% & 0.33\% \\
\wmove{Na5} & \cellcolor{heatmapcolor!1!white}0.58\% & \cellcolor{heatmapcolor!1!white}0.59\% & \cellcolor{heatmapcolor!1!white}0.87\% & 0.03\% & 0.27\% & 0.10\% & \cellcolor{heatmapcolor!1!white}1.49\% & 0.23\% \\
\wmove{b6} & 0.06\% & 0.06\% & 0.12\% & 0.02\% & 0.01\% & 0.05\% & 0.14\% & \cellcolor{heatmapcolor!1!white}0.62\% \\
\wmove{Na3} & 0.14\% & 0.35\% & \cellcolor{heatmapcolor!1!white}1.00\% & 0.03\% & 0.16\% & 0.05\% & 0.05\% & 0.18\% \\
\wmove{R2a4} & 0.03\% & 0.08\% & 0.13\% & 0.15\% & 0.09\% & 0.03\% & 0.02\% & 0.06\% \\
\wmove{R8a4} & 0.07\% & 0.13\% & \cellcolor{heatmapcolor!1!white}0.73\% & 0.04\% & 0.01\% & 0.02\% & 0.01\% & 0.03\% \\
\wmove{Kf8} & 0.05\% & \cellcolor{heatmapcolor!2!white}2.16\% & 0.25\% & 0.03\% & 0.09\% & 0.08\% & 0.02\% & 0.04\% \\
\wmove{R8a3} & 0.05\% & 0.14\% & \cellcolor{heatmapcolor!2!white}1.61\% & 0.15\% & 0.04\% & 0.11\% & 0.03\% & 0.06\% \\
\wmove{R8a5} & 0.05\% & 0.20\% & \cellcolor{heatmapcolor!1!white}0.90\% & 0.05\% & 0.03\% & 0.07\% & 0.04\% & 0.09\% \\
\wmove{R2a6} & 0.12\% & 0.12\% & 0.15\% & 0.03\% & 0.05\% & 0.12\% & 0.02\% & 0.02\% \\
\wmove{f5} & \cellcolor{heatmapcolor!1!white}0.80\% & 0.05\% & 0.07\% & 0.01\% & 0.02\% & 0.03\% & 0.03\% & 0.06\% \\
\wmove{Kg7} & 0.04\% & 0.24\% & 0.05\% & 0.25\% & 0.15\% & 0.07\% & 0.03\% & 0.04\% \\
\wmove{R2a3} & 0.01\% & 0.07\% & 0.12\% & 0.03\% & 0.06\% & 0.08\% & 0.03\% & 0.08\% \\
\wmove{R2a7} & 0.09\% & 0.09\% & 0.12\% & 0.04\% & 0.04\% & 0.05\% & 0.01\% & 0.02\% \\
\wmove{c5} & 0.39\% & 0.26\% & 0.14\% & 0.07\% & 0.03\% & 0.02\% & 0.02\% & 0.05\% \\
\wmove{R2a5} & 0.09\% & 0.08\% & 0.07\% & 0.03\% & 0.07\% & 0.07\% & 0.05\% & 0.05\% \\
\wmove{Kh8} & 0.14\% & 0.23\% & 0.19\% & 0.04\% & 0.08\% & 0.09\% & 0.03\% & 0.04\% \\
\wmove{R8a6} & 0.03\% & 0.13\% & 0.35\% & 0.04\% & 0.04\% & 0.17\% & 0.03\% & 0.06\% \\
\wmove{e5} & 0.21\% & 0.04\% & 0.15\% & 0.00\% & 0.04\% & 0.03\% & 0.04\% & 0.06\% \\
\wmove{Rf8} & 0.08\% & \cellcolor{heatmapcolor!1!white}0.52\% & 0.13\% & 0.02\% & 0.01\% & 0.02\% & 0.01\% & 0.02\% \\
\wmove{Kh7} & 0.08\% & 0.46\% & 0.05\% & 0.05\% & 0.13\% & 0.10\% & 0.03\% & 0.03\% \\
\wmove{g5} & 0.49\% & 0.41\% & 0.29\% & 0.01\% & 0.04\% & 0.01\% & 0.03\% & 0.05\% \\
\wmove{Rb8} & 0.05\% & 0.05\% & 0.10\% & 0.02\% & 0.02\% & 0.06\% & 0.01\% & 0.02\% \\
\wmove{R8a7} & 0.04\% & 0.09\% & 0.17\% & 0.02\% & 0.02\% & 0.05\% & 0.01\% & 0.02\% \\
\wmove{Nb6} & 0.31\% & 0.09\% & 0.14\% & 0.02\% & 0.02\% & 0.10\% & 0.09\% & 0.06\% \\
\wmove{Rd8} & 0.06\% & 0.05\% & 0.10\% & 0.02\% & 0.02\% & 0.04\% & 0.01\% & 0.02\% \\
\wmove{Re8} & 0.06\% & 0.05\% & 0.09\% & 0.02\% & 0.01\% & 0.03\% & 0.01\% & 0.01\% \\
\wmove{Rc8} & 0.05\% & 0.06\% & 0.13\% & 0.02\% & 0.02\% & 0.06\% & 0.01\% & 0.02\% \\
\wmove{Ne5} & 0.12\% & 0.04\% & 0.10\% & 0.01\% & 0.01\% & 0.05\% & 0.03\% & 0.04\% \\
\bottomrule
\end{tabular}%
}
\end{table*}

\begin{table*}[h!]
\centering
\caption{Move probabilities by layer for puzzle \href{https://lichess.org/training/00aDl}{\texttt{00aDl}} (Part 2: Layer 7 to Final)}
\label{tab:puzzle_00aDl_probs2}
\resizebox{\textwidth}{!}{%
\begin{tabular}{lrrrrrrrr}
\toprule
\textbf{Move} & \textbf{Layer 7} & \textbf{Layer 8} & \textbf{Layer 9} & \textbf{Layer 10} & \textbf{Layer 11} & \textbf{Layer 12} & \textbf{Layer 13} & \textbf{Final} \\
\midrule
\wmove{Qxb1+} & \cellcolor{heatmapcolor!25!white}25.28\% & \cellcolor{heatmapcolor!28!white}27.69\% & \cellcolor{heatmapcolor!25!white}24.79\% & \cellcolor{heatmapcolor!33!white}33.23\% & \cellcolor{heatmapcolor!41!white}40.50\% & \cellcolor{heatmapcolor!42!white}41.78\% & \cellcolor{heatmapcolor!57!white}56.96\% & \cellcolor{heatmapcolor!36!white}35.99\% \\
\wmove{Nd2} & \cellcolor{heatmapcolor!1!white}0.68\% & 0.26\% & \cellcolor{heatmapcolor!8!white}7.69\% & \cellcolor{heatmapcolor!2!white}1.87\% & 0.16\% & 0.17\% & 0.21\% & 0.32\% \\
\wmove{b5} & \cellcolor{heatmapcolor!17!white}16.94\% & \cellcolor{heatmapcolor!27!white}26.56\% & \cellcolor{heatmapcolor!17!white}17.03\% & \cellcolor{heatmapcolor!25!white}25.45\% & \cellcolor{heatmapcolor!36!white}35.78\% & \cellcolor{heatmapcolor!34!white}33.58\% & \cellcolor{heatmapcolor!9!white}8.85\% & \cellcolor{heatmapcolor!36!white}36.12\% \\
\wmove{Nxb2} & \cellcolor{heatmapcolor!31!white}31.09\% & \cellcolor{heatmapcolor!18!white}17.79\% & \cellcolor{heatmapcolor!13!white}12.84\% & \cellcolor{heatmapcolor!12!white}12.49\% & \cellcolor{heatmapcolor!5!white}4.64\% & \cellcolor{heatmapcolor!1!white}0.73\% & \cellcolor{heatmapcolor!1!white}0.70\% & 0.35\% \\
\wmove{Ne3} & 0.02\% & 0.01\% & 0.04\% & 0.03\% & 0.04\% & 0.04\% & 0.20\% & 0.32\% \\
\wmove{Qxb2+} & \cellcolor{heatmapcolor!11!white}10.75\% & \cellcolor{heatmapcolor!12!white}11.81\% & \cellcolor{heatmapcolor!12!white}11.53\% & \cellcolor{heatmapcolor!8!white}8.22\% & \cellcolor{heatmapcolor!3!white}3.26\% & 0.36\% & 0.50\% & 0.29\% \\
\wmove{Rxb2} & \cellcolor{heatmapcolor!10!white}9.97\% & \cellcolor{heatmapcolor!10!white}10.30\% & \cellcolor{heatmapcolor!9!white}8.75\% & \cellcolor{heatmapcolor!7!white}6.71\% & \cellcolor{heatmapcolor!2!white}2.18\% & 0.32\% & 0.49\% & 0.44\% \\
\wmove{Nd6} & \cellcolor{heatmapcolor!1!white}1.16\% & \cellcolor{heatmapcolor!1!white}1.37\% & \cellcolor{heatmapcolor!6!white}5.72\% & \cellcolor{heatmapcolor!3!white}2.96\% & \cellcolor{heatmapcolor!5!white}5.18\% & \cellcolor{heatmapcolor!11!white}10.97\% & \cellcolor{heatmapcolor!10!white}9.70\% & \cellcolor{heatmapcolor!5!white}5.41\% \\
\wmove{Na5} & \cellcolor{heatmapcolor!1!white}0.95\% & \cellcolor{heatmapcolor!1!white}0.83\% & \cellcolor{heatmapcolor!4!white}4.04\% & \cellcolor{heatmapcolor!2!white}1.76\% & \cellcolor{heatmapcolor!1!white}0.93\% & \cellcolor{heatmapcolor!2!white}2.08\% & \cellcolor{heatmapcolor!6!white}5.94\% & \cellcolor{heatmapcolor!2!white}1.85\% \\
\wmove{b6} & \cellcolor{heatmapcolor!1!white}1.28\% & \cellcolor{heatmapcolor!1!white}1.19\% & \cellcolor{heatmapcolor!2!white}1.84\% & \cellcolor{heatmapcolor!2!white}1.99\% & \cellcolor{heatmapcolor!2!white}2.01\% & \cellcolor{heatmapcolor!1!white}0.87\% & \cellcolor{heatmapcolor!3!white}2.58\% & \cellcolor{heatmapcolor!1!white}0.62\% \\
\wmove{Na3} & 0.46\% & 0.31\% & \cellcolor{heatmapcolor!2!white}2.31\% & \cellcolor{heatmapcolor!1!white}0.77\% & 0.49\% & \cellcolor{heatmapcolor!1!white}0.84\% & \cellcolor{heatmapcolor!2!white}1.79\% & \cellcolor{heatmapcolor!1!white}0.94\% \\
\wmove{R2a4} & 0.16\% & 0.24\% & 0.46\% & 0.35\% & \cellcolor{heatmapcolor!1!white}0.63\% & \cellcolor{heatmapcolor!2!white}1.68\% & \cellcolor{heatmapcolor!2!white}2.24\% & \cellcolor{heatmapcolor!2!white}2.28\% \\
\wmove{R8a4} & 0.07\% & 0.19\% & 0.27\% & 0.29\% & 0.29\% & \cellcolor{heatmapcolor!1!white}0.92\% & \cellcolor{heatmapcolor!1!white}1.06\% & \cellcolor{heatmapcolor!2!white}2.18\% \\
\wmove{Kf8} & 0.03\% & 0.05\% & 0.11\% & 0.22\% & 0.25\% & 0.21\% & 0.16\% & 0.40\% \\
\wmove{R8a3} & 0.11\% & 0.20\% & 0.47\% & 0.32\% & 0.19\% & \cellcolor{heatmapcolor!1!white}0.60\% & \cellcolor{heatmapcolor!1!white}0.74\% & \cellcolor{heatmapcolor!1!white}0.64\% \\
\wmove{R8a5} & 0.08\% & 0.12\% & 0.11\% & 0.13\% & 0.26\% & 0.22\% & 0.45\% & \cellcolor{heatmapcolor!1!white}1.35\% \\
\wmove{R2a6} & 0.02\% & 0.02\% & 0.05\% & 0.04\% & 0.08\% & 0.22\% & \cellcolor{heatmapcolor!1!white}0.65\% & \cellcolor{heatmapcolor!1!white}0.93\% \\
\wmove{f5} & 0.14\% & 0.17\% & 0.06\% & 0.25\% & \cellcolor{heatmapcolor!1!white}0.50\% & 0.21\% & 0.49\% & \cellcolor{heatmapcolor!1!white}0.54\% \\
\wmove{Kg7} & 0.04\% & 0.03\% & 0.05\% & 0.08\% & 0.02\% & 0.04\% & 0.20\% & \cellcolor{heatmapcolor!1!white}0.77\% \\
\wmove{R2a3} & 0.10\% & 0.16\% & 0.42\% & 0.19\% & 0.21\% & \cellcolor{heatmapcolor!1!white}0.72\% & \cellcolor{heatmapcolor!1!white}0.64\% & \cellcolor{heatmapcolor!1!white}0.65\% \\
\wmove{R2a7} & 0.02\% & 0.02\% & 0.05\% & 0.06\% & 0.12\% & 0.27\% & 0.47\% & \cellcolor{heatmapcolor!1!white}0.71\% \\
\wmove{c5} & 0.04\% & 0.07\% & 0.10\% & 0.25\% & 0.15\% & 0.34\% & \cellcolor{heatmapcolor!1!white}0.68\% & \cellcolor{heatmapcolor!1!white}0.70\% \\
\wmove{R2a5} & 0.06\% & 0.08\% & 0.16\% & 0.11\% & 0.32\% & \cellcolor{heatmapcolor!1!white}0.51\% & \cellcolor{heatmapcolor!1!white}0.67\% & \cellcolor{heatmapcolor!1!white}0.52\% \\
\wmove{Kh8} & 0.06\% & 0.06\% & 0.24\% & \cellcolor{heatmapcolor!1!white}0.66\% & 0.35\% & 0.17\% & 0.29\% & 0.41\% \\
\wmove{R8a6} & 0.03\% & 0.03\% & 0.05\% & 0.07\% & 0.06\% & 0.18\% & 0.27\% & \cellcolor{heatmapcolor!1!white}0.61\% \\
\wmove{e5} & 0.11\% & 0.10\% & 0.14\% & 0.09\% & 0.07\% & 0.08\% & 0.41\% & \cellcolor{heatmapcolor!1!white}0.54\% \\
\wmove{Rf8} & 0.02\% & 0.01\% & 0.03\% & 0.07\% & 0.04\% & 0.08\% & 0.18\% & 0.40\% \\
\wmove{Kh7} & 0.04\% & 0.04\% & 0.16\% & 0.35\% & 0.35\% & 0.12\% & 0.22\% & \cellcolor{heatmapcolor!1!white}0.51\% \\
\wmove{g5} & 0.10\% & 0.10\% & 0.10\% & 0.32\% & 0.40\% & 0.45\% & 0.41\% & 0.41\% \\
\wmove{Rb8} & 0.02\% & 0.01\% & 0.03\% & 0.04\% & 0.03\% & 0.08\% & 0.16\% & 0.47\% \\
\wmove{R8a7} & 0.02\% & 0.02\% & 0.04\% & 0.07\% & 0.07\% & 0.28\% & 0.26\% & 0.47\% \\
\wmove{Nb6} & 0.07\% & 0.07\% & 0.14\% & 0.13\% & 0.18\% & 0.27\% & 0.43\% & 0.32\% \\
\wmove{Rd8} & 0.02\% & 0.01\% & 0.03\% & 0.07\% & 0.03\% & 0.09\% & 0.26\% & 0.42\% \\
\wmove{Re8} & 0.02\% & 0.01\% & 0.03\% & 0.11\% & 0.04\% & 0.12\% & 0.23\% & 0.39\% \\
\wmove{Rc8} & 0.02\% & 0.01\% & 0.04\% & 0.13\% & 0.04\% & 0.15\% & 0.27\% & 0.39\% \\
\wmove{Ne5} & 0.04\% & 0.06\% & 0.07\% & 0.11\% & 0.15\% & 0.27\% & 0.23\% & 0.34\% \\
\bottomrule
\end{tabular}%
}
\end{table*}

%% file: Figures/concepts_individual/concepts_figures.tex
\begin{figure}[htbp]
  \centering
  \begin{subfigure}[t]{0.31\textwidth}
    \caption{Material Total midgame\vphantom{gjpqy}}
    \centering
    \includegraphics[width=\textwidth]{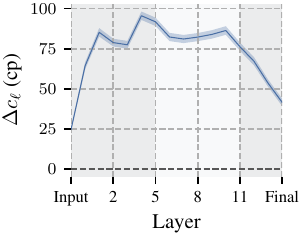}
  \end{subfigure}
  \hfill
  \begin{subfigure}[t]{0.31\textwidth}
    \caption{Material Total endgame\vphantom{gjpqy}}
    \centering
    \includegraphics[width=\textwidth]{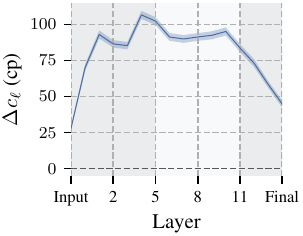}
  \end{subfigure}
  \hfill
  \begin{subfigure}[t]{0.31\textwidth}
    \caption{Material Total phased\vphantom{gjpqy}}
    \centering
    \includegraphics[width=\textwidth]{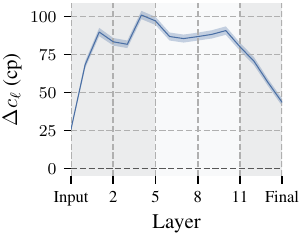}
  \end{subfigure}
  \\[0.3cm]
  \begin{subfigure}[t]{0.31\textwidth}
    \caption{Imbalance Total midgame\vphantom{gjpqy}}
    \centering
    \includegraphics[width=\textwidth]{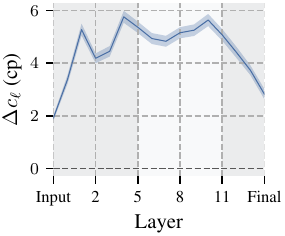}
  \end{subfigure}
  \hfill
  \begin{subfigure}[t]{0.31\textwidth}
    \caption{Imbalance Total endgame\vphantom{gjpqy}}
    \centering
    \includegraphics[width=\textwidth]{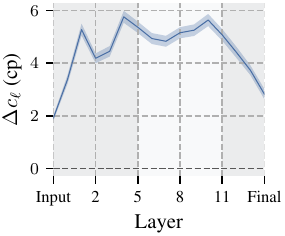}
  \end{subfigure}
  \hfill
  \begin{subfigure}[t]{0.31\textwidth}
    \caption{Imbalance Total phased\vphantom{gjpqy}}
    \centering
    \includegraphics[width=\textwidth]{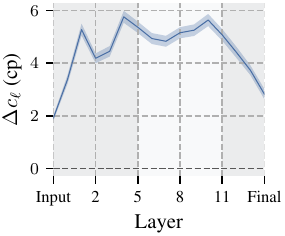}
  \end{subfigure}
  \\[0.3cm]
  \begin{subfigure}[t]{0.31\textwidth}
    \caption{Pawns Total midgame\vphantom{gjpqy}}
    \centering
    \includegraphics[width=\textwidth]{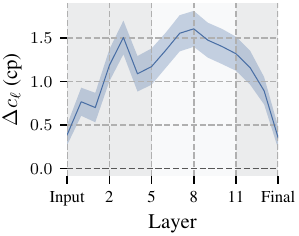}
  \end{subfigure}
  \hfill
  \begin{subfigure}[t]{0.31\textwidth}
    \caption{Pawns Total endgame\vphantom{gjpqy}}
    \centering
    \includegraphics[width=\textwidth]{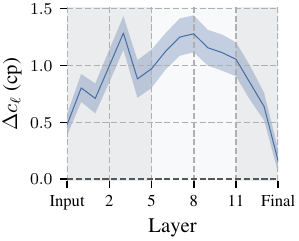}
  \end{subfigure}
  \hfill
  \begin{subfigure}[t]{0.31\textwidth}
    \caption{Pawns Total phased\vphantom{gjpqy}}
    \centering
    \includegraphics[width=\textwidth]{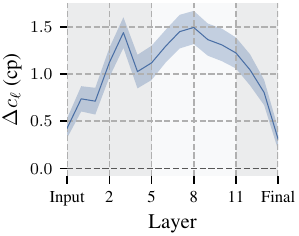}
  \end{subfigure}
  \\[0.3cm]
  \begin{subfigure}[t]{0.31\textwidth}
    \caption{Knights mine midgame\vphantom{gjpqy}}
    \centering
    \includegraphics[width=\textwidth]{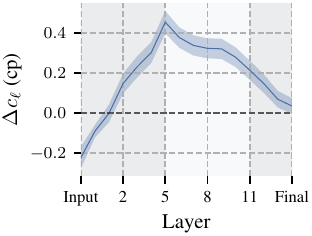}
  \end{subfigure}
  \hfill
  \begin{subfigure}[t]{0.31\textwidth}
    \caption{Knights mine endgame\vphantom{gjpqy}}
    \centering
    \includegraphics[width=\textwidth]{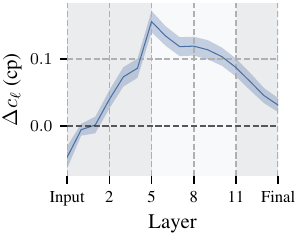}
  \end{subfigure}
  \hfill
  \begin{subfigure}[t]{0.31\textwidth}
    \caption{Knights mine phased\vphantom{gjpqy}}
    \centering
    \includegraphics[width=\textwidth]{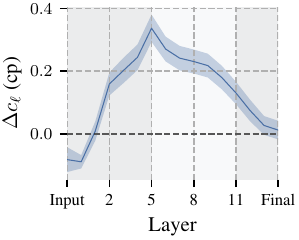}
  \end{subfigure}
  \\[0.3cm]
  \begin{subfigure}[t]{0.31\textwidth}
    \caption{Knights opp. midgame\vphantom{gjpqy}}
    \centering
    \includegraphics[width=\textwidth]{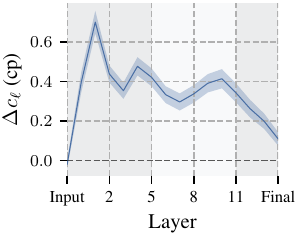}
  \end{subfigure}
  \hfill
  \begin{subfigure}[t]{0.31\textwidth}
    \caption{Knights opp. endgame\vphantom{gjpqy}}
    \centering
    \includegraphics[width=\textwidth]{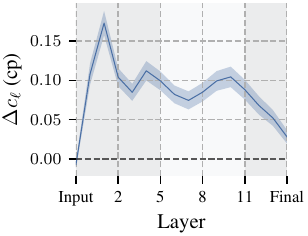}
  \end{subfigure}
  \hfill
  \begin{subfigure}[t]{0.31\textwidth}
    \caption{Knights opp. phased\vphantom{gjpqy}}
    \centering
    \includegraphics[width=\textwidth]{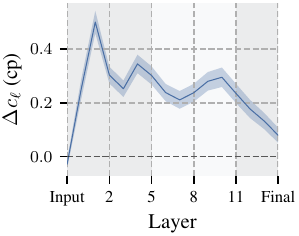}
  \end{subfigure}
  \caption{Stockfish evaluation concepts across layers (Part 1 of 7). Lines show mean probability-weighted concept delta across positions; shaded regions show 95\% confidence intervals. All concepts are evaluated from the current player's perspective.}
  \label{fig:concepts-part1}
\end{figure}

\begin{figure}[htbp]
  \centering
  \begin{subfigure}[t]{0.31\textwidth}
    \caption{Knights Total midgame\vphantom{gjpqy}}
    \centering
    \includegraphics[width=\textwidth]{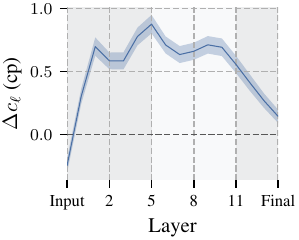}
  \end{subfigure}
  \hfill
  \begin{subfigure}[t]{0.31\textwidth}
    \caption{Knights Total endgame\vphantom{gjpqy}}
    \centering
    \includegraphics[width=\textwidth]{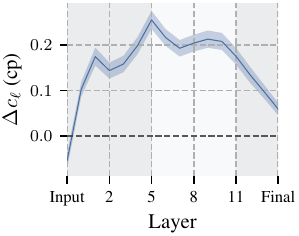}
  \end{subfigure}
  \hfill
  \begin{subfigure}[t]{0.31\textwidth}
    \caption{Knights Total phased\vphantom{gjpqy}}
    \centering
    \includegraphics[width=\textwidth]{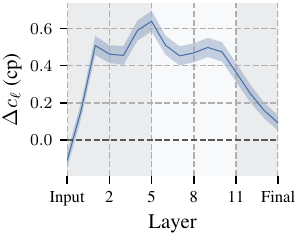}
  \end{subfigure}
  \\[0.3cm]
  \begin{subfigure}[t]{0.31\textwidth}
    \caption{Bishops mine midgame\vphantom{gjpqy}}
    \centering
    \includegraphics[width=\textwidth]{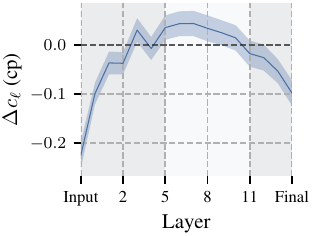}
  \end{subfigure}
  \hfill
  \begin{subfigure}[t]{0.31\textwidth}
    \caption{Bishops mine endgame\vphantom{gjpqy}}
    \centering
    \includegraphics[width=\textwidth]{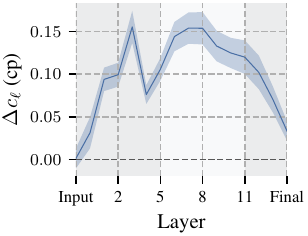}
  \end{subfigure}
  \hfill
  \begin{subfigure}[t]{0.31\textwidth}
    \caption{Bishops mine phased\vphantom{gjpqy}}
    \centering
    \includegraphics[width=\textwidth]{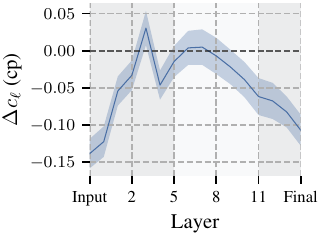}
  \end{subfigure}
  \\[0.3cm]
  \begin{subfigure}[t]{0.31\textwidth}
    \caption{Bishops opp. midgame\vphantom{gjpqy}}
    \centering
    \includegraphics[width=\textwidth]{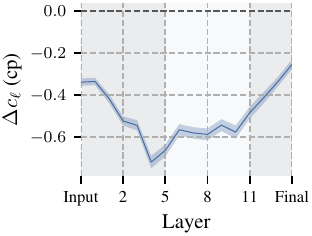}
  \end{subfigure}
  \hfill
  \begin{subfigure}[t]{0.31\textwidth}
    \caption{Bishops opp. endgame\vphantom{gjpqy}}
    \centering
    \includegraphics[width=\textwidth]{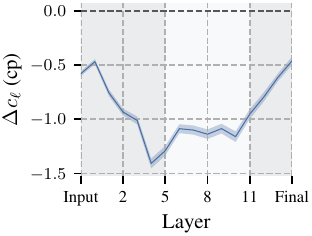}
  \end{subfigure}
  \hfill
  \begin{subfigure}[t]{0.31\textwidth}
    \caption{Bishops opp. phased\vphantom{gjpqy}}
    \centering
    \includegraphics[width=\textwidth]{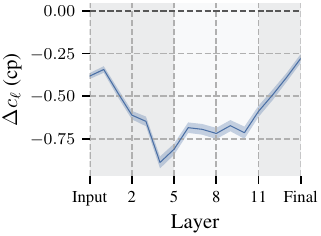}
  \end{subfigure}
  \\[0.3cm]
  \begin{subfigure}[t]{0.31\textwidth}
    \caption{Bishops Total midgame\vphantom{gjpqy}}
    \centering
    \includegraphics[width=\textwidth]{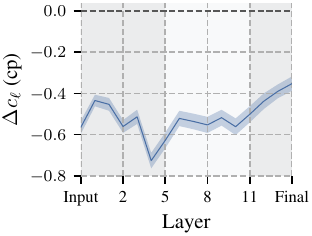}
  \end{subfigure}
  \hfill
  \begin{subfigure}[t]{0.31\textwidth}
    \caption{Bishops Total endgame\vphantom{gjpqy}}
    \centering
    \includegraphics[width=\textwidth]{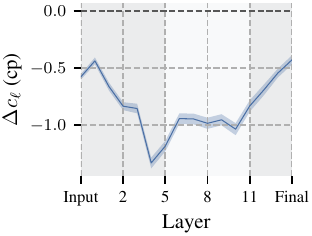}
  \end{subfigure}
  \hfill
  \begin{subfigure}[t]{0.31\textwidth}
    \caption{Bishops Total phased\vphantom{gjpqy}}
    \centering
    \includegraphics[width=\textwidth]{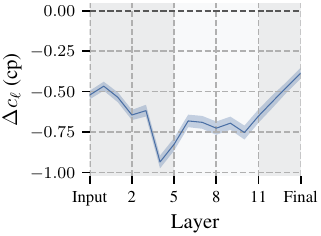}
  \end{subfigure}
  \\[0.3cm]
  \begin{subfigure}[t]{0.31\textwidth}
    \caption{Rooks mine midgame\vphantom{gjpqy}}
    \centering
    \includegraphics[width=\textwidth]{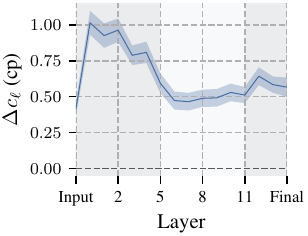}
  \end{subfigure}
  \hfill
  \begin{subfigure}[t]{0.31\textwidth}
    \caption{Rooks mine endgame\vphantom{gjpqy}}
    \centering
    \includegraphics[width=\textwidth]{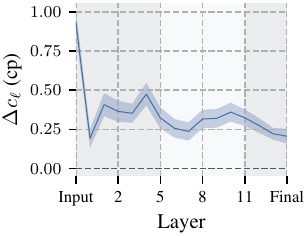}
  \end{subfigure}
  \hfill
  \begin{subfigure}[t]{0.31\textwidth}
    \caption{Rooks mine phased\vphantom{gjpqy}}
    \centering
    \includegraphics[width=\textwidth]{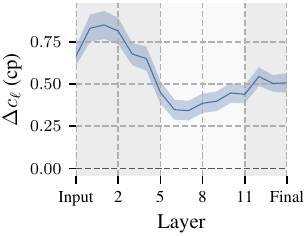}
  \end{subfigure}
  \caption{Stockfish evaluation concepts across layers (Part 2 of 7). Lines show mean probability-weighted concept delta across positions; shaded regions show 95\% confidence intervals. All concepts are evaluated from the current player's perspective.}
  \label{fig:concepts-part2}
\end{figure}

\begin{figure}[htbp]
  \centering
  \begin{subfigure}[t]{0.31\textwidth}
    \caption{Rooks opp. midgame\vphantom{gjpqy}}
    \centering
    \includegraphics[width=\textwidth]{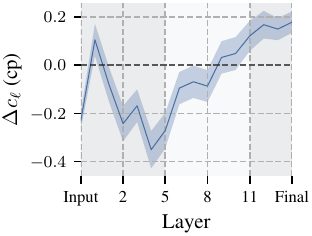}
  \end{subfigure}
  \hfill
  \begin{subfigure}[t]{0.31\textwidth}
    \caption{Rooks opp. endgame\vphantom{gjpqy}}
    \centering
    \includegraphics[width=\textwidth]{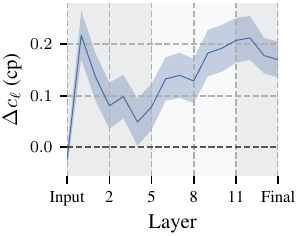}
  \end{subfigure}
  \hfill
  \begin{subfigure}[t]{0.31\textwidth}
    \caption{Rooks opp. phased\vphantom{gjpqy}}
    \centering
    \includegraphics[width=\textwidth]{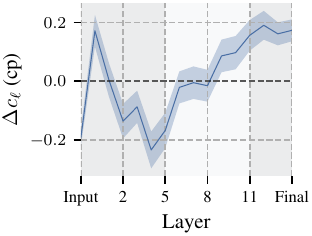}
  \end{subfigure}
  \\[0.3cm]
  \begin{subfigure}[t]{0.31\textwidth}
    \caption{Rooks Total midgame\vphantom{gjpqy}}
    \centering
    \includegraphics[width=\textwidth]{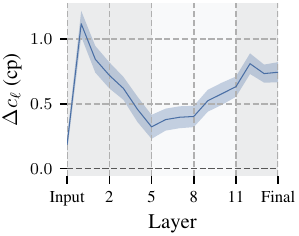}
  \end{subfigure}
  \hfill
  \begin{subfigure}[t]{0.31\textwidth}
    \caption{Rooks Total endgame\vphantom{gjpqy}}
    \centering
    \includegraphics[width=\textwidth]{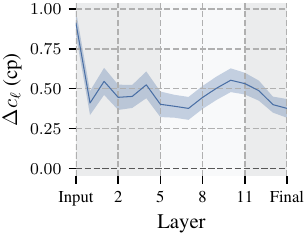}
  \end{subfigure}
  \hfill
  \begin{subfigure}[t]{0.31\textwidth}
    \caption{Rooks Total phased\vphantom{gjpqy}}
    \centering
    \includegraphics[width=\textwidth]{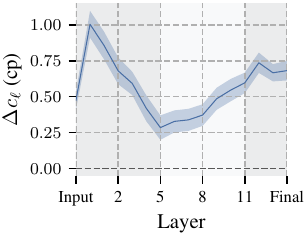}
  \end{subfigure}
  \\[0.3cm]
  \begin{subfigure}[t]{0.31\textwidth}
    \caption{Queens mine midgame\vphantom{gjpqy}}
    \centering
    \includegraphics[width=\textwidth]{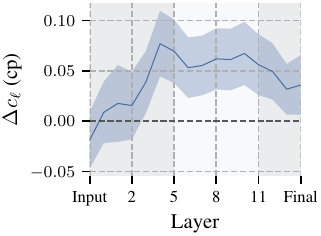}
  \end{subfigure}
  \hfill
  \begin{subfigure}[t]{0.31\textwidth}
    \caption{Queens mine endgame\vphantom{gjpqy}}
    \centering
    \includegraphics[width=\textwidth]{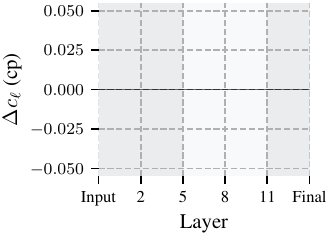}
  \end{subfigure}
  \hfill
  \begin{subfigure}[t]{0.31\textwidth}
    \caption{Queens mine phased\vphantom{gjpqy}}
    \centering
    \includegraphics[width=\textwidth]{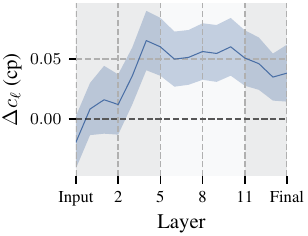}
  \end{subfigure}
  \\[0.3cm]
  \begin{subfigure}[t]{0.31\textwidth}
    \caption{Queens opp. midgame\vphantom{gjpqy}}
    \centering
    \includegraphics[width=\textwidth]{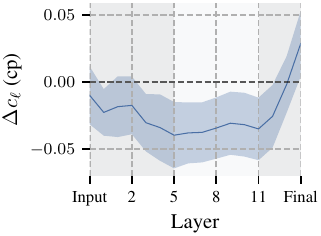}
  \end{subfigure}
  \hfill
  \begin{subfigure}[t]{0.31\textwidth}
    \caption{Queens opp. endgame\vphantom{gjpqy}}
    \centering
    \includegraphics[width=\textwidth]{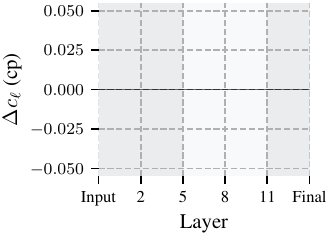}
  \end{subfigure}
  \hfill
  \begin{subfigure}[t]{0.31\textwidth}
    \caption{Queens opp. phased\vphantom{gjpqy}}
    \centering
    \includegraphics[width=\textwidth]{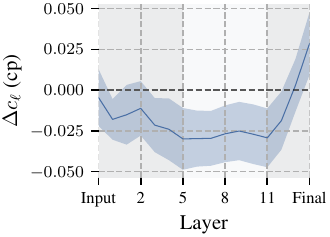}
  \end{subfigure}
  \\[0.3cm]
  \begin{subfigure}[t]{0.31\textwidth}
    \caption{Queens Total midgame\vphantom{gjpqy}}
    \centering
    \includegraphics[width=\textwidth]{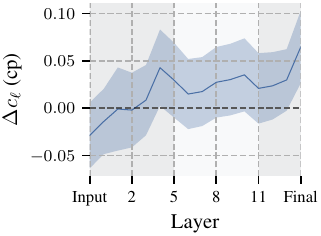}
  \end{subfigure}
  \hfill
  \begin{subfigure}[t]{0.31\textwidth}
    \caption{Queens Total endgame\vphantom{gjpqy}}
    \centering
    \includegraphics[width=\textwidth]{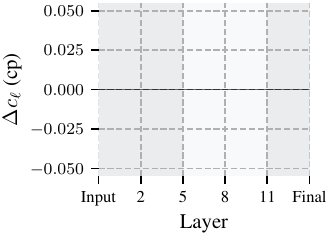}
  \end{subfigure}
  \hfill
  \begin{subfigure}[t]{0.31\textwidth}
    \caption{Queens Total phased\vphantom{gjpqy}}
    \centering
    \includegraphics[width=\textwidth]{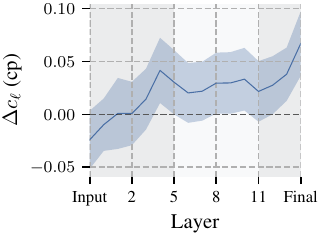}
  \end{subfigure}
  \caption{Stockfish evaluation concepts across layers (Part 3 of 7). Lines show mean probability-weighted concept delta across positions; shaded regions show 95\% confidence intervals. All concepts are evaluated from the current player's perspective.}
  \label{fig:concepts-part3}
\end{figure}

\begin{figure}[htbp]
  \centering
  \begin{subfigure}[t]{0.31\textwidth}
    \caption{Mobility mine midgame\vphantom{gjpqy}}
    \centering
    \includegraphics[width=\textwidth]{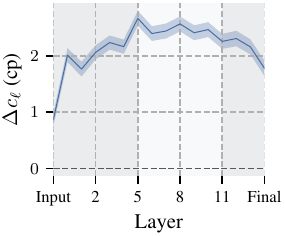}
  \end{subfigure}
  \hfill
  \begin{subfigure}[t]{0.31\textwidth}
    \caption{Mobility mine endgame\vphantom{gjpqy}}
    \centering
    \includegraphics[width=\textwidth]{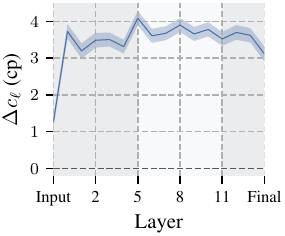}
  \end{subfigure}
  \hfill
  \begin{subfigure}[t]{0.31\textwidth}
    \caption{Mobility mine phased\vphantom{gjpqy}}
    \centering
    \includegraphics[width=\textwidth]{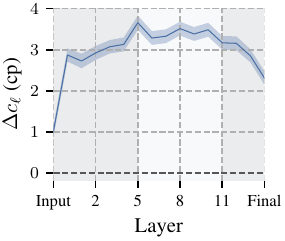}
  \end{subfigure}
  \\[0.3cm]
  \begin{subfigure}[t]{0.31\textwidth}
    \caption{Mobility opp. midgame\vphantom{gjpqy}}
    \centering
    \includegraphics[width=\textwidth]{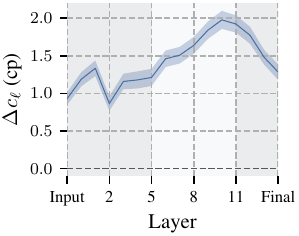}
  \end{subfigure}
  \hfill
  \begin{subfigure}[t]{0.31\textwidth}
    \caption{Mobility opp. endgame\vphantom{gjpqy}}
    \centering
    \includegraphics[width=\textwidth]{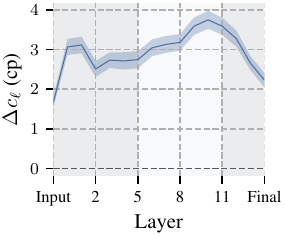}
  \end{subfigure}
  \hfill
  \begin{subfigure}[t]{0.31\textwidth}
    \caption{Mobility opp. phased\vphantom{gjpqy}}
    \centering
    \includegraphics[width=\textwidth]{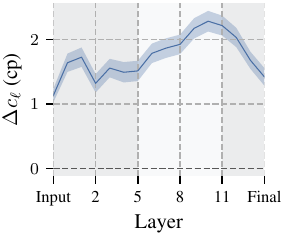}
  \end{subfigure}
  \\[0.3cm]
  \begin{subfigure}[t]{0.31\textwidth}
    \caption{Mobility Total midgame\vphantom{gjpqy}}
    \centering
    \includegraphics[width=\textwidth]{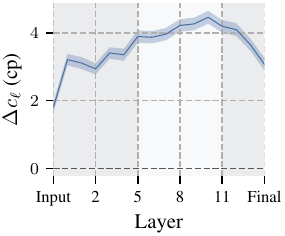}
  \end{subfigure}
  \hfill
  \begin{subfigure}[t]{0.31\textwidth}
    \caption{Mobility Total endgame\vphantom{gjpqy}}
    \centering
    \includegraphics[width=\textwidth]{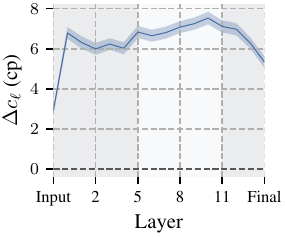}
  \end{subfigure}
  \hfill
  \begin{subfigure}[t]{0.31\textwidth}
    \caption{Mobility Total phased\vphantom{gjpqy}}
    \centering
    \includegraphics[width=\textwidth]{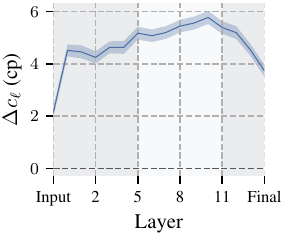}
  \end{subfigure}
  \\[0.3cm]
  \begin{subfigure}[t]{0.31\textwidth}
    \caption{King safety mine midgame\vphantom{gjpqy}}
    \centering
    \includegraphics[width=\textwidth]{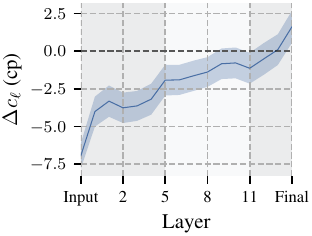}
  \end{subfigure}
  \hfill
  \begin{subfigure}[t]{0.31\textwidth}
    \caption{King safety mine endgame\vphantom{gjpqy}}
    \centering
    \includegraphics[width=\textwidth]{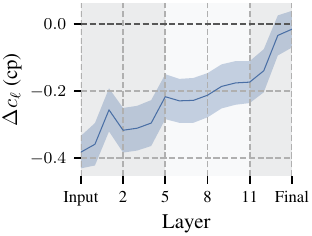}
  \end{subfigure}
  \hfill
  \begin{subfigure}[t]{0.31\textwidth}
    \caption{King safety mine phased\vphantom{gjpqy}}
    \centering
    \includegraphics[width=\textwidth]{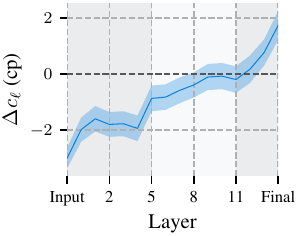}
  \end{subfigure}
  \\[0.3cm]
  \begin{subfigure}[t]{0.31\textwidth}
    \caption{King safety opp. midgame\vphantom{gjpqy}}
    \centering
    \includegraphics[width=\textwidth]{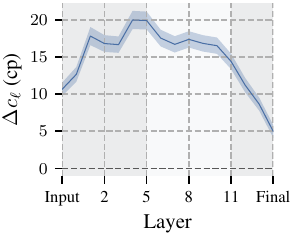}
  \end{subfigure}
  \hfill
  \begin{subfigure}[t]{0.31\textwidth}
    \caption{King safety opp. endgame\vphantom{gjpqy}}
    \centering
    \includegraphics[width=\textwidth]{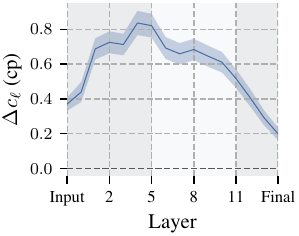}
  \end{subfigure}
  \hfill
  \begin{subfigure}[t]{0.31\textwidth}
    \caption{King safety opp. phased\vphantom{gjpqy}}
    \centering
    \includegraphics[width=\textwidth]{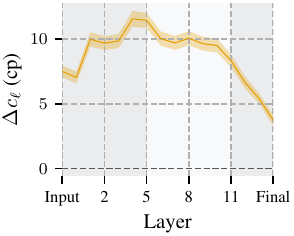}
  \end{subfigure}
  \caption{Stockfish evaluation concepts across layers (Part 4 of 7). Lines show mean probability-weighted concept delta across positions; shaded regions show 95\% confidence intervals. All concepts are evaluated from the current player's perspective.}
  \label{fig:concepts-part4}
\end{figure}

\begin{figure}[htbp]
  \centering
  \begin{subfigure}[t]{0.31\textwidth}
    \caption{King safety Total midgame\vphantom{gjpqy}}
    \centering
    \includegraphics[width=\textwidth]{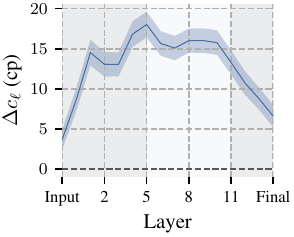}
  \end{subfigure}
  \hfill
  \begin{subfigure}[t]{0.31\textwidth}
    \caption{King safety Total endgame\vphantom{gjpqy}}
    \centering
    \includegraphics[width=\textwidth]{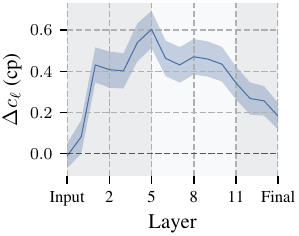}
  \end{subfigure}
  \hfill
  \begin{subfigure}[t]{0.31\textwidth}
    \caption{King safety Total phased\vphantom{gjpqy}}
    \centering
    \includegraphics[width=\textwidth]{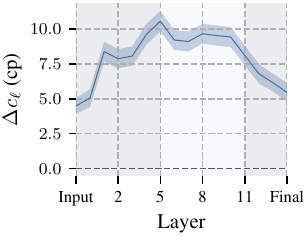}
  \end{subfigure}
  \\[0.3cm]
  \begin{subfigure}[t]{0.31\textwidth}
    \caption{Threats mine midgame\vphantom{gjpqy}}
    \centering
    \includegraphics[width=\textwidth]{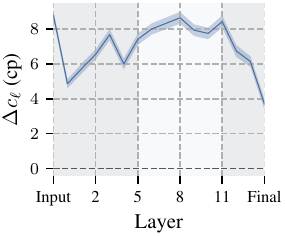}
  \end{subfigure}
  \hfill
  \begin{subfigure}[t]{0.31\textwidth}
    \caption{Threats mine endgame\vphantom{gjpqy}}
    \centering
    \includegraphics[width=\textwidth]{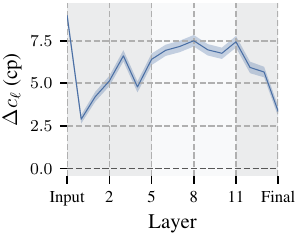}
  \end{subfigure}
  \hfill
  \begin{subfigure}[t]{0.31\textwidth}
    \caption{Threats mine phased\vphantom{gjpqy}}
    \centering
    \includegraphics[width=\textwidth]{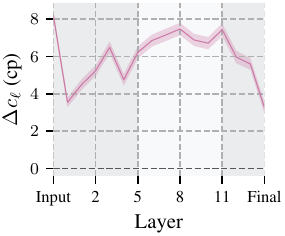}
  \end{subfigure}
  \\[0.3cm]
  \begin{subfigure}[t]{0.31\textwidth}
    \caption{Threats opp. midgame\vphantom{gjpqy}}
    \centering
    \includegraphics[width=\textwidth]{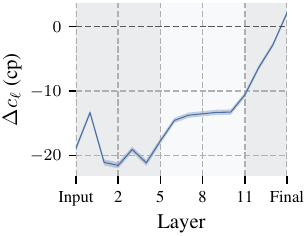}
  \end{subfigure}
  \hfill
  \begin{subfigure}[t]{0.31\textwidth}
    \caption{Threats opp. endgame\vphantom{gjpqy}}
    \centering
    \includegraphics[width=\textwidth]{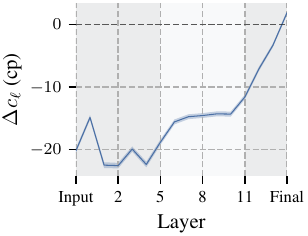}
  \end{subfigure}
  \hfill
  \begin{subfigure}[t]{0.31\textwidth}
    \caption{Threats opp. phased\vphantom{gjpqy}}
    \centering
    \includegraphics[width=\textwidth]{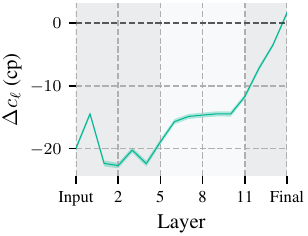}
  \end{subfigure}
  \\[0.3cm]
  \begin{subfigure}[t]{0.31\textwidth}
    \caption{Threats Total midgame\vphantom{gjpqy}}
    \centering
    \includegraphics[width=\textwidth]{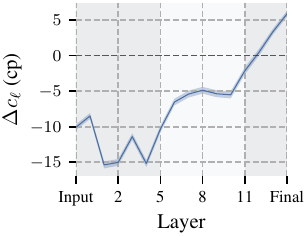}
  \end{subfigure}
  \hfill
  \begin{subfigure}[t]{0.31\textwidth}
    \caption{Threats Total endgame\vphantom{gjpqy}}
    \centering
    \includegraphics[width=\textwidth]{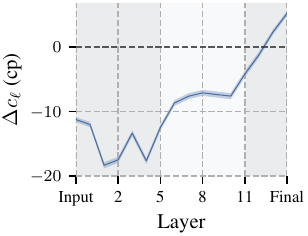}
  \end{subfigure}
  \hfill
  \begin{subfigure}[t]{0.31\textwidth}
    \caption{Threats Total phased\vphantom{gjpqy}}
    \centering
    \includegraphics[width=\textwidth]{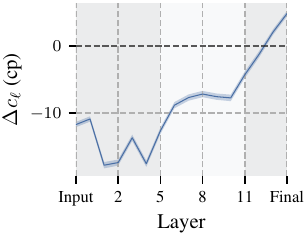}
  \end{subfigure}
  \\[0.3cm]
  \begin{subfigure}[t]{0.31\textwidth}
    \caption{Passed Pawns mine midgame\vphantom{gjpqy}}
    \centering
    \includegraphics[width=\textwidth]{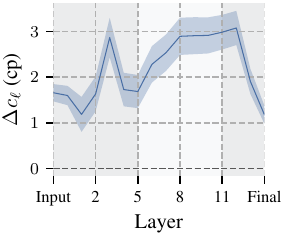}
  \end{subfigure}
  \hfill
  \begin{subfigure}[t]{0.31\textwidth}
    \caption{Passed Pawns mine endgame\vphantom{gjpqy}}
    \centering
    \includegraphics[width=\textwidth]{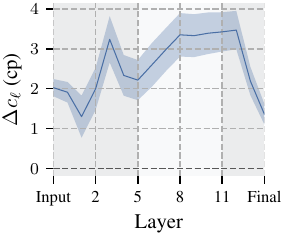}
  \end{subfigure}
  \hfill
  \begin{subfigure}[t]{0.31\textwidth}
    \caption{Passed Pawns mine phased\vphantom{gjpqy}}
    \centering
    \includegraphics[width=\textwidth]{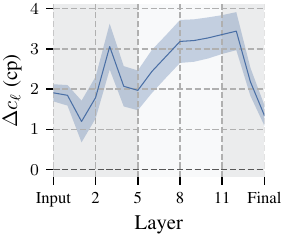}
  \end{subfigure}
  \caption{Stockfish evaluation concepts across layers (Part 5 of 7). Lines show mean probability-weighted concept delta across positions; shaded regions show 95\% confidence intervals. All concepts are evaluated from the current player's perspective.}
  \label{fig:concepts-part5}
\end{figure}

\begin{figure}[htbp]
  \centering
  \begin{subfigure}[t]{0.31\textwidth}
    \caption{Passed Pawns opp. midgame\vphantom{gjpqy}}
    \centering
    \includegraphics[width=\textwidth]{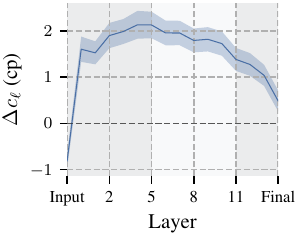}
  \end{subfigure}
  \hfill
  \begin{subfigure}[t]{0.31\textwidth}
    \caption{Passed Pawns opp. endgame\vphantom{gjpqy}}
    \centering
    \includegraphics[width=\textwidth]{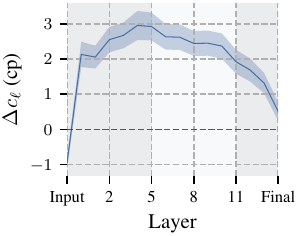}
  \end{subfigure}
  \hfill
  \begin{subfigure}[t]{0.31\textwidth}
    \caption{Passed Pawns opp. phased\vphantom{gjpqy}}
    \centering
    \includegraphics[width=\textwidth]{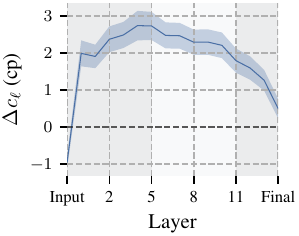}
  \end{subfigure}
  \\[0.3cm]
  \begin{subfigure}[t]{0.31\textwidth}
    \caption{Passed Pawns Total midgame\vphantom{gjpqy}}
    \centering
    \includegraphics[width=\textwidth]{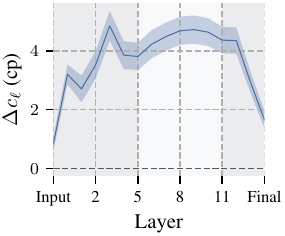}
  \end{subfigure}
  \hfill
  \begin{subfigure}[t]{0.31\textwidth}
    \caption{Passed Pawns Total endgame\vphantom{gjpqy}}
    \centering
    \includegraphics[width=\textwidth]{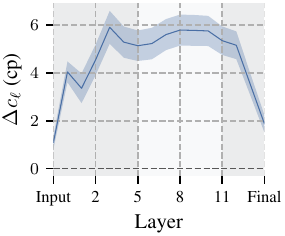}
  \end{subfigure}
  \hfill
  \begin{subfigure}[t]{0.31\textwidth}
    \caption{Passed Pawns Total phased\vphantom{gjpqy}}
    \centering
    \includegraphics[width=\textwidth]{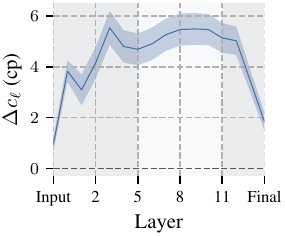}
  \end{subfigure}
  \\[0.3cm]
  \begin{subfigure}[t]{0.31\textwidth}
    \caption{Space mine midgame\vphantom{gjpqy}}
    \centering
    \includegraphics[width=\textwidth]{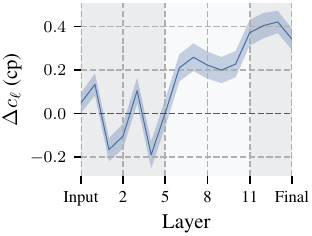}
  \end{subfigure}
  \hfill
  \begin{subfigure}[t]{0.31\textwidth}
    \caption{Space mine endgame\vphantom{gjpqy}}
    \centering
    \includegraphics[width=\textwidth]{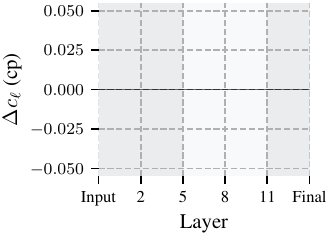}
  \end{subfigure}
  \hfill
  \begin{subfigure}[t]{0.31\textwidth}
    \caption{Space mine phased\vphantom{gjpqy}}
    \centering
    \includegraphics[width=\textwidth]{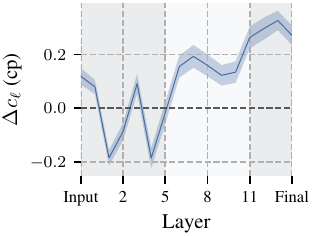}
  \end{subfigure}
  \\[0.3cm]
  \begin{subfigure}[t]{0.31\textwidth}
    \caption{Space opp. midgame\vphantom{gjpqy}}
    \centering
    \includegraphics[width=\textwidth]{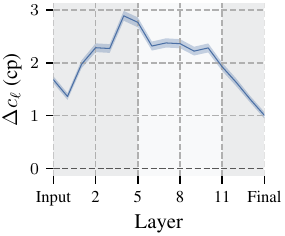}
  \end{subfigure}
  \hfill
  \begin{subfigure}[t]{0.31\textwidth}
    \caption{Space opp. endgame\vphantom{gjpqy}}
    \centering
    \includegraphics[width=\textwidth]{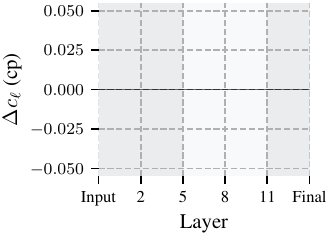}
  \end{subfigure}
  \hfill
  \begin{subfigure}[t]{0.31\textwidth}
    \caption{Space opp. phased\vphantom{gjpqy}}
    \centering
    \includegraphics[width=\textwidth]{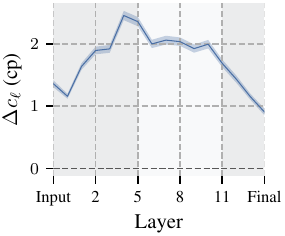}
  \end{subfigure}
  \\[0.3cm]
  \begin{subfigure}[t]{0.31\textwidth}
    \caption{Space Total midgame\vphantom{gjpqy}}
    \centering
    \includegraphics[width=\textwidth]{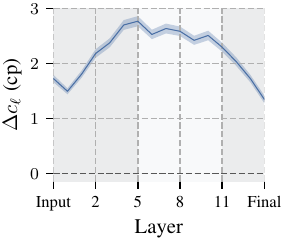}
  \end{subfigure}
  \hfill
  \begin{subfigure}[t]{0.31\textwidth}
    \caption{Space Total endgame\vphantom{gjpqy}}
    \centering
    \includegraphics[width=\textwidth]{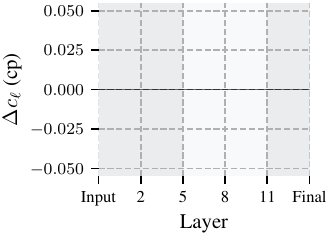}
  \end{subfigure}
  \hfill
  \begin{subfigure}[t]{0.31\textwidth}
    \caption{Space Total phased\vphantom{gjpqy}}
    \centering
    \includegraphics[width=\textwidth]{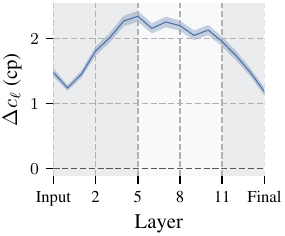}
  \end{subfigure}
  \caption{Stockfish evaluation concepts across layers (Part 6 of 7). Lines show mean probability-weighted concept delta across positions; shaded regions show 95\% confidence intervals. All concepts are evaluated from the current player's perspective.}
  \label{fig:concepts-part6}
\end{figure}

\begin{figure}[htbp]
  \centering
  \begin{subfigure}[t]{0.31\textwidth}
    \caption{Total Total midgame\vphantom{gjpqy}}
    \centering
    \includegraphics[width=\textwidth]{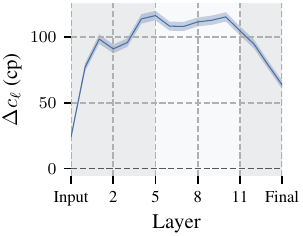}
  \end{subfigure}
  \hfill
  \begin{subfigure}[t]{0.31\textwidth}
    \caption{Total Total endgame\vphantom{gjpqy}}
    \centering
    \includegraphics[width=\textwidth]{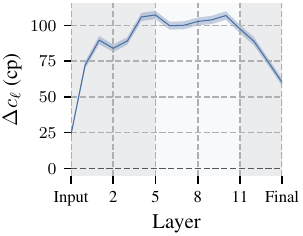}
  \end{subfigure}
  \hfill
  \begin{subfigure}[t]{0.31\textwidth}
    \caption{Total Total phased\vphantom{gjpqy}}
    \centering
    \includegraphics[width=\textwidth]{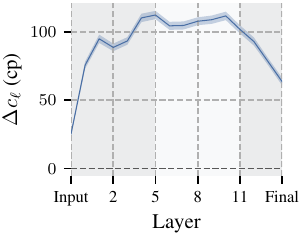}
  \end{subfigure}
  \\[0.3cm]
  \begin{subfigure}[t]{0.31\textwidth}
    \caption{Total evaluation\vphantom{gjpqy}}
    \centering
    \includegraphics[width=\textwidth]{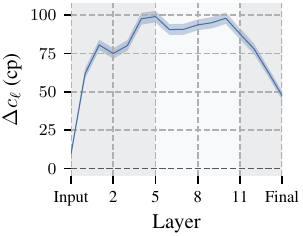}
  \end{subfigure}
  \hfill
  \begin{subfigure}[t]{0.31\textwidth}
    \caption{Total $-$ material phased\vphantom{gjpqy}}
    \centering
    \includegraphics[width=\textwidth]{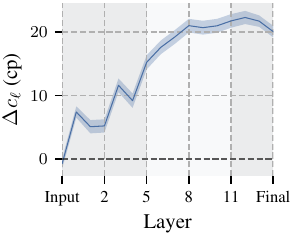}
  \end{subfigure}
  \caption{Stockfish evaluation concepts across layers (Part 7 of 7). Panel (d) shows Stockfish's final static evaluation, which differs from the phased total in (c) by scaling down drawish endgames and adding a bonus for the side to move. Panel (e) shows the phased total in (c) without the material term, which dominates the total. Lines show mean probability-weighted concept delta across positions; shaded regions show 95\% confidence intervals. All concepts are evaluated from the current player's perspective.}
  \label{fig:concepts-part7}
\end{figure}

%% file: Figures/steering/steering_figures.tex
\begin{figure}[htbp]
  \centering
  \begin{subfigure}[t]{0.31\textwidth}
    \caption{Material Total midgame\vphantom{gjpqy}}
    \centering
    \includegraphics[width=\textwidth]{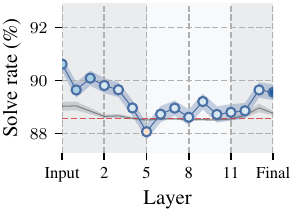}
  \end{subfigure}
  \hfill
  \begin{subfigure}[t]{0.31\textwidth}
    \caption{Material Total endgame\vphantom{gjpqy}}
    \centering
    \includegraphics[width=\textwidth]{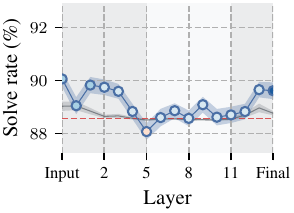}
  \end{subfigure}
  \hfill
  \begin{subfigure}[t]{0.31\textwidth}
    \caption{Material Total phased\vphantom{gjpqy}}
    \centering
    \includegraphics[width=\textwidth]{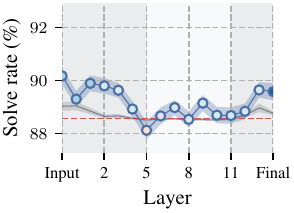}
  \end{subfigure}
  \\[0.1cm]
  \begin{subfigure}[t]{0.31\textwidth}
    \caption{Imbalance Total midgame\vphantom{gjpqy}}
    \centering
    \includegraphics[width=\textwidth]{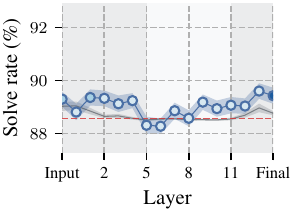}
  \end{subfigure}
  \hfill
  \begin{subfigure}[t]{0.31\textwidth}
    \caption{Imbalance Total endgame\vphantom{gjpqy}}
    \centering
    \includegraphics[width=\textwidth]{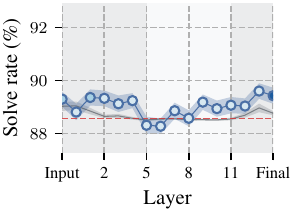}
  \end{subfigure}
  \hfill
  \begin{subfigure}[t]{0.31\textwidth}
    \caption{Imbalance Total phased\vphantom{gjpqy}}
    \centering
    \includegraphics[width=\textwidth]{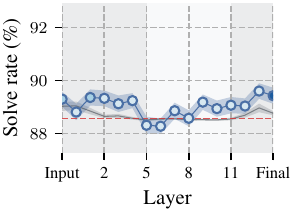}
  \end{subfigure}
  \\[0.1cm]
  \begin{subfigure}[t]{0.31\textwidth}
    \caption{Pawns Total midgame\vphantom{gjpqy}}
    \centering
    \includegraphics[width=\textwidth]{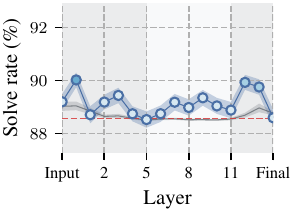}
  \end{subfigure}
  \hfill
  \begin{subfigure}[t]{0.31\textwidth}
    \caption{Pawns Total endgame\vphantom{gjpqy}}
    \centering
    \includegraphics[width=\textwidth]{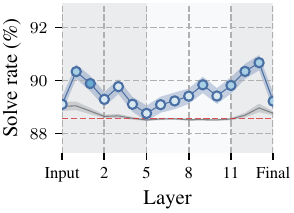}
  \end{subfigure}
  \hfill
  \begin{subfigure}[t]{0.31\textwidth}
    \caption{Pawns Total phased\vphantom{gjpqy}}
    \centering
    \includegraphics[width=\textwidth]{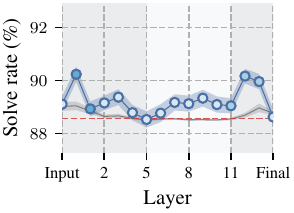}
  \end{subfigure}
  \\[0.1cm]
  \begin{subfigure}[t]{0.31\textwidth}
    \caption{Knights mine midgame\vphantom{gjpqy}}
    \centering
    \includegraphics[width=\textwidth]{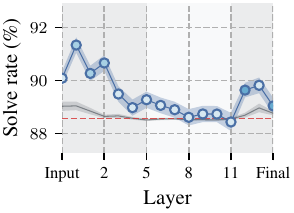}
  \end{subfigure}
  \hfill
  \begin{subfigure}[t]{0.31\textwidth}
    \caption{Knights mine endgame\vphantom{gjpqy}}
    \centering
    \includegraphics[width=\textwidth]{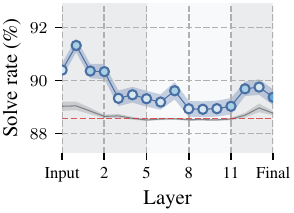}
  \end{subfigure}
  \hfill
  \begin{subfigure}[t]{0.31\textwidth}
    \caption{Knights mine phased\vphantom{gjpqy}}
    \centering
    \includegraphics[width=\textwidth]{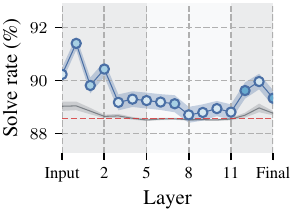}
  \end{subfigure}
  \\[0.1cm]
  \begin{subfigure}[t]{0.31\textwidth}
    \caption{Knights opp. midgame\vphantom{gjpqy}}
    \centering
    \includegraphics[width=\textwidth]{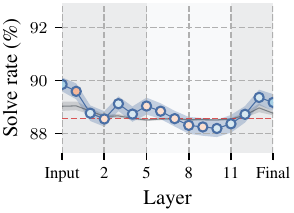}
  \end{subfigure}
  \hfill
  \begin{subfigure}[t]{0.31\textwidth}
    \caption{Knights opp. endgame\vphantom{gjpqy}}
    \centering
    \includegraphics[width=\textwidth]{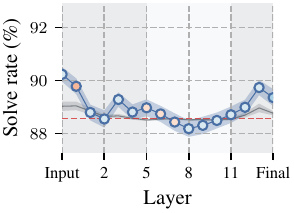}
  \end{subfigure}
  \hfill
  \begin{subfigure}[t]{0.31\textwidth}
    \caption{Knights opp. phased\vphantom{gjpqy}}
    \centering
    \includegraphics[width=\textwidth]{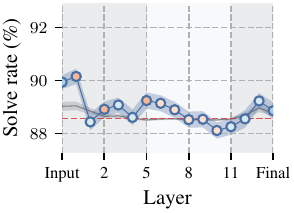}
  \end{subfigure}
  \\[0.1cm]
  \includegraphics[width=0.85\textwidth]{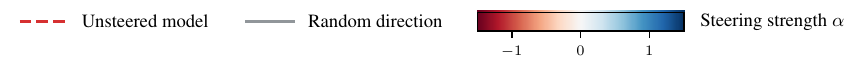}
  \caption{Concept steering effects across layers (Part 1 of 7). Each panel shows the held-out solve rate for the given concept at each intervention layer: the strength is selected on a random half of the puzzles and evaluated on the other half, averaged over $200$ random splits, and the shaded band shows one standard deviation across splits. Marker colors indicate the most frequently selected strength, \steerpos{blue} for positive and \steerneg{red} for negative values. The \randomdirection{gray line and band} show the mean and $95\%$ CI across $50$ random directions under the same protocol. Panels without a curve are terms that are zero in every position, so no steering vector exists. The \baselineunsteered{red dashed line} shows the unsteered baseline.}
  \label{fig:steering-part1}
\end{figure}

\begin{figure}[htbp]
  \centering
  \begin{subfigure}[t]{0.31\textwidth}
    \caption{Knights Total midgame\vphantom{gjpqy}}
    \centering
    \includegraphics[width=\textwidth]{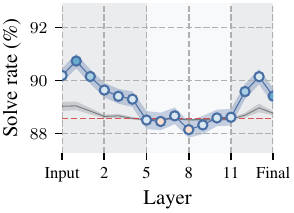}
  \end{subfigure}
  \hfill
  \begin{subfigure}[t]{0.31\textwidth}
    \caption{Knights Total endgame\vphantom{gjpqy}}
    \centering
    \includegraphics[width=\textwidth]{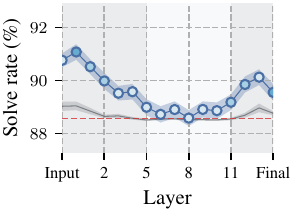}
  \end{subfigure}
  \hfill
  \begin{subfigure}[t]{0.31\textwidth}
    \caption{Knights Total phased\vphantom{gjpqy}}
    \centering
    \includegraphics[width=\textwidth]{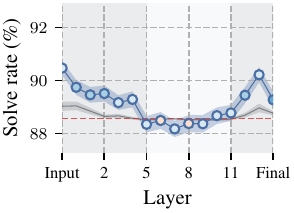}
  \end{subfigure}
  \\[0.1cm]
  \begin{subfigure}[t]{0.31\textwidth}
    \caption{Bishops mine midgame\vphantom{gjpqy}}
    \centering
    \includegraphics[width=\textwidth]{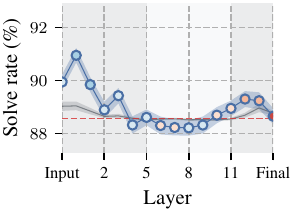}
  \end{subfigure}
  \hfill
  \begin{subfigure}[t]{0.31\textwidth}
    \caption{Bishops mine endgame\vphantom{gjpqy}}
    \centering
    \includegraphics[width=\textwidth]{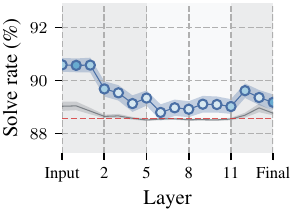}
  \end{subfigure}
  \hfill
  \begin{subfigure}[t]{0.31\textwidth}
    \caption{Bishops mine phased\vphantom{gjpqy}}
    \centering
    \includegraphics[width=\textwidth]{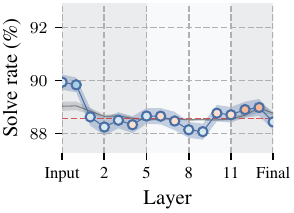}
  \end{subfigure}
  \\[0.1cm]
  \begin{subfigure}[t]{0.31\textwidth}
    \caption{Bishops opp. midgame\vphantom{gjpqy}}
    \centering
    \includegraphics[width=\textwidth]{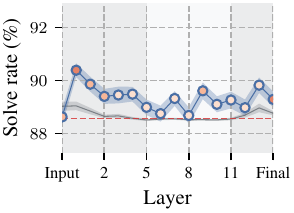}
  \end{subfigure}
  \hfill
  \begin{subfigure}[t]{0.31\textwidth}
    \caption{Bishops opp. endgame\vphantom{gjpqy}}
    \centering
    \includegraphics[width=\textwidth]{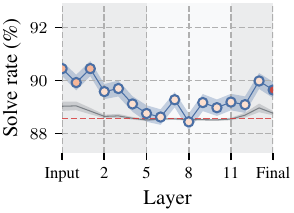}
  \end{subfigure}
  \hfill
  \begin{subfigure}[t]{0.31\textwidth}
    \caption{Bishops opp. phased\vphantom{gjpqy}}
    \centering
    \includegraphics[width=\textwidth]{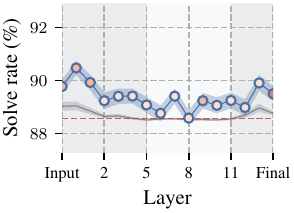}
  \end{subfigure}
  \\[0.1cm]
  \begin{subfigure}[t]{0.31\textwidth}
    \caption{Bishops Total midgame\vphantom{gjpqy}}
    \centering
    \includegraphics[width=\textwidth]{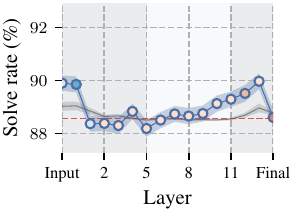}
  \end{subfigure}
  \hfill
  \begin{subfigure}[t]{0.31\textwidth}
    \caption{Bishops Total endgame\vphantom{gjpqy}}
    \centering
    \includegraphics[width=\textwidth]{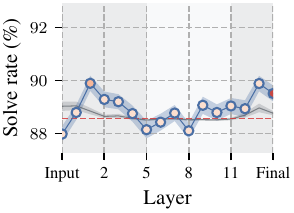}
  \end{subfigure}
  \hfill
  \begin{subfigure}[t]{0.31\textwidth}
    \caption{Bishops Total phased\vphantom{gjpqy}}
    \centering
    \includegraphics[width=\textwidth]{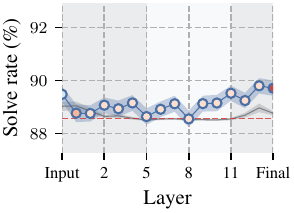}
  \end{subfigure}
  \\[0.1cm]
  \begin{subfigure}[t]{0.31\textwidth}
    \caption{Rooks mine midgame\vphantom{gjpqy}}
    \centering
    \includegraphics[width=\textwidth]{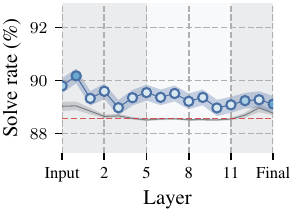}
  \end{subfigure}
  \hfill
  \begin{subfigure}[t]{0.31\textwidth}
    \caption{Rooks mine endgame\vphantom{gjpqy}}
    \centering
    \includegraphics[width=\textwidth]{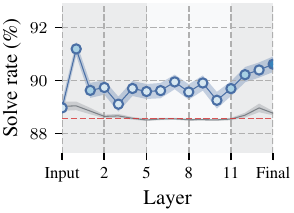}
  \end{subfigure}
  \hfill
  \begin{subfigure}[t]{0.31\textwidth}
    \caption{Rooks mine phased\vphantom{gjpqy}}
    \centering
    \includegraphics[width=\textwidth]{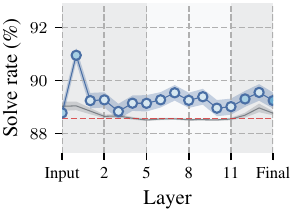}
  \end{subfigure}
  \\[0.1cm]
  \includegraphics[width=0.85\textwidth]{Figures/steering/appendix_steering_legend.pdf}
  \caption{Concept steering effects across layers (Part 2 of 7). Each panel shows the held-out solve rate for the given concept at each intervention layer: the strength is selected on a random half of the puzzles and evaluated on the other half, averaged over $200$ random splits, and the shaded band shows one standard deviation across splits. Marker colors indicate the most frequently selected strength, \steerpos{blue} for positive and \steerneg{red} for negative values. The \randomdirection{gray line and band} show the mean and $95\%$ CI across $50$ random directions under the same protocol. Panels without a curve are terms that are zero in every position, so no steering vector exists. The \baselineunsteered{red dashed line} shows the unsteered baseline.}
  \label{fig:steering-part2}
\end{figure}

\begin{figure}[htbp]
  \centering
  \begin{subfigure}[t]{0.31\textwidth}
    \caption{Rooks opp. midgame\vphantom{gjpqy}}
    \centering
    \includegraphics[width=\textwidth]{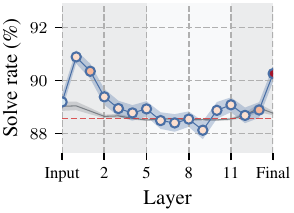}
  \end{subfigure}
  \hfill
  \begin{subfigure}[t]{0.31\textwidth}
    \caption{Rooks opp. endgame\vphantom{gjpqy}}
    \centering
    \includegraphics[width=\textwidth]{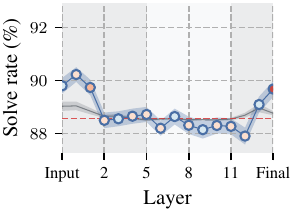}
  \end{subfigure}
  \hfill
  \begin{subfigure}[t]{0.31\textwidth}
    \caption{Rooks opp. phased\vphantom{gjpqy}}
    \centering
    \includegraphics[width=\textwidth]{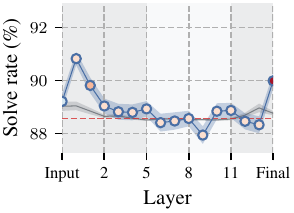}
  \end{subfigure}
  \\[0.1cm]
  \begin{subfigure}[t]{0.31\textwidth}
    \caption{Rooks Total midgame\vphantom{gjpqy}}
    \centering
    \includegraphics[width=\textwidth]{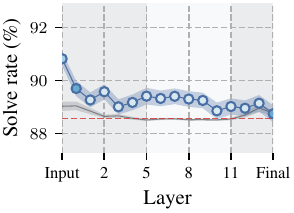}
  \end{subfigure}
  \hfill
  \begin{subfigure}[t]{0.31\textwidth}
    \caption{Rooks Total endgame\vphantom{gjpqy}}
    \centering
    \includegraphics[width=\textwidth]{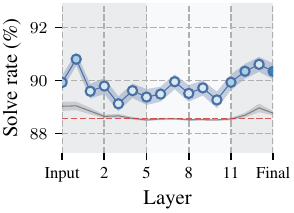}
  \end{subfigure}
  \hfill
  \begin{subfigure}[t]{0.31\textwidth}
    \caption{Rooks Total phased\vphantom{gjpqy}}
    \centering
    \includegraphics[width=\textwidth]{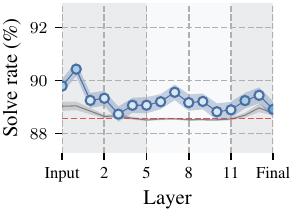}
  \end{subfigure}
  \\[0.1cm]
  \begin{subfigure}[t]{0.31\textwidth}
    \caption{Queens mine midgame\vphantom{gjpqy}}
    \centering
    \includegraphics[width=\textwidth]{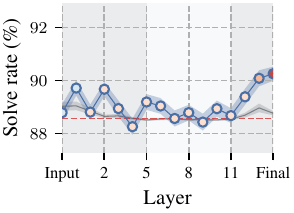}
  \end{subfigure}
  \hfill
  \begin{subfigure}[t]{0.31\textwidth}
    \caption{Queens mine endgame\vphantom{gjpqy}}
    \centering
    \includegraphics[width=\textwidth]{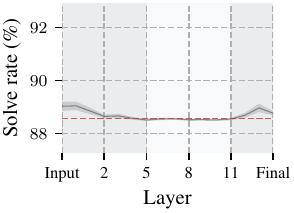}
  \end{subfigure}
  \hfill
  \begin{subfigure}[t]{0.31\textwidth}
    \caption{Queens mine phased\vphantom{gjpqy}}
    \centering
    \includegraphics[width=\textwidth]{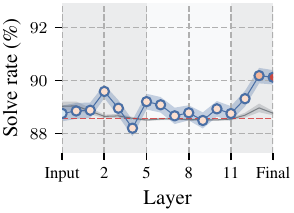}
  \end{subfigure}
  \\[0.1cm]
  \begin{subfigure}[t]{0.31\textwidth}
    \caption{Queens opp. midgame\vphantom{gjpqy}}
    \centering
    \includegraphics[width=\textwidth]{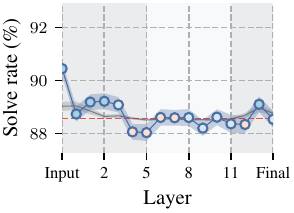}
  \end{subfigure}
  \hfill
  \begin{subfigure}[t]{0.31\textwidth}
    \caption{Queens opp. endgame\vphantom{gjpqy}}
    \centering
    \includegraphics[width=\textwidth]{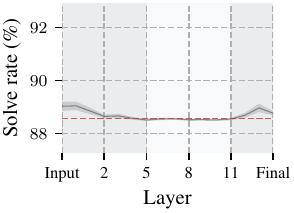}
  \end{subfigure}
  \hfill
  \begin{subfigure}[t]{0.31\textwidth}
    \caption{Queens opp. phased\vphantom{gjpqy}}
    \centering
    \includegraphics[width=\textwidth]{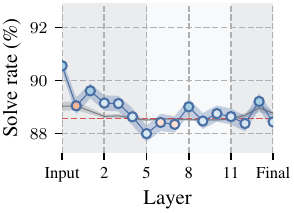}
  \end{subfigure}
  \\[0.1cm]
  \begin{subfigure}[t]{0.31\textwidth}
    \caption{Queens Total midgame\vphantom{gjpqy}}
    \centering
    \includegraphics[width=\textwidth]{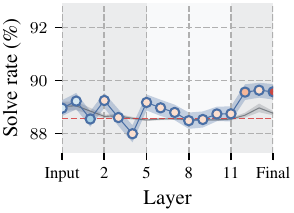}
  \end{subfigure}
  \hfill
  \begin{subfigure}[t]{0.31\textwidth}
    \caption{Queens Total endgame\vphantom{gjpqy}}
    \centering
    \includegraphics[width=\textwidth]{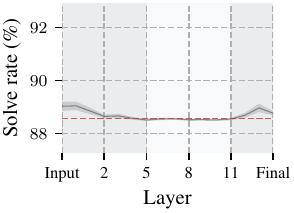}
  \end{subfigure}
  \hfill
  \begin{subfigure}[t]{0.31\textwidth}
    \caption{Queens Total phased\vphantom{gjpqy}}
    \centering
    \includegraphics[width=\textwidth]{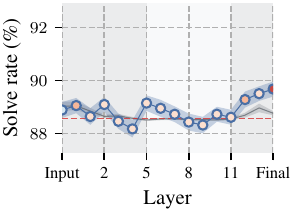}
  \end{subfigure}
  \\[0.1cm]
  \includegraphics[width=0.85\textwidth]{Figures/steering/appendix_steering_legend.pdf}
  \caption{Concept steering effects across layers (Part 3 of 7). Each panel shows the held-out solve rate for the given concept at each intervention layer: the strength is selected on a random half of the puzzles and evaluated on the other half, averaged over $200$ random splits, and the shaded band shows one standard deviation across splits. Marker colors indicate the most frequently selected strength, \steerpos{blue} for positive and \steerneg{red} for negative values. The \randomdirection{gray line and band} show the mean and $95\%$ CI across $50$ random directions under the same protocol. Panels without a curve are terms that are zero in every position, so no steering vector exists. The \baselineunsteered{red dashed line} shows the unsteered baseline.}
  \label{fig:steering-part3}
\end{figure}

\begin{figure}[htbp]
  \centering
  \begin{subfigure}[t]{0.31\textwidth}
    \caption{Mobility mine midgame\vphantom{gjpqy}}
    \centering
    \includegraphics[width=\textwidth]{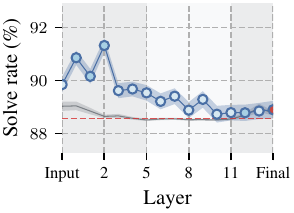}
  \end{subfigure}
  \hfill
  \begin{subfigure}[t]{0.31\textwidth}
    \caption{Mobility mine endgame\vphantom{gjpqy}}
    \centering
    \includegraphics[width=\textwidth]{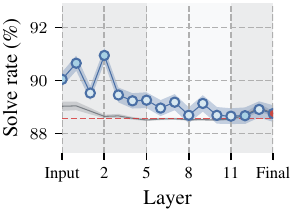}
  \end{subfigure}
  \hfill
  \begin{subfigure}[t]{0.31\textwidth}
    \caption{Mobility mine phased\vphantom{gjpqy}}
    \centering
    \includegraphics[width=\textwidth]{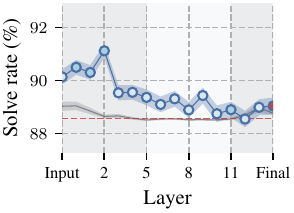}
  \end{subfigure}
  \\[0.1cm]
  \begin{subfigure}[t]{0.31\textwidth}
    \caption{Mobility opp. midgame\vphantom{gjpqy}}
    \centering
    \includegraphics[width=\textwidth]{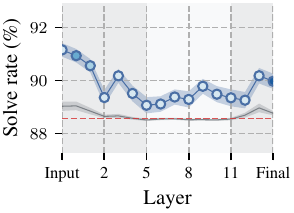}
  \end{subfigure}
  \hfill
  \begin{subfigure}[t]{0.31\textwidth}
    \caption{Mobility opp. endgame\vphantom{gjpqy}}
    \centering
    \includegraphics[width=\textwidth]{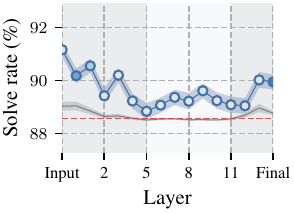}
  \end{subfigure}
  \hfill
  \begin{subfigure}[t]{0.31\textwidth}
    \caption{Mobility opp. phased\vphantom{gjpqy}}
    \centering
    \includegraphics[width=\textwidth]{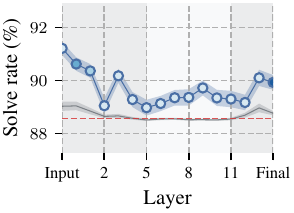}
  \end{subfigure}
  \\[0.1cm]
  \begin{subfigure}[t]{0.31\textwidth}
    \caption{Mobility Total midgame\vphantom{gjpqy}}
    \centering
    \includegraphics[width=\textwidth]{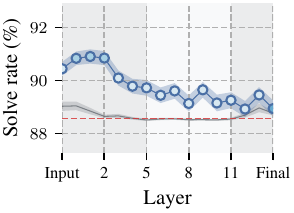}
  \end{subfigure}
  \hfill
  \begin{subfigure}[t]{0.31\textwidth}
    \caption{Mobility Total endgame\vphantom{gjpqy}}
    \centering
    \includegraphics[width=\textwidth]{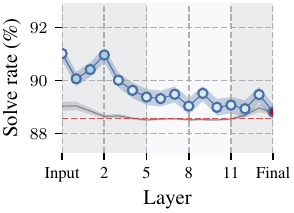}
  \end{subfigure}
  \hfill
  \begin{subfigure}[t]{0.31\textwidth}
    \caption{Mobility Total phased\vphantom{gjpqy}}
    \centering
    \includegraphics[width=\textwidth]{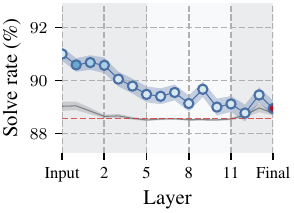}
  \end{subfigure}
  \\[0.1cm]
  \begin{subfigure}[t]{0.31\textwidth}
    \caption{King safety mine midgame\vphantom{gjpqy}}
    \centering
    \includegraphics[width=\textwidth]{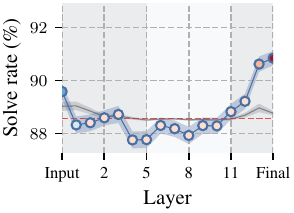}
  \end{subfigure}
  \hfill
  \begin{subfigure}[t]{0.31\textwidth}
    \caption{King safety mine endgame\vphantom{gjpqy}}
    \centering
    \includegraphics[width=\textwidth]{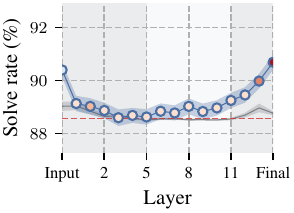}
  \end{subfigure}
  \hfill
  \begin{subfigure}[t]{0.31\textwidth}
    \caption{King safety mine phased\vphantom{gjpqy}}
    \centering
    \includegraphics[width=\textwidth]{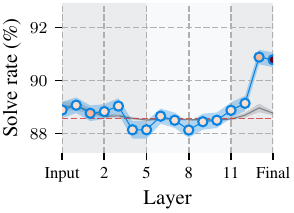}
  \end{subfigure}
  \\[0.1cm]
  \begin{subfigure}[t]{0.31\textwidth}
    \caption{King safety opp. midgame\vphantom{gjpqy}}
    \centering
    \includegraphics[width=\textwidth]{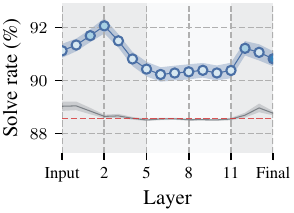}
  \end{subfigure}
  \hfill
  \begin{subfigure}[t]{0.31\textwidth}
    \caption{King safety opp. endgame\vphantom{gjpqy}}
    \centering
    \includegraphics[width=\textwidth]{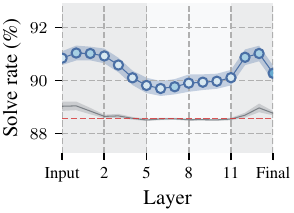}
  \end{subfigure}
  \hfill
  \begin{subfigure}[t]{0.31\textwidth}
    \caption{King safety opp. phased\vphantom{gjpqy}}
    \centering
    \includegraphics[width=\textwidth]{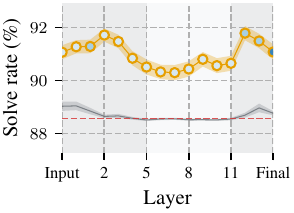}
  \end{subfigure}
  \\[0.1cm]
  \includegraphics[width=0.85\textwidth]{Figures/steering/appendix_steering_legend.pdf}
  \caption{Concept steering effects across layers (Part 4 of 7). Each panel shows the held-out solve rate for the given concept at each intervention layer: the strength is selected on a random half of the puzzles and evaluated on the other half, averaged over $200$ random splits, and the shaded band shows one standard deviation across splits. Marker colors indicate the most frequently selected strength, \steerpos{blue} for positive and \steerneg{red} for negative values. The \randomdirection{gray line and band} show the mean and $95\%$ CI across $50$ random directions under the same protocol. Panels without a curve are terms that are zero in every position, so no steering vector exists. The \baselineunsteered{red dashed line} shows the unsteered baseline.}
  \label{fig:steering-part4}
\end{figure}

\begin{figure}[htbp]
  \centering
  \begin{subfigure}[t]{0.31\textwidth}
    \caption{King safety Total midgame\vphantom{gjpqy}}
    \centering
    \includegraphics[width=\textwidth]{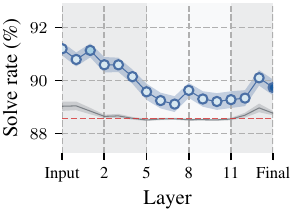}
  \end{subfigure}
  \hfill
  \begin{subfigure}[t]{0.31\textwidth}
    \caption{King safety Total endgame\vphantom{gjpqy}}
    \centering
    \includegraphics[width=\textwidth]{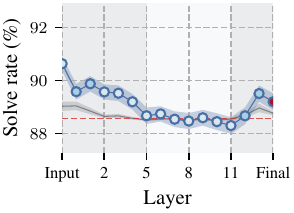}
  \end{subfigure}
  \hfill
  \begin{subfigure}[t]{0.31\textwidth}
    \caption{King safety Total phased\vphantom{gjpqy}}
    \centering
    \includegraphics[width=\textwidth]{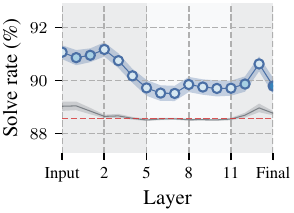}
  \end{subfigure}
  \\[0.1cm]
  \begin{subfigure}[t]{0.31\textwidth}
    \caption{Threats mine midgame\vphantom{gjpqy}}
    \centering
    \includegraphics[width=\textwidth]{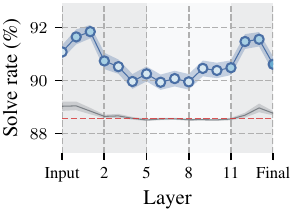}
  \end{subfigure}
  \hfill
  \begin{subfigure}[t]{0.31\textwidth}
    \caption{Threats mine endgame\vphantom{gjpqy}}
    \centering
    \includegraphics[width=\textwidth]{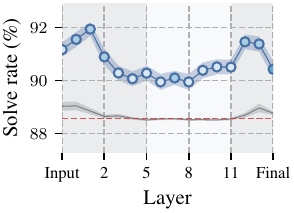}
  \end{subfigure}
  \hfill
  \begin{subfigure}[t]{0.31\textwidth}
    \caption{Threats mine phased\vphantom{gjpqy}}
    \centering
    \includegraphics[width=\textwidth]{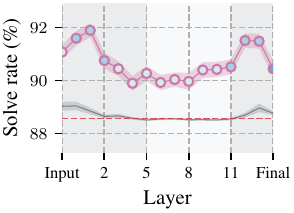}
  \end{subfigure}
  \\[0.1cm]
  \begin{subfigure}[t]{0.31\textwidth}
    \caption{Threats opp. midgame\vphantom{gjpqy}}
    \centering
    \includegraphics[width=\textwidth]{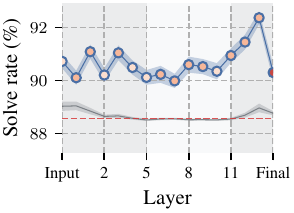}
  \end{subfigure}
  \hfill
  \begin{subfigure}[t]{0.31\textwidth}
    \caption{Threats opp. endgame\vphantom{gjpqy}}
    \centering
    \includegraphics[width=\textwidth]{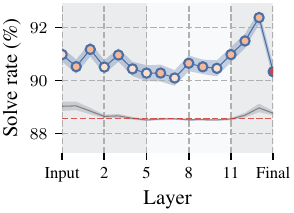}
  \end{subfigure}
  \hfill
  \begin{subfigure}[t]{0.31\textwidth}
    \caption{Threats opp. phased\vphantom{gjpqy}}
    \centering
    \includegraphics[width=\textwidth]{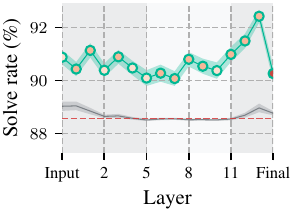}
  \end{subfigure}
  \\[0.1cm]
  \begin{subfigure}[t]{0.31\textwidth}
    \caption{Threats Total midgame\vphantom{gjpqy}}
    \centering
    \includegraphics[width=\textwidth]{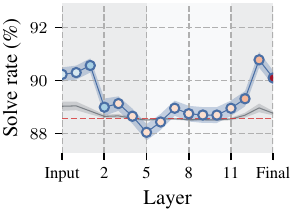}
  \end{subfigure}
  \hfill
  \begin{subfigure}[t]{0.31\textwidth}
    \caption{Threats Total endgame\vphantom{gjpqy}}
    \centering
    \includegraphics[width=\textwidth]{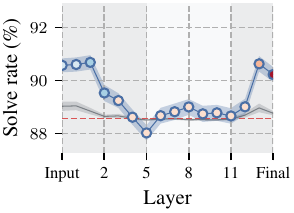}
  \end{subfigure}
  \hfill
  \begin{subfigure}[t]{0.31\textwidth}
    \caption{Threats Total phased\vphantom{gjpqy}}
    \centering
    \includegraphics[width=\textwidth]{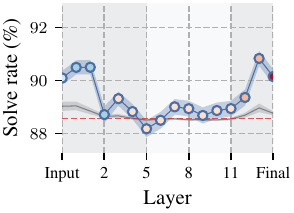}
  \end{subfigure}
  \\[0.1cm]
  \begin{subfigure}[t]{0.31\textwidth}
    \caption{Passed Pawns mine midgame\vphantom{gjpqy}}
    \centering
    \includegraphics[width=\textwidth]{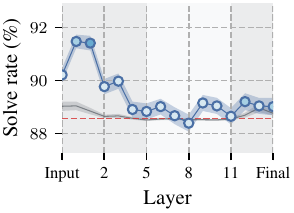}
  \end{subfigure}
  \hfill
  \begin{subfigure}[t]{0.31\textwidth}
    \caption{Passed Pawns mine endgame\vphantom{gjpqy}}
    \centering
    \includegraphics[width=\textwidth]{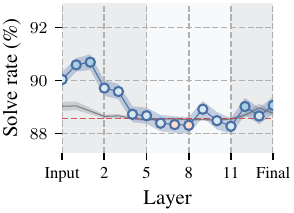}
  \end{subfigure}
  \hfill
  \begin{subfigure}[t]{0.31\textwidth}
    \caption{Passed Pawns mine phased\vphantom{gjpqy}}
    \centering
    \includegraphics[width=\textwidth]{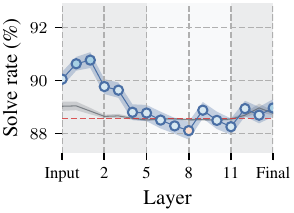}
  \end{subfigure}
  \\[0.1cm]
  \includegraphics[width=0.85\textwidth]{Figures/steering/appendix_steering_legend.pdf}
  \caption{Concept steering effects across layers (Part 5 of 7). Each panel shows the held-out solve rate for the given concept at each intervention layer: the strength is selected on a random half of the puzzles and evaluated on the other half, averaged over $200$ random splits, and the shaded band shows one standard deviation across splits. Marker colors indicate the most frequently selected strength, \steerpos{blue} for positive and \steerneg{red} for negative values. The \randomdirection{gray line and band} show the mean and $95\%$ CI across $50$ random directions under the same protocol. Panels without a curve are terms that are zero in every position, so no steering vector exists. The \baselineunsteered{red dashed line} shows the unsteered baseline.}
  \label{fig:steering-part5}
\end{figure}

\begin{figure}[htbp]
  \centering
  \begin{subfigure}[t]{0.31\textwidth}
    \caption{Passed Pawns opp. midgame\vphantom{gjpqy}}
    \centering
    \includegraphics[width=\textwidth]{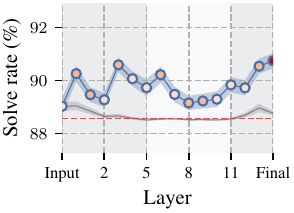}
  \end{subfigure}
  \hfill
  \begin{subfigure}[t]{0.31\textwidth}
    \caption{Passed Pawns opp. endgame\vphantom{gjpqy}}
    \centering
    \includegraphics[width=\textwidth]{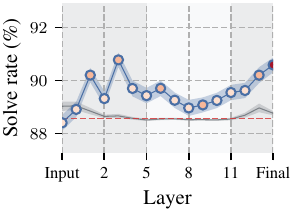}
  \end{subfigure}
  \hfill
  \begin{subfigure}[t]{0.31\textwidth}
    \caption{Passed Pawns opp. phased\vphantom{gjpqy}}
    \centering
    \includegraphics[width=\textwidth]{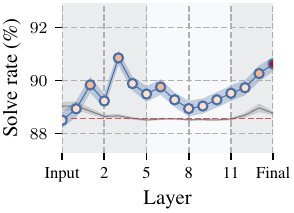}
  \end{subfigure}
  \\[0.1cm]
  \begin{subfigure}[t]{0.31\textwidth}
    \caption{Passed Pawns Total midgame\vphantom{gjpqy}}
    \centering
    \includegraphics[width=\textwidth]{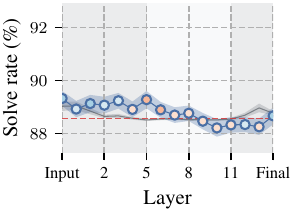}
  \end{subfigure}
  \hfill
  \begin{subfigure}[t]{0.31\textwidth}
    \caption{Passed Pawns Total endgame\vphantom{gjpqy}}
    \centering
    \includegraphics[width=\textwidth]{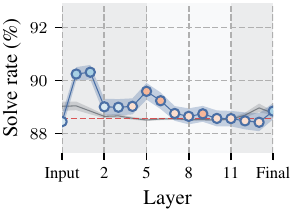}
  \end{subfigure}
  \hfill
  \begin{subfigure}[t]{0.31\textwidth}
    \caption{Passed Pawns Total phased\vphantom{gjpqy}}
    \centering
    \includegraphics[width=\textwidth]{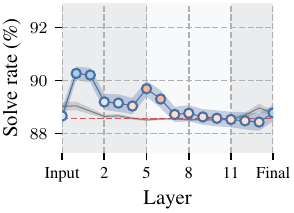}
  \end{subfigure}
  \\[0.1cm]
  \begin{subfigure}[t]{0.31\textwidth}
    \caption{Space mine midgame\vphantom{gjpqy}}
    \centering
    \includegraphics[width=\textwidth]{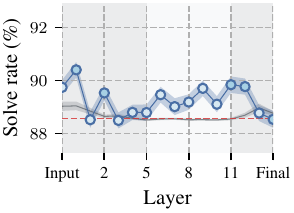}
  \end{subfigure}
  \hfill
  \begin{subfigure}[t]{0.31\textwidth}
    \caption{Space mine endgame\vphantom{gjpqy}}
    \centering
    \includegraphics[width=\textwidth]{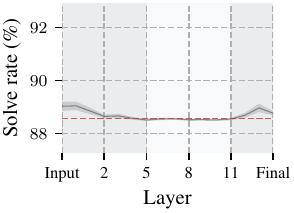}
  \end{subfigure}
  \hfill
  \begin{subfigure}[t]{0.31\textwidth}
    \caption{Space mine phased\vphantom{gjpqy}}
    \centering
    \includegraphics[width=\textwidth]{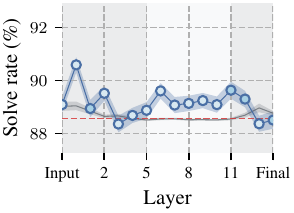}
  \end{subfigure}
  \\[0.1cm]
  \begin{subfigure}[t]{0.31\textwidth}
    \caption{Space opp. midgame\vphantom{gjpqy}}
    \centering
    \includegraphics[width=\textwidth]{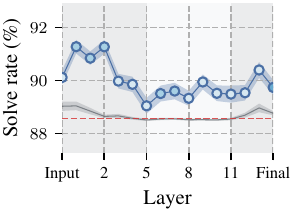}
  \end{subfigure}
  \hfill
  \begin{subfigure}[t]{0.31\textwidth}
    \caption{Space opp. endgame\vphantom{gjpqy}}
    \centering
    \includegraphics[width=\textwidth]{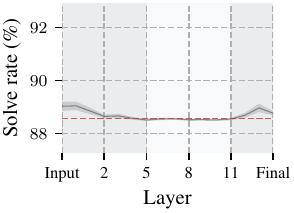}
  \end{subfigure}
  \hfill
  \begin{subfigure}[t]{0.31\textwidth}
    \caption{Space opp. phased\vphantom{gjpqy}}
    \centering
    \includegraphics[width=\textwidth]{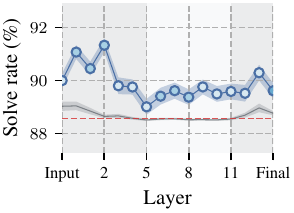}
  \end{subfigure}
  \\[0.1cm]
  \begin{subfigure}[t]{0.31\textwidth}
    \caption{Space Total midgame\vphantom{gjpqy}}
    \centering
    \includegraphics[width=\textwidth]{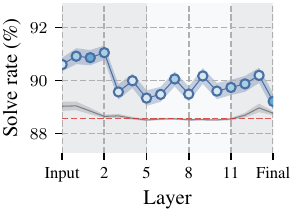}
  \end{subfigure}
  \hfill
  \begin{subfigure}[t]{0.31\textwidth}
    \caption{Space Total endgame\vphantom{gjpqy}}
    \centering
    \includegraphics[width=\textwidth]{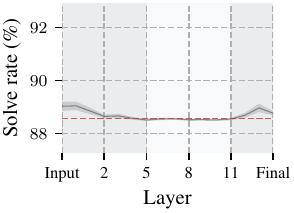}
  \end{subfigure}
  \hfill
  \begin{subfigure}[t]{0.31\textwidth}
    \caption{Space Total phased\vphantom{gjpqy}}
    \centering
    \includegraphics[width=\textwidth]{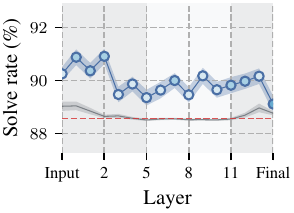}
  \end{subfigure}
  \\[0.1cm]
  \includegraphics[width=0.85\textwidth]{Figures/steering/appendix_steering_legend.pdf}
  \caption{Concept steering effects across layers (Part 6 of 7). Each panel shows the held-out solve rate for the given concept at each intervention layer: the strength is selected on a random half of the puzzles and evaluated on the other half, averaged over $200$ random splits, and the shaded band shows one standard deviation across splits. Marker colors indicate the most frequently selected strength, \steerpos{blue} for positive and \steerneg{red} for negative values. The \randomdirection{gray line and band} show the mean and $95\%$ CI across $50$ random directions under the same protocol. Panels without a curve are terms that are zero in every position, so no steering vector exists. The \baselineunsteered{red dashed line} shows the unsteered baseline.}
  \label{fig:steering-part6}
\end{figure}

\begin{figure}[htbp]
  \centering
  \begin{subfigure}[t]{0.31\textwidth}
    \caption{Total Total midgame\vphantom{gjpqy}}
    \centering
    \includegraphics[width=\textwidth]{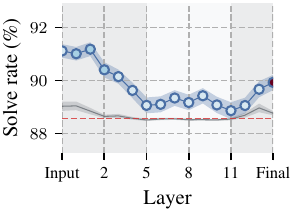}
  \end{subfigure}
  \hfill
  \begin{subfigure}[t]{0.31\textwidth}
    \caption{Total Total endgame\vphantom{gjpqy}}
    \centering
    \includegraphics[width=\textwidth]{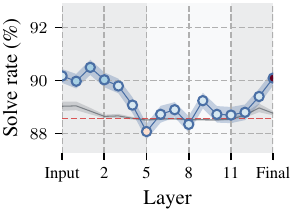}
  \end{subfigure}
  \hfill
  \begin{subfigure}[t]{0.31\textwidth}
    \caption{Total Total phased\vphantom{gjpqy}}
    \centering
    \includegraphics[width=\textwidth]{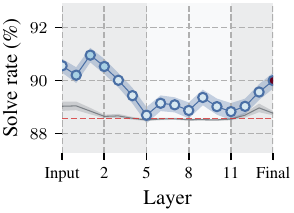}
  \end{subfigure}
  \\[0.1cm]
  \begin{subfigure}[t]{0.31\textwidth}
    \caption{Total evaluation\vphantom{gjpqy}}
    \centering
    \includegraphics[width=\textwidth]{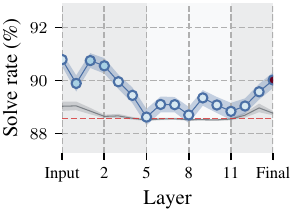}
  \end{subfigure}
  \\[0.1cm]
  \includegraphics[width=0.85\textwidth]{Figures/steering/appendix_steering_legend.pdf}
  \caption{Concept steering effects across layers (Part 7 of 7). Each panel shows the held-out solve rate for the given concept at each intervention layer: the strength is selected on a random half of the puzzles and evaluated on the other half, averaged over $200$ random splits, and the shaded band shows one standard deviation across splits. Marker colors indicate the most frequently selected strength, \steerpos{blue} for positive and \steerneg{red} for negative values. The \randomdirection{gray line and band} show the mean and $95\%$ CI across $50$ random directions under the same protocol. Panels without a curve are terms that are zero in every position, so no steering vector exists. The \baselineunsteered{red dashed line} shows the unsteered baseline.}
  \label{fig:steering-part7}
\end{figure}